\documentclass[letterpaper]{article} %
\usepackage{aaai2026} 
\nocopyright %
\usepackage{times}  %
\usepackage{helvet}  %
\usepackage{courier}  %
\usepackage[hyphens]{url}  %
\usepackage{graphicx} %
\usepackage{natbib}  %
\usepackage{caption} %
\usepackage{algorithm} %
\usepackage[noend]{algpseudocode} %
\usepackage{amsmath} %
\usepackage{placeins} %
\usepackage{dsfont} %

\usepackage{booktabs} %

\newcommand{\cameraready}[1]{#1}

\usepackage{subcaption}
\usepackage{makecell}
\usepackage{newfloat}
\usepackage{listings}
\usepackage{xcolor}

\DeclareCaptionStyle{ruled}{labelfont=normalfont,labelsep=colon,strut=off} %
\floatstyle{ruled}
\newfloat{listing}{tb}{lst}{}
\floatname{listing}{Listing}
\title{When Interpretability Is Unequally Distributed:\\ Fairness in Hybrid Interpretable Models}

\author {
    Ziba Jabbar Zare\textsuperscript{\rm 1},
    Ulrich Aïvodji\textsuperscript{\rm 2},
    Julien Ferry\textsuperscript{\rm 3},
    Thibaut Vidal\textsuperscript{\rm 3}
}
\affiliations {
    \textsuperscript{\rm 1}Department of Mechanical, Industrial and Aerospace Engineering, Concordia University, Canada\\
    \textsuperscript{\rm 2}École de technologie supérieure, Mila, Canada\\
    \textsuperscript{\rm 3}CIRRELT \& SCALE-AI Chair in Data-Driven Supply Chains, Department of Mathematics and Industrial Engineering, Polytechnique Montréal, Canada\\
    ziba.jabbarzare@mail.concordia.ca, ulrich.aivodji@etsmtl.ca, julien.ferry@polymtl.ca, thibaut.vidal@polymtl.ca
}

\lstdefinelanguage{RuleListsLanguage}{
  keywords={if, then, else, and},
  keywordstyle=\color{blue}\bfseries,
  ndkeywords={},
  ndkeywordstyle=\color{darkgray}\bfseries,
  identifierstyle=\color{black},
  sensitive=false,
  comment=[l]{//},
  morecomment=[s]{/*}{*/},
  commentstyle=\color{purple}\ttfamily,
  stringstyle=\color{red}\ttfamily,
  morestring=[b]',
  morestring=[b]"
}
\definecolor{RuleListsLanguageBackgroundColor}{rgb}{0.96, 0.96, 0.96}

\begin{document}

\maketitle

\begin{abstract}
Hybrid interpretable models combine a transparent component with a black-box model by assigning some examples to the former and deferring the rest to the latter. While this design enables flexible tradeoffs between accuracy and interpretability, it also raises a distinct procedural fairness concern: some demographic groups may systematically receive interpretable decisions, while others are disproportionately routed to a black box.
We formalize this issue as \emph{Interpretability Coverage Disparity} (ICD), a demographic-parity-style measure applied to the routing decision of hybrid interpretable models. Using tools from predictive multiplicity, we study ICD across four hybrid interpretable learning methods, three standard fairness benchmark datasets, and multiple sensitive attributes. Our experiments reveal substantial ICD in intermediate transparency regimes, where both the interpretable and black-box components are actively used. We further show that simple coverage-disparity constraints can significantly reduce ICD in exact hybrid learning methods, with marginal impact on accuracy and sparsity. In several settings, ICD mitigation also improves standard algorithmic fairness metrics. These results show that hybrid interpretable models should be audited not only for predictive fairness, but also for how they allocate interpretability across individuals and groups.
\end{abstract}

\begin{links}
    \link{Code}{github.com/vidalt/Fair-Hyb-Interpretability}
    %\link{Paper with appendices}{arxiv.org/abs/2605.28626}
\end{links}

\section{Introduction}

Explainability broadly refers to the accessibility and comprehensibility of information about how an AI system operates. It is increasingly emphasized in regulatory frameworks, particularly for high-risk applications such as hiring, healthcare, and criminal justice, where it is essential to support informed decision-making and enable meaningful human oversight. A common strategy to enhance explainability in machine learning is the use of post-hoc explanation techniques. These methods aim to approximate or explain the behavior of complex black-box models after training, often by using surrogate models that provide a simpler and more interpretable representation of the underlying decision process \cite{guidottiSurveyMethodsExplaining2018}. However, post-hoc explanations can be unreliable and susceptible to manipulation, raising concerns about their faithfulness~\cite{aivodjiFairwashingRiskRationalization2019}.

An alternative is to design inherently interpretable models, whose structure directly reveals how inputs are mapped to predictions. Examples include decision trees \cite{breimanClassificationRegressionTrees1984}, rule lists \cite{DBLP:journals/ml/Rivest87}, rule sets \cite{rijnbeekFindingShortAccurate2010}, and scoring systems \cite{ustunSupersparseLinearInteger2016}. While these models are generally easier for humans to understand, they may exhibit reduced predictive performance in complex settings.
This limitation has motivated research on \emph{hybrid interpretable models}~\citep{panInterpretableCompanionsBlackBox2020}, which combine an interpretable component with a black-box component. A gating mechanism assigns each example to either the interpretable component or the black box. By varying the proportion of examples assigned to the interpretable component, often referred to as the model's \emph{transparency}, hybrid interpretable models can realize different trade-offs between interpretability and predictive performance. However, this design also raises a largely overlooked question: even when hybrid models improve the aggregate accuracy-interpretability trade-off, they may distribute interpretability unevenly across individuals or demographic groups.

A large body of work on fairness in machine learning focuses on fairness criteria defined with respect to a predictive model’s outputs. Existing approaches typically intervene at one of three points in the machine learning pipeline: by modifying the training data through preprocessing, by incorporating fairness constraints during model training, or by adjusting model outputs through post-processing~\citep{kamiran2009classifying,zemel2013learning,hardtEqualityOpportunitySupervised2016, zhang2018mitigating}.
Although it has received considerably less attention, fairness has also been studied with respect to other dimensions of predictive models. In particular, fairness in explanation quality has been examined for several post-hoc explanation methods, including feature-attribution methods~\cite{daiFairnessExplanationQuality2022, balagopalanRoadExplainabilityPaved2022} and counterfactual explanations~\cite{bell2024fairness}. These works reveal demographic disparities in explanation quality, measured through properties such as sparsity, robustness, time to recourse, and effort to recourse. They show that fairness issues can arise not only from what predictions are made, but also from how decisions are explained or acted upon.

Our work follows this procedural perspective, but studies a distinct source of disparity arising in hybrid interpretable models. In these models, interpretability is not provided by a separate post-hoc explanation method; it is allocated by the model itself through its gating mechanism. This raises the question of whether different individuals or demographic groups have comparable access to the interpretable component.
Our main contributions are as follows:
\begin{itemize}
    \item We introduce \emph{Interpretability Coverage Disparity} (ICD), a demographic-parity-style disparity measure defined over interpretability assignment in hybrid interpretable models. We also leverage notions from predictive multiplicity to characterize individual-level \emph{Interpretability Coverage Arbitrariness} (ICA) in whether examples are assigned to the interpretable or black-box component.

    \item We conduct extensive experiments over sets of high-performing hybrid interpretable models learned by four state-of-the-art methods. Our results show that substantial ICA or ICD can arise on real-world datasets, leading to situations in which (i) some individuals are arbitrarily assigned to the black-box component, and (ii) some demographic groups are predominantly routed to the black-box component, while others are predominantly covered by the interpretable component.

    \item We extend existing hybrid interpretable model learning methods to mitigate ICD, and investigate the impact of the resulting constraints on model sparsity, predictive performance, and algorithmic fairness.
\end{itemize}

The remainder of the paper is organized as follows. We first review the relevant background and related work in Sections~\ref{sec:background} and~\ref{sec:related_works}, respectively. We then formalize our proposed group-level and individual-level metrics for quantifying unfairness in interpretability coverage in Section~\ref{sec:ICD}. In Section~\ref{sec:rashomon_sets_hybrid_models_experiments}, we empirically investigate the extent to which ICD and ICA arise when state-of-the-art hybrid interpretable models are applied to real-world datasets. Next, in Section~\ref{sec:ICD_mitigation}, we address ICD by adapting existing learning methods and assessing the empirical tradeoffs induced by mitigation. Finally, we conclude in Section~\ref{sec:conclusion}.

\section{Technical Background}
\label{sec:background}

\paragraph{Notation.}
Let $\mathcal{X}$ denote the input space and $\mathcal{Y}$ the output space. Given a dataset $S=\{(\mathbf{x}_i,y_i)\}_{i=1}^{n}
\subset \mathcal{X}\times\mathcal{Y}$, the objective of a supervised learning algorithm is to learn a function $h:\mathcal{X}\to\mathcal{Y}$ from a hypothesis space $\mathcal{H}$ by minimizing an empirical loss, such as the $0/1$ loss:
\[
    \ell_{0/1}(h;S)
    =
    \frac{1}{n}\sum_{i=1}^{n}
    \mathds{1}(h(\mathbf{x}_i)\neq y_i).
\]
Following prior work on hybrid interpretable models and algorithmic fairness, we focus on binary classification tasks, i.e., $\mathcal{Y}=\{0,1\}$. This choice allows us to rely on established methods and metrics from these research areas while leaving our main contribution unchanged: ICD concerns whether a decision is handled by the interpretable component, not the predicted outcome
itself.

\paragraph{Hybrid Interpretable Models.} A hybrid interpretable model is a predictive model defined by a triplet $\langle \Omega, h_s, h_c \rangle$ in which $\Omega \subset \mathcal{X}$ defines the \emph{interpretable region} of the input space, in which examples are handled by a simple component $h_s$ belonging to some interpretable hypothesis class $\mathcal{H}_s$, and examples outside $\Omega$ are deferred to a black-box model $h_c$ from a more complex hypothesis class $\mathcal{H}_c$, i.e.,:
\[
\forall \mathbf{x} \in \mathcal{X}, \quad \langle \Omega, h_s, h_c \rangle (\mathbf{x}) = 
\begin{cases}
   h_s(\mathbf{x}) &\quad\text{if } \mathbf{x}\in \Omega,\\
   h_c(\mathbf{x}) &\quad\text{otherwise.} \\
 \end{cases}
\]
The \emph{transparency} of such a model on a given dataset $S$ is then the proportion of examples covered by the interpretable part:
\[
C_{\Omega} = \frac{\lvert \Omega \cap S \rvert}{n},
\]
such that a pure black-box corresponds to $C_{\Omega}=0$ while a single interpretable model to $C_{\Omega}=1$.

In practice, existing hybrid interpretable models~\citep{panInterpretableCompanionsBlackBox2020,wangHybridPredictiveModels2021,HybridCorel_Julien} use a set of rules $r$ as the interpretable component, which simultaneously defines the interpretable region $\Omega$ as the support of its rules: any example not captured by at least one rule is deferred to the black-box. We use $S_r$ to denote the subset of $S$ captured by the rules in $r$, such that the transparency can be written as $\smash{C_{r}=\frac{\lvert S_r \rvert}{n}}$. Figure~\ref{fig:example_hybrid_interpretable_model} illustrates a hybrid interpretable model whose interpretable component $h_s$ is a rule list, and whose support defines the interpretable region $\Omega$.

\begin{figure}[t!]
    \centering
\begin{lstlisting}[language=RuleListsLanguage,backgroundcolor = \color{RuleListsLanguageBackgroundColor}, basicstyle=\scriptsize,numbers=none]
if ["Prior-Crimes=0" and "Age>=30"] then 0
else if ["Prior-Crimes>5" and "Age=24-30"] then 1
else if ["Prior-Crimes=1-3" and "Age>=30"] then 0
else
    RandomForest()
\end{lstlisting} 
    \vspace{2pt}
    \caption{Example hybrid interpretable model from~\citet{HybridCorel_Julien}, trained on the COMPAS dataset (for recidivism prediction).}%
    \label{fig:example_hybrid_interpretable_model}
\end{figure}

\paragraph{Algorithmic Fairness.} Undesirable disparities in predictive models may stem from biased data, historical inequalities, or biases introduced during model development ~\citep{mehrabiSurveyBiasFairness2022}. Several notions have been proposed to quantify and mitigate such disparities. They are commonly divided into \textit{individual fairness}, which requires similar individuals to receive similar predictions~\citep{dworkFairnessAwareness2011}, and \textit{group fairness}, which requires some form of parity across demographic groups. The latter has received particular attention because it can be  operationalized through aggregate statistics over protected groups, and because it echoes legal notions of disparate impact in anti-discrimination law
~\citep{uniformGuidelines1978,feldmanCertifyingRemovingDisparate2015,barocasBigDataDisparate2016}.

In our experiments, we will assess the effect of ICD mitigation on algorithmic fairness using two popular group fairness metrics. Let $\mathcal{P}$ denote a set of protected groups, where each $p \in \mathcal{P}$ defines a subset $S_p \subseteq S$ of examples sharing a given sensitive attribute value, such as gender, race, or age. 
The maximum statistical parity (SP)~\citep{dworkFairnessAwareness2011} disparity is
\[ 
d_{SP}(h,S) := \max_{(p,p') \in \mathcal{P}^2}\left(\frac{\sum_{i \in S_p} h(\mathbf{x}_i)}{|S_p|} - \frac{\sum_{i \in S_{p'}} h(\mathbf{x}_i)}{|S_{p'}|}\right). \label{eq:sp} 
\]
It measures the largest difference in positive prediction rates across protected groups. We also consider maximum equal opportunity (EO)~\citep{hardtEqualityOpportunitySupervised2016} disparity, defined as
\[ 
d_{EO}(h,S) := \max_{(p,p') \in \mathcal{P}^2} \left(\frac{\sum_{i \in S_p^+} h(\mathbf{x}_i)}{|S_p^+|} - \frac{\sum_{i \in S_{p'}^+} h(\mathbf{x}_i)}{|S_{p'}^+|}\right), \label{eq:eo} 
\]
where $S_{p}^+ = \{i \in S_p : y_i=1\}$ denotes the set of positive examples in group $p$. This metric measures the largest difference in true positive rates across protected groups.

\paragraph{Predictive Multiplicity.}
In many settings, several models can achieve nearly indistinguishable predictive performance while disagreeing on the predictions assigned to individual instances. This coexistence of competing, similarly accurate models is known as \emph{predictive multiplicity}. The set of such near-optimal models is referred to as the \emph{Rashomon set}~\citep{breiman2001statistical}:
\[
\mathcal{R}(\mathcal{H},\varepsilon,S) := \{h \in \mathcal{H}~:~\ell_{0/1}(h;S) \leq (1 + \varepsilon) \cdot \ell_{0/1}(h^\star;S)\},
\]
where $h^\star$ is a reference model minimizing $\ell_{0/1}$ over $S$ and $\varepsilon$ controls the tolerated loss increase relative to $h^\star$. Although models in the Rashomon set have comparable predictive performance, they can differ substantially with respect to other desiderata. Assessing these desiderata across $\mathcal{R}(\mathcal{H},\varepsilon,S)$ therefore provides a more comprehensive evaluation than focusing on a single model, which may be only one arbitrary representative of this set.

Some \emph{enumeration-based} methods have been proposed to exactly construct $\mathcal{R}(\mathcal{H},\varepsilon,S)$ for specific and relatively simple hypothesis classes, including rule lists~\citep{DBLP:journals/corr/abs-2204-11285},
rule sets~\citep{ciaperoni2024efficient}, and decision trees~\citep{DBLP:conf/nips/XinZ0TSR22,NEURIPS2025_3f3f2d55}. However, existing methods for rule-based models tend to scale poorly with the Rashomon parameter $\varepsilon$, the number of candidate rules, and the maximum number of rules. Some \emph{enumeration-free} methods have therefore been proposed to characterize properties of interest within the Rashomon set of a given hypothesis class, without explicitly constructing the full set. These methods, however, are again limited to particular properties and hypothesis classes, such as linear models~\citep{coker2021theory,DBLP:conf/aaai/Watson-DanielsP23,fisher2019all,coston2021characterizing},
additive models~\citep{zhong2024exploring}, or models whose learning problem can be cast as a mathematical programming formulation~\citep{FairnessSparsityRashomon2025}.

For more complex models, including black-box models, simple model-sampling tools have been shown to provide useful approximations of the Rashomon set, notably through bootstrapping~\citep{DBLP:conf/aaai/CooperLCBSGK0Z24}. Although these methods typically recover only a subset of the full Rashomon set, they still allow meaningful statistics to be computed over collections of \emph{good} models.

\section{Related Works} 
\label{sec:related_works}

While fairness and explainability have traditionally been studied as separate dimensions of trustworthy AI, recent work has begun to examine their intersection. In particular, an emerging line of research investigates whether explanations themselves are fair, namely, whether their quality, reliability, and informativeness are consistent across individuals and demographic groups. This notion, referred to as \textit{fairness in explanation quality}, was formalized by \citet{daiFairnessExplanationQuality2022}. Subsequent work has shown that explanation methods can exhibit disparities across subgroups, even when the underlying predictive model is fair. For instance, \citet{balagopalanRoadExplainabilityPaved2022} analyze subgroup differences in post-hoc explanations, while \citet{dhainiGenderBiasExplainability2025} demonstrate the presence of gender-based disparities in feature attribution methods for language models. Furthermore, \citet{zhaoFairnessExplainabilityBridging2022} propose optimization-based approaches to mitigate disparities in explanation quality. 

A closely related line of work examines fairness through the lens of \textit{algorithmic recourse}, also known as \textit{recourse disparity}. These works examine whether individuals from different demographic subgroups have equally feasible and actionable paths to overturn an unfavorable prediction. In the context of counterfactual explanations, recourse cost is typically defined as the distance between an input instance and a corresponding counterfactual (or the decision boundary), and recourse fairness is characterized by disparities in average recourse cost across groups \cite{guptaEqualizingRecourseGroups2019,slackCounterfactualExplanationsCan2021,kavourasFairnessAwareCounterfactuals2023}. Empirical studies have shown that such disparities can arise even when the predictive performance of the model at hand is comparable across groups, which motivated recent works on mitigating them~\citep{DBLP:journals/corr/abs-2211-14858,DBLP:journals/corr/abs-2601-20449}.

Together, these developments highlight an important distinction between \textit{outcome-based fairness}, which focuses on the fairness of model predictions, and \textit{procedure-oriented fairness}, which concerns the fairness of the mechanisms through which decisions are produced, explained, or acted upon. Fairness in explanation quality and recourse fairness both fall into the latter category, as they evaluate whether the reasoning provided by an explanation method is equally reliable and informative across groups. 

In this work, we introduce complementary procedural fairness notions tailored to hybrid interpretable models: \textit{Interpretability Coverage Disparity} (ICD) and \emph{Interpretability Coverage Arbitrariness} (ICA). Unlike prior work on explanation quality or recourse, ICD and ICA do not assess the quality of an explanation produced after a prediction. Instead, they measure how interpretability itself is allocated by the model, through the routing decision that determines whether an example is handled by the interpretable component or deferred to the black-box component. This distinction is important because, in hybrid interpretable models, access to interpretability is not provided by an external explanation method; it is an intrinsic property of the predictive model. ICD therefore shifts attention from whether explanations are equally good to whether different individuals and groups are equally likely to receive interpretable decisions in the first place.

\section{Interpretability Coverage Unfairness}
\label{sec:ICD}

We now formalize metrics to quantify unfairness in interpretability coverage for hybrid interpretable models. We first consider disparities at the group level, before turning to arbitrariness at the individual level.

\subsection{Group-level Disparity}

We first introduce the \emph{Interpretability Coverage} of a protected group $p \in \mathcal{P}$ as the fraction of individuals from group $p$ that fall within the interpretable region $\Omega$:
\begin{align}
    \mathrm{IC}_p := \frac{\lvert \Omega \cap S_p \rvert}{\lvert S_p \rvert}.
\end{align}
Higher values of $\mathrm{IC}_p$ indicate that individuals from group $p$ are more frequently handled by the interpretable component of the hybrid interpretable model. Note that the overall model \emph{transparency}, defined in Section~\ref{sec:background}, corresponds to interpretability coverage measured over the entire dataset.

We now define the \emph{Interpretability Coverage Disparity} (ICD) as the maximum pairwise difference in interpretability coverage across protected groups:
\begin{align}
    \mathrm{ICD}
    := \max_{(p,p') \in \mathcal{P}^2}
    \left(
     \mathrm{IC}_p
    -
     \mathrm{IC}_{p'}
    \right).
\end{align}
Equivalently, ICD is the difference between the largest and smallest group-specific interpretability coverage:
\begin{align}
    \mathrm{ICD}
    =
    \max_{p \in \mathcal{P}} \mathrm{IC}_p
    -
    \min_{p \in \mathcal{P}} \mathrm{IC}_p.
\end{align}
ICD captures the largest gap in access to interpretability between any two protected groups. A value of zero indicates that all groups are covered by the interpretable component at the same rate, whereas larger values indicate that some groups are disproportionately routed to the black-box component. In our experiments, we also report the per-group interpretability coverage when relevant, since ICD alone does not identify which groups are relatively advantaged or disadvantaged, and only reflects the largest observed disparity.

Importantly, ICD depends only on the allocation of individuals to the interpretable and black-box regions, and not on the predictions made by either component of the hybrid model. It is therefore a procedural fairness notion: rather than measuring disparities in predictive outcomes, it quantifies the extent to which individuals from different protected groups have unequal access to interpretable decisions.

For rule-based hybrid interpretable models, this notation can be made more explicit. Let $S_{r,p}$ denote the set of training examples from protected group $p \in \mathcal{P}$ that are classified by the rules in $r$. We thus have $\mathrm{IC}_p = \frac{\lvert S_{r,p} \rvert}{\lvert S_p \rvert}$ and:
\begin{align}
    \mathrm{ICD}
    := \max_{(p,p') \in \mathcal{P}^2}
    \left(
    \frac{\lvert S_{r,p} \rvert}{\lvert S_p \rvert}
    -
    \frac{\lvert S_{r,p'} \rvert}{\lvert S_{p'} \rvert}
    \right).\label{eq:icd_computation_rulebased}
\end{align}

\cameraready{Although ICD is primarily intended for a given set of protected groups, extending it to settings such as intersectional fairness could be insightful. The main challenge, namely efficiently navigating exponentially many subgroup intersections, is not ICD-specific, so existing fairness tools may be adaptable. We leave this extension to future work.}

\subsection{Individual-level Arbitrariness}

Inspired by previous work on predictive multiplicity, we further characterize the \emph{arbitrariness} of interpretability coverage at the level of an individual example $\mathbf{x}$. Let $\mathcal{H}$ denote the hypothesis class of hybrid interpretable models, and let $\mathcal{R}(\mathcal{H},\varepsilon,S)$ be its $\varepsilon$-Rashomon set on a dataset $S$. We first define the interpretability coverage frequency of $\mathbf{x}$ as
\begin{align}
    \mathrm{ICF}(\mathbf{x}) :=
    \frac{
    \left\lvert
    \left\{
    \langle \Omega, h_s, h_c \rangle \in
    \mathcal{R}(\mathcal{H},\varepsilon,S)
    ~:~ \mathbf{x} \in \Omega
    \right\}
    \right\rvert
    }{
    \left\lvert \mathcal{R}(\mathcal{H},\varepsilon,S) \right\rvert
    }.
    \label{def:ICF}
\end{align}
This quantity measures how often $\mathbf{x}$ is handled by the interpretable component across near-optimal hybrid interpretable models.
We then define the \emph{Interpretability Coverage Arbitrariness} (ICA) of $\mathbf{x}$ as
\begin{align}
    \mathrm{ICA}(\mathbf{x}) :=
    1 - 2 \left\lvert \mathrm{ICF}(\mathbf{x}) - \frac{1}{2} \right\rvert .
    \label{def:ICA}
\end{align}
Values close to $0$ indicate low arbitrariness, since $\mathbf{x}$ is consistently assigned either to the black-box component or to the interpretable component across the Rashomon set. In contrast, values close to $1$ indicate high arbitrariness, since $\mathbf{x}$ is assigned to the interpretable region for roughly half of the models and to the black-box region for the other half.

In practice, building the exact Rashomon set is not tractable for black-box models, and even more so for hybrid interpretable models, whose interpretable component training often depends on a black-box trained beforehand. Accordingly, we replace $\mathcal{R}(\mathcal{H},\varepsilon,S)$ in Definitions~\eqref{def:ICF} and~\eqref{def:ICA} with an approximation $\smash{\widehat{\mathcal{R}}(\mathcal{H},\varepsilon,S)}$. This approximation, described in Section~\ref{sec:rashomon_sets_hybrid_models_experiments}, does not aim to enumerate all near-optimal models, but rather to provide a large and diverse collection of high-performing models for estimating interpretability-coverage statistics.

\section{Experimental Evaluation of ICD and ICA} \label{sec:rashomon_sets_hybrid_models_experiments}

We now assess to what extent ICD and ICA arise when using state-of-the-art hybrid interpretable models trained over real-world datasets. We first describe the considered setup before presenting the main results.

\subsection{Experimental Setup} \label{sec:expes_setup}

\paragraph{Datasets.}
In our experiments, we consider three datasets of varying sizes and prediction tasks. All datasets are preprocessed by discretizing numerical features using quantile-based binning and applying one-hot encoding to categorical variables. To ensure consistency across hybrid model implementations and avoid dependence on the rule generation procedure, we pre-mine a set of candidate rules using the FP-Growth algorithm \cite{hanMiningFrequentPatterns2000}. The resulting rules have a maximum cardinality of 2 and a minimum support threshold of 1\%. Following the setup of~\citet{HybridCorel_Julien}, we retain a maximum of $300$ rules, keeping the ones with the largest support.

All group-level disparities, including ICD and standard algorithmic fairness metrics, are evaluated with respect to three sensitive attributes: Age, Gender, and Race. For each attribute, we consider up to three subgroups corresponding to the most frequent categories in the data, grouping less frequent categories into an ``Other'' category when necessary: %

\begin{itemize}
\item \textbf{COMPAS} \cite{mattuMachineBias2016}: \textit{(7,214 samples, 230 mined rules)} contains records of criminal defendants from Florida (2013–2014). The prediction task is to determine whether a defendant will recidivate within two years. Demographic subgroups are defined as follows: 
Age: \{18--25, 26--29, $\geq$30\}, 
Gender: \{Male, Female\}, 
Race: \{African-American, Caucasian, Other\}.

\item \textbf{UCI Adult Income} \cite{UCIMachineLearning}: \textit{(48,842 samples, 300 mined rules)} consists of demographic and occupational attributes from the 1994 U.S. Census. The task is to predict whether an individual earns more than \$50K per year. Demographic subgroups are defined as: 
Age: \{Low, Middle, High\} (quantile-based bins), 
Gender: \{Male, Female\}, 
Race: \{White, Black, Other\}.

\item \textbf{ACS Employment} \cite{dingRetiringAdultNew2021}: \textit{(208,358 samples, 300 mined rules)} is derived from the American Community Survey (2014–2018) and aims to predict whether an individual is employed based on socioeconomic features. Demographic subgroups are defined as: 
Age: \{Low, Middle, High\} (quantile-based bins), 
Gender: \{Male, Female\}, 
Race: \{White alone, Black or African American alone, Other\}.
\end{itemize}

\paragraph{Learning algorithms.}
We consider four state-of-the-art hybrid interpretable learning methods: Hybrid Rule Set (HyRS), Companion Rule List (CRL), HybridCORELSPre, and HybridCORELSPost. HyRS~\cite{wangHybridPredictiveModels2021} and CRL~\cite{panInterpretableCompanionsBlackBox2020} are heuristic methods based on simulated annealing. Both take as input a pre-trained black-box model and learn a rule-based interpretable component that covers part of the input space, while deferring the remaining instances to the black-box. They optimize trade-offs between accuracy, transparency, and sparsity, but do not provide optimality guarantees and cannot enforce a prescribed transparency level.

By contrast, HybridCORELSPre and HybridCORELSPost~\citep{HybridCorel_Julien} rely on an exact branch-and-bound procedure to explore ordered rule prefixes. Given a pre-trained black-box, HybridCORELSPost optimizes a weighted sum of hybrid-model accuracy and sparsity subject to a hard transparency constraint. HybridCORELSPre follows the same principle, but learns the interpretable prefix before training the black-box component; it therefore optimizes an upper bound on the final hybrid-model accuracy when fitting the prefix. %

\paragraph{Hyperparameters.} Across all datasets, our experimental setup is designed to explore a broad range of transparency levels. We therefore vary only the hyperparameters that directly control, or strongly influence, interpretability coverage, while keeping the remaining hyperparameters at their default values. For HybridCORELSPre and HybridCORELSPost, transparency is explicitly controlled via a minimum-transparency constraint. We consider 10 values, $min\_transp \in \{0.1, 0.2, \dots, 0.9, 0.95\}$.
For HyRS, we vary the hyperparameter $\beta$, which governs the trade-off between accuracy and transparency. We consider 10 values logarithmically spaced between $10^{-3}$ and $10^{0}$, following \citet{wangHybridPredictiveModels2021}. For CRL, we vary the regularization parameter $\lambda$, which controls model sparsity and consequently affects transparency. We consider 10 logarithmically spaced values in the range $[10^{-3}, 10^{-1}]$.

The black-box component of each hybrid interpretable model is instantiated as a random forest classifier from the scikit-learn library~\cite{scikit-learn}, with \texttt{min\_samples\_split = 10} and \texttt{max\_depth = 10}.

\paragraph{Bootstrapping setup.}
To better quantify the extent to which ICD or ICA can arise, we evaluate them over sets of \emph{good models}. Since exactly enumerating the Rashomon set of hybrid models is computationally infeasible, we approximate it using a bootstrap-based procedure. \cameraready{Since this procedure approximates the true Rashomon set, the resulting model distribution provides an estimate of the true distribution, and our results should be interpreted accordingly.}

We first split each dataset into training ($80\%$) and test ($20\%$) sets. For each method and hyperparameter configuration, we first train a reference model on the original training set. We then generate $1000$ bootstrap samples from the training set and retrain the same configuration on each resampled dataset. This yields $1001$ trained models per configuration, including the reference model. Since we consider 10 configurations per method, this results in $10{,}010$ trained models for each method. For each bootstrap run, the black-box component is trained with a different random seed.

The models obtained through this procedure are not necessarily unique. We therefore post-process the set of hybrid interpretable models obtained for each learning algorithm to remove duplicates. In practice, we consider two hybrid interpretable models from the same learning algorithm to be duplicates when their interpretable components are identical, that is, composed of the same rules in the same order, and when their black-box components produce similar predictions on the training set. Finally, the resulting unique models are grouped into four bins according to their training-set transparency values and filtered to keep only those whose training accuracy satisfies the Rashomon parameter $\varepsilon$ (relative to the reference model within each bin). All downstream analyses are performed on the held-out test set. %

Using the described bootstrap sampling procedure, we obtain large and diverse approximate Rashomon sets across transparency bins for all evaluated hybrid interpretable methods. Additional analyses provided in Figures~\ref{Fig:RS_growth_HybridCORELSPost}--\ref{Fig:RS_growth_CRL} in Appendix~\ref{sec:appendix_boots} further illustrate how the size of the approximate Rashomon sets evolves as the Rashomon parameter $\varepsilon$ increases, highlighting that this approach is able to recover a substantial set of near-optimal models, enabling meaningful analyses of predictive multiplicity and fairness properties over Rashomon sets.

\paragraph{Computational resources.}
To keep experiments computationally manageable, we impose a 5-minute limit on the training of each interpretable component, run on a single thread, with a maximum memory usage of 8 GB. The training of the black-box component is not limited, but typically takes only a few seconds. All experiments are run on a homogeneous set of cluster nodes equipped with Intel 6972P CPUs at 2.4 GHz.

\begin{figure*}[t]
\centering
\includegraphics[width=0.5\textwidth]{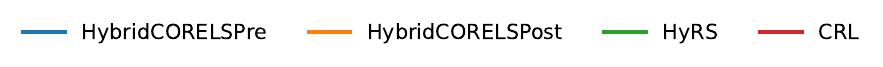}
\begin{subfigure}{0.85\textwidth}
    \centering
    \includegraphics[width=0.32\linewidth]{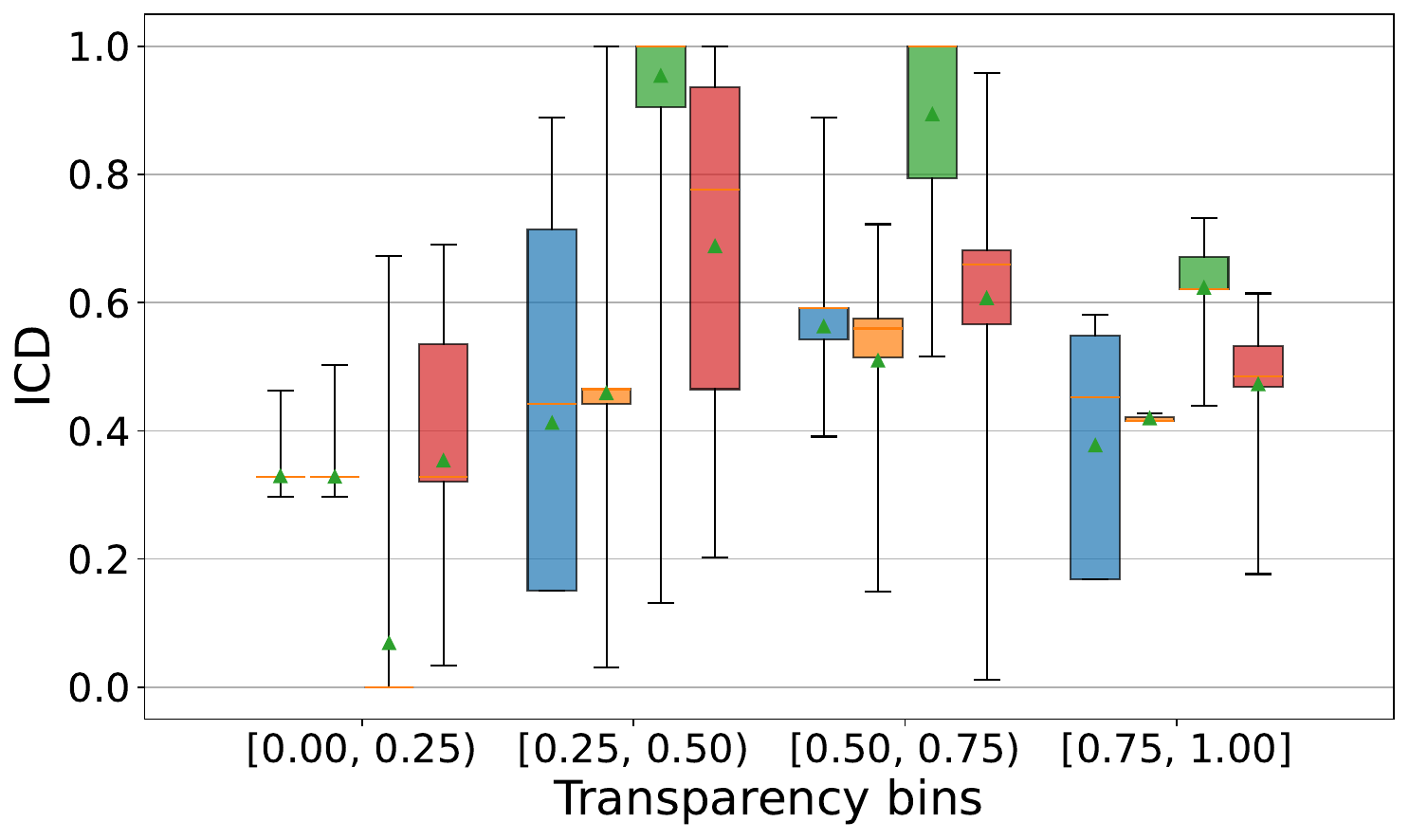}
    \hfill
    \includegraphics[width=0.32\linewidth]{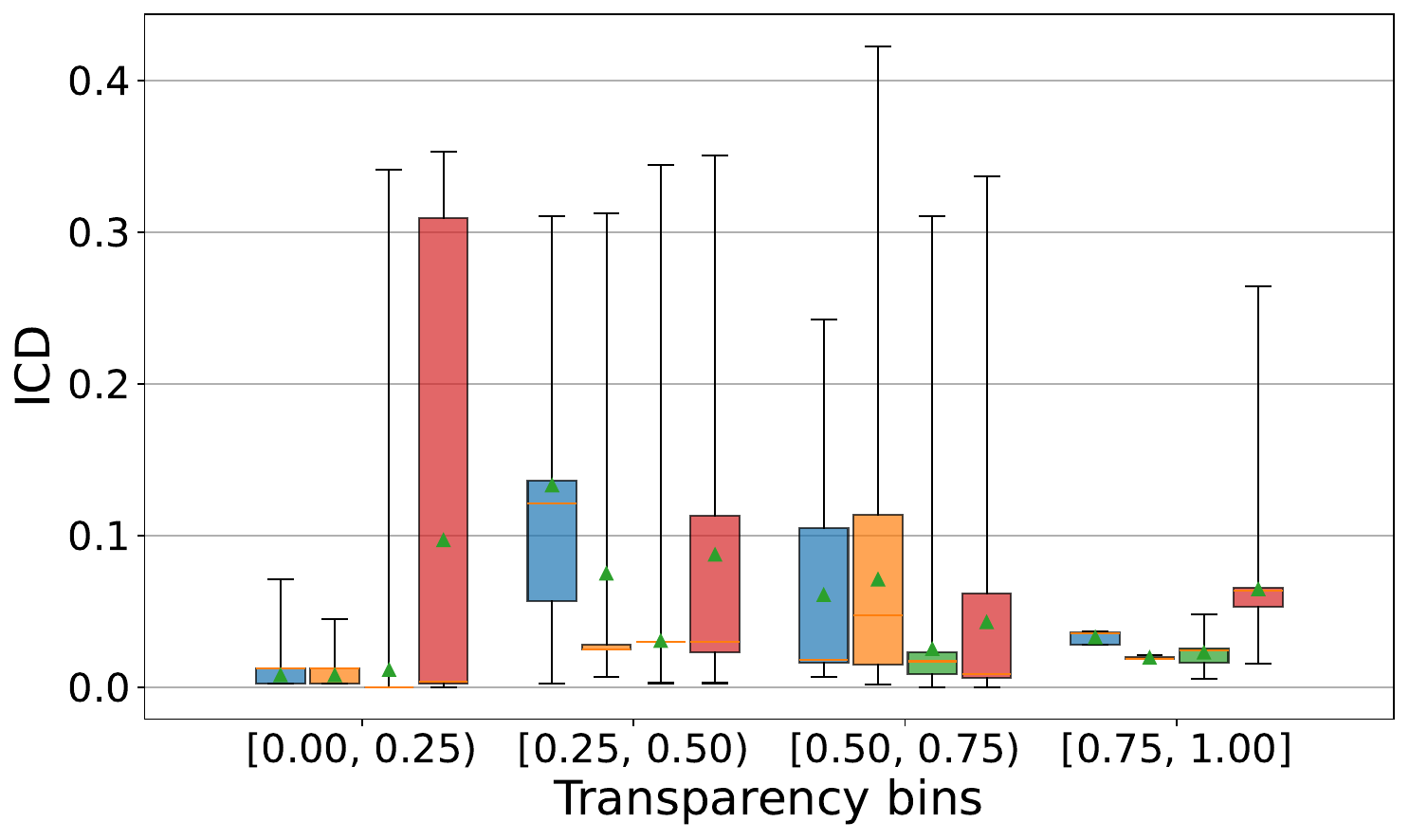}
    \hfill
    \includegraphics[width=0.32\linewidth]{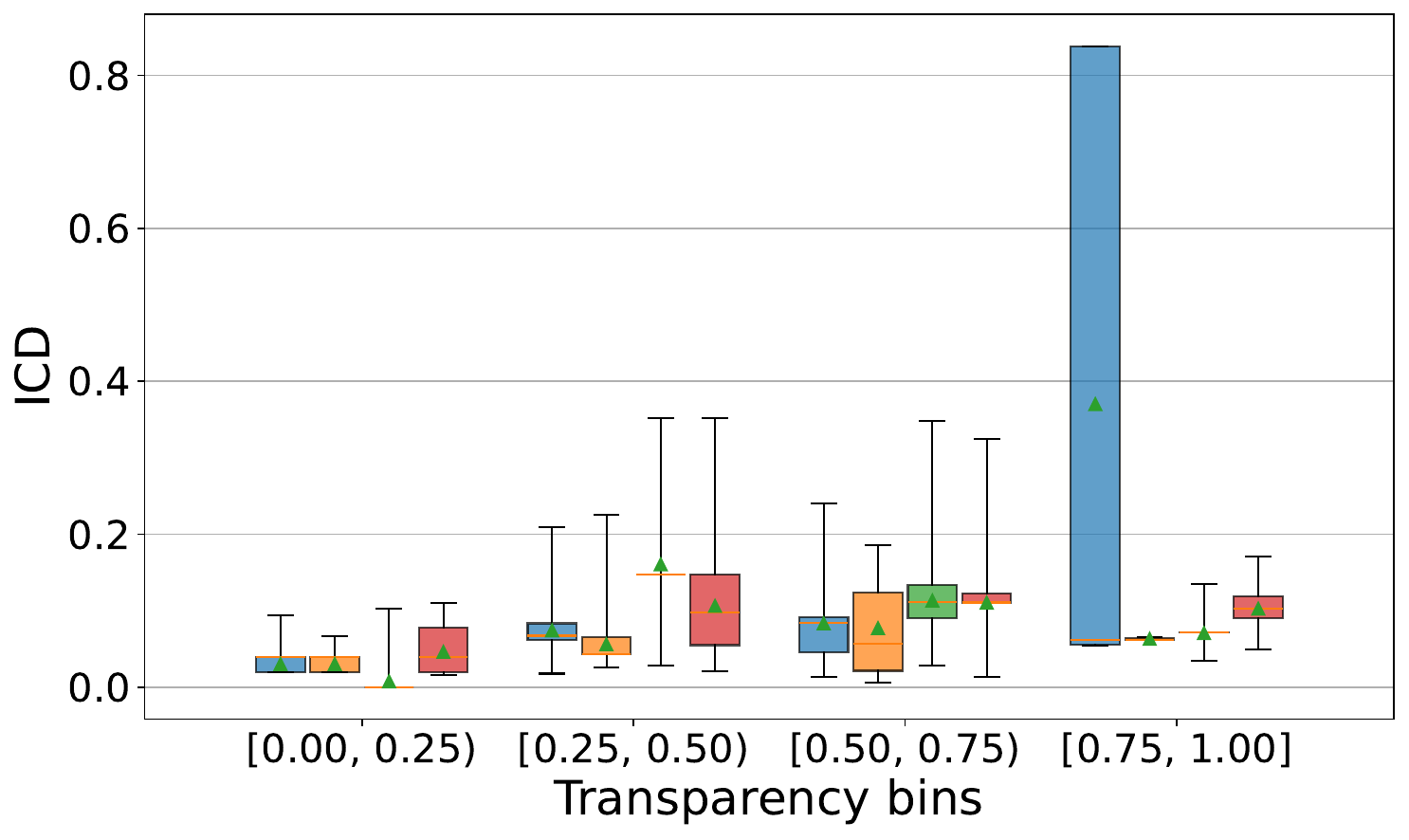}
    \caption{ACS Employment dataset}\label{fig:box_ICF_max_delta_1_acs_employment}
\end{subfigure}

\vspace{0.5em}

\begin{subfigure}{0.85\textwidth}
    \centering
    \includegraphics[width=0.32\linewidth]{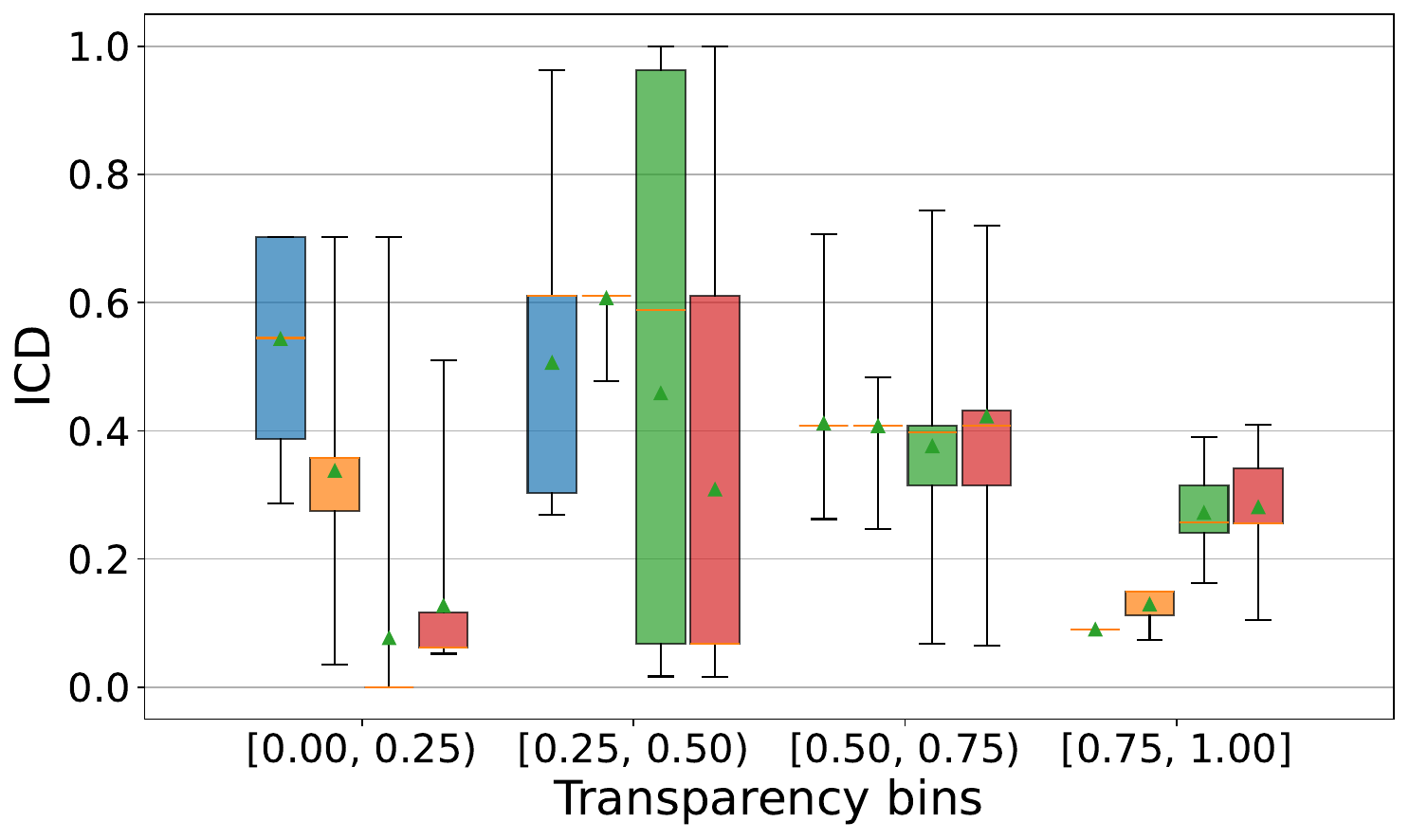}
    \includegraphics[width=0.32\linewidth]{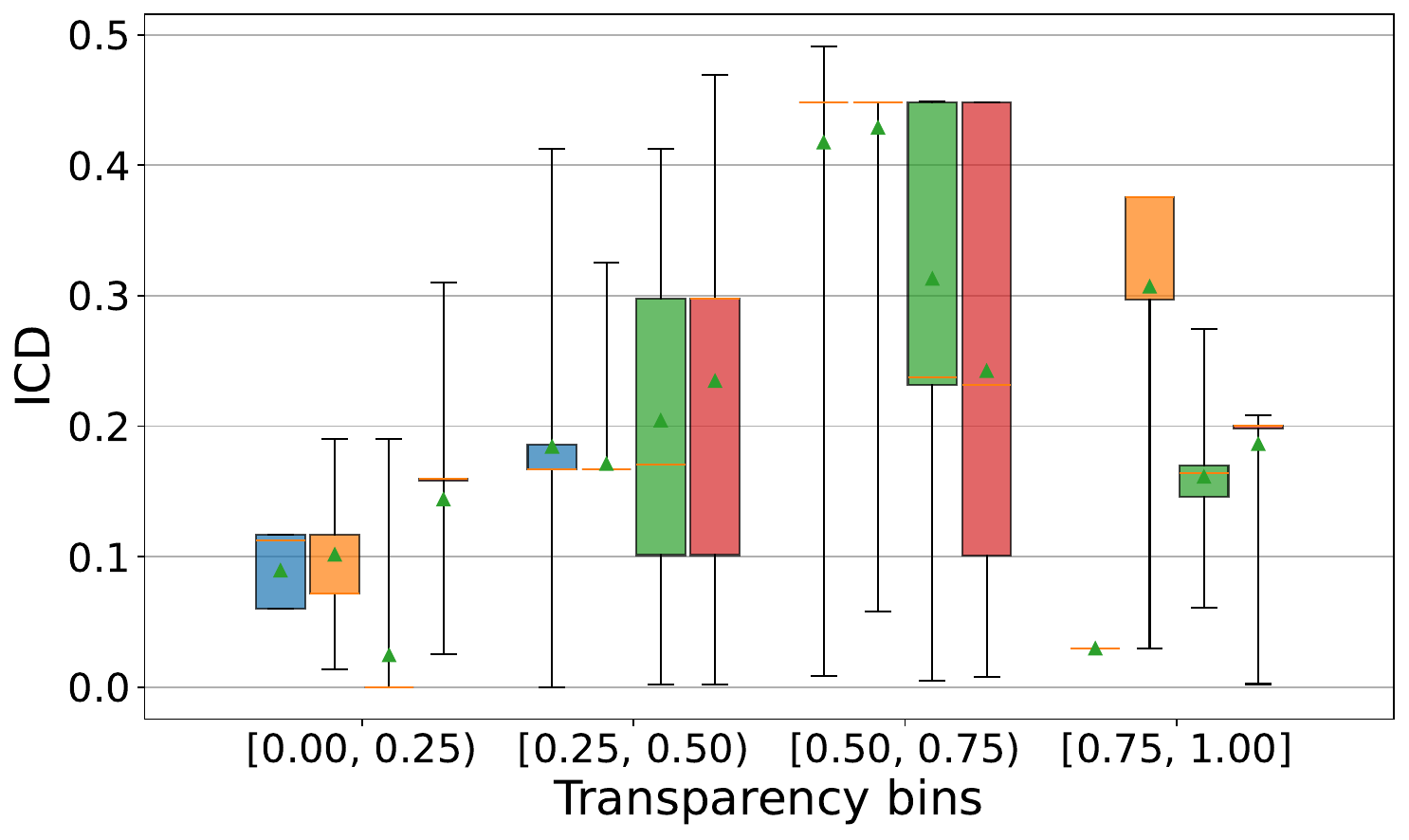}
    \includegraphics[width=0.32\linewidth]{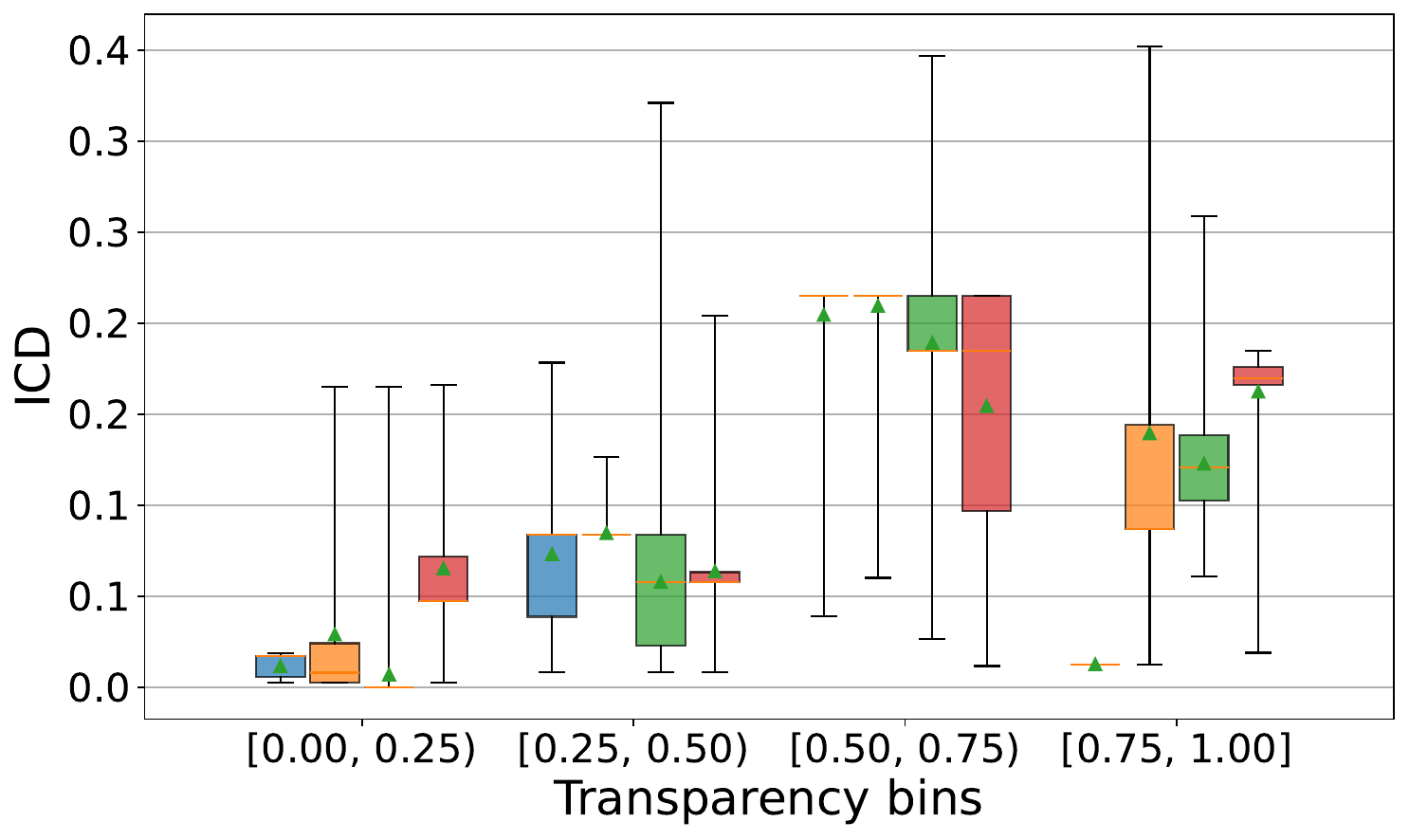}
    \caption{UCI Adult Income dataset}\label{fig:box_ICF_max_delta_1_adult}
\end{subfigure}

\vspace{0.5em}

\begin{subfigure}{0.85\textwidth}
    \centering
    \includegraphics[width=0.32\linewidth]{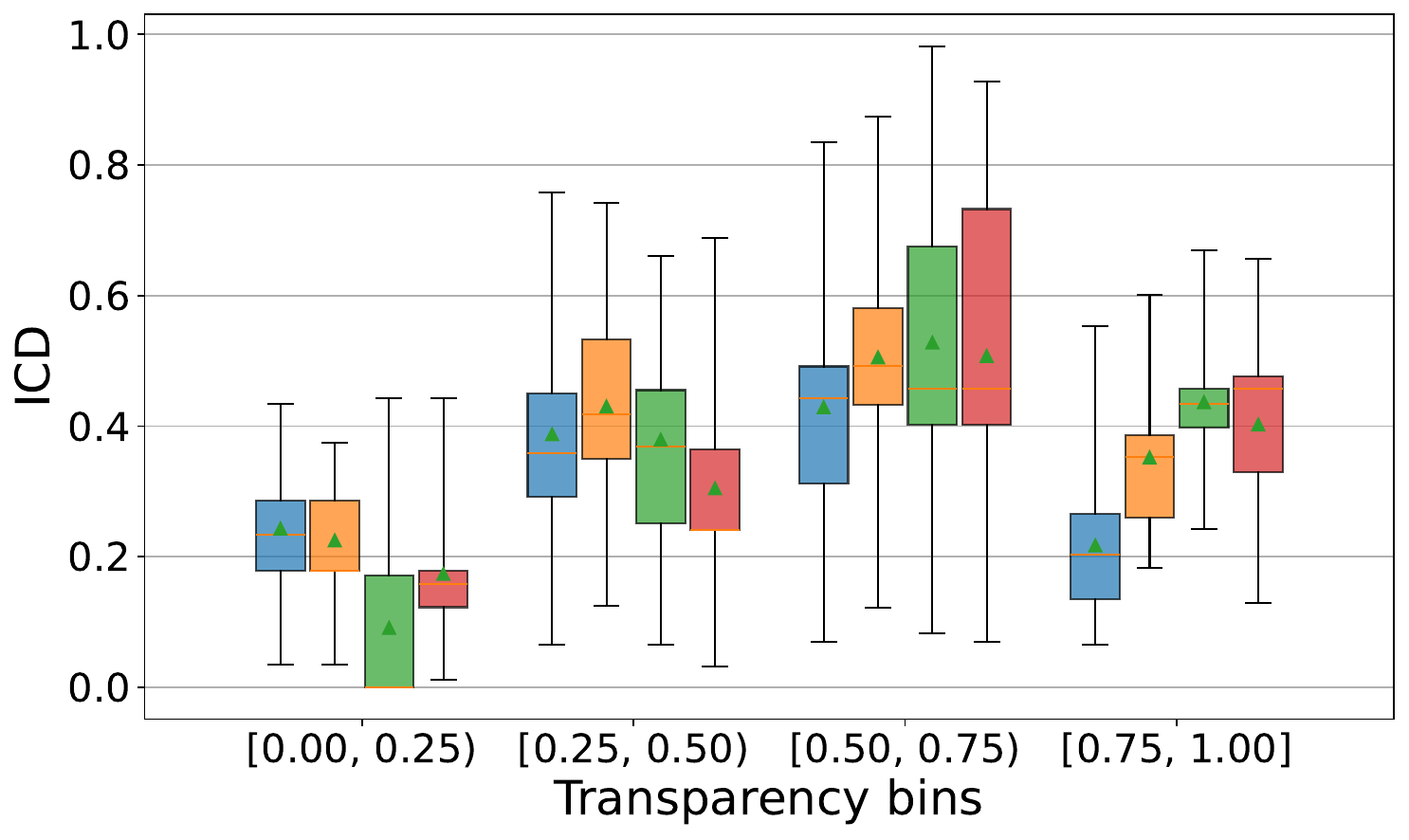}
    \includegraphics[width=0.32\linewidth]{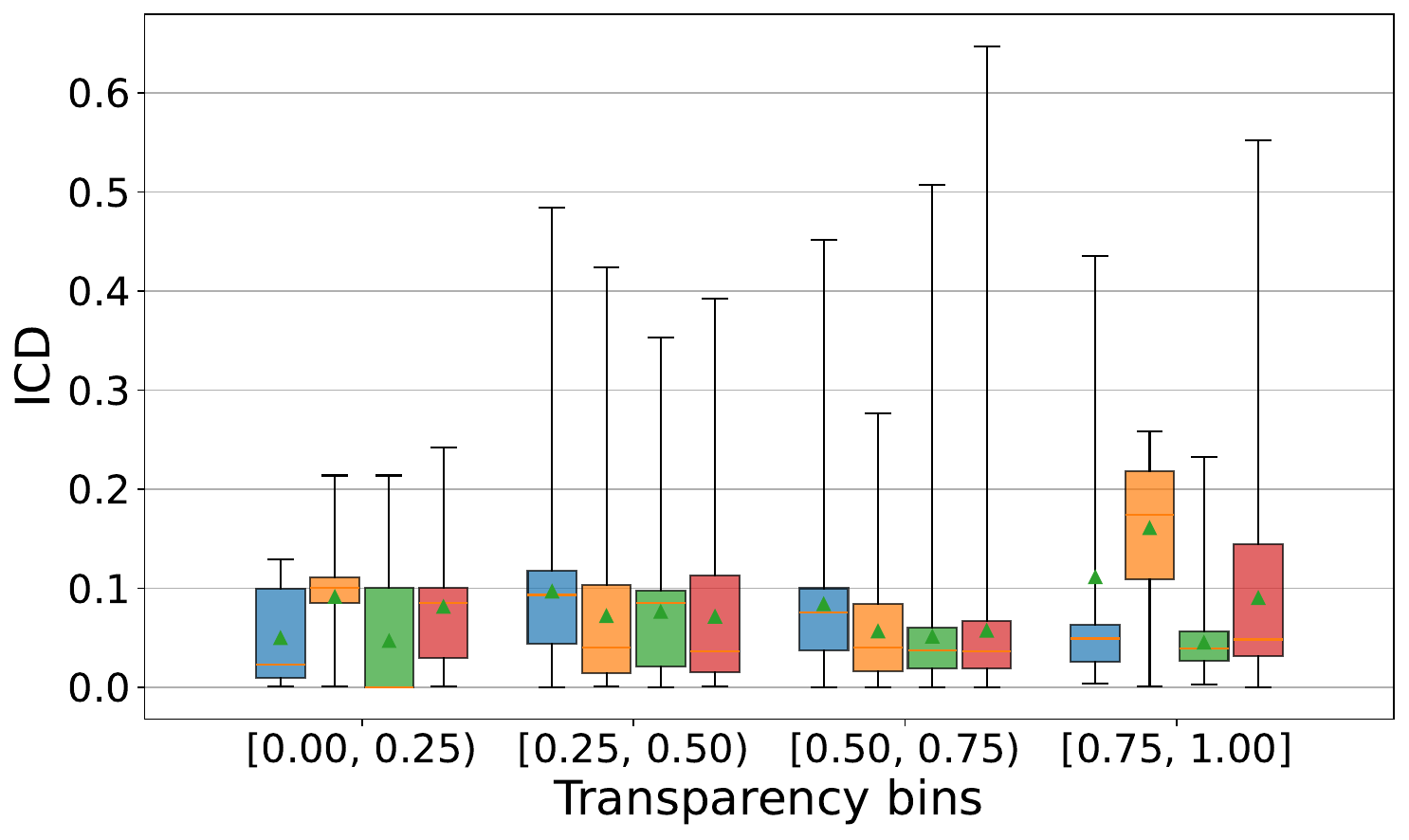}
    \includegraphics[width=0.32\linewidth]{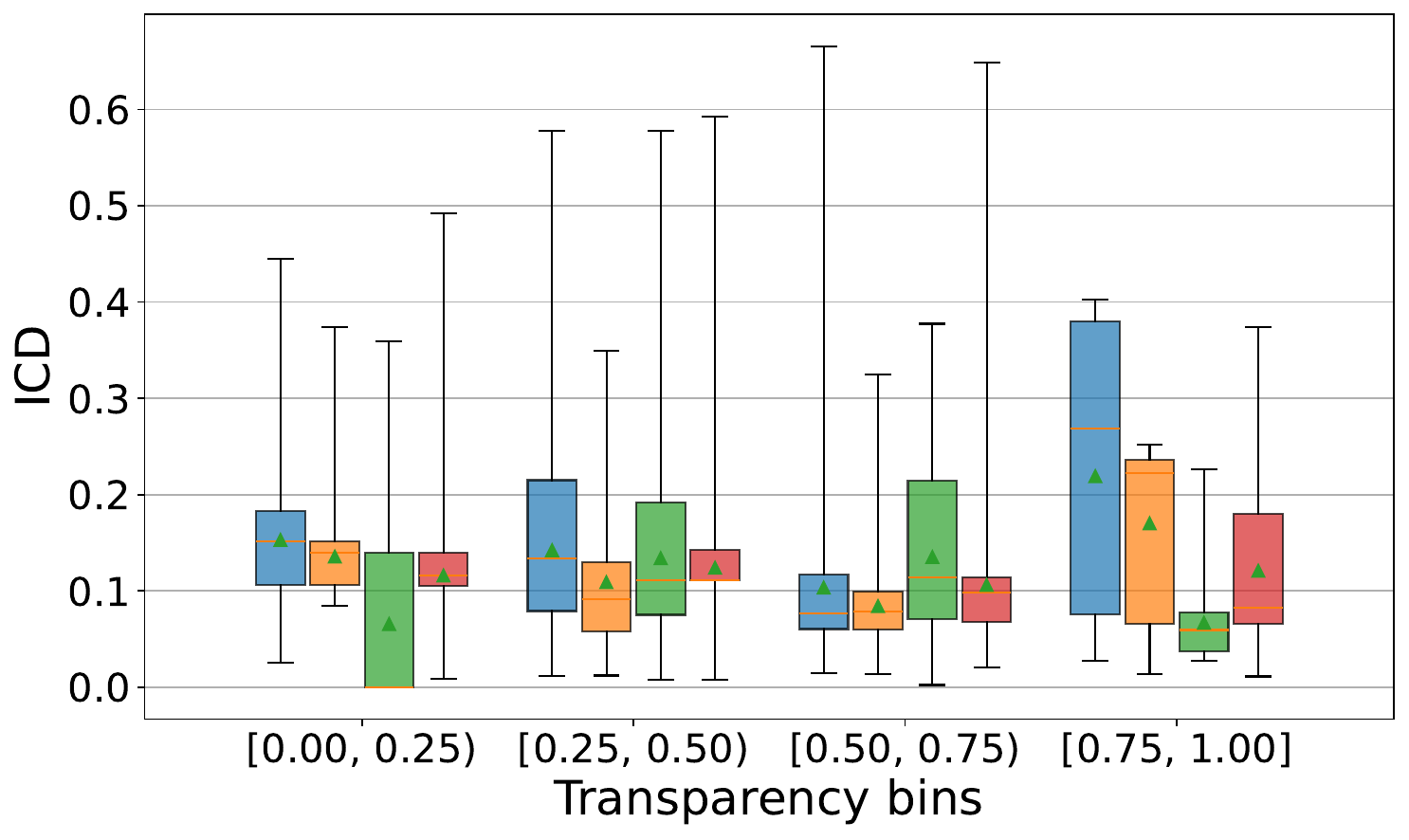}
    \caption{COMPAS dataset}\label{fig:box_ICF_max_delta_1_compas}
\end{subfigure}
\caption{Distribution of test set ICD across Rashomon sets for all transparency bins $(\varepsilon = 0.01)$. Results are shown for all datasets and sensitive attributes (Age, Gender, Race), ordered from left to right columns.} \label{fig:box_ICF_max_delta_1}
\end{figure*}

\subsection{Results} \label{sec:ICD_results}

We now present the main empirical findings through a subset of representative results. Unless otherwise specified, all analyses are performed on approximate Rashomon sets with $\varepsilon = 0.01$. %
Comprehensive results for all hybrid interpretable methods, datasets, sensitive attributes, and experimental settings are deferred to the Appendix~\ref{sec:appendix_ICD_results}.

\paragraph{Result 1. Hybrid interpretable models exhibit significant disparities in interpretability coverage across demographic groups.} 

Figure~\ref{fig:box_ICF_max_delta_1} provides a comprehensive overview of the distribution of ICD across transparency bins using box plots. As defined in Section~\ref{sec:ICD},  ICD reflects the largest observed subgroup difference in interpretability access for a given model. Each box plot summarizes the distribution of ICD values across Rashomon sets, where the box indicates the interquartile range, the horizontal line represents the median, the triangle denotes the mean, and the whiskers correspond to the minimum and maximum values. 

\begin{figure}[t!]
\centering
    \includegraphics[width=0.85\linewidth]{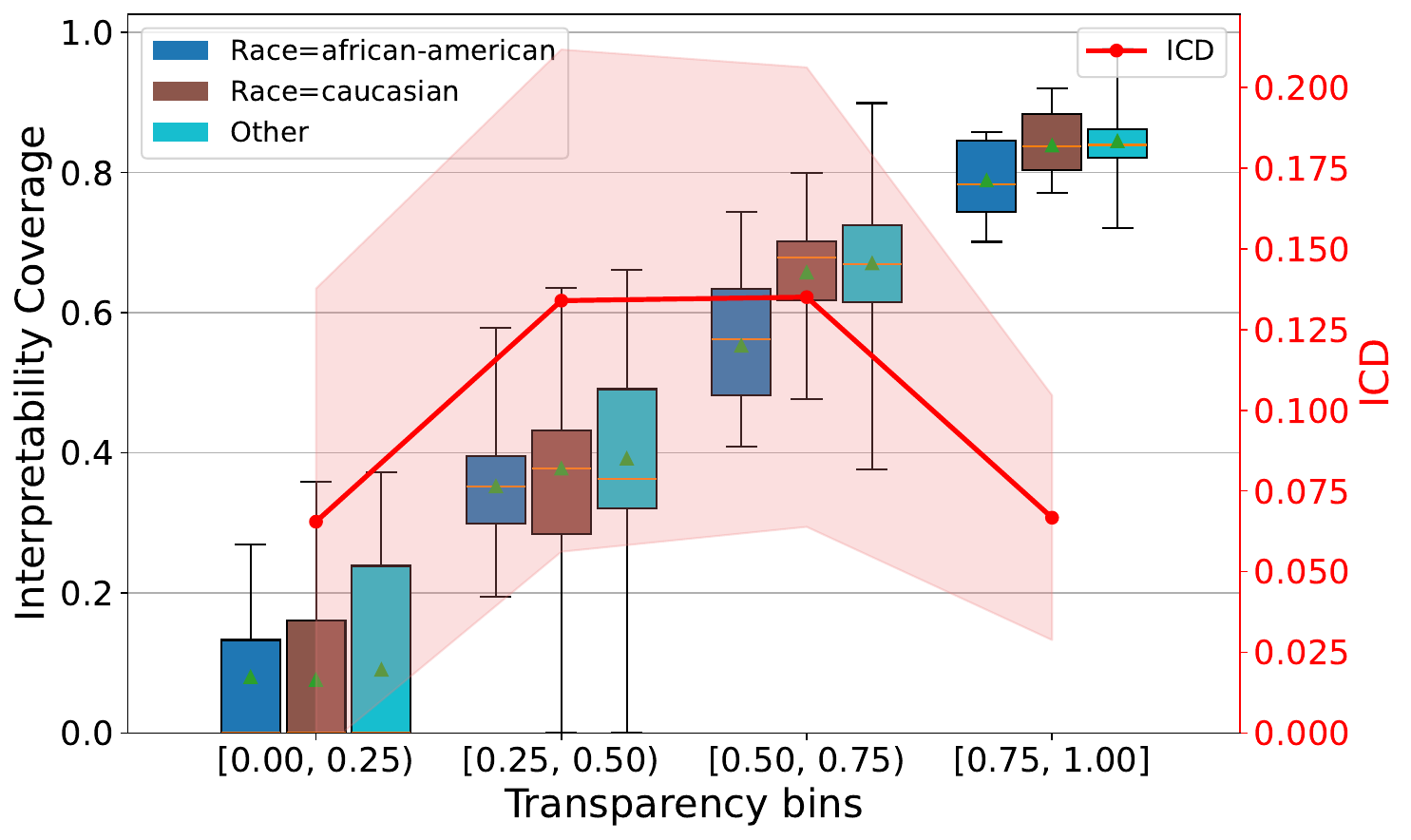}

\caption{Distribution of test set Interpretability Coverage (IC) for demographic subgroups of the Race attribute on the COMPAS dataset, for the different transparency bins, across Rashomon sets of HyRS $(\varepsilon = 0.01)$. 
The red curve represents the average ICD across models within each bin, and the shaded region indicates the standard deviation.}
\label{fig:ICF_HyRS_compas_Race}
\end{figure}
The results reveal substantial interpretability coverage disparities across nearly all experimental settings, confirming that hybrid interpretable models can distribute interpretability unevenly across demographic groups. In several cases, ICD values approach $1.0$, indicating extreme disparities where certain demographic groups are systematically deferred to the black-box while others are always handled by the interpretable component of the hybrid model.

\begin{figure*}[t]
\centering
\includegraphics[width=0.5\textwidth]{Plots/ICF/allmethods_ICF_shared_legend.pdf}

\begin{subfigure}{0.28\textwidth}
    \centering
    \includegraphics[width=\linewidth]{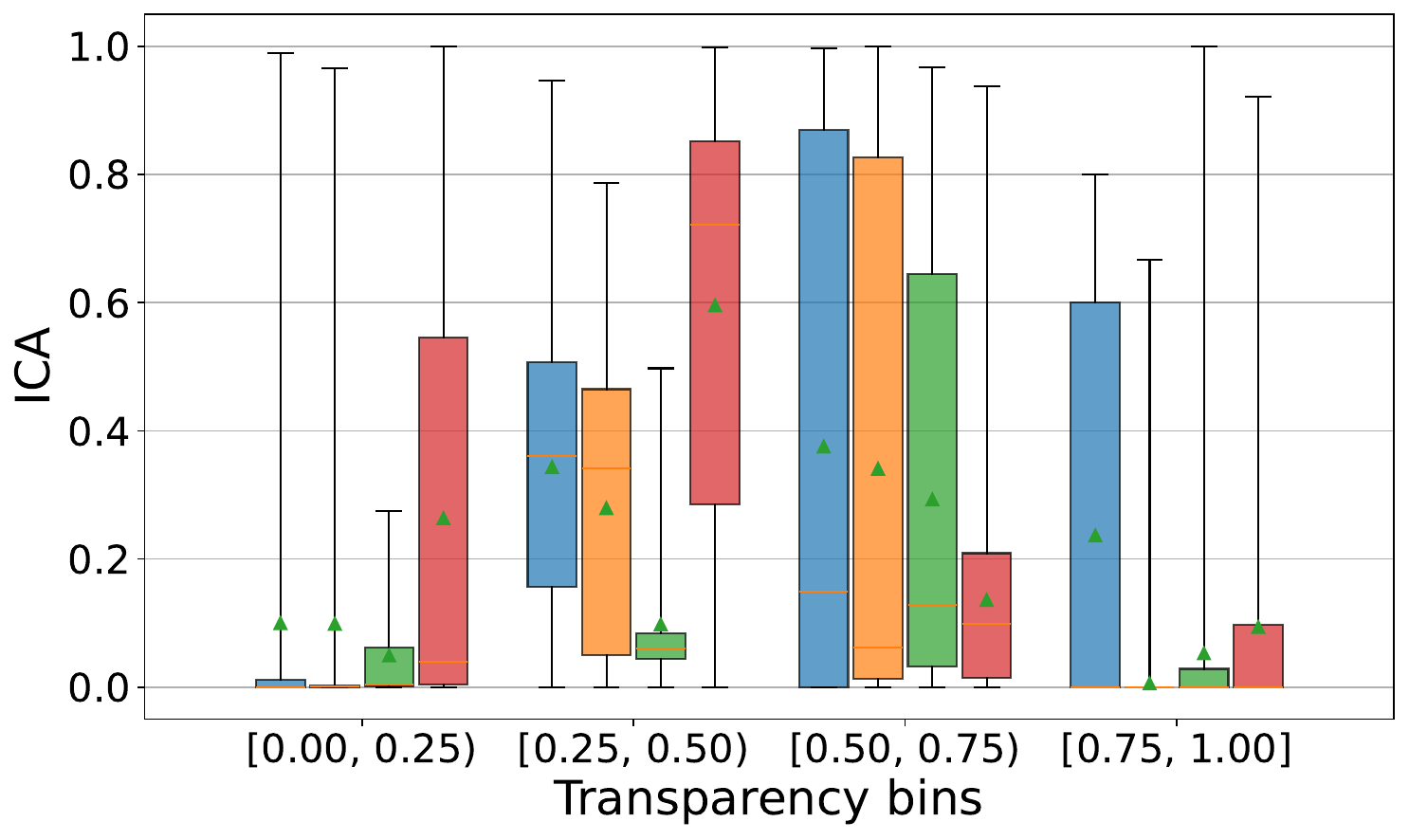}
    \caption{ACS Employment dataset}
\end{subfigure}
\begin{subfigure}{0.28\textwidth}
    \centering
    \includegraphics[width=\linewidth]{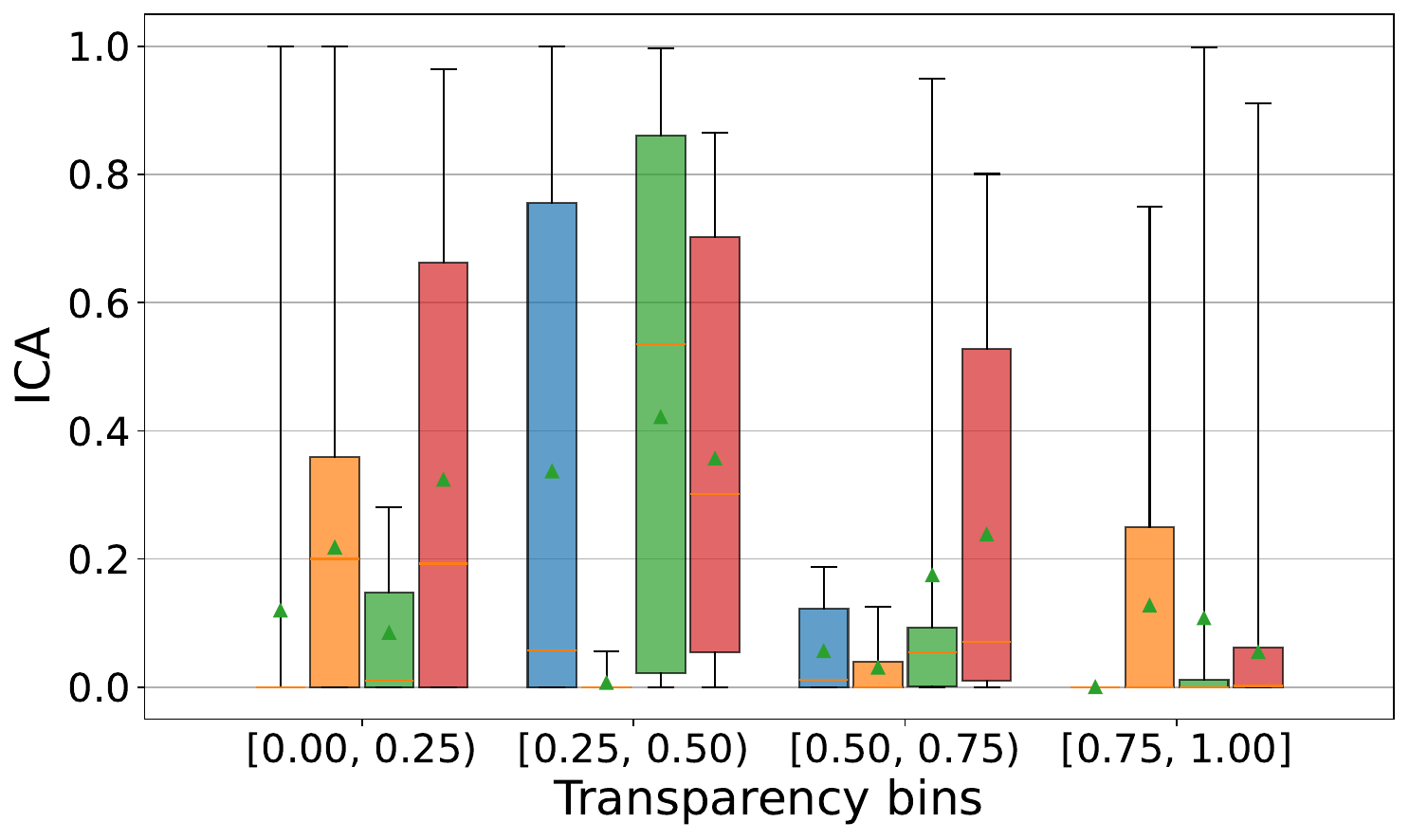}
    \caption{UCI Adult Income dataset}
\end{subfigure}
\begin{subfigure}{0.28\textwidth}
    \centering
    \includegraphics[width=\linewidth]{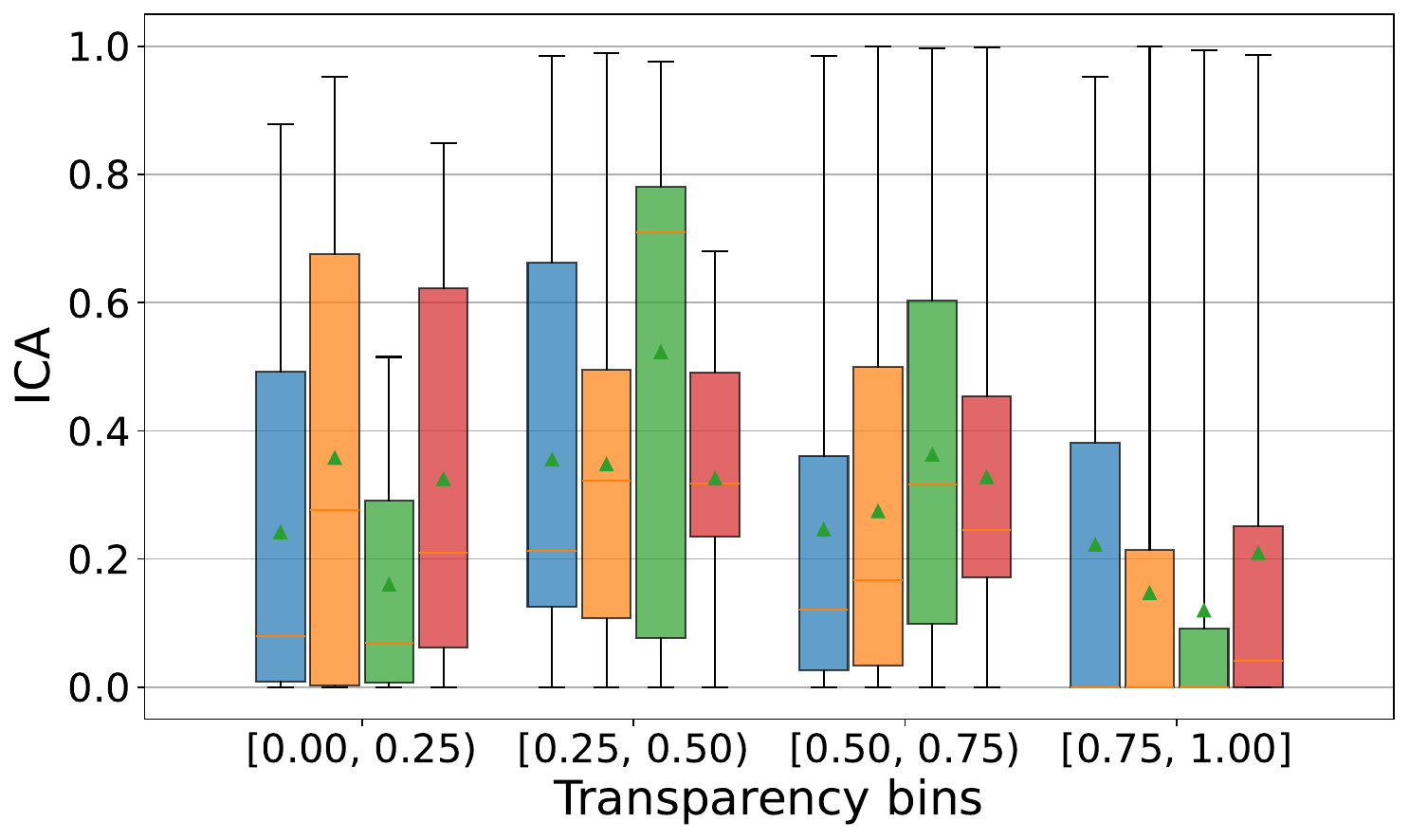}
    \caption{COMPAS dataset}
\end{subfigure}
\caption{Distribution of test set ICA across Rashomon sets for all transparency bins $(\varepsilon = 0.01)$. Results are shown for all datasets. }
\label{Fig:ICA_box}

\end{figure*}

To further investigate whether the observed disparities consistently favor some demographic subgroups in receiving access to interpretable decisions, Figure~\ref{fig:ICF_HyRS_compas_Race} presents the distribution of interpretability coverage (IC) across demographic subgroups of the Race attribute for the HyRS method on the COMPAS dataset. Analogous figures for the remaining hybrid interpretable methods, datasets, and sensitive attributes are provided in Appendix~\ref{sec:appendix_ICD_results}. The results show that within the same transparency bin, certain subgroups consistently receive higher interpretability coverage than others, indicating that some populations are more frequently routed to the interpretable component of the hybrid model than others.
For example, as seen in Figure~\ref{fig:ICF_HyRS_compas_Race}, the African-American subgroup generally exhibits lower interpretability coverage, with smaller maximum and average IC values compared to the other Race groups across transparency bins.

Another important observation from Figures~\ref{fig:box_ICF_max_delta_1} and~\ref{fig:ICF_HyRS_compas_Race} concerns how ICD evolves as model transparency increases. 
According to the mechanism of hybrid interpretable models, the lowest and highest transparency bins correspond to regimes in which predictions are largely dominated by a single component. In the first bin, most instances are routed to the black-box model, whereas in the last bin, most instances are assigned to the interpretable component. The intermediate transparency bins (transparency values between 0.25 and 0.75) therefore better characterize the truly hybrid regime, where both components actively contribute to prediction. The largest ICD values are observed in these bins, as the curves tend to have a \emph{bell-like} shape, where ICD increases from low to intermediate transparency bins and then decreases or stabilizes at higher transparency bins. %

To statistically validate this behavior, we conduct a non-parametric pairwise testing procedure across transparency bins. For each dataset, method, and sensitive attribute, we compare the distributions of ICD between adjacent bins, namely $(Q_1$ vs. $Q_2)$, $(Q_2$ vs. $Q_3)$, and $(Q_3$ vs. $Q_4)$. We employ the Mann–Whitney U test, which compares two distributions without assuming normality, to test the null hypothesis that the ICD distributions of adjacent bins are not significantly different \cite{mannTestWhetherOne1947}. Since multiple comparisons are performed within each experimental setting, Holm’s correction is applied to control the family-wise error rate \cite{holmSimpleSequentiallyRejective1979}. For each adjacent comparison, we determine whether ICD exhibits a statistically significant increase, a statistically significant decrease, or no significant change at $(p \leq 0.05)$ after correction.

Table~\ref{tab:bell_like_icd} summarizes, for each method and for two Rashomon parameters ($\varepsilon=0.01$ and $\varepsilon=0.05$), the proportion of experimental settings exhibiting a \emph{bell-like} pattern in a statistically significant manner, all remaining transition patterns being categorized as \emph{mixed}.
Overall, \emph{bell-like} patterns are frequently observed across datasets and methods, supporting the hypothesis that ICD is amplified in intermediate transparency regimes, where hybridization is most pronounced, and decreases when predictions become increasingly dominated by one component or the other. %
Because the size of the approximate Rashomon set depends directly on~$\varepsilon$, smaller tolerance values yield relatively fewer Rashomon models in higher transparency bins, limiting the statistical power of the pairwise comparisons and making bell-like patterns harder to detect reliably. Increasing the tolerance to $\varepsilon = 0.05$ yields larger Rashomon sets and, consequently, more stable statistical comparisons across bins. Consequently, the bell-like trend is strengthened when a richer set of near-optimal models is available for analysis.

\begin{table}[t]
\centering
\caption{Prevalence of statistically significant bell-like ICD patterns for two Rashomon parameters $\varepsilon$. 
}
\label{tab:bell_like_icd}
\resizebox{\columnwidth}{!}{%
\begin{tabular}{lcccc}
\toprule
 & \multicolumn{2}{c}{$\varepsilon = 0.01$} 
 & \multicolumn{2}{c}{$\varepsilon = 0.05$} \\
\cmidrule(lr){2-3} \cmidrule(lr){4-5}
Method & Bell-like & Mixed & Bell-like & Mixed \\
\midrule
HybridCORELSPost & 89\% & 11\% & 78\% & 22\% \\
HybridCORELSPre  & 78\% & 22\% & 100\% & 0\% \\
HyRS             & 89\% & 11\% & 100\% & 0\% \\
CRL              & 67\% & 33\% & 100\% & 0\% \\
\bottomrule
\end{tabular}%
}
\end{table}

\paragraph{Result 2. Hybrid interpretable models exhibit assignment arbitrariness across Rashomon sets.}

\begin{figure*}[t]
\centering
\includegraphics[width=0.8\textwidth]{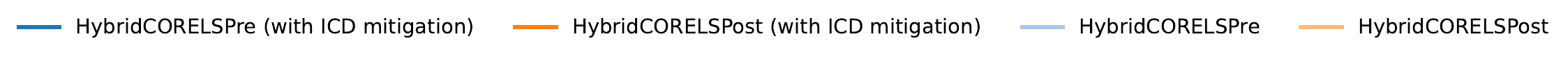}

\begin{subfigure}{\textwidth}
    \centering
    \includegraphics[width=0.28\linewidth]{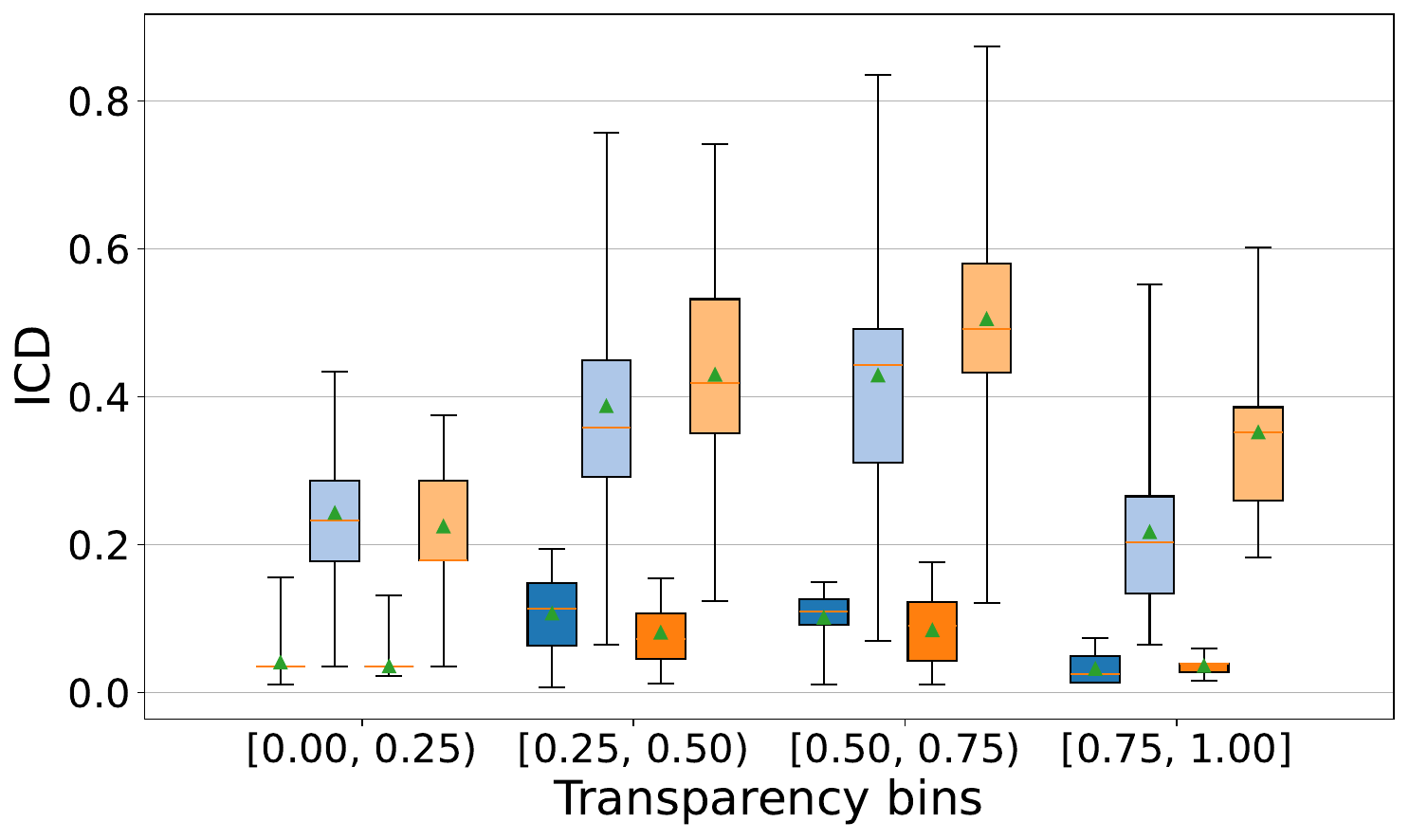}
    \includegraphics[width=0.28\linewidth]{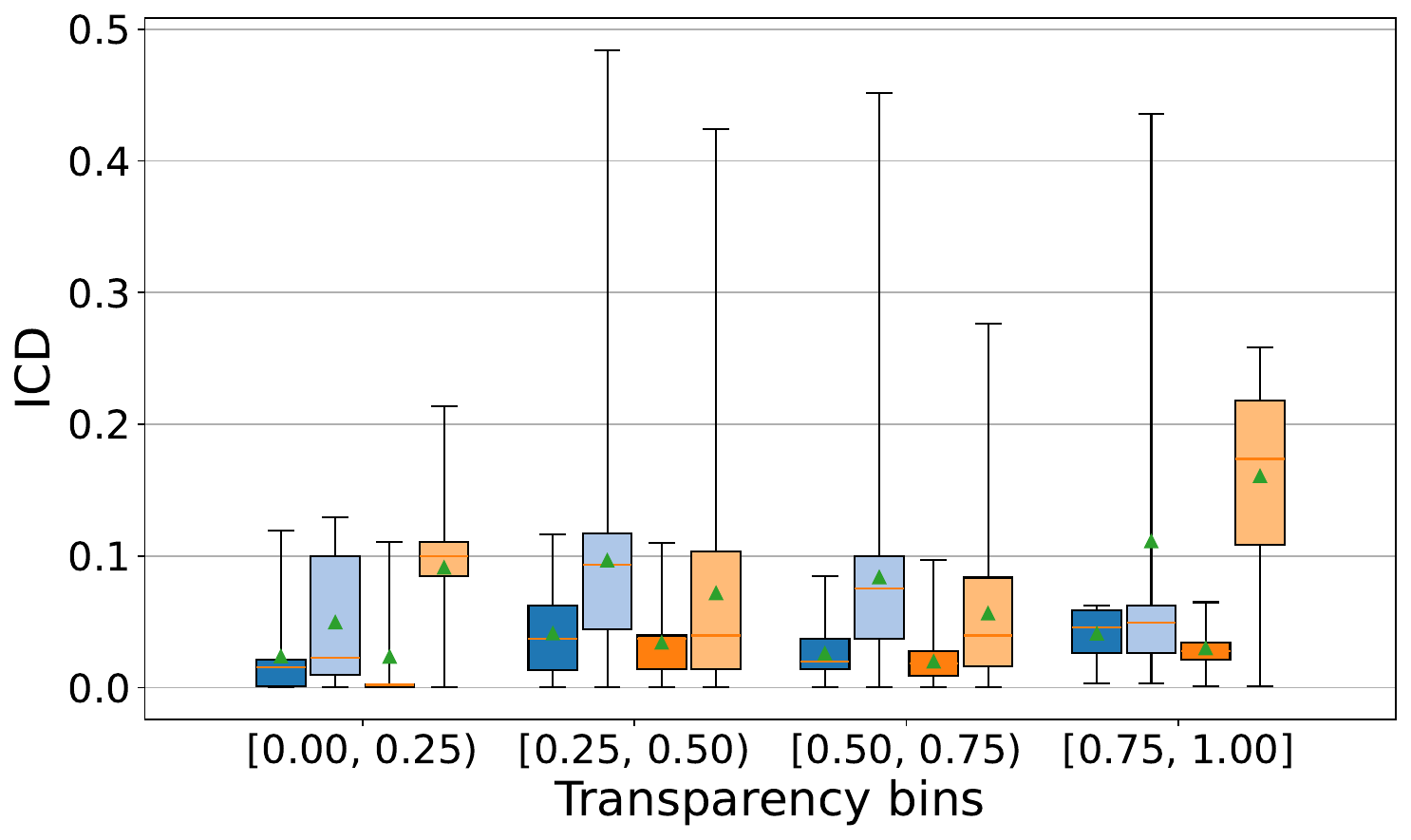}
    \includegraphics[width=0.28\linewidth]{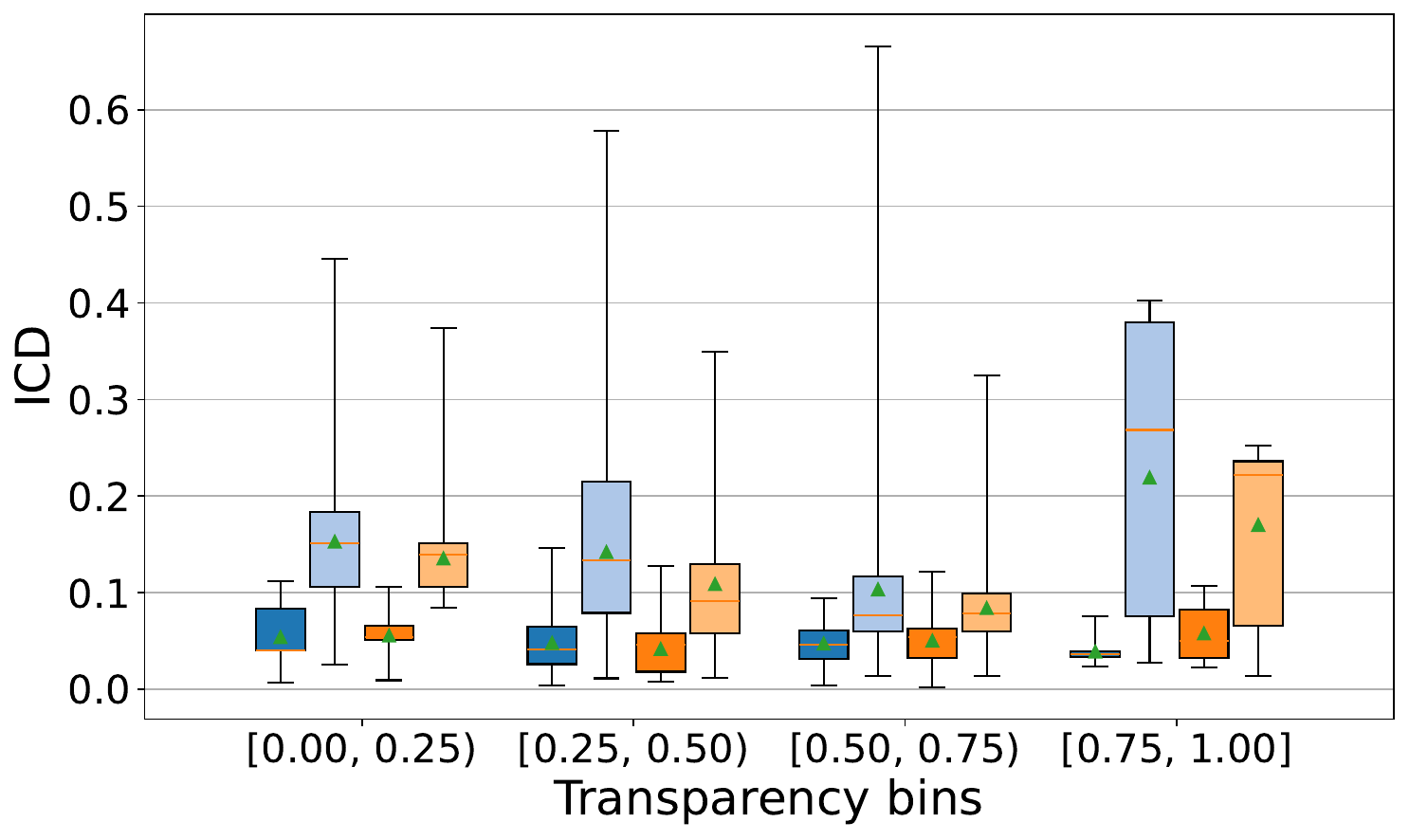}
    \caption{Test set ICD.}\label{fig:Fair_max_ICF_compas}
\end{subfigure}

\vspace{0.5em}

\begin{subfigure}{\textwidth}
    \centering
    \includegraphics[width=0.28\linewidth]{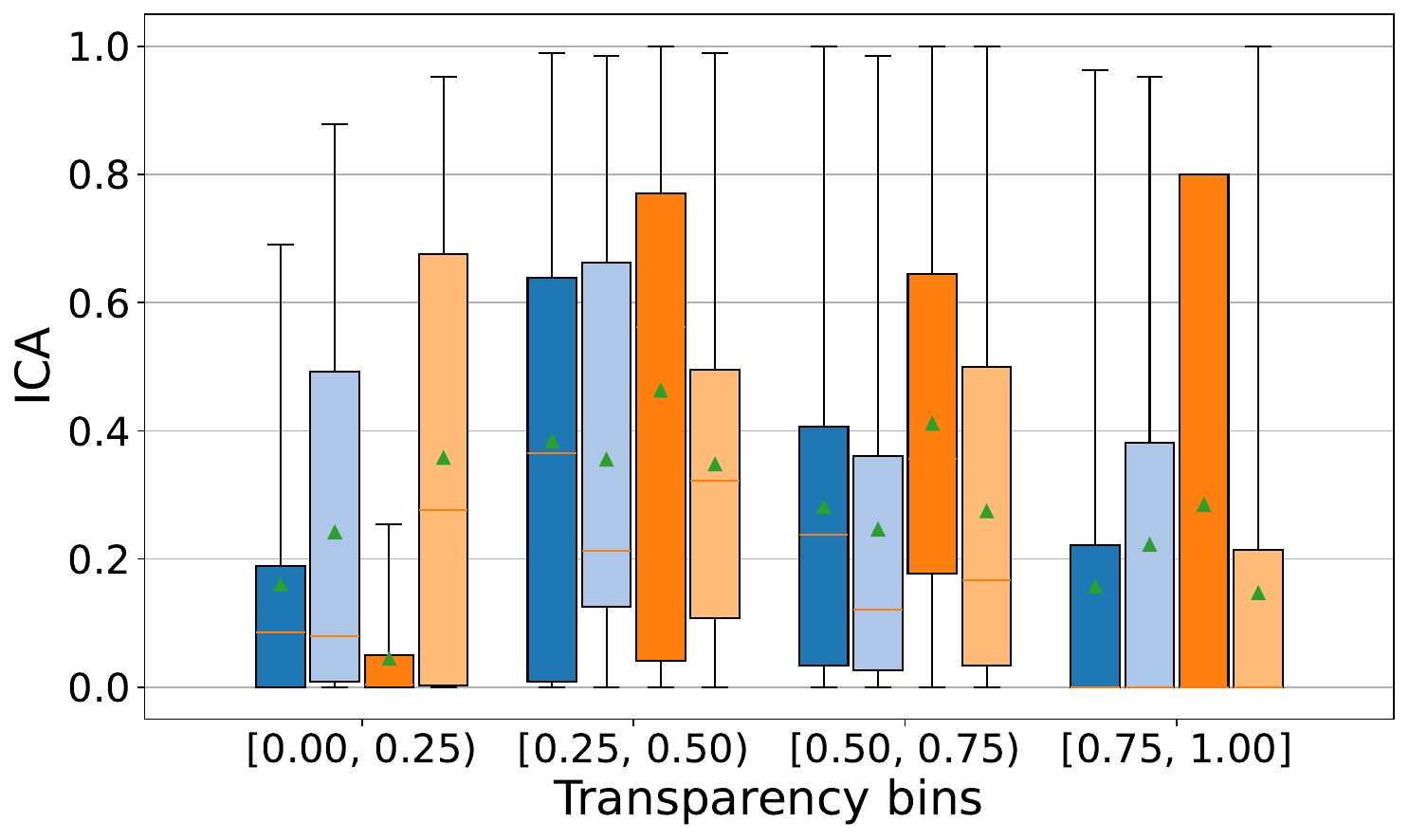}
    \includegraphics[width=0.28\linewidth]{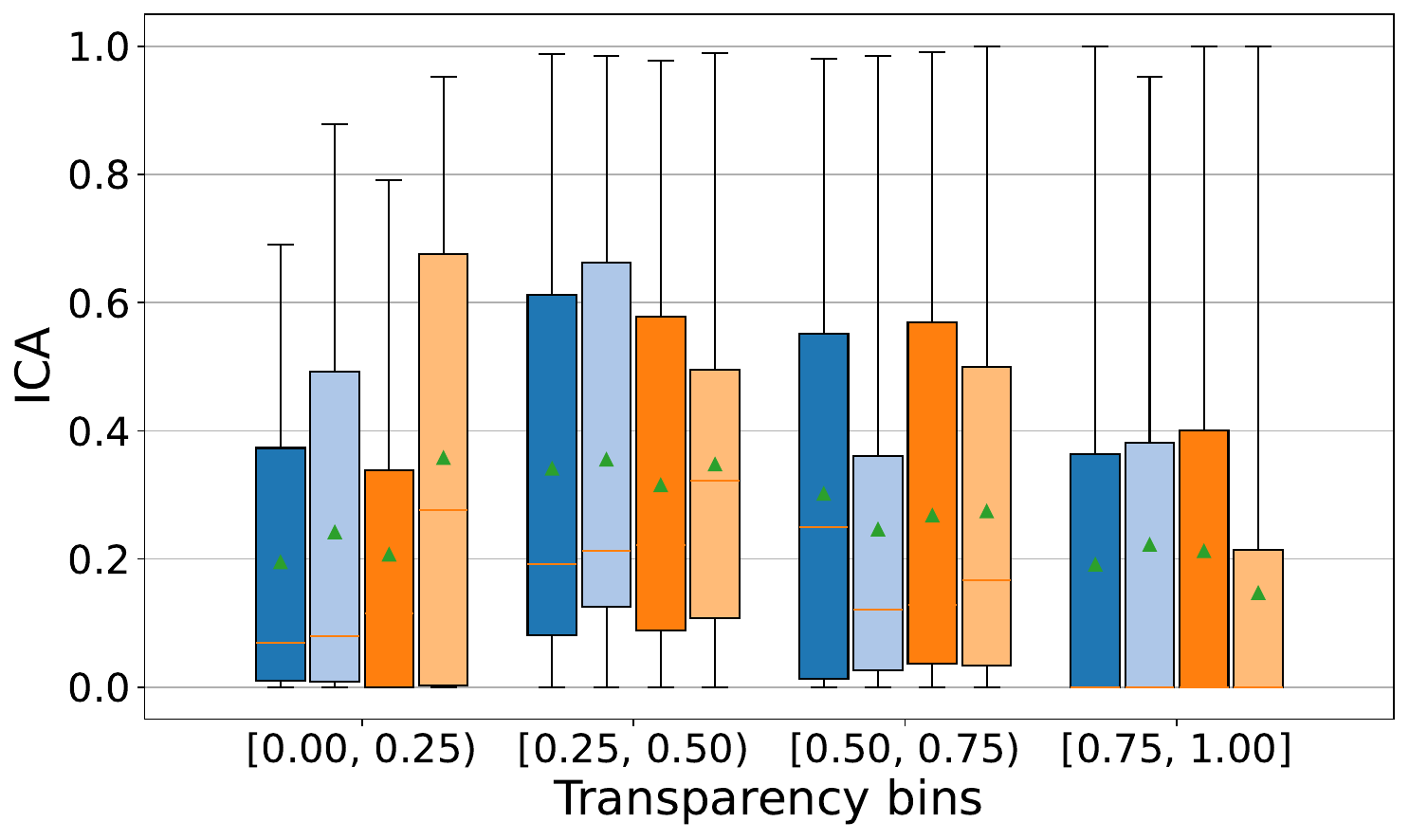}
    \includegraphics[width=0.28\linewidth]{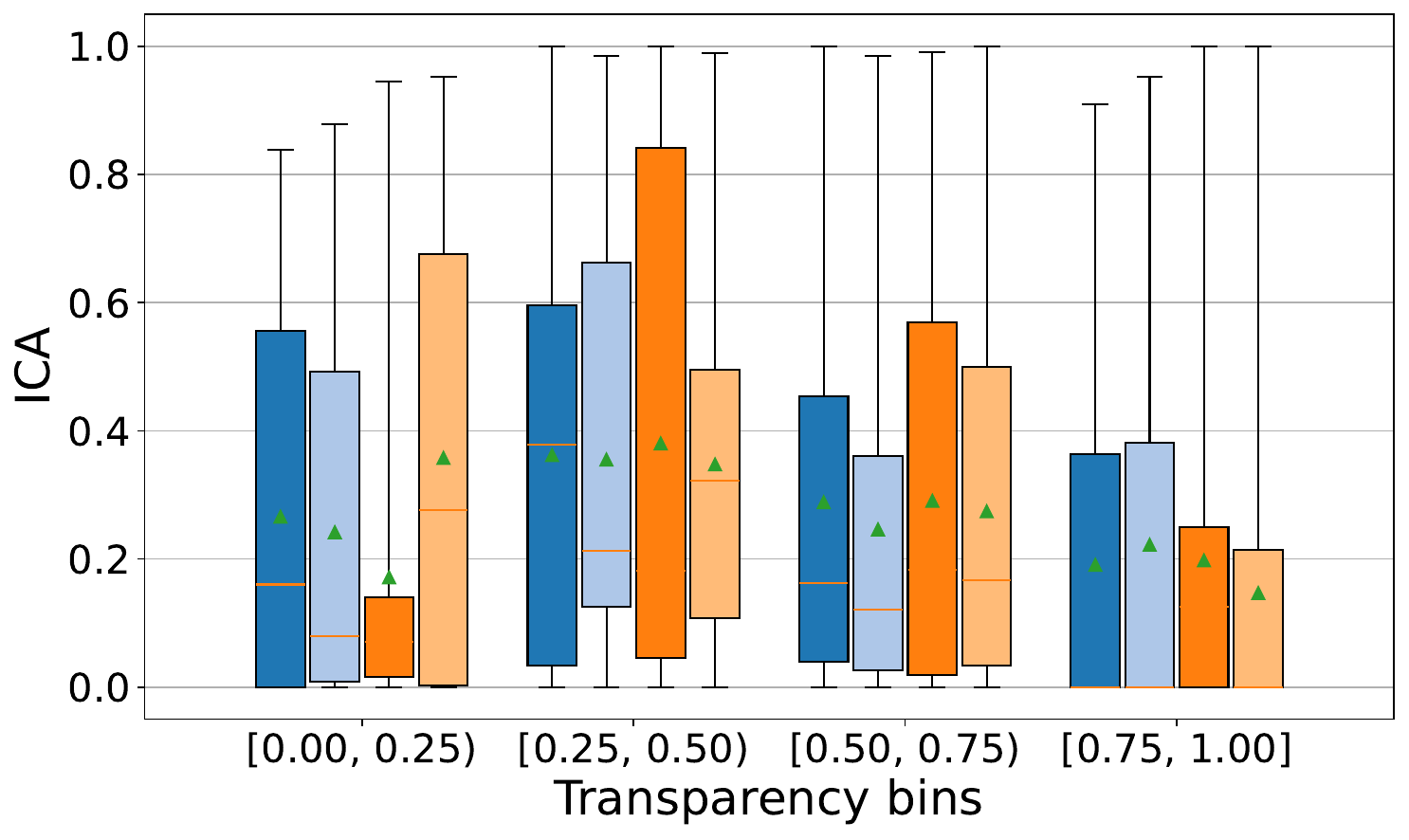}
    \caption{Test set ICA.} \label{fig:Fair_ICA_compas}
\end{subfigure}

\caption{Distribution of our proposed metrics across Rashomon sets over transparency bins $(\varepsilon = 0.01)$ for HybridCORELSPre and HybridCORELSPost, with or without our proposed ICD mitigation $(\eta = 0.05)$. Each subplot corresponds to a mitigation applied to a sensitive attribute (Age, Gender, Race), ordered from left to right. Results are reported for the COMPAS dataset.}
\end{figure*}

Beyond group-level disparities in interpretability access, Figure~\ref{Fig:ICA_box} displays the distribution of ICA values over all test set examples for the three considered datasets and each transparency bin.
The results reveal that hybrid interpretable models also exhibit assignment arbitrariness at the individual level. More precisely, arbitrariness is generally limited in the lowest and highest transparency bins, where predictions are predominantly handled by a single component of the hybrid model. In contrast, substantially higher arbitrariness emerges in intermediate transparency bins, where both interpretable and black-box components contribute to prediction. In these regimes, a larger proportion of individuals exhibit instability in whether they receive interpretable or black-box decisions across equally accurate models. This behavior is observed across all methods, supporting the hypothesis that hybridization amplifies assignment uncertainty. %

\section{Mitigating ICD} \label{sec:ICD_mitigation}

\subsection{Methodology}

\paragraph{Modified Learning Algorithms.} The experimental analysis in the previous section revealed substantial ICD across demographic groups, methods, and datasets, as well as high variability within Rashomon sets over transparency bins. These observations suggest that near-optimal models with similar predictive performance can nonetheless induce substantially different levels of ICD. This variability further motivates the use of predictive multiplicity to explicitly search for and select fairer models while preserving high predictive performance. To this aim, we mitigate the observed ICD by incorporating maximum ICD constraints into HybridCORELS~\citep{HybridCorel_Julien} branch-and-bound algorithms. In a nutshell, given a set of protected groups $\mathcal{P}$ and a maximum ICD value $\eta$, each time a new best prefix is found, we check whether it satisfies the maximum ICD constraint. If it does, the prefix is retained and used to update the current incumbent; otherwise, it is discarded. This integration follows the same principle as the overall transparency constraint used in HybridCORELS. Importantly, the bounds and symmetry-aware data structures used in HybridCORELSPre and HybridCORELSPost remain valid and tight under the new constraint.
The corresponding pseudocodes are provided in Appendix~\ref{app:detailed_pseudocodes}.%

\paragraph{Experimental Setup.} We run experiments for both HybridCORELSPre and HybridCORELSPost, following the setup introduced in Section~\ref{sec:expes_setup}. These experiments have two main objectives. First, we assess whether ICD mitigation is effective, and whether enforcing ICD constraints on the bootstrapped training sets also translates into improved ICD on the test set. Second, we evaluate the effect of this mitigation on other dimensions: we report sparsity, measured as the number of rules in the prefix of each hybrid interpretable model, as well as test-set accuracy and algorithmic fairness, measured using SP and EO.
Similar to Section~\ref{sec:ICD_results}, all experiments are conducted on approximate Rashomon sets with $\varepsilon = 0.01$ across transparency bins. Unless otherwise specified, mitigation is performed by enforcing maximum ICD constraints during training with $\eta = 0.05$, while all reported metrics are evaluated on the test set to assess the generalization ability of the proposed mitigation approach. %

\subsection{Results} \label{sec:mitigation_results}

Comprehensive results for all datasets, sensitive attributes, and experimental settings are provided in Appendix~\ref{sec:appendix_mitigation_results}. We hereafter report the key trends using a subset of these results.

\begin{figure*}[t]
\centering
\includegraphics[width=0.8\textwidth]{Plots/ICF/HybridCORELS_fairness_shared_legend.pdf}

\begin{subfigure}{\textwidth}
    \centering
    \includegraphics[width=0.28\linewidth]{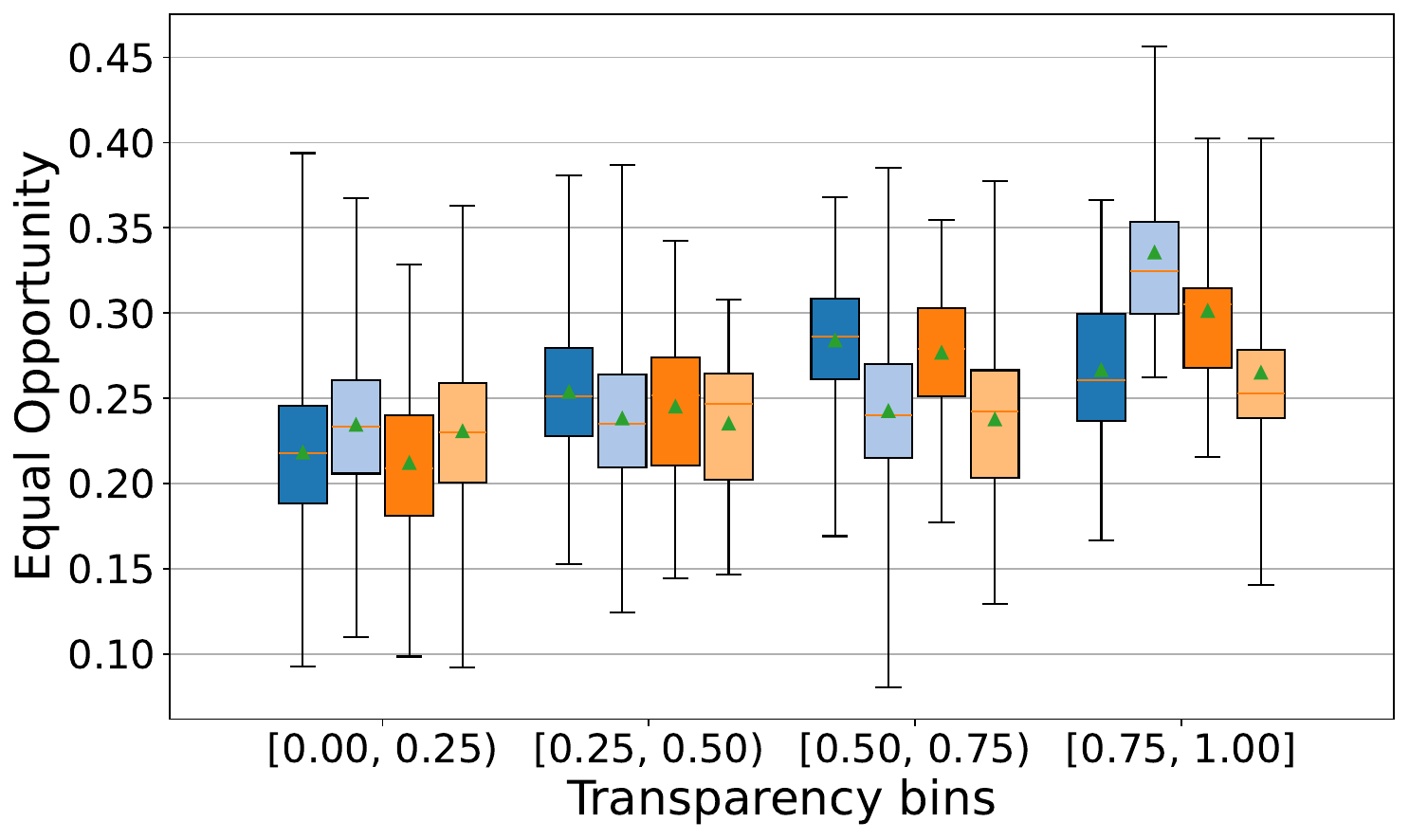}
    \includegraphics[width=0.28\linewidth]{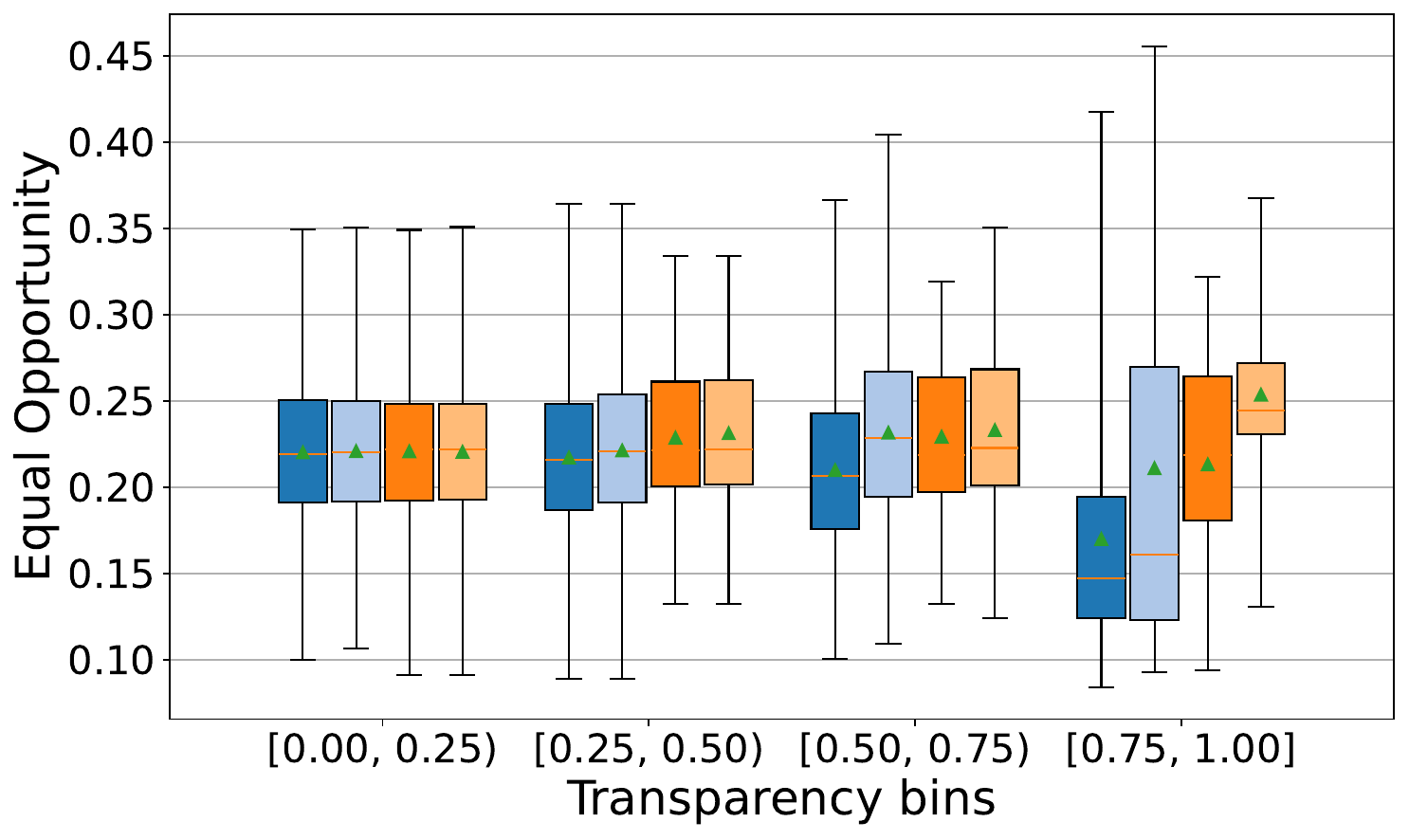}
    \includegraphics[width=0.28\linewidth]{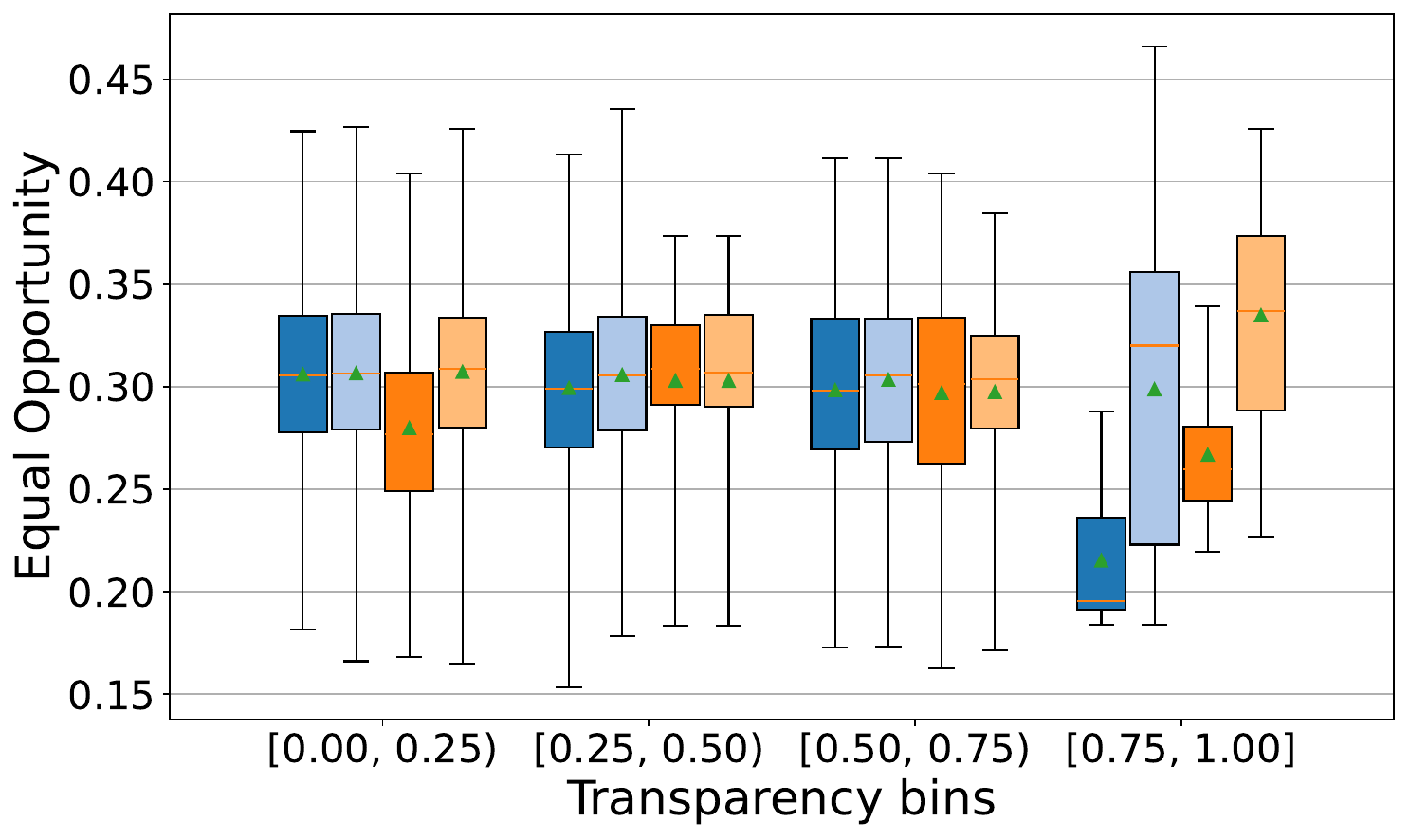}
    \caption{Test set EO.}\label{fig:Fair_max_EO_comaps}
\end{subfigure}

\vspace{0.5em}

\begin{subfigure}{\textwidth}
    \centering
    \includegraphics[width=0.28\linewidth]{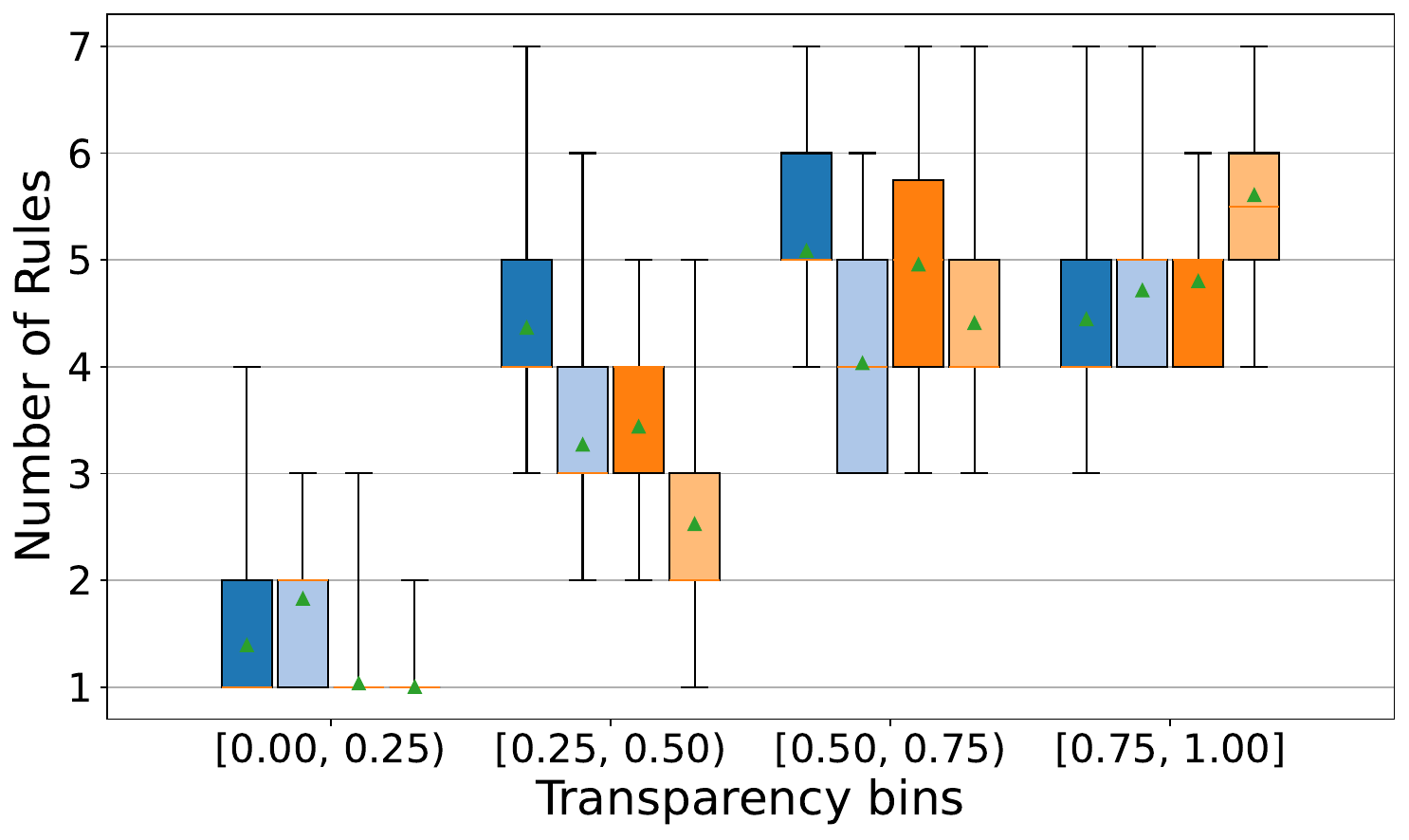}
    \includegraphics[width=0.28\linewidth]{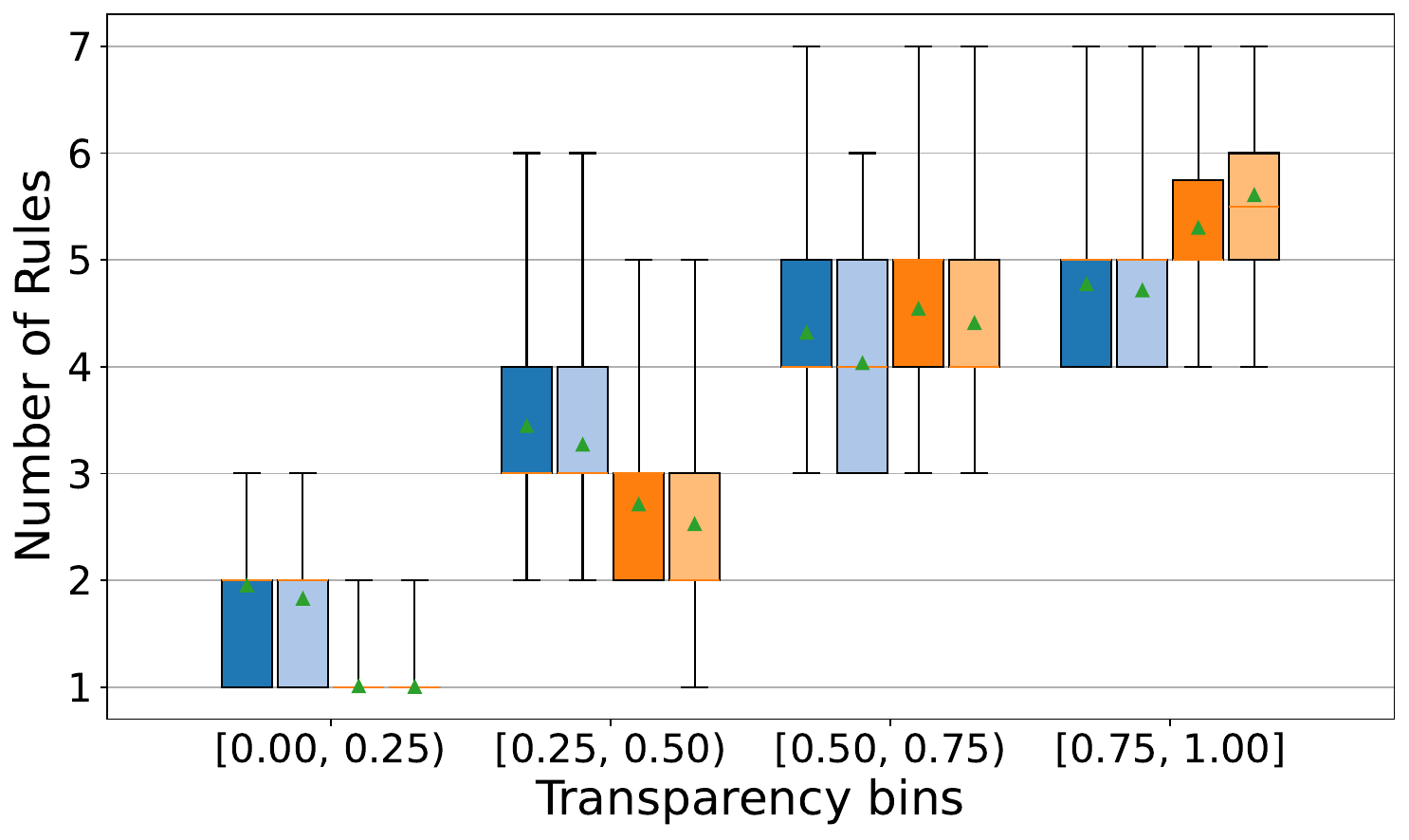}
    \includegraphics[width=0.28\linewidth]{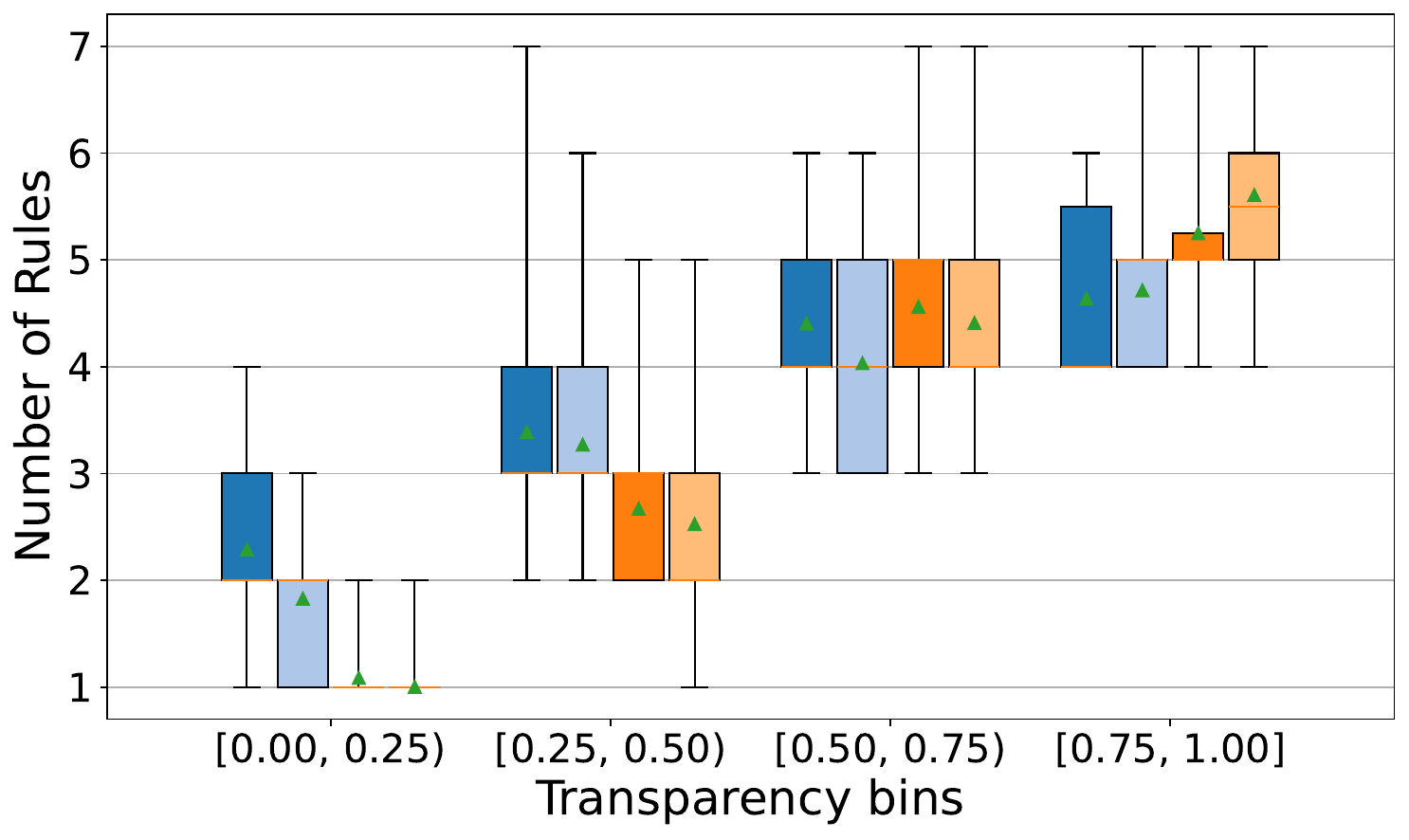}
    \caption{Model sparsity (number of rules in the interpretable component).}\label{fig:Fair_Sparsity_compas}
\end{subfigure}

\vspace{0.5em}

\begin{subfigure}{\textwidth}
    \centering
    \includegraphics[width=0.28\linewidth]{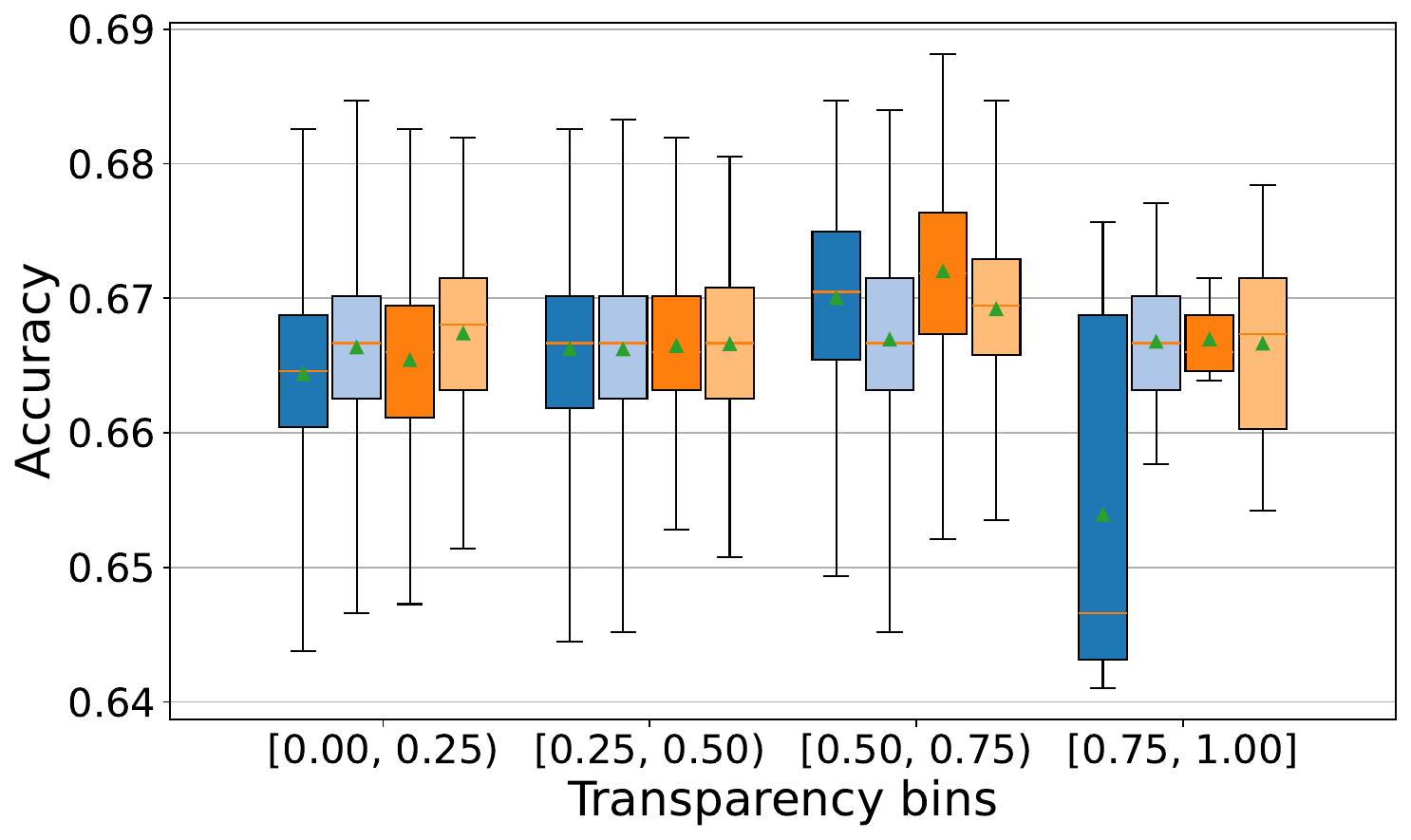}
    \includegraphics[width=0.28\linewidth]{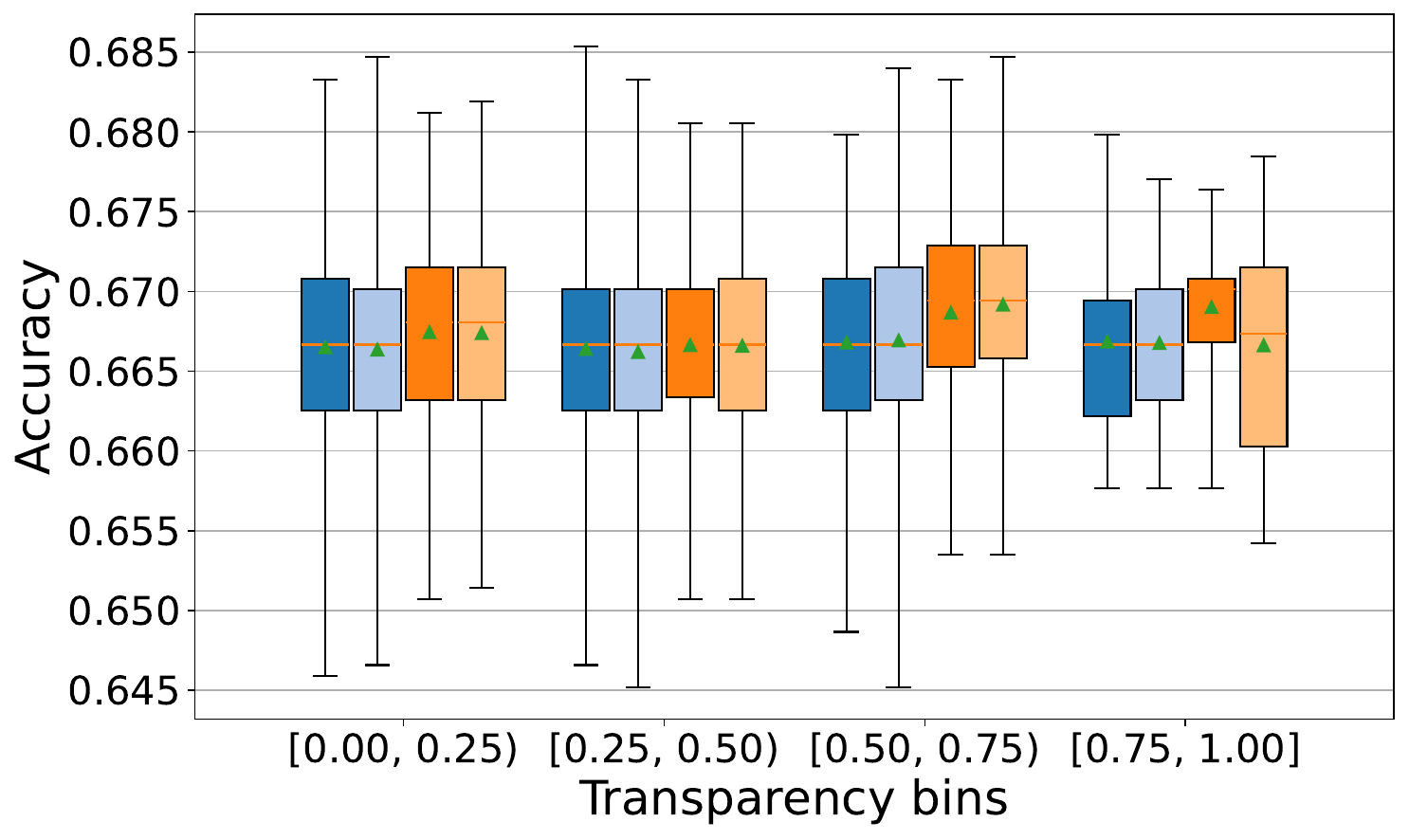}
    \includegraphics[width=0.28\linewidth]{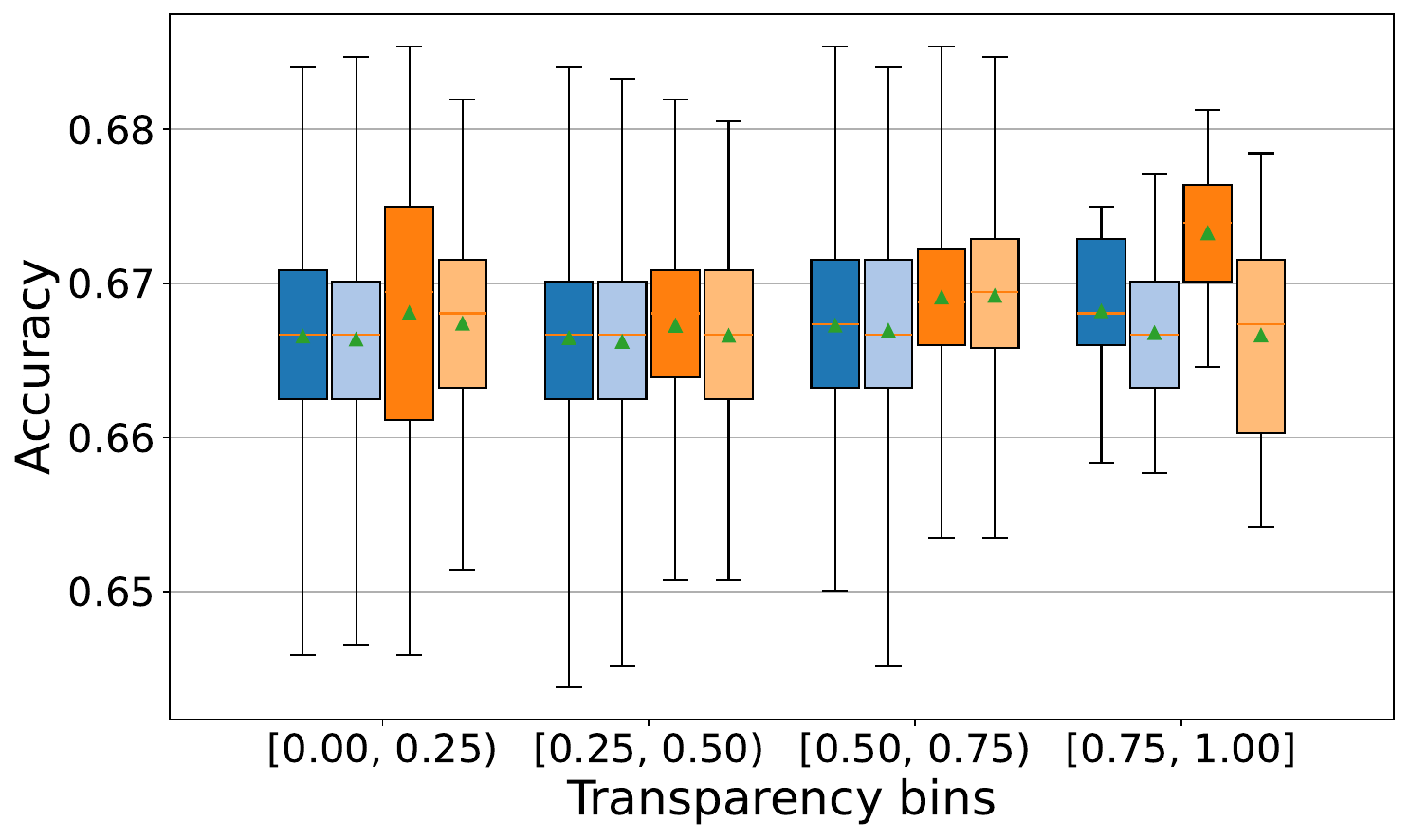}
    \caption{Test set accuracy.}\label{fig:Fair_Acc_compas}
\end{subfigure}

\caption{Distribution of several desiderata across Rashomon sets over transparency bins $(\varepsilon = 0.01)$ for HybridCORELSPre and HybridCORELSPost, with or without our proposed ICD mitigation $(\eta = 0.05)$.  Each subplot corresponds to a mitigation applied to a sensitive attribute (Age, Gender, Race), ordered from left to right. Results are reported for the COMPAS dataset.}
\end{figure*}

\paragraph{Result 3. ICD mitigation generalizes effectively to unseen test data.}

Figure~\ref{fig:Fair_max_ICF_compas} illustrates the distribution of test set ICD within Rashomon sets of different transparency bins, on the COMPAS dataset, with or without our proposed ICD mitigation. Across all sensitive attributes, mitigation consistently reduces both the magnitude and variability of ICD distributions. In particular, post-mitigation Rashomon sets exhibit substantially lower median and maximum ICD values, along with narrower distributions across transparency bins, indicating more stable and equitable interpretability allocation across demographic groups. These results demonstrate that enforcing ICD constraints during training generalizes effectively to unseen test data.

\paragraph{Result 4. ICD mitigation has no clear impact on assignment arbitrariness.}
Figure~\ref{fig:Fair_ICA_compas} displays the distribution of test set ICA within Rashomon sets of different transparency bins, on the COMPAS dataset, with or without our proposed ICD mitigation. Overall, the ICD mitigation procedure produces no substantial change in the distribution of assignment arbitrariness across datasets and transparency bins: with or without ICD mitigation, the ICA values remain largely similar. This indicates that although ICD mitigation is highly effective at reducing group-level interpretability coverage disparities, it has limited impact on individual-level assignment arbitrariness across Rashomon sets.

\paragraph{Result 5. ICD mitigation frequently improves algorithmic fairness metrics.}
Figure~\ref{fig:Fair_max_EO_comaps} displays the distribution of test set EO within Rashomon sets of different transparency bins, on the COMPAS dataset, with or without our proposed ICD mitigation. Analogous results for the SP metric are provided in Appendix~\ref{sec:appendix_mitigation_results}. %
The experiments show frequent improvements in both EO and SP after mitigation, despite these metrics not being directly targeted during training. In many settings, mitigated models exhibit lower median disparities and reduced variability across Rashomon sets, particularly noticeable decreases in both the spread and the upper range of EO values in higher transparency bins. However, the improvement is not uniform across all datasets and bins, and certain settings show relatively limited or inconsistent changes after mitigation. These observations suggest that reducing ICD can often indirectly improve algorithmic fairness metrics, further illustrating how improving procedural fairness might benefit outcome-based fairness.

\begin{figure*}[t]
\centering
\includegraphics[width=0.6\textwidth]{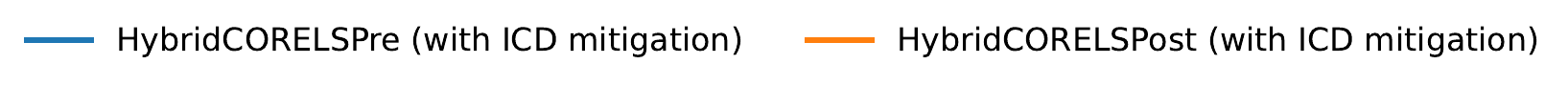}

\begin{subfigure}{0.24\textwidth}
    \centering
    \includegraphics[width=\linewidth]{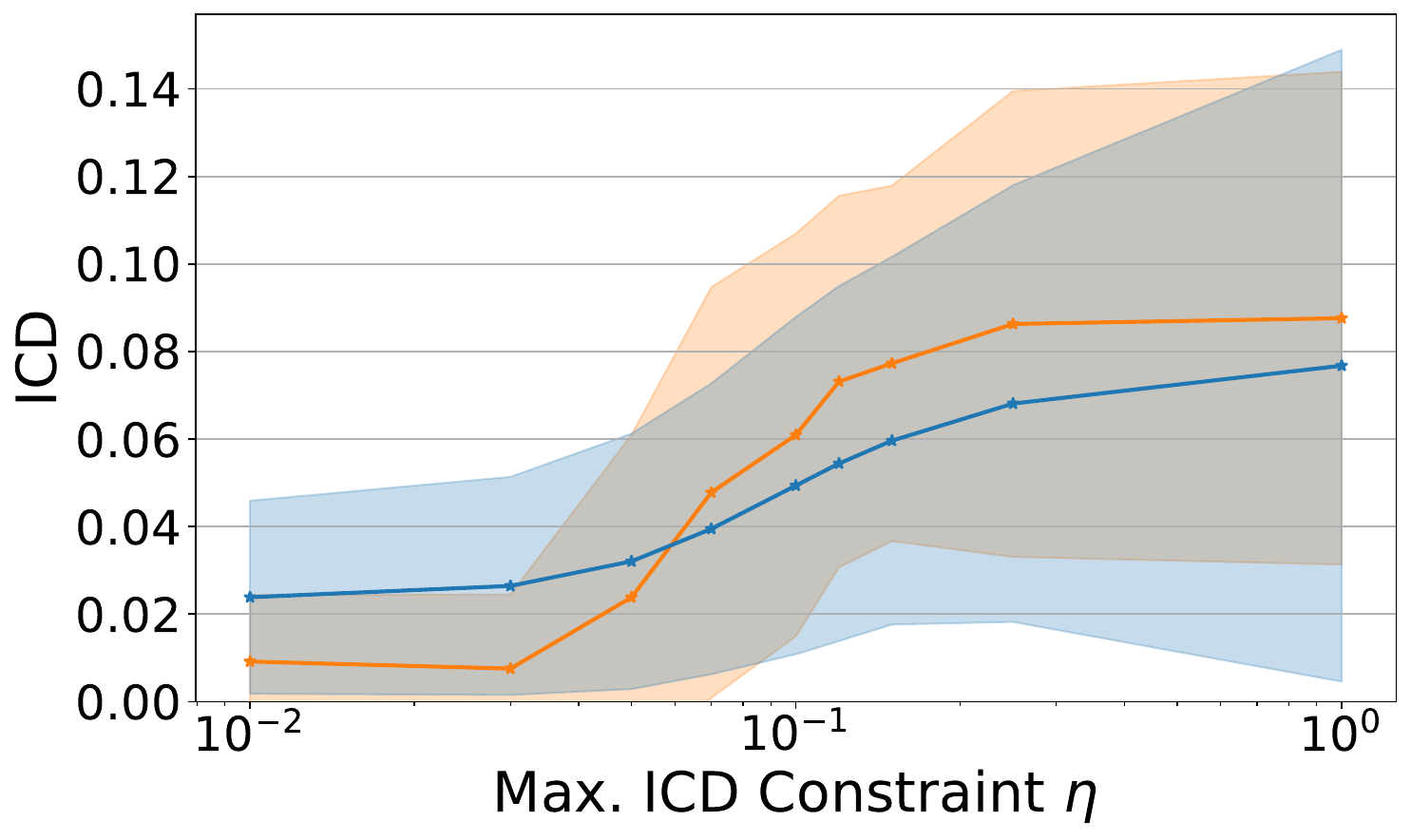}
\end{subfigure}
\hfill
\begin{subfigure}{0.24\textwidth}
    \centering
    \includegraphics[width=\linewidth]{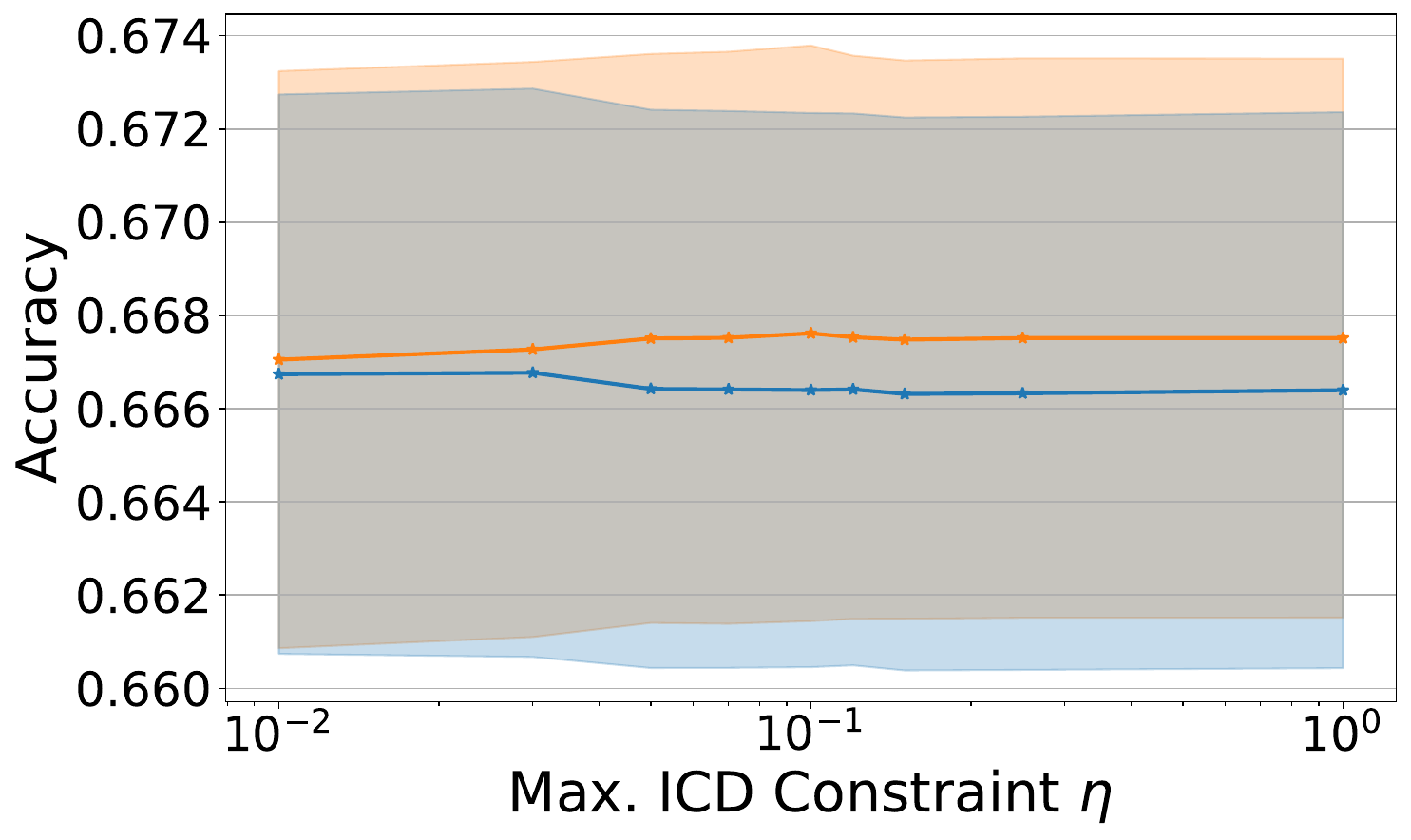}
\end{subfigure}
\hfill
\begin{subfigure}{0.24\textwidth}
    \centering
    \includegraphics[width=\linewidth]{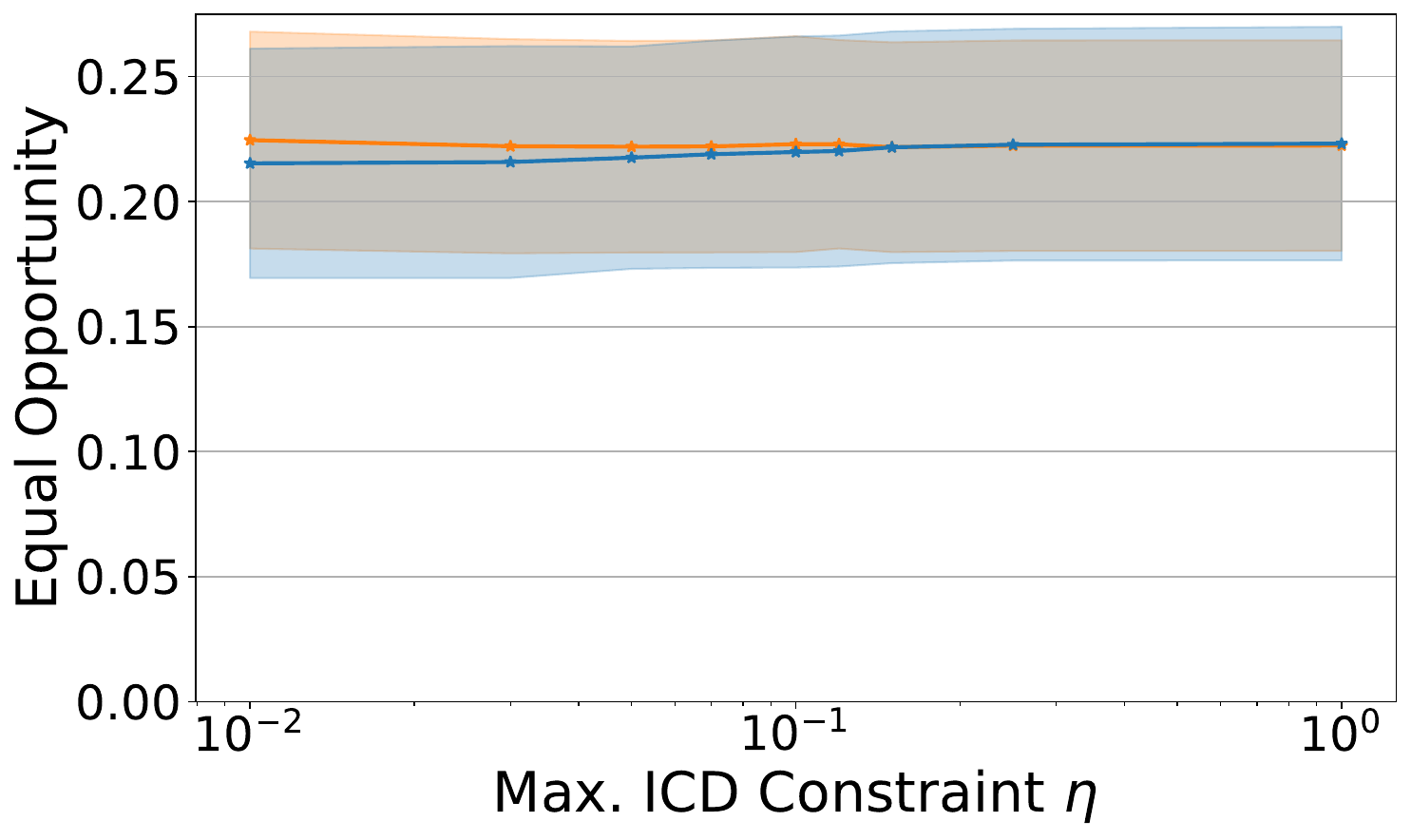}
\end{subfigure}\hfill
\begin{subfigure}{0.24\textwidth}
    \centering
    \includegraphics[width=\linewidth]{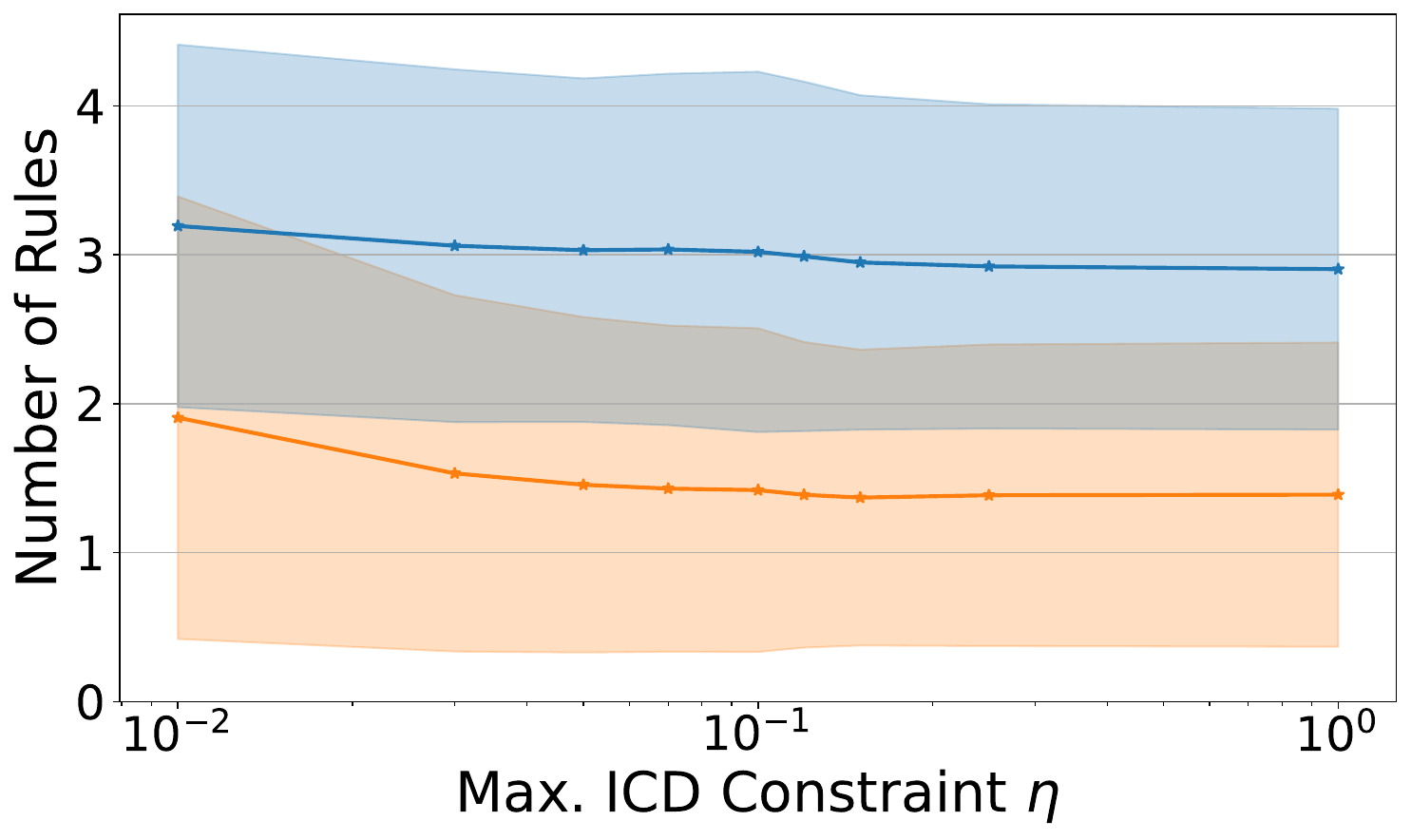}
\end{subfigure}
\caption{Test set ICD, accuracy and EO, and model sparsity for HybridCORELSPre and HybridCORELSPost with ICD mitigation across Rashomon sets over all transparency values in $[0,1]$ ($\varepsilon = 0.01$). For different levels of mitigation induced by the maximum ICD constraint ($\eta$), curves report the average and standard deviation across Rashomon models. Results are provided for the COMPAS dataset and Gender as sensitive attribute.}

\label{allmetrics_MaxCov_1Q}

\end{figure*}

\paragraph{Result 6. ICD mitigation preserves predictive performance and sparsity.}\label{subsubsec:icd-vs-accuracy}
Figure~\ref{fig:Fair_Sparsity_compas} (respectively,~\ref{fig:Fair_Acc_compas}) reports the distribution of the number of rules in the learned prefix (respectively, of test accuracy) within Rashomon sets of different transparency bins, on the COMPAS dataset, with or without our proposed ICD mitigation. Corresponding results for the other datasets are provided in Appendix~\ref{sec:appendix_mitigation_results}. 
Since the number of rules quantifies the \emph{sparsity} of the interpretable component, this analysis evaluates whether ICD mitigation increases model complexity. The results show that enforcing ICD constraints leads to only minor increases in the number of rules, with sparsity remaining unchanged in many settings.
Enforcing ICD constraints also has only a limited impact on predictive performance. Although some settings exhibit slight decreases in test accuracy after mitigation, the overall accuracy distributions remain highly comparable to those without mitigation across all transparency bins. In some configurations, enforcing ICD constraints can improve test accuracy, suggesting a slight regularization effect. %

\paragraph{Result 7. Varying the enforced ICD constraint confirms the previous observations.}

To further analyze the effect of ICD mitigation strength, Figure~\ref{allmetrics_MaxCov_1Q} illustrates how test set ICD, accuracy, EO, as well as sparsity evolve as the maximum ICD constraint is gradually relaxed from a highly restrictive setting $(\eta = 0.01)$ to the unconstrained case $(\eta = 1)$, corresponding to no mitigation. This analysis is conducted on the COMPAS dataset with mitigation applied to the Gender attribute. We consider all models within the approximate Rashomon set satisfying the desired performance threshold $(\varepsilon = 0.01)$ without partitioning them into transparency bins. Appendix~\ref{sec:appendix_mitigation_results} provides bin-level analyses.

In terms of predictive performance, test accuracy remains remarkably stable across the full range of mitigation strengths, with no substantial degradation even under the strictest maximum ICD constraints. %
As expected, relaxing the maximum ICD constraint consistently increases both the average and the variability of ICD across Rashomon models. %
Combined with relatively unchanged predictive performance across mitigation levels, these results suggest that models with substantially lower ICD can be selected from the Rashomon set without sacrificing predictive accuracy. %

For algorithmic fairness metrics, measured through EO and SP, the aggregated effect across all Rashomon models appears relatively modest. However, the bin-level analyses in Appendix~\ref{sec:appendix_mitigation_results} (Figures~\ref{EO_MaxCov_4Q} and~\ref{SP_MaxCov_4Q}) show that the largest improvements occur in higher-transparency bins, where interpretability coverage is greatest and the mitigation constraint most strongly affects routing. For instance, in the highest-transparency bin, tightening $\eta$ from $0.25$ to $0.05$ reduces the average EO of HybridCORELSPost from about $0.25$ to about $0.18$, and its average SP from about $0.22$ to about $0.16$. %

Finally, the sparsity analysis shows only a slight increase in the number of rules under the strictest mitigation settings, consistent with previous observations. 
\cameraready{Overall, these findings provide evidence that ICD can often be substantially reduced with limited changes in predictive performance and sparsity, while frequently yielding improvements in outcome-based fairness metrics.}

\section{Conclusion} \label{sec:conclusion}

In this work, we formalized ICD and ICA as procedural fairness concerns arising at the group and individual levels in hybrid interpretable models, where some individuals receive decisions from an interpretable component while others are deferred to a black box. Our empirical results show that state-of-the-art hybrid interpretable methods can exhibit substantial ICD across protected groups when applied to real-world datasets. We also showed that predictive multiplicity provides a practical path to mitigation: within sets of similarly accurate models, one can often identify models with substantially lower ICD, with little impact on predictive performance or sparsity and, in some cases, indirect improvements in standard algorithmic fairness metrics. \cameraready{Further work will assess how consistently these trends extend across datasets, and evaluate, through user studies, whether greater interpretability coverage for unprivileged groups systematically translates into improved human understanding.}

\cameraready{Future work also includes investigating which characteristics make an example more or less likely to be affected by ICA, as well as designing mechanisms to mitigate such arbitrariness. Because ICA is defined with respect to a set of models, it cannot be directly evaluated during the training of a single model, making its mitigation a challenging open problem.
Mechanisms inspired by the stability literature~\citep{hara2023average} may help address it.}

\cameraready{Understanding the upstream causes of ICD is another important research direction, as it may arise from complex, dataset-dependent correlations among features, ground-truth labels, and sensitive attributes. Investigating its downstream effects on other dimensions of trustworthy machine learning is also important.} For instance, ICD might cause disparate privacy risks, since the examples deferred to the black-box are vulnerable to inference attacks~\citep{10.1145/3624010} against both the interpretable part they went through, and the black-box component ultimately handling them. ICD may also interact with robustness, as examples deferred to the black-box could be harder to audit for adversarial vulnerability~\citep{DBLP:journals/corr/abs-2005-08087}, especially when robustness must be assessed locally within a decision region.

\section*{Acknowledgments}
This research was enabled by support provided by Calcul Québec and the Digital Research Alliance of Canada, as well as funding from the SCALE-AI Chair in Data-Driven Supply Chains and the \emph{Fonds de recherche du Québec} -- \emph{Nature et technologies (FRQNT)} through a Team Research Project \emph{(327090)}. Ziba Jabbar Zare is also supported by the \emph{Fonds de recherche du Québec} -- \emph{Nature et technologies (FRQNT)} through a B2X Doctoral scholarship \emph{(350463)}. The authors would like to thank the anonymous reviewers for their valuable suggestions.

\bibliography{aaai2026}

\newpage
\appendix
\onecolumn

\def\highlighted#1{\textcolor{blue}{#1}}
\def\oneobj{z}
\def\oneant{a}
\def\oneconseq{q}
\def\defconseq{q_0}
\def\singlerule{r}
\def\priorityqueue{Q}
\def\specializationcoefficient{\alpha}
\def\dataset{S}
\newcommand\emploss[1]{\widehat{\mathcal{L}}_{#1}}
\def\mintransp{C_\text{min}}
\def\corelsreg{\lambda}
\def\prefix{r}
\def\minedrules{\Upsilon}
\def\lb{\mathsf{lb}}
\def\blackbox{h_c}
\def\regcoeffhycorels{\beta}
\def\obj{\mathsf{obj}}
\def\errs[#1,#2]{\emploss{#2}(#1)}
\def\incons{\mathsf{incons}}
\def\lenrulelist[#1]{\lvert #1 \rvert}

\section{Detailed Pseudo-Codes}\label{app:detailed_pseudocodes}

The pseudo-codes of our modified versions of HybridCORELSPre and HybridCORELSPost are provided in Algorithms~\ref{alg:hybridcorels_pre} and~\ref{alg:hybridcorels_post}, with our modifications highlighted in \highlighted{blue}. In a nutshell, in the branch-and-bound procedure, we additionally verify that the corresponding prefix satisfies the maximum ICD constraint (using Equation~\eqref{eq:icd_computation_rulebased}) before updating the current best solution. This verification is performed along with HybridCORELS' original verification of the overall coverage, at line~7
of both HybridCORELSPre and HybridCORELSPost.

Note that both HybridCORELS algorithms return prefixes (i.e., the interpretable part of the hybrid interpretable model, with the black-box part being trained before or after running the algorithm, in an agnostic manner). They take as input an initial best known prefix $\prefix^0$ satisfying the transparency constraint.
A simple choice for the initial prefix $\prefix^0$ satisfying the transparency constraint is a constant majority prediction: $\prefix^0 \gets [(True \to \oneconseq{}_0)]$ (whose transparency is $1.0$).
As was done by~\citet{HybridCorel_Julien}, we use this trivial initial solution in all our experiments.

Observe that only HybridCORELSPost requires access to the pre-trained black-box $\blackbox$, which is part of its inputs, and is used in the computation of the objective function.

\begin{algorithm*}[t]
\caption{HybridCORELSPre \highlighted{(with ICD mitigation)}}\label{alg:hybridcorels_pre}

\noindent \textbf{Input}: Training data $\dataset{}$ with
set of pre-mined antecedents $\minedrules$; 
minimum transparency value $\mintransp$;
initial prefix $\prefix^0$ such that $\frac{\lvert \dataset_{\prefix^0} \rvert}{\lvert \dataset \rvert} \geq \mintransp$;
\highlighted{set of protected groups $\mathcal{P}$;
maximum coverage disparity $\eta$}
\hspace*{\algorithmicindent} 

\textbf{Output}: 
$(\prefix^*, \oneobj{}^*)$ in which $\prefix^*$ is a prefix with the minimum objective function value $\oneobj{}^*$

\begin{algorithmic}[1]
    \State $(\prefix^c, \oneobj{}^c) \gets  (\prefix^0, \oneobj{}^0)$
    \State $\priorityqueue \gets queue(())$     \Comment{Initially the queue contains the empty prefix $()$}
    \While{$\priorityqueue$ not empty} \Comment{Stop when the queue is empty}
        \State $\prefix{} \gets \priorityqueue.pop()$
        \If{ \label{line:lb_pre} $\lb(\prefix,\dataset) < \oneobj{}^c$}
            \State $\oneobj{} \gets \frac{\errs[\prefix,\dataset_r] + \incons(\dataset \setminus \dataset_\prefix)}{\lvert \dataset \rvert} + \corelsreg \cdot \lenrulelist[\prefix] + \beta \cdot \frac{\lvert \dataset \setminus \dataset_r \rvert}{\lvert \dataset \rvert}$\label{line:objective_pre_eval}\Comment{Compute objective $\obj_\text{pre}(\prefix,\dataset)$}
           \ \If{$\oneobj{} < \oneobj{}^c$ and $\frac{\lvert \dataset_{\prefix} \rvert}{\lvert \dataset \rvert} \geq \mintransp$ \highlighted{and $\max_{(p,p')\in\mathcal{P}^2}{\left(\frac{\lvert \dataset_{\prefix,p} \rvert}{\lvert \dataset_p \rvert}-\frac{\lvert \dataset_{\prefix,p'} \rvert}{\lvert \dataset_{p'} \rvert}\right)} \leq \eta$}}\label{line:check_constraint_pre}
                \State $(\prefix^c, \oneobj{}^c) \gets (\prefix, \oneobj{})$  \label{algl:bestUpdated_pre}\Comment{Update best prefix and objective}
            \EndIf

        \For {$\oneant{}$ in $\minedrules \setminus \{ a_i \mid \exists \singlerule_i \in \prefix{}, \singlerule_i = \oneant_i \to \oneconseq_i \}$} \Comment{Antecedent $\oneant{}$ not involved in $\prefix{}$}
            \State $\singlerule_{new} \gets (\oneant{} \to \oneconseq{})$ \Comment{Set $\oneant{}$'s consequent $\oneconseq{}$ to minimize training error}
             \State $\priorityqueue.push(\prefix{} \cup \singlerule_{new})$ \Comment{Enqueue extension of $\prefix{}$ with new rule $\singlerule_{new}$}
        \EndFor
        
    \EndIf
\EndWhile
\State $(\prefix^*, \oneobj{}^*) \gets (\prefix^c, \oneobj{}^c)$
\end{algorithmic}
\end{algorithm*}

\begin{algorithm*}[t]
\caption{HybridCORELSPost \highlighted{(with ICD mitigation)}}\label{alg:hybridcorels_post}

\noindent \textbf{Input}: Training data $\dataset{}$ with
set of pre-mined antecedents $\minedrules$; 
minimum transparency value $\mintransp$;
initial prefix $\prefix^0$ such that $\frac{\lvert \dataset_{\prefix^0} \rvert}{\lvert \dataset \rvert} \geq \mintransp$;
pre-trained black-box model $\blackbox$;
\highlighted{set of protected groups $\mathcal{P}$;
maximum coverage disparity $\eta$}
\hspace*{\algorithmicindent} 

\textbf{Output}: 
$(\prefix^*, \oneobj{}^*)$ in which $\prefix^*$ is a prefix with the minimum objective function value $\oneobj{}^*$

\begin{algorithmic}[1]
    \State $(\prefix^c, \oneobj{}^c) \gets  (\prefix^0, \oneobj{}^0)$
    \State $\priorityqueue \gets queue(())$     \Comment{Initially the queue contains the empty prefix $()$}
    \While{$\priorityqueue$ not empty} \Comment{Stop when the queue is empty}
        \State $\prefix{} \gets \priorityqueue.pop()$
        \If{ \label{line:lb_post} $\lb(\prefix,\dataset) < \oneobj{}^c$}
            \State $\oneobj{} \gets \frac{\errs[\prefix, \dataset_\prefix]+\errs[h_c, \dataset \setminus \dataset_\prefix]}{\lvert \dataset \rvert} + \corelsreg \cdot \lenrulelist[\prefix] + \regcoeffhycorels \cdot \frac{\lvert \dataset \setminus \dataset_\prefix \rvert}{\lvert \dataset \rvert}$\label{line:objective_post_eval}\Comment{Compute objective $\obj_\text{post}(\prefix,\dataset)$}
           \ \If{$\oneobj{} < \oneobj{}^c$ and $\frac{\lvert \dataset_{\prefix} \rvert}{\lvert \dataset \rvert} \geq \mintransp$ \highlighted{and $\max_{(p,p')\in\mathcal{P}^2}{\left(\frac{\lvert \dataset_{\prefix,p} \rvert}{\lvert \dataset_p \rvert}-\frac{\lvert \dataset_{\prefix,p'} \rvert}{\lvert \dataset_{p'} \rvert}\right)}\leq \eta$}}\label{line:check_constraint_post}
                \State $(\prefix^c, \oneobj{}^c) \gets (\prefix, \oneobj{})$  \label{algl:bestUpdated_post}\Comment{Update best prefix and objective}
            \EndIf

        \For {$\oneant{}$ in $\minedrules \setminus \{ a_i \mid \exists \singlerule_i \in \prefix{}, \singlerule_i = \oneant_i \to \oneconseq_i \}$} \Comment{Antecedent $\oneant{}$ not involved in $\prefix{}$}
            \State $\singlerule_{new} \gets (\oneant{} \to \oneconseq{})$ \Comment{Set $\oneant{}$'s consequent $\oneconseq{}$ to minimize training error}
             \State $\priorityqueue.push(\prefix{} \cup \singlerule_{new})$ \Comment{Enqueue extension of $\prefix{}$ with new rule $\singlerule_{new}$}
        \EndFor
        
    \EndIf
\EndWhile
\State $(\prefix^*, \oneobj{}^*) \gets (\prefix^c, \oneobj{}^c)$
\end{algorithmic}
\end{algorithm*}

\section {Approximate Rashomon Sets Analysis} \label{sec:appendix_boots}

\begin{figure*}[h!]
\centering
\includegraphics[width=0.5\textwidth]{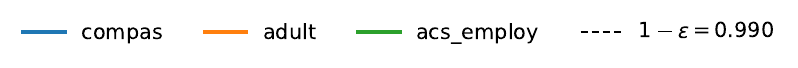}

\begin{subfigure}{0.35\textwidth}
    \centering
    \includegraphics[width=\linewidth]{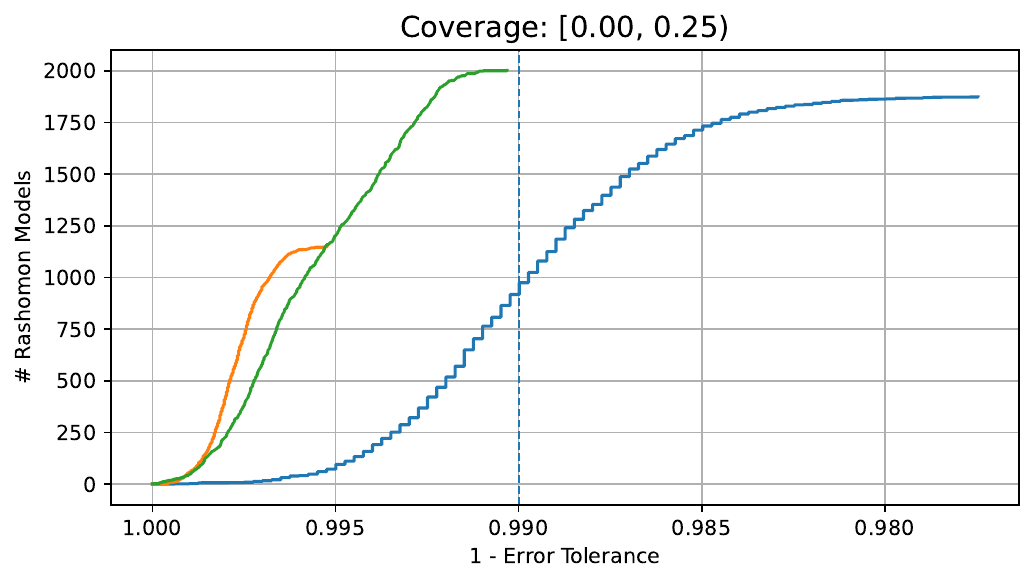}
    \caption{$Q_1$}
\end{subfigure}
\begin{subfigure}{0.35\textwidth}
    \centering
    \includegraphics[width=\linewidth]{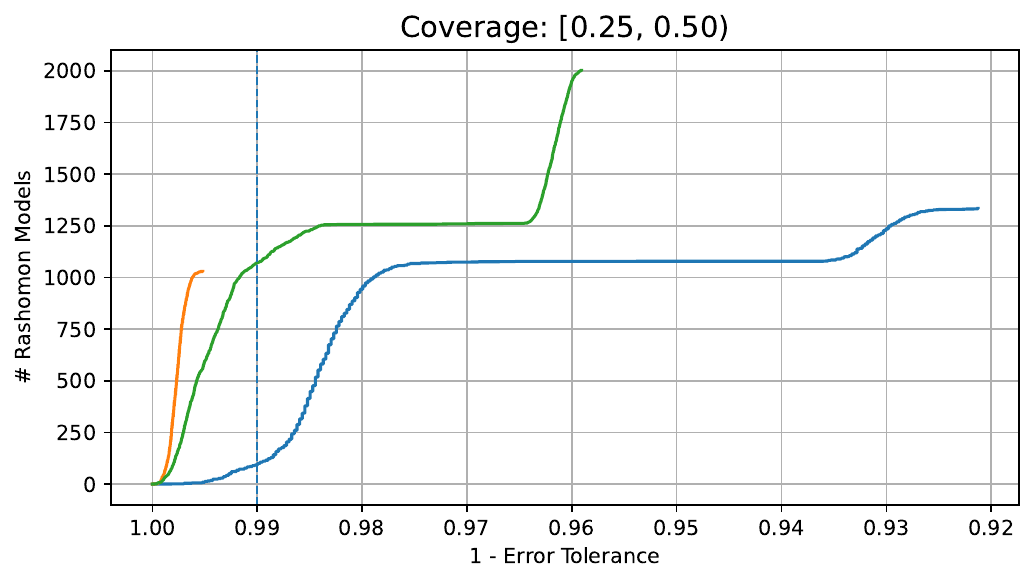}
    \caption{$Q_2$}
\end{subfigure}

\vspace{0.5em} %

\begin{subfigure}{0.35\textwidth}
    \centering
    \includegraphics[width=\linewidth]{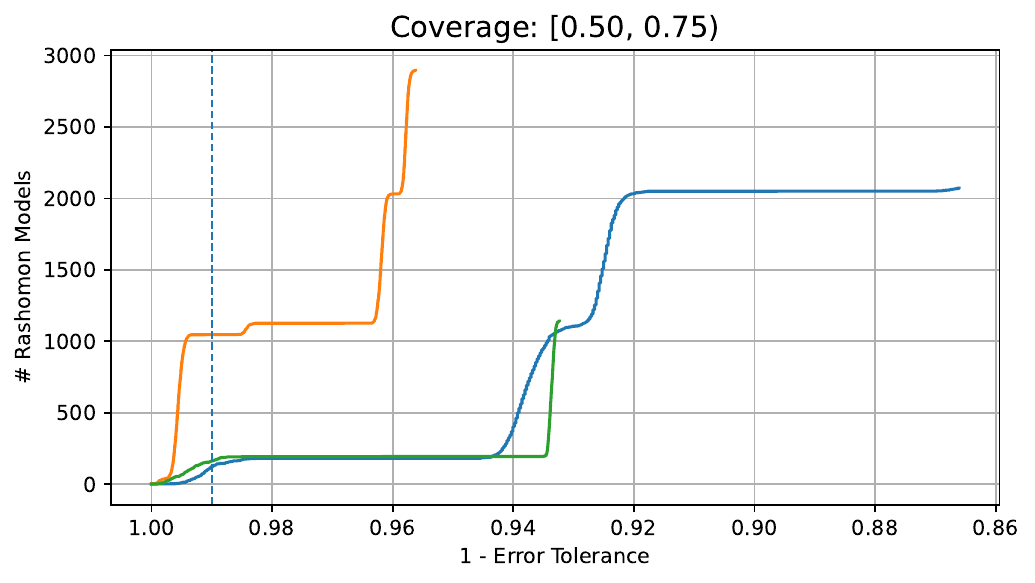}
    \caption{$Q_3$}
\end{subfigure}
\begin{subfigure}{0.35\textwidth}
    \centering
    \includegraphics[width=\linewidth]{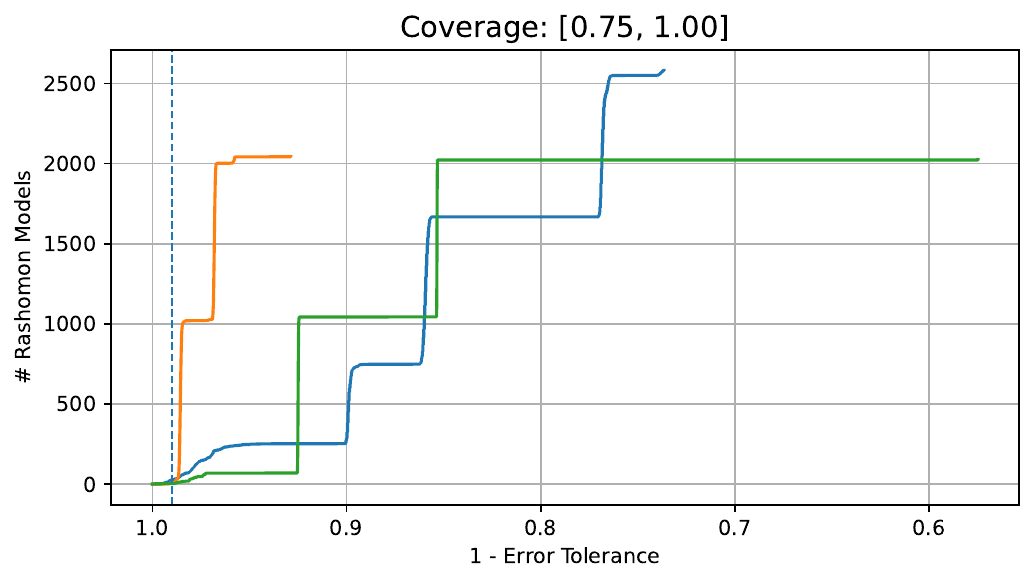}
    \caption{$Q_4$}
\end{subfigure}

\caption{Approximate Rashomon set growth across all transparency bins for HybridCORELSPost: number of unique models with accuracy $\geq (1 - \varepsilon)$ times maximum accuracy. The dashed vertical line shows the Rashomon bound at $\varepsilon = 0.01$. Results are provided for all datasets.}
\label{Fig:RS_growth_HybridCORELSPost}

\end{figure*}

\begin{figure*}[h!]
\centering
\includegraphics[width=0.5\textwidth]{Plots/RS/RS_shared_legend.pdf}

\begin{subfigure}{0.35\textwidth}
    \centering
    \includegraphics[width=\linewidth]{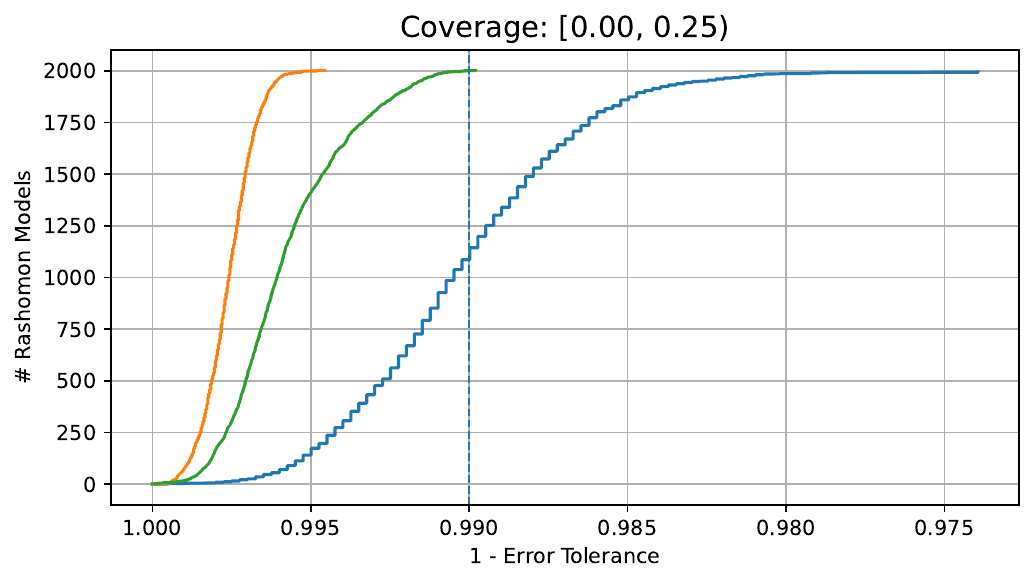}
    \caption{$Q_1$}
\end{subfigure}
\hspace{0.01\textwidth}
\begin{subfigure}{0.35\textwidth}
    \centering
    \includegraphics[width=\linewidth]{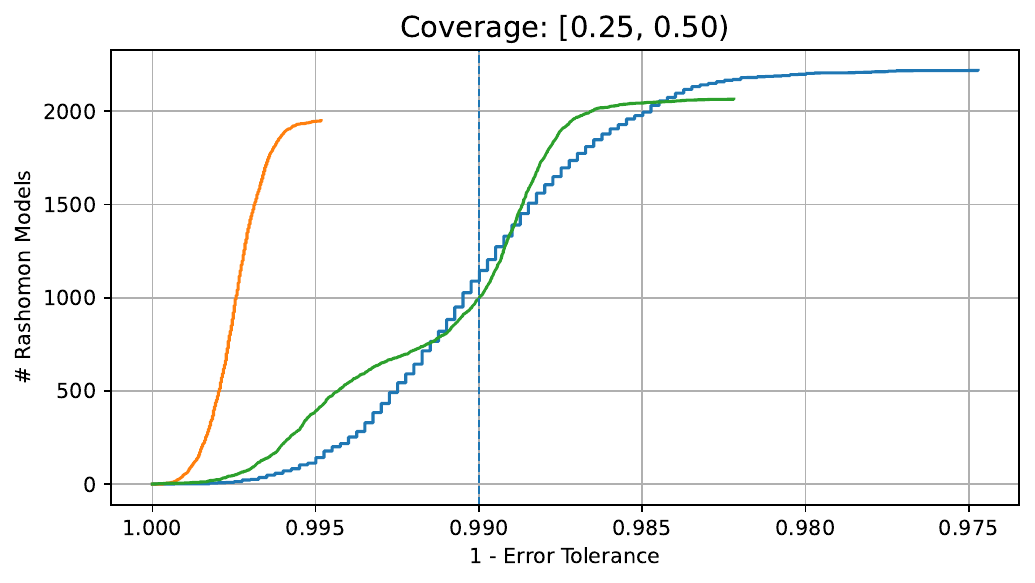}
    \caption{$Q_2$}
\end{subfigure}

\vspace{0.5em} %

\begin{subfigure}{0.35\textwidth}
    \centering
    \includegraphics[width=\linewidth]{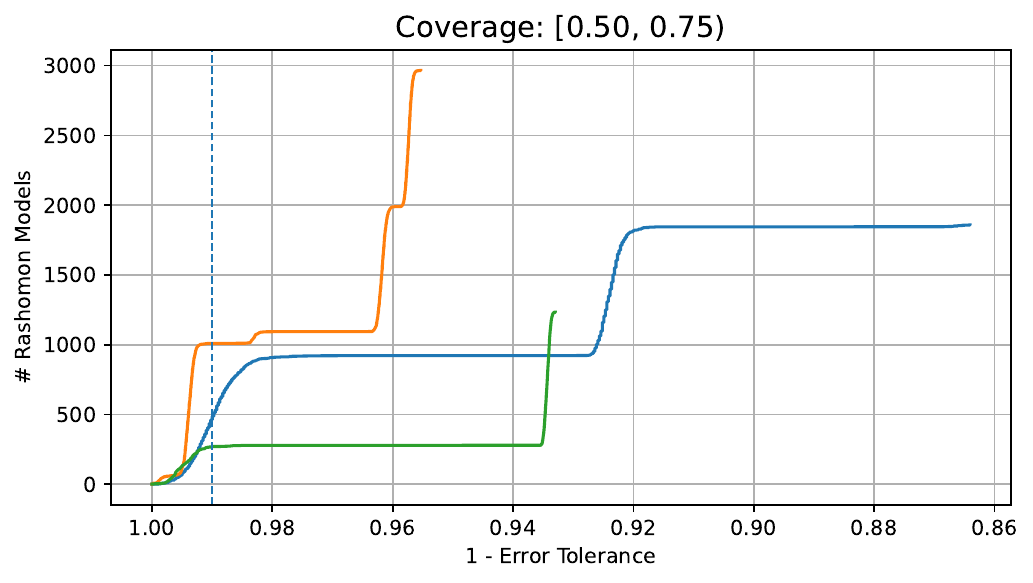}
    \caption{$Q_3$}
\end{subfigure}
\hspace{0.01\textwidth}
\begin{subfigure}{0.35\textwidth}
    \centering
    \includegraphics[width=\linewidth]{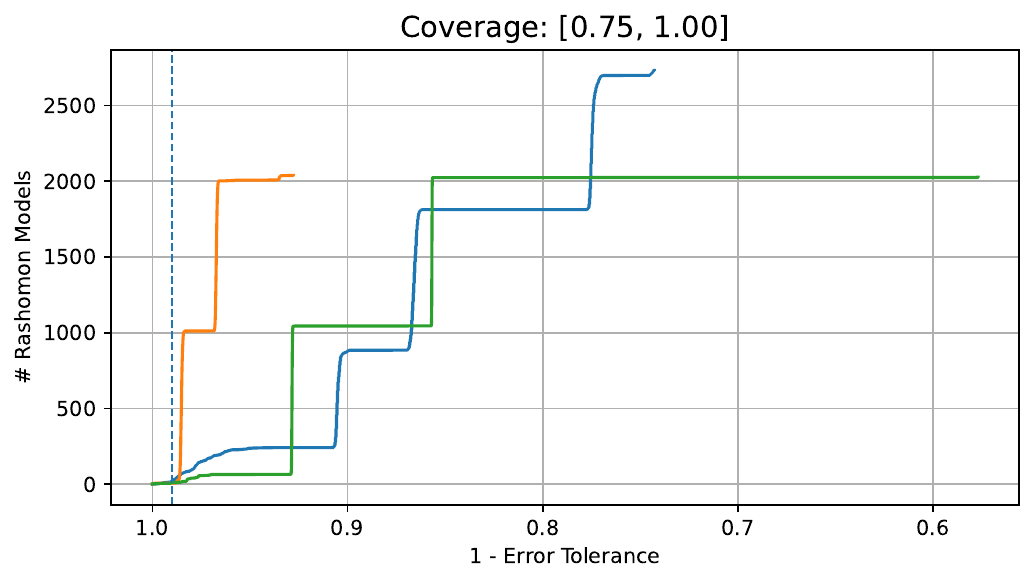}
    \caption{$Q_4$}
\end{subfigure}
\caption{Approximate Rashomon set growth across all transparency bins for HybridCORELSPre: number of unique models with accuracy $\geq (1 - \varepsilon)$ times maximum accuracy. The dashed vertical line shows the Rashomon bound at $\varepsilon = 0.01$. Results are provided for all datasets.}
\label{RS_growth_HybridCORELSPre}
\end{figure*}

\begin{figure*}[h!]
\centering
\includegraphics[width=0.5\textwidth]{Plots/RS/RS_shared_legend.pdf}

\begin{subfigure}{0.35\textwidth}
    \centering
    \includegraphics[width=\linewidth]{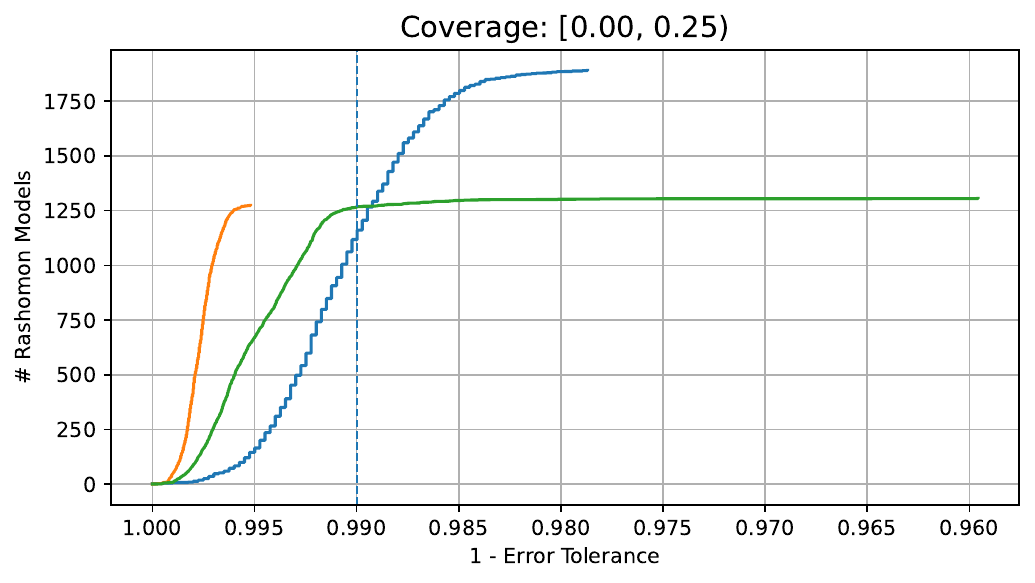}
    \caption{$Q_1$}
\end{subfigure}
\begin{subfigure}{0.35\textwidth}
    \centering
    \includegraphics[width=\linewidth]{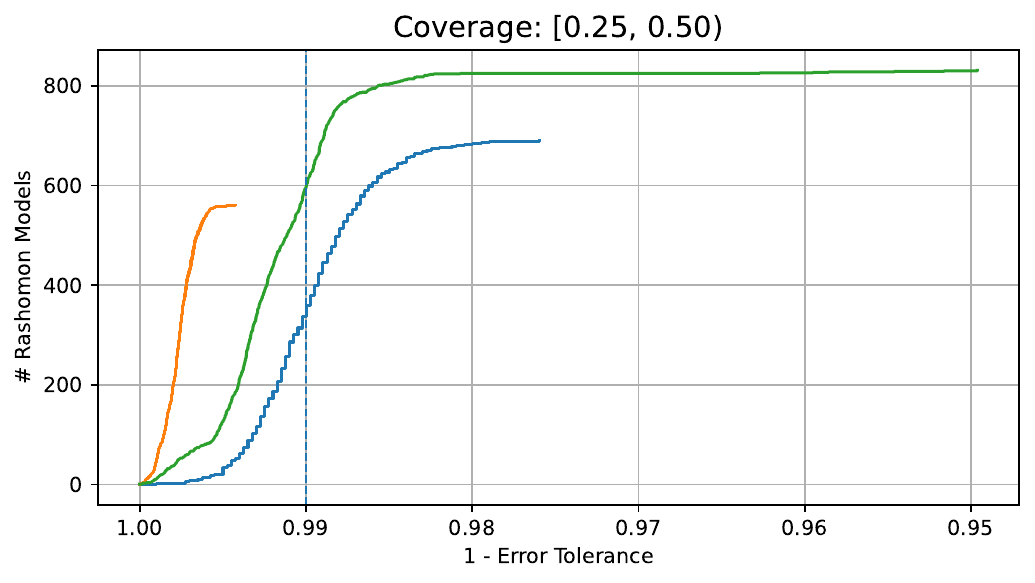}
    \caption{$Q_2$}
\end{subfigure}

\vspace{0.5em} %

\begin{subfigure}{0.35\textwidth}
    \centering
    \includegraphics[width=\linewidth]{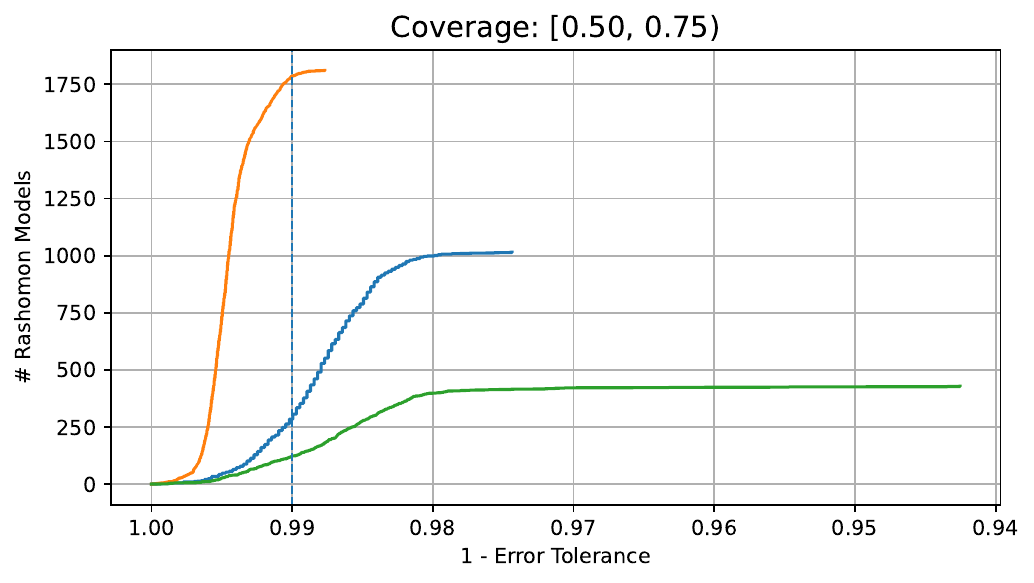}
    \caption{$Q_3$}
\end{subfigure}
\begin{subfigure}{0.35\textwidth}
    \centering
    \includegraphics[width=\linewidth]{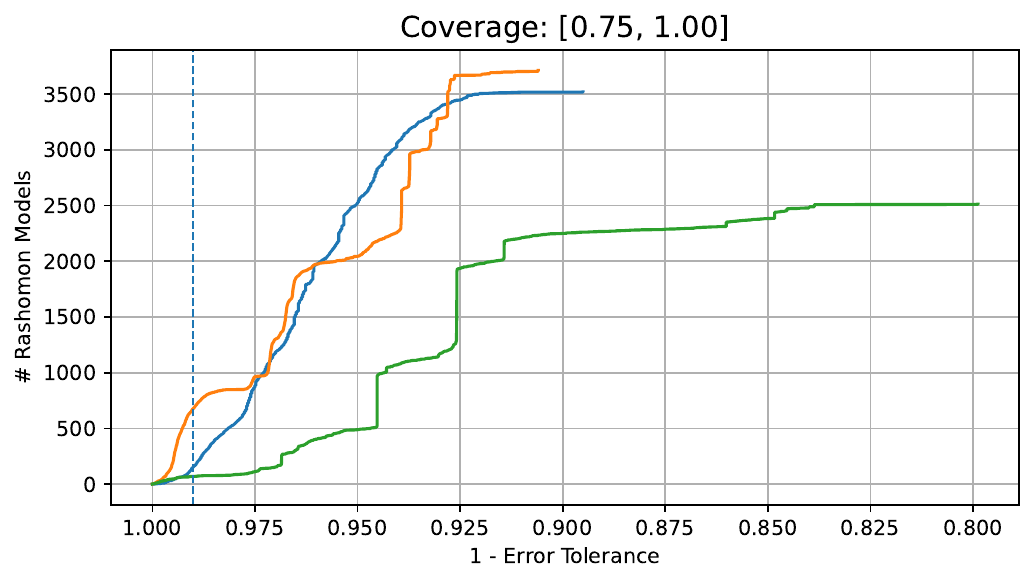}
    \caption{$Q_4$}
\end{subfigure}
\caption{Approximate Rashomon set growth across all transparency bins for HyRS: number of unique models with accuracy $\geq (1 - \varepsilon)$ times maximum accuracy. The dashed vertical line shows the Rashomon bound at $\varepsilon = 0.01$. Results are provided for all datasets.}
\label{Fig:RS_growth_HyRS}

\end{figure*}

\begin{figure*}[h!]
\centering
\includegraphics[width=0.5\textwidth]{Plots/RS/RS_shared_legend.pdf}

\begin{subfigure}{0.35\textwidth}
    \centering
    \includegraphics[width=\linewidth]{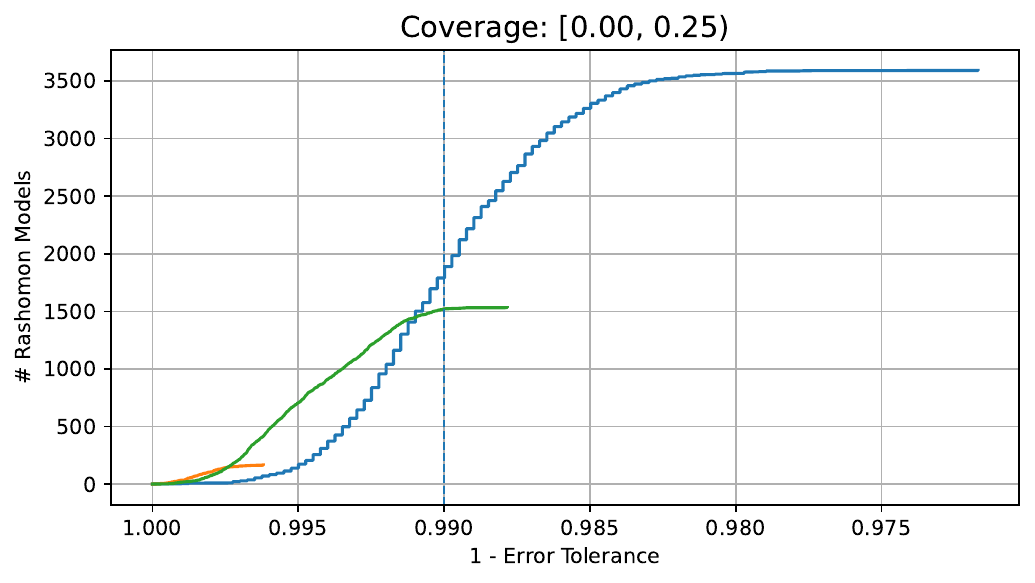}
    \caption{$Q_1$}
\end{subfigure}
\begin{subfigure}{0.35\textwidth}
    \centering
    \includegraphics[width=\linewidth]{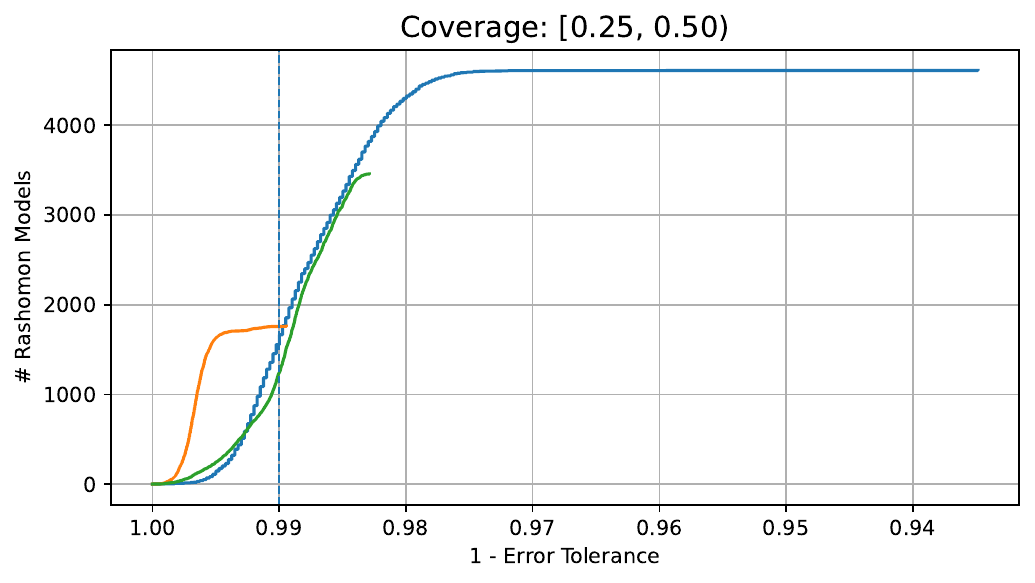}
    \caption{$Q_2$}
\end{subfigure}

\vspace{0.5em} %

\begin{subfigure}{0.35\textwidth}
    \centering
    \includegraphics[width=\linewidth]{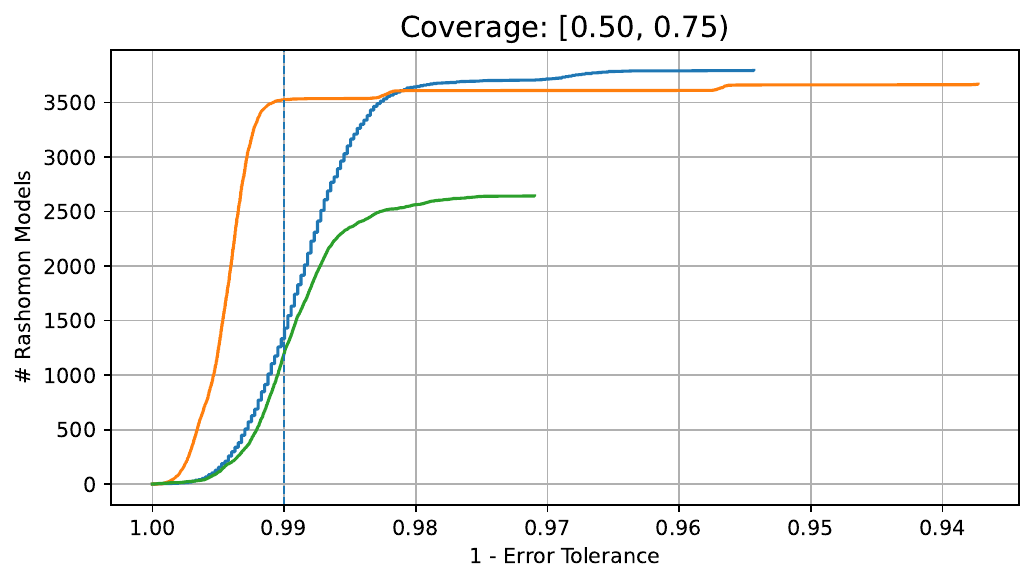}
    \caption{$Q_3$}
\end{subfigure}
\begin{subfigure}{0.35\textwidth}
    \centering
    \includegraphics[width=\linewidth]{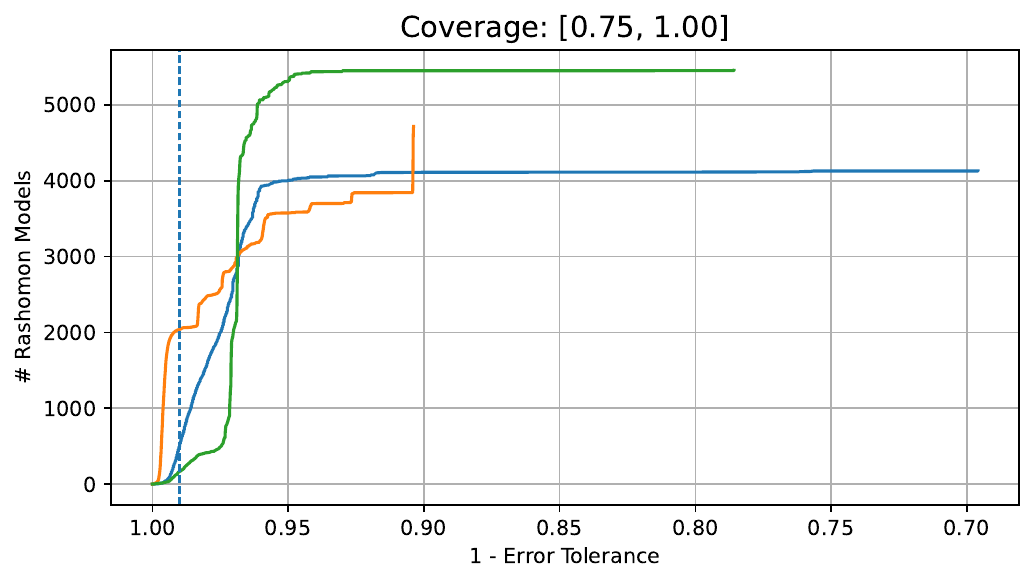}
    \caption{$Q_4$}
\end{subfigure}
\caption{Approximate Rashomon set growth across all transparency bins for CRL: number of unique models with accuracy $\geq (1 - \varepsilon)$ times maximum accuracy. The dashed vertical line shows the Rashomon bound at $\varepsilon = 0.01$. Results are provided for all datasets.}
\label{Fig:RS_growth_CRL}

\end{figure*}

To approximate the Rashomon sets of hybrid interpretable models, we generate collections of near-optimal models through bootstrap resampling of the training data. Figures \ref{Fig:RS_growth_HybridCORELSPost}--\ref{Fig:RS_growth_CRL} illustrate how the number of unique models contained in the approximate Rashomon set evolves as the Rashomon parameter $\varepsilon$ increases across transparency bins.
The results show that bootstrap sampling provides a sufficiently large and diverse set of near-optimal models to support meaningful statistical analysis. 

Across all bins, the number of models in the approximate Rashomon set increases as the admissible error tolerance grows, since larger values of $\varepsilon$ allow more alternative models to be included within the Rashomon set. After a certain threshold, the curves plateau, indicating that few or no additional unique models are discovered beyond this point. This stabilization suggests that the bootstrap procedure can recover a substantial portion of the practically relevant solution space.

The plots also reveal notable differences across transparency bins. In lower-coverage bins (e.g., $Q_1$), near-optimal models are concentrated within a relatively narrow error range, reflecting the high predictive performance achieved when most instances are deferred to the black-box component. In contrast, higher-coverage bins, particularly $Q_4$, exhibit a substantially wider error range before reaching saturation. This behavior is consistent with the reduced predictive accuracy typically observed in high-transparency regimes, where a larger proportion of instances are handled by the interpretable component. Therefore, the size and growth rate of the approximate Rashomon sets vary across both transparency bins and datasets. In lower-transparency bins and for certain datasets, a large number of near-optimal models emerge even under very small error tolerances, indicating a dense concentration of high-performing alternative solutions. In contrast, other bins and datasets require larger error tolerances before comparable model diversity is observed. 

\FloatBarrier
\section{Additional Experimental Results for Section~\ref{sec:ICD_results}} \label{sec:appendix_ICD_results}

\begin{figure*}[h!]
\centering

\begin{subfigure}{0.85\textwidth}
    \centering
    \includegraphics[width=0.32\linewidth]{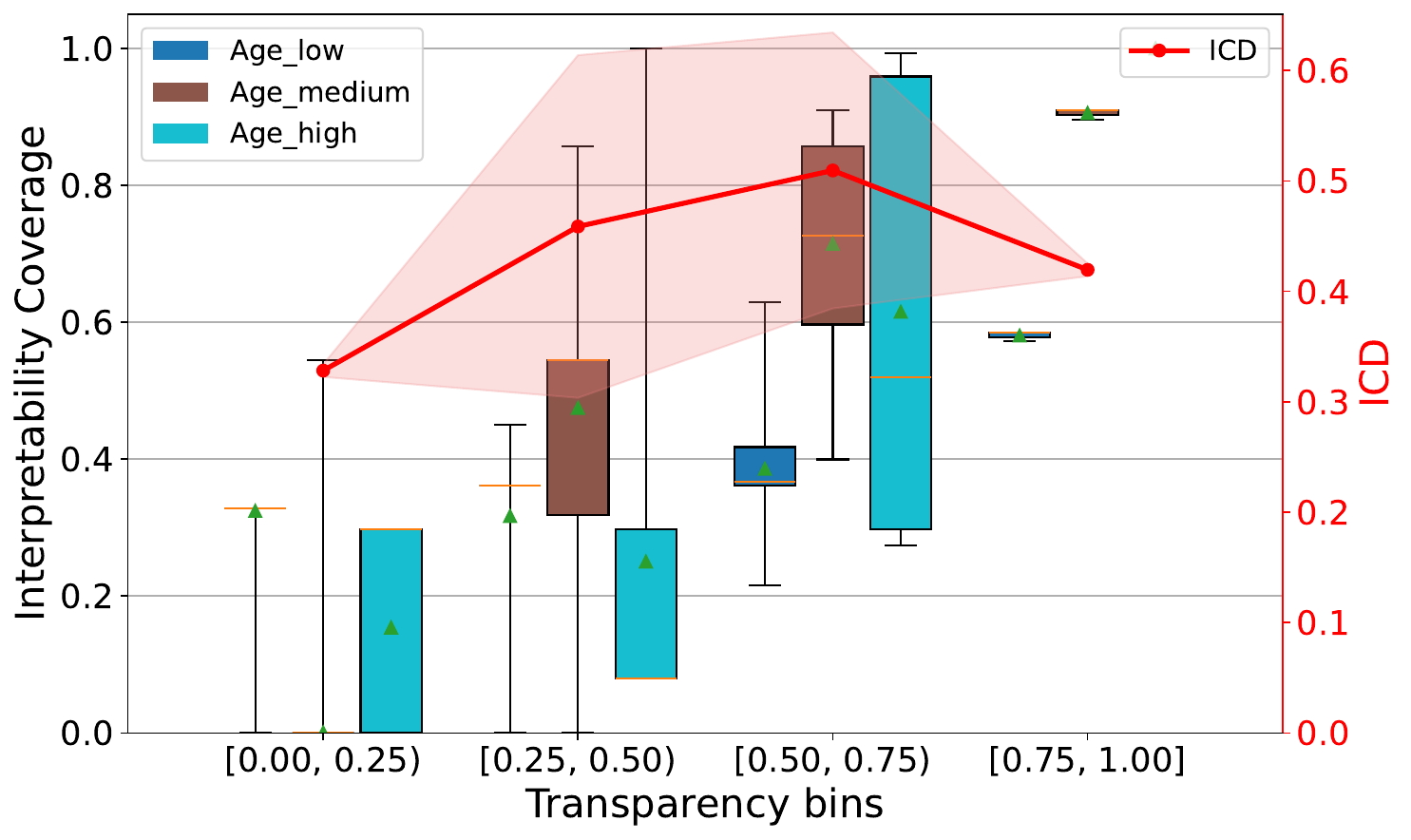}
    \hfill
    \includegraphics[width=0.32\linewidth]{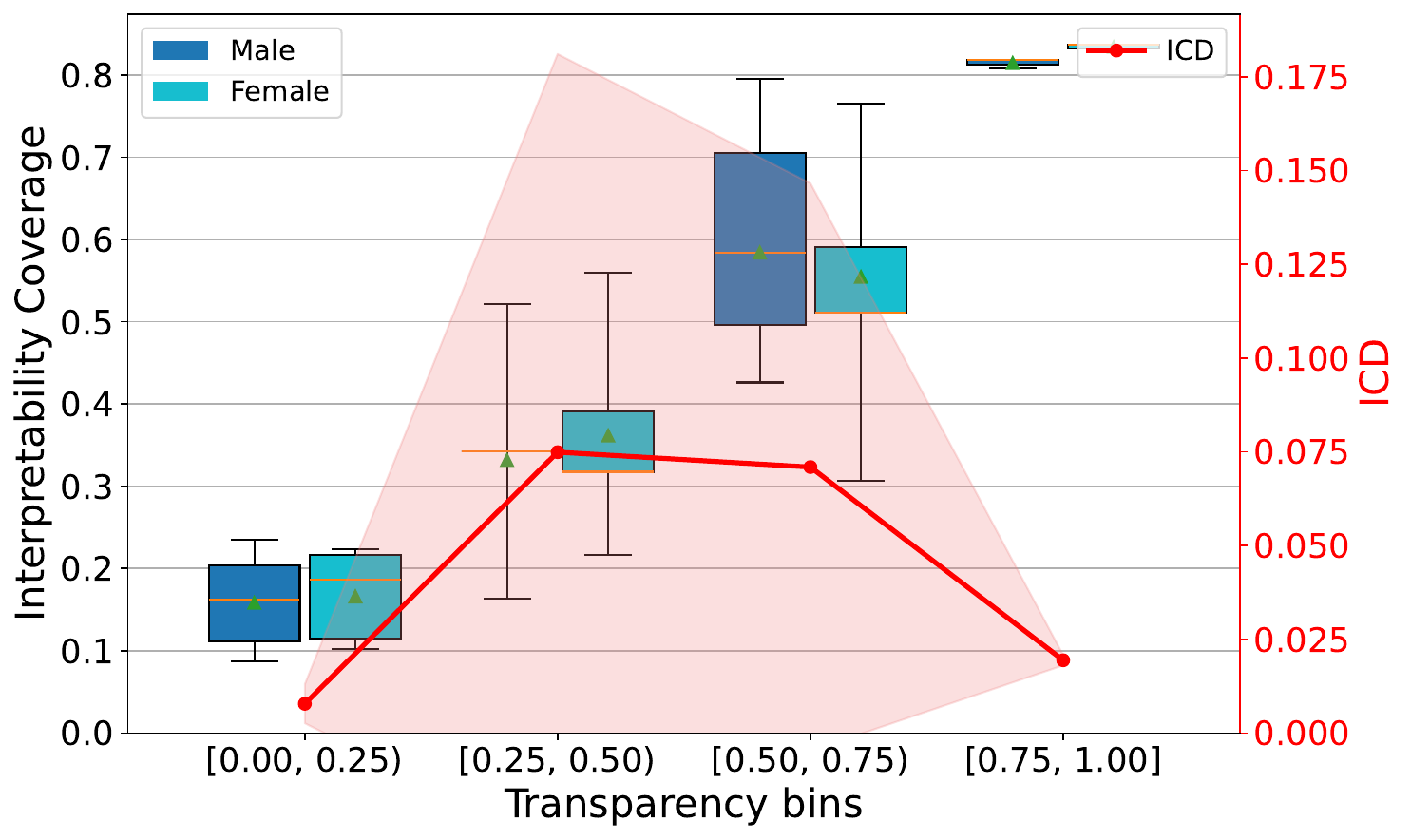}
    \hfill
    \includegraphics[width=0.32\linewidth]{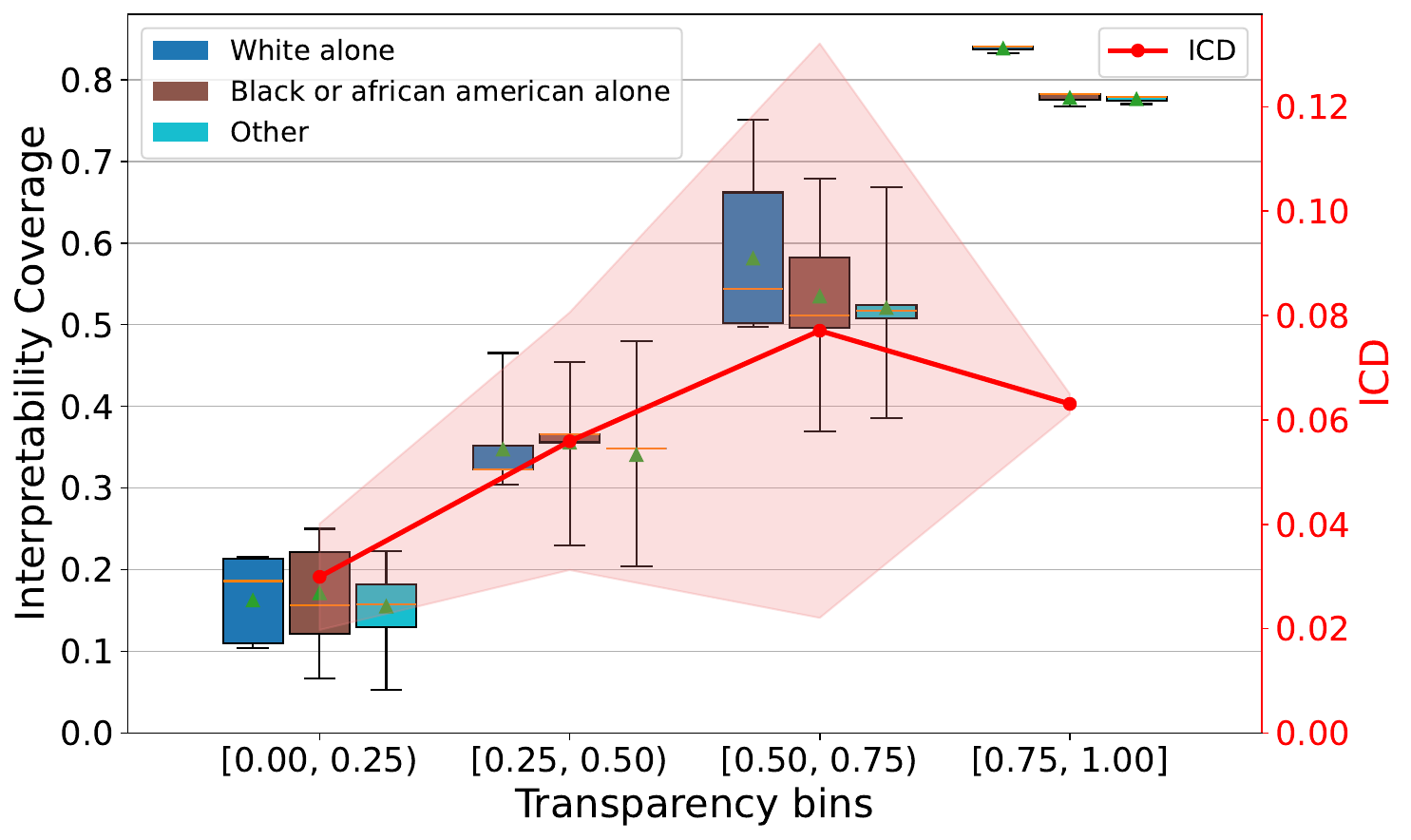}
    \caption{ACS Employment dataset}\label{fig:ICF_plot_post_acs_employ}
\end{subfigure}

\vspace{0.5em}

\begin{subfigure}{0.85\textwidth}
    \centering
    \includegraphics[width=0.32\linewidth]{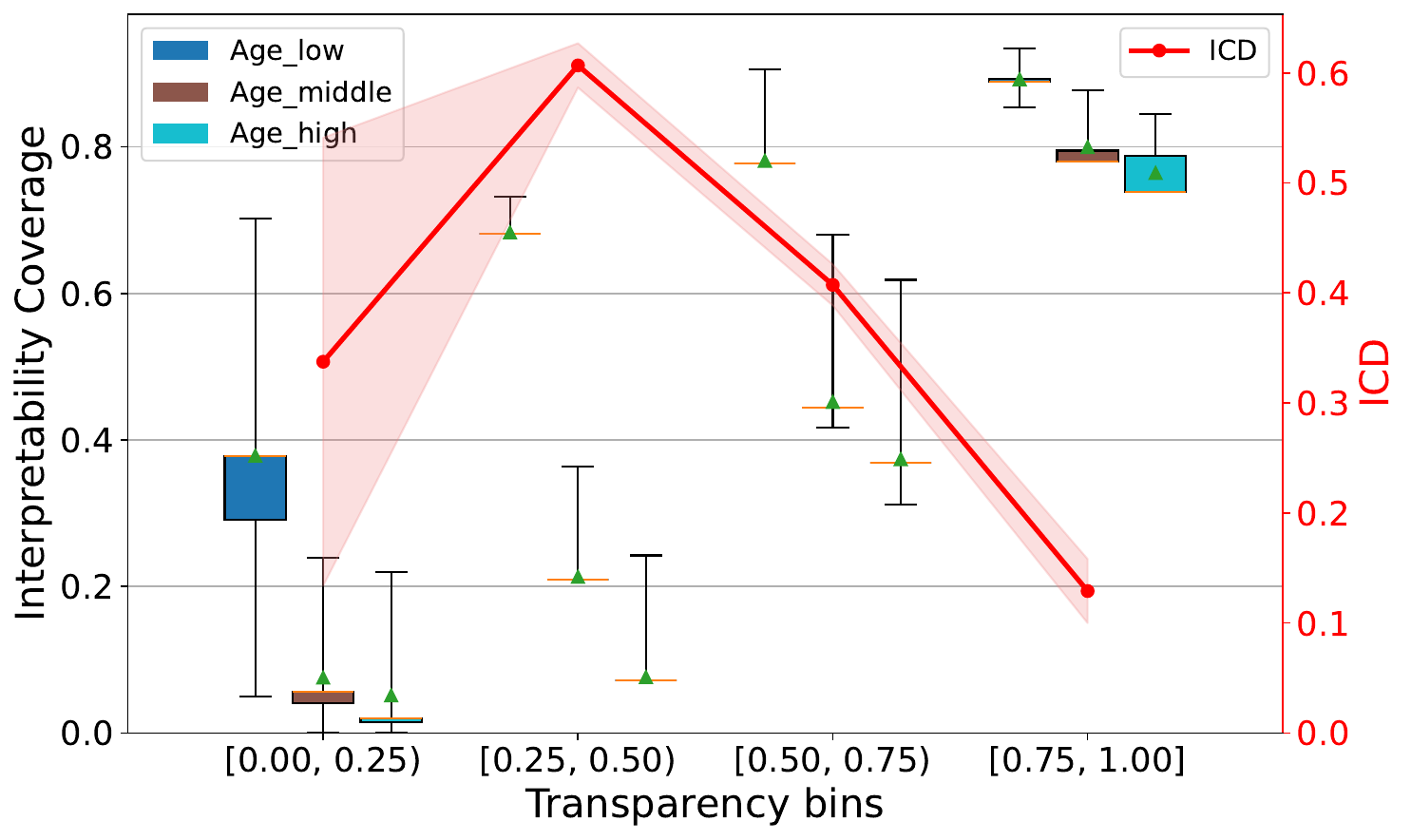}
    \includegraphics[width=0.32\linewidth]{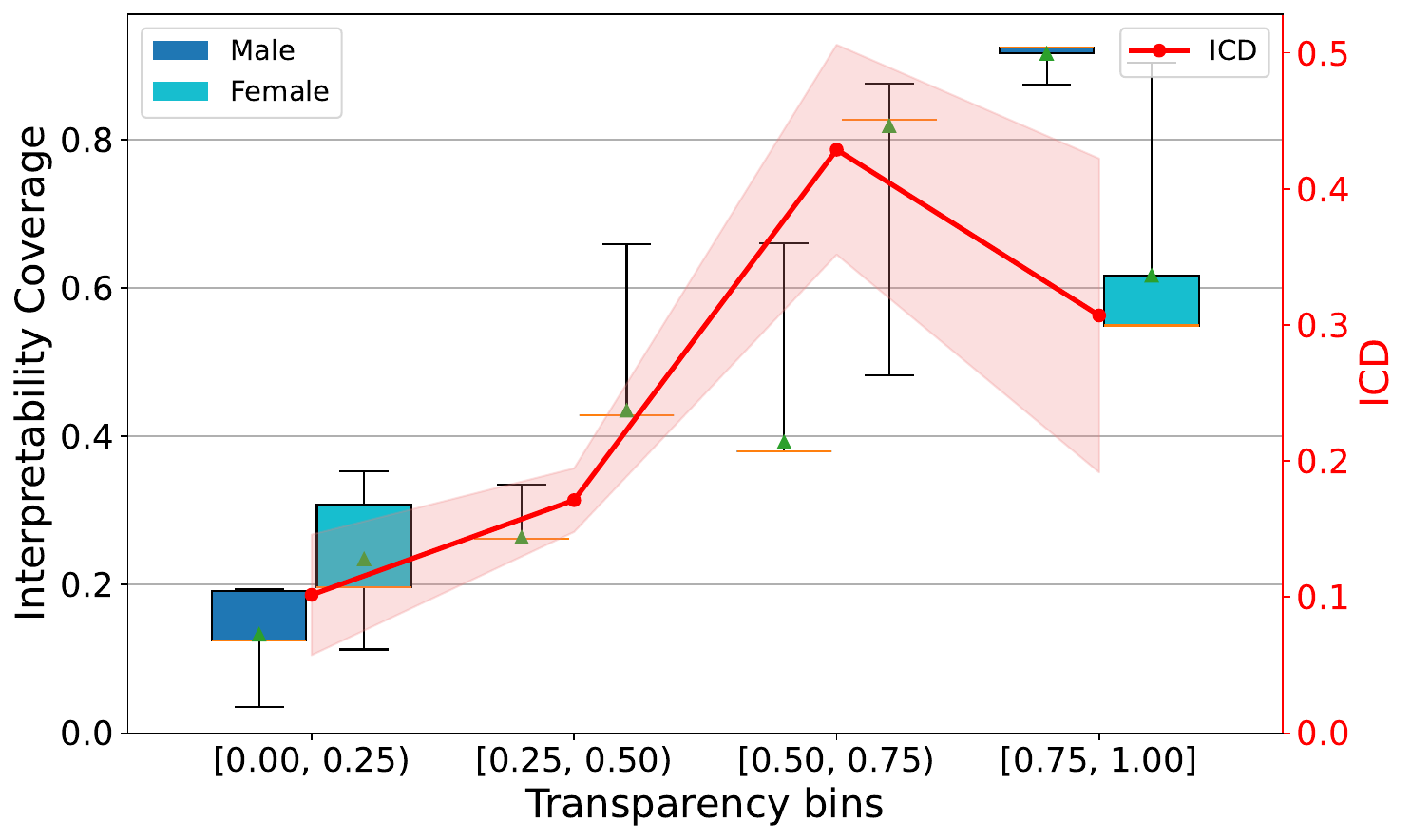}
    \includegraphics[width=0.32\linewidth]{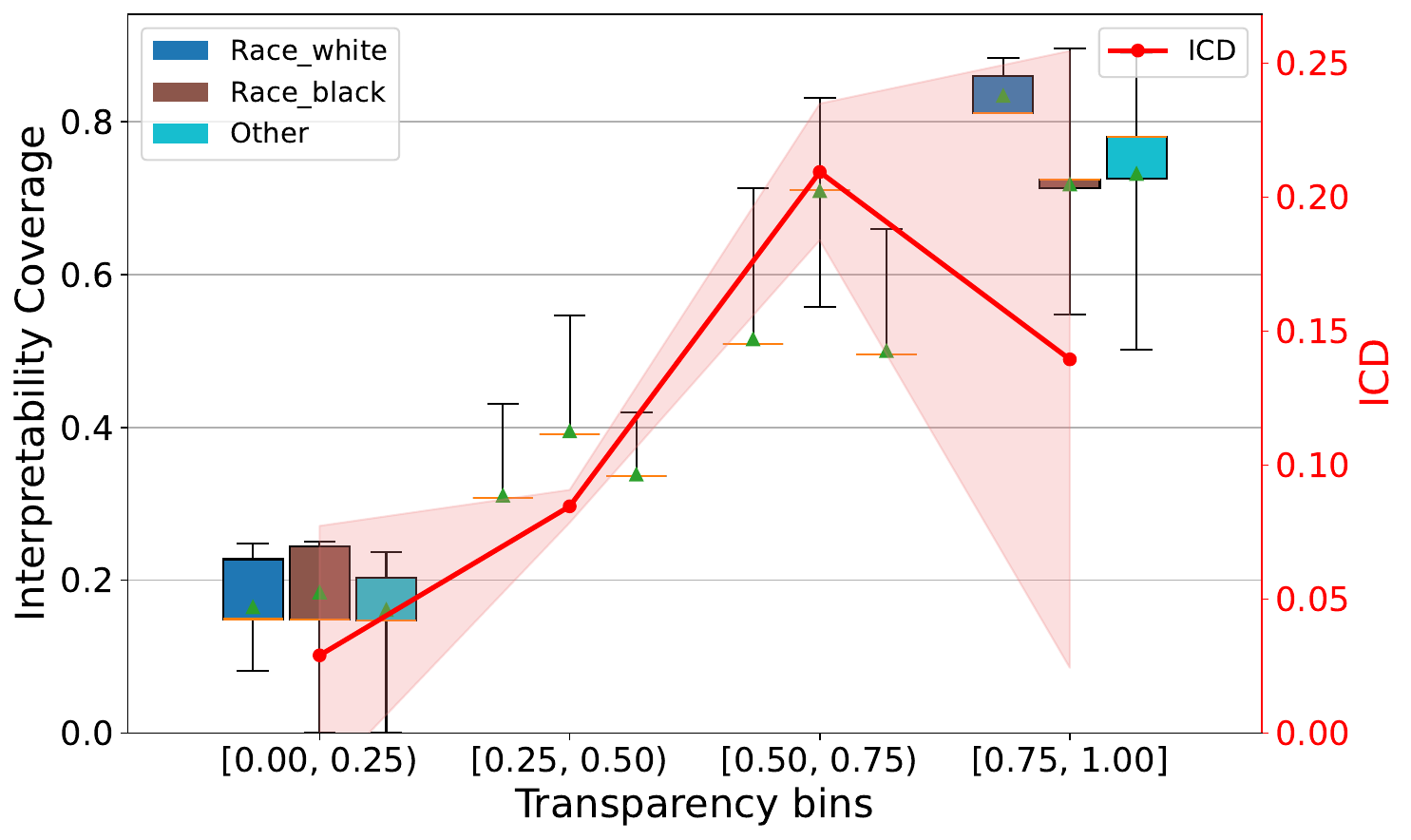}
    \caption{UCI Adult Income dataset}\label{fig:ICF_plot_post_adult}
\end{subfigure}

\vspace{0.5em}

\begin{subfigure}{0.85\textwidth}
    \centering
    \includegraphics[width=0.32\linewidth]{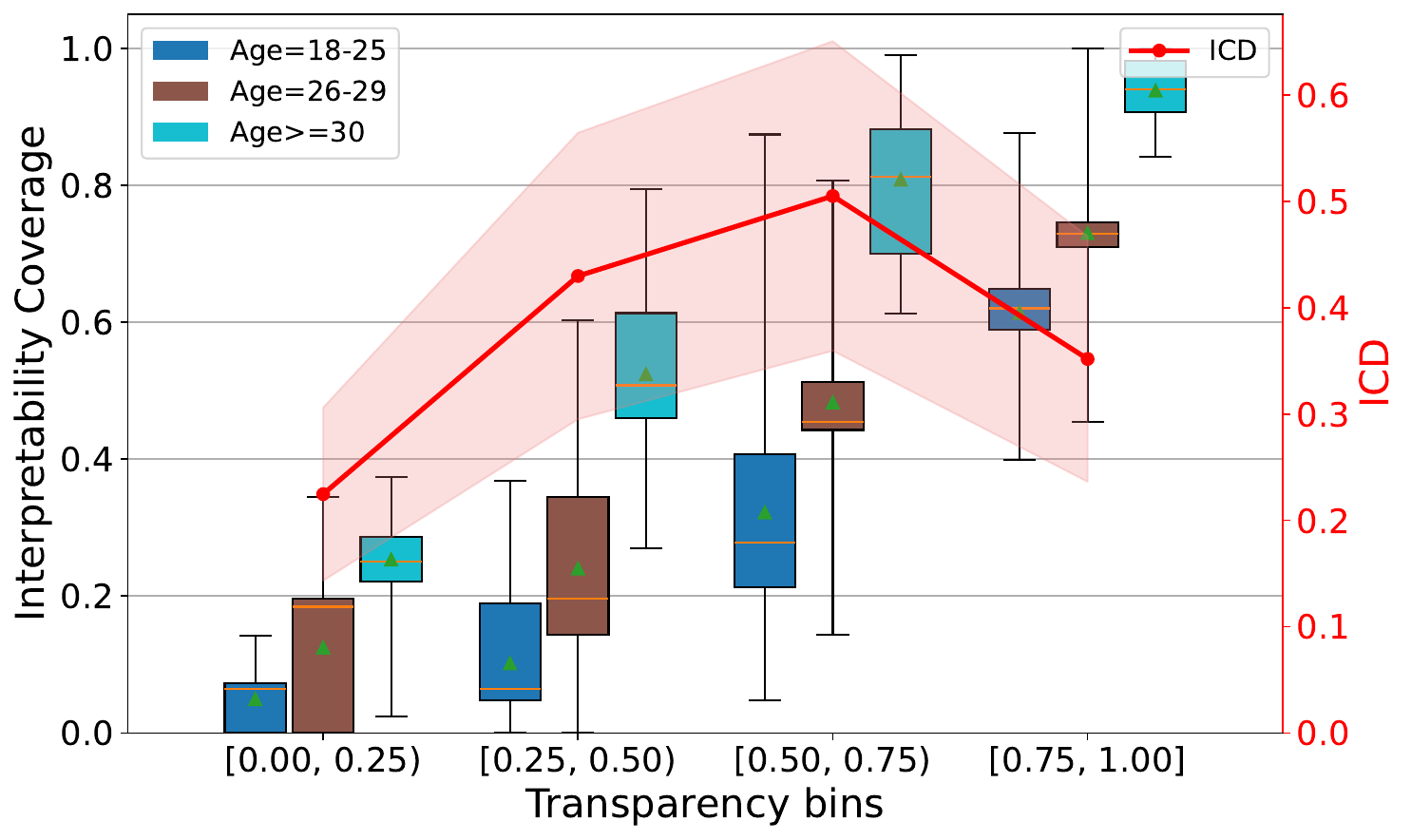}
    \includegraphics[width=0.32\linewidth]{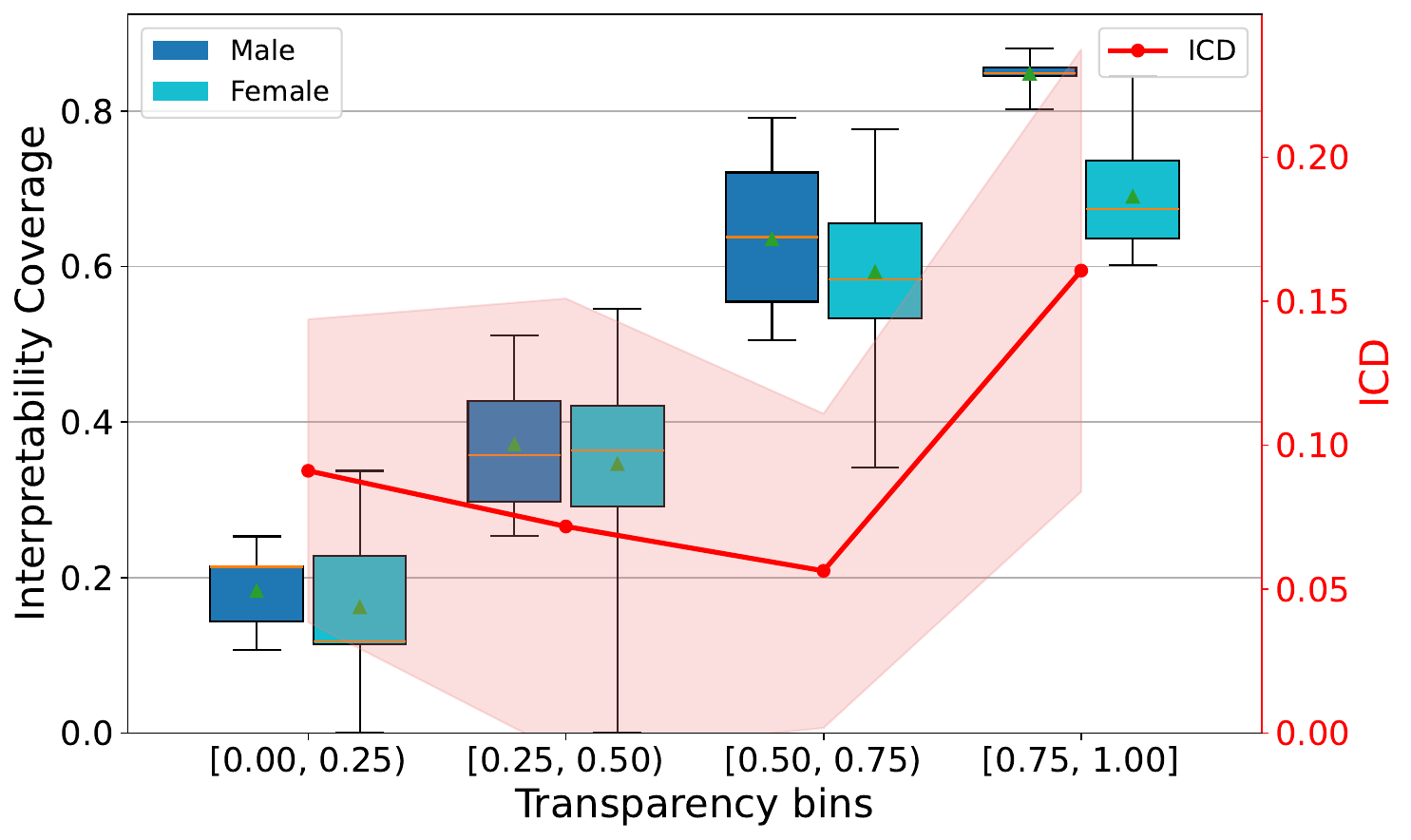}
    \includegraphics[width=0.32\linewidth]{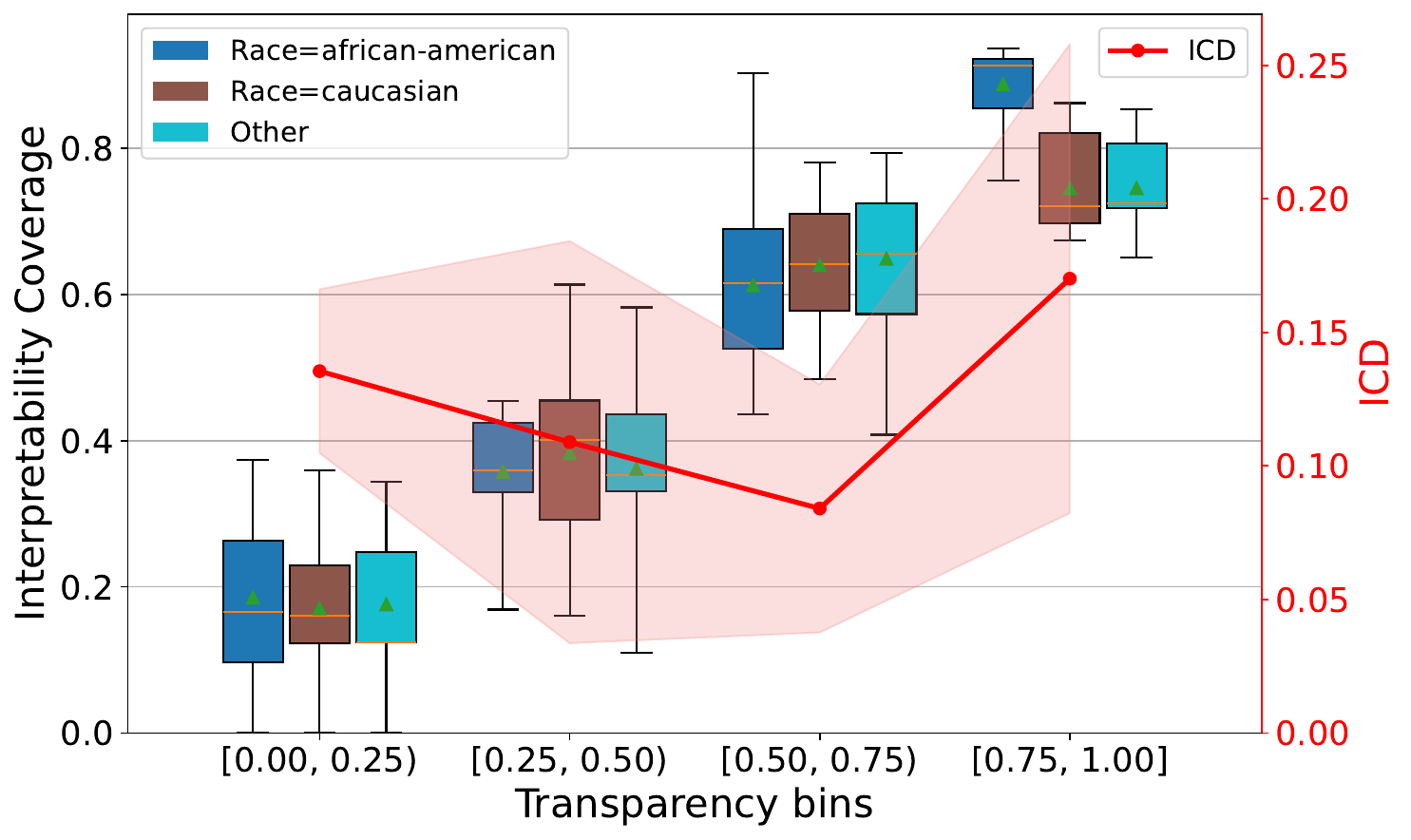}
    \caption{COMPAS dataset}\label{fig:ICF_plot_post_compas}
\end{subfigure}
\caption{Distribution of test set Interpretability Coverage (IC) across Rashomon sets of HybridCORELSPost for all transparency bins $(\varepsilon = 0.01)$. Results are shown for all datasets and sensitive attributes (Age, Gender, Race), ordered from left to right columns. The red curve represents the average ICD across models within each bin, while the shaded region indicates the corresponding standard deviation.} \label{fig:ICF_plot_post}
\end{figure*}

\begin{figure*}[t]
\centering

\begin{subfigure}{0.85\textwidth}
    \centering
    \includegraphics[width=0.32\linewidth]{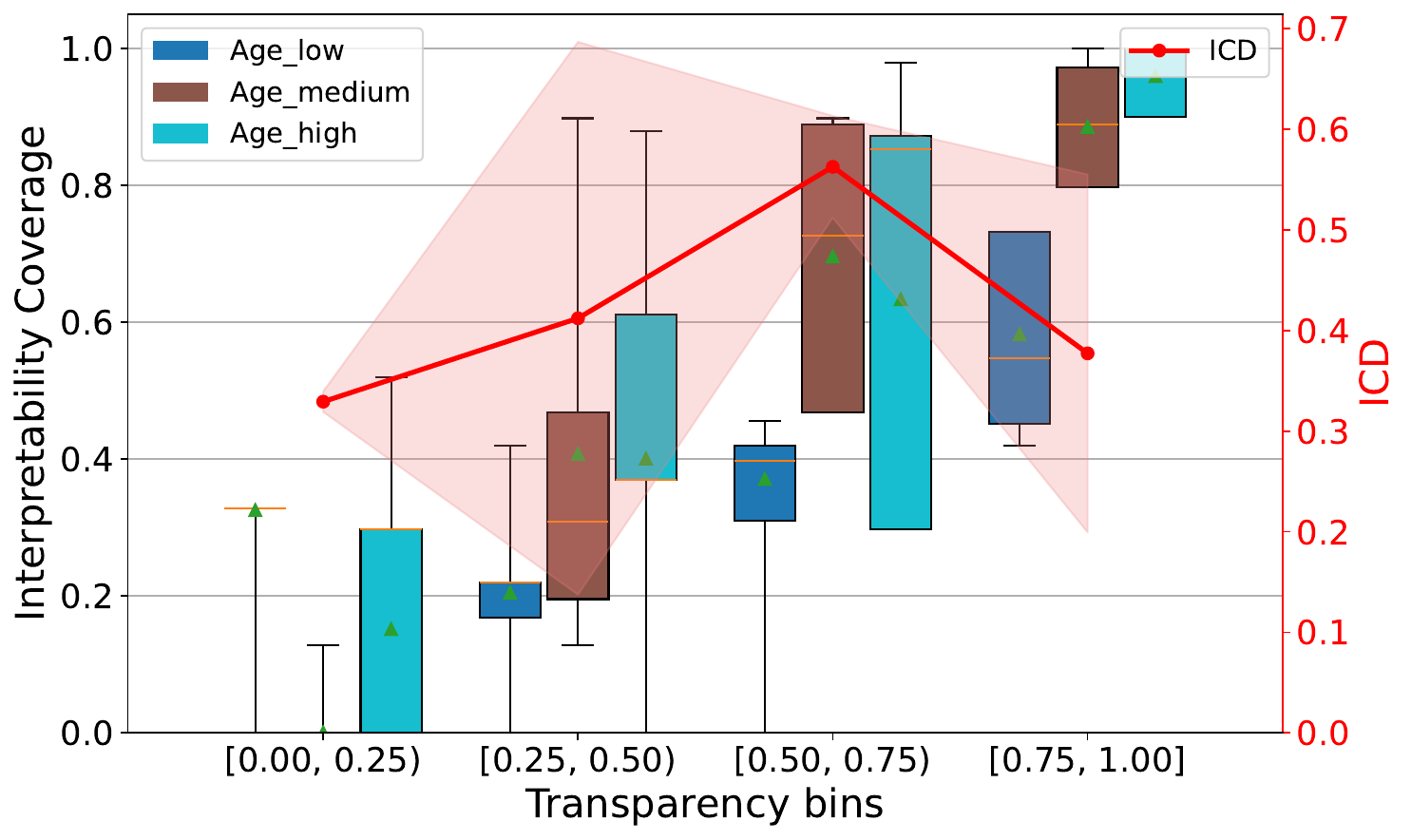}
    \hfill
    \includegraphics[width=0.32\linewidth]{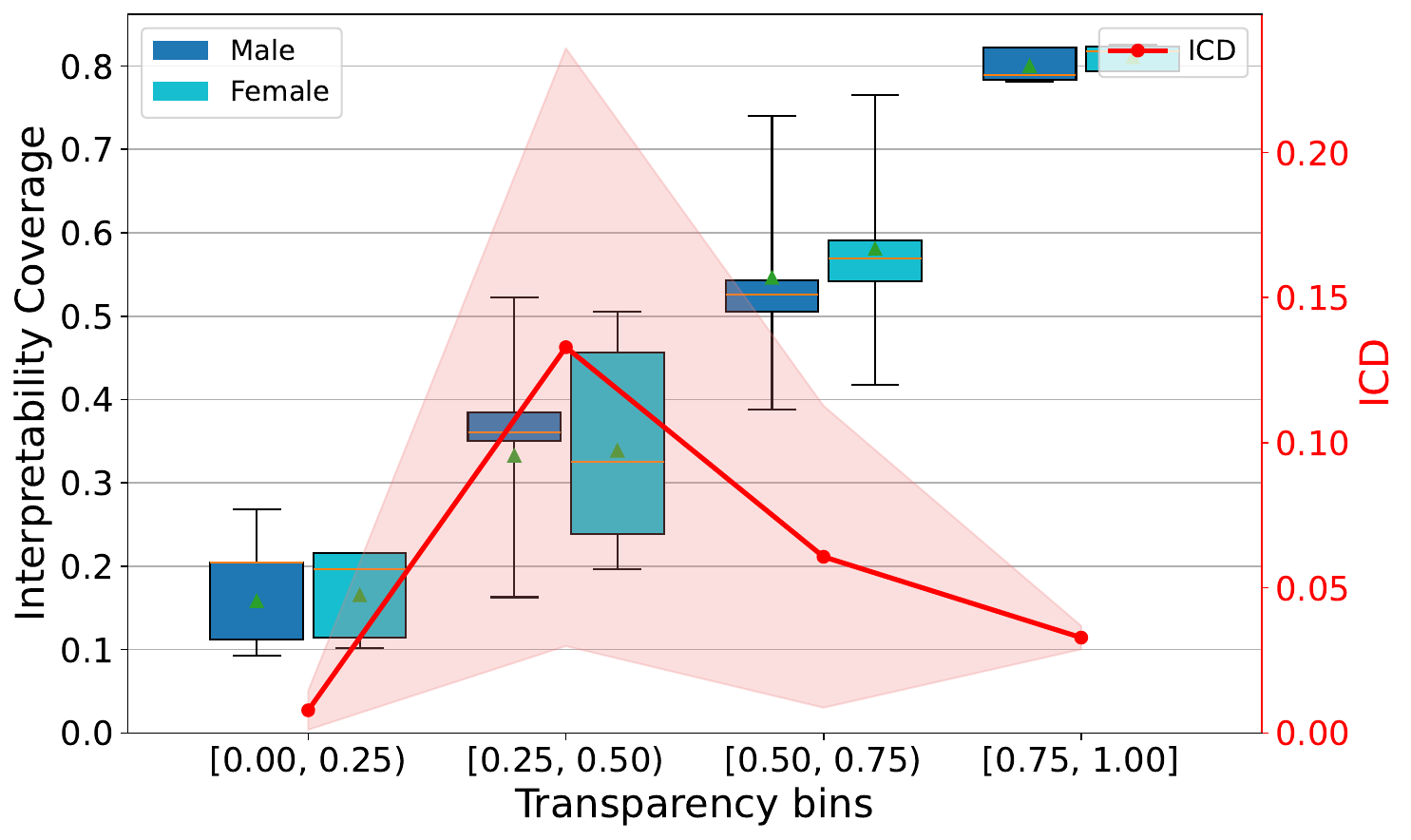}
    \hfill
    \includegraphics[width=0.32\linewidth]{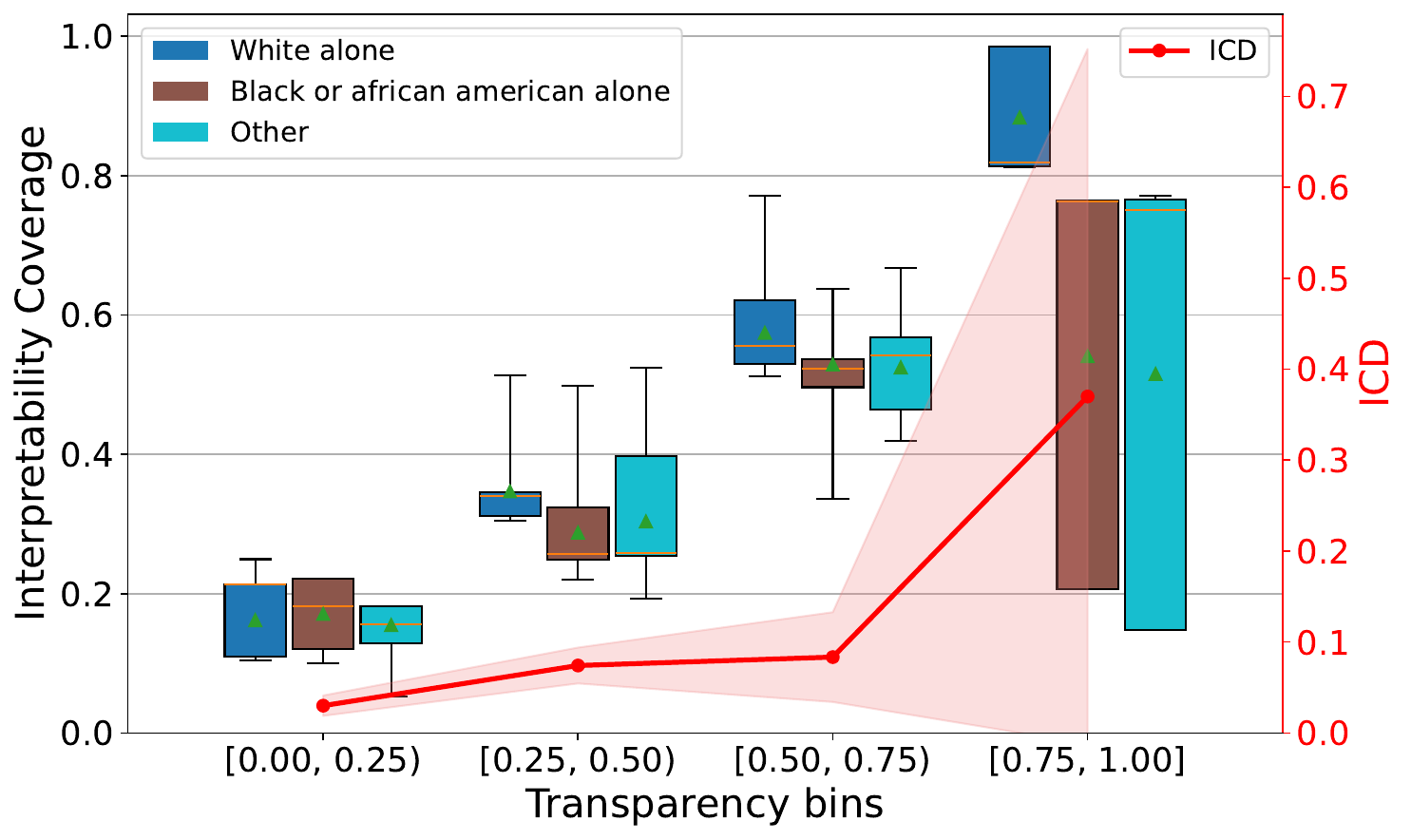}
    \caption{ACS Employment dataset}\label{fig:ICF_plot_Pre_acs_employ}
\end{subfigure}

\vspace{0.5em}

\begin{subfigure}{0.85\textwidth}
    \centering
    \includegraphics[width=0.32\linewidth]{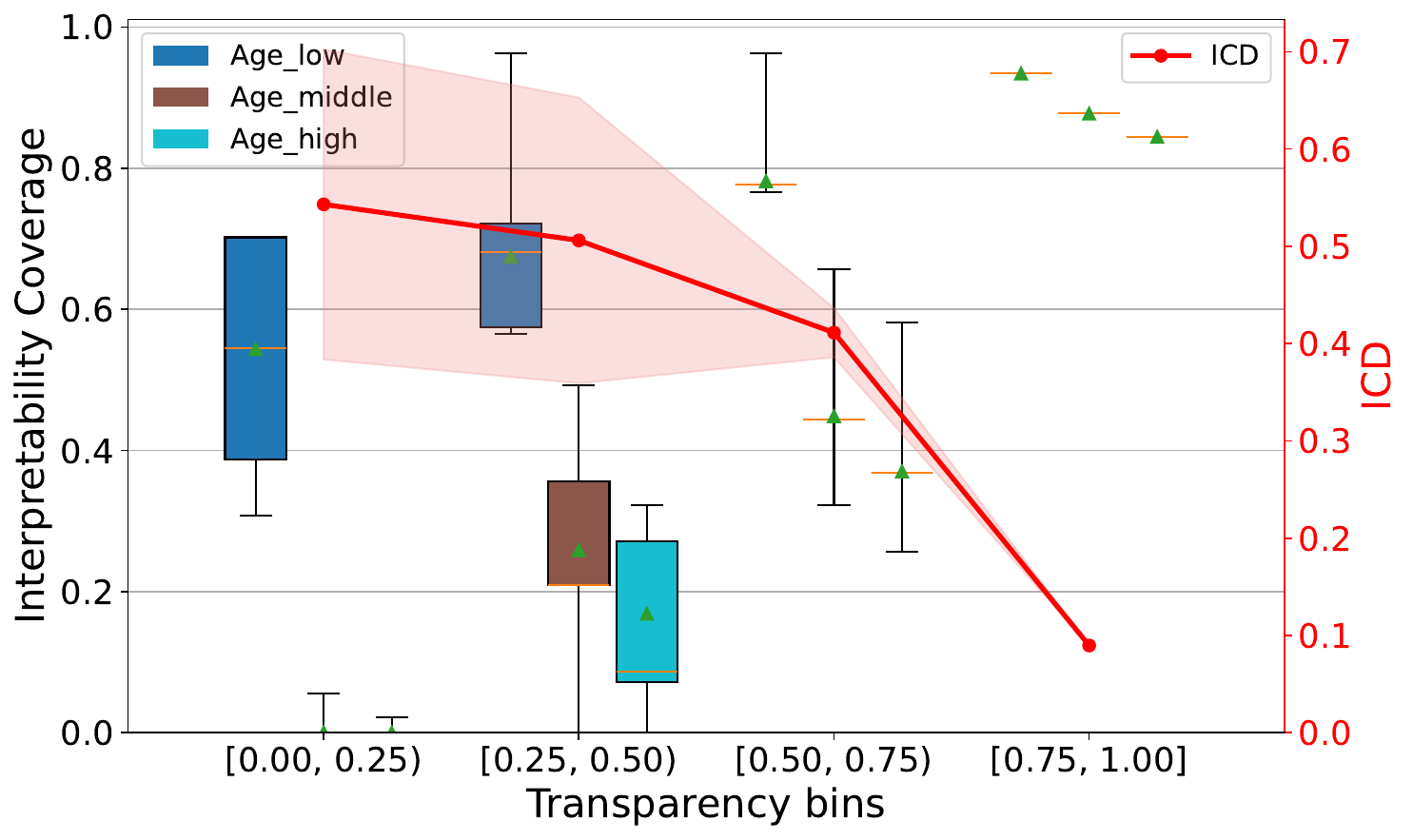}
    \includegraphics[width=0.32\linewidth]{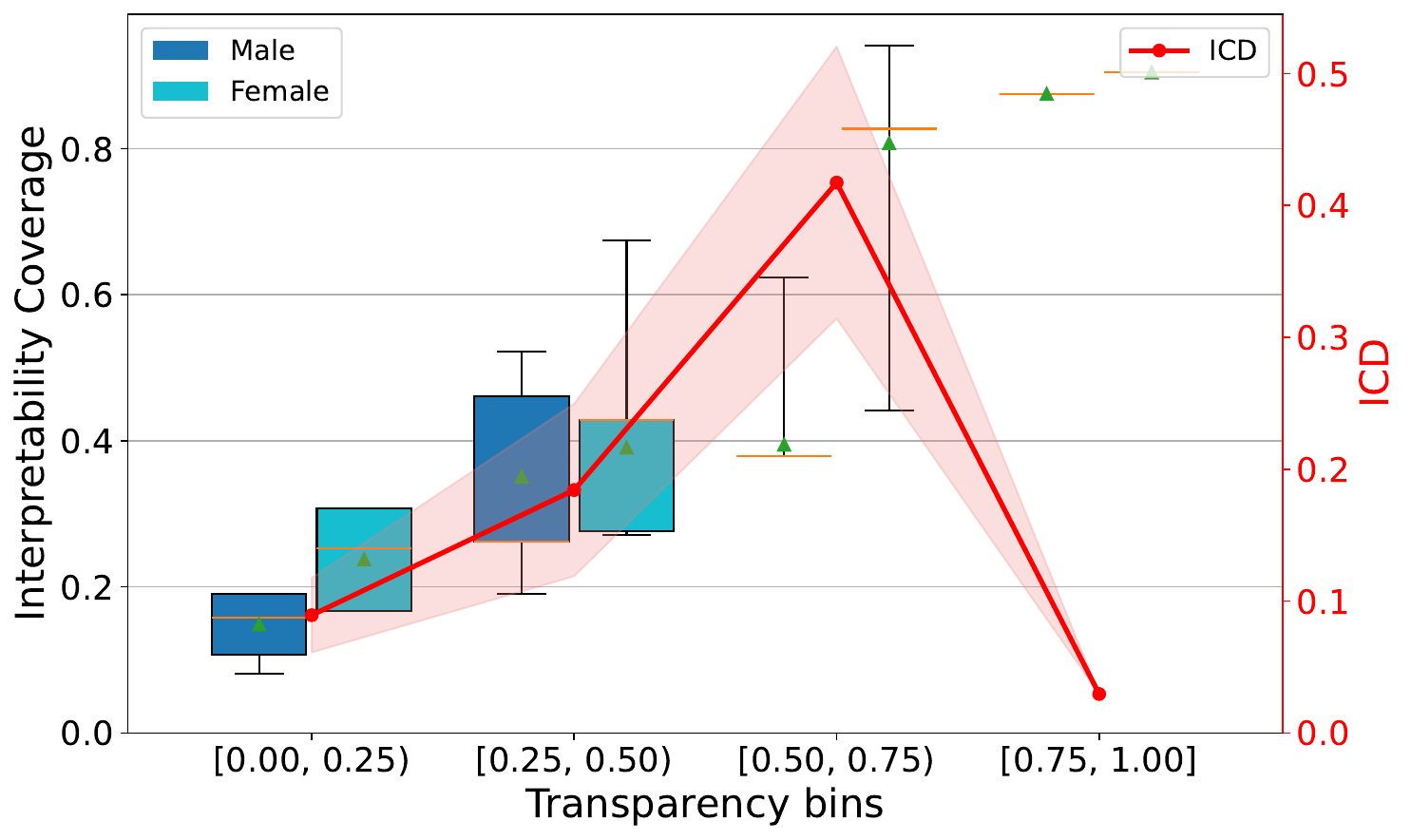}
    \includegraphics[width=0.32\linewidth]{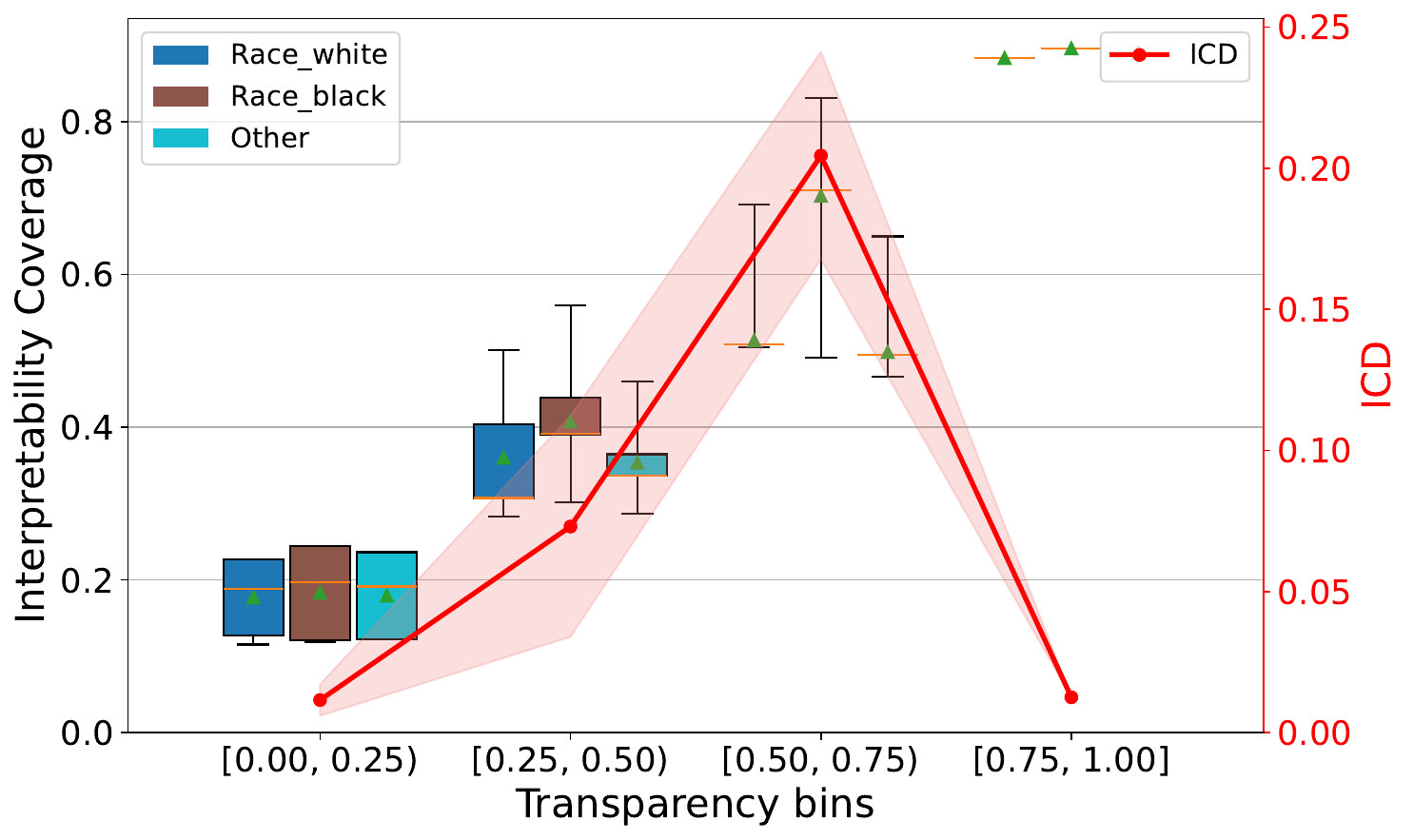}
    \caption{UCI Adult Income dataset}\label{fig:ICF_plot_Pre_adult}
\end{subfigure}

\vspace{0.5em}

\begin{subfigure}{0.85\textwidth}
    \centering
    \includegraphics[width=0.32\linewidth]{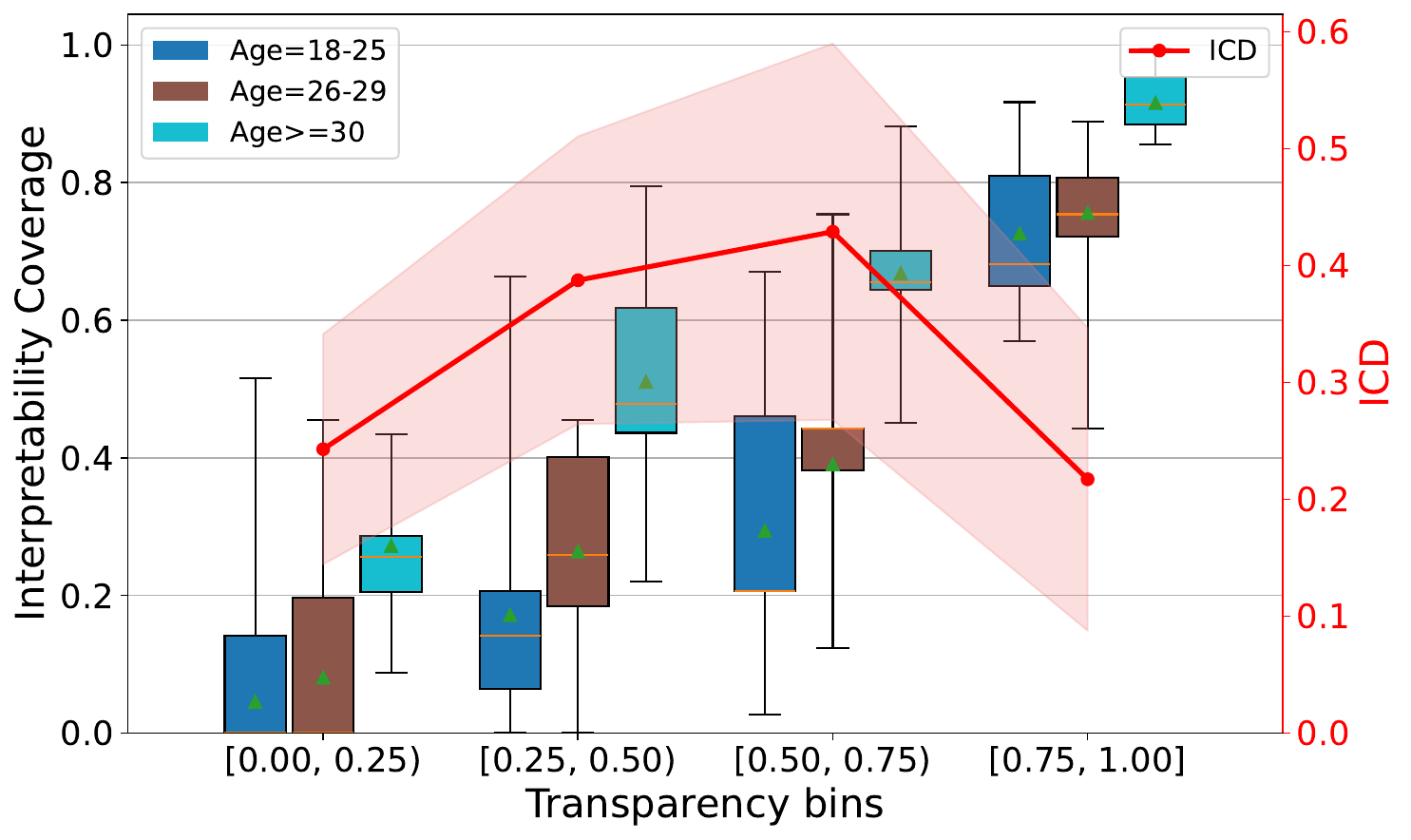}
    \includegraphics[width=0.32\linewidth]{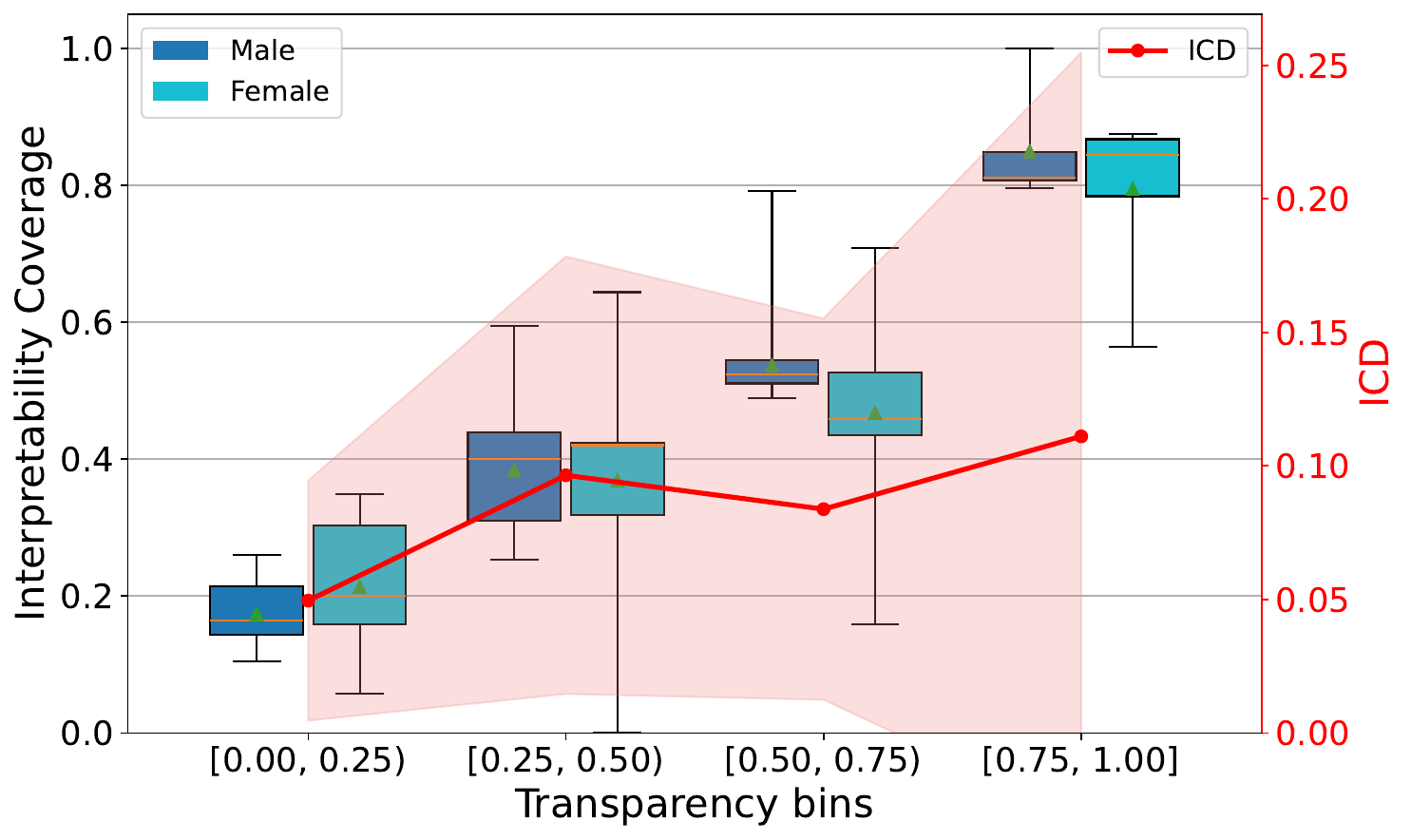}
    \includegraphics[width=0.32\linewidth]{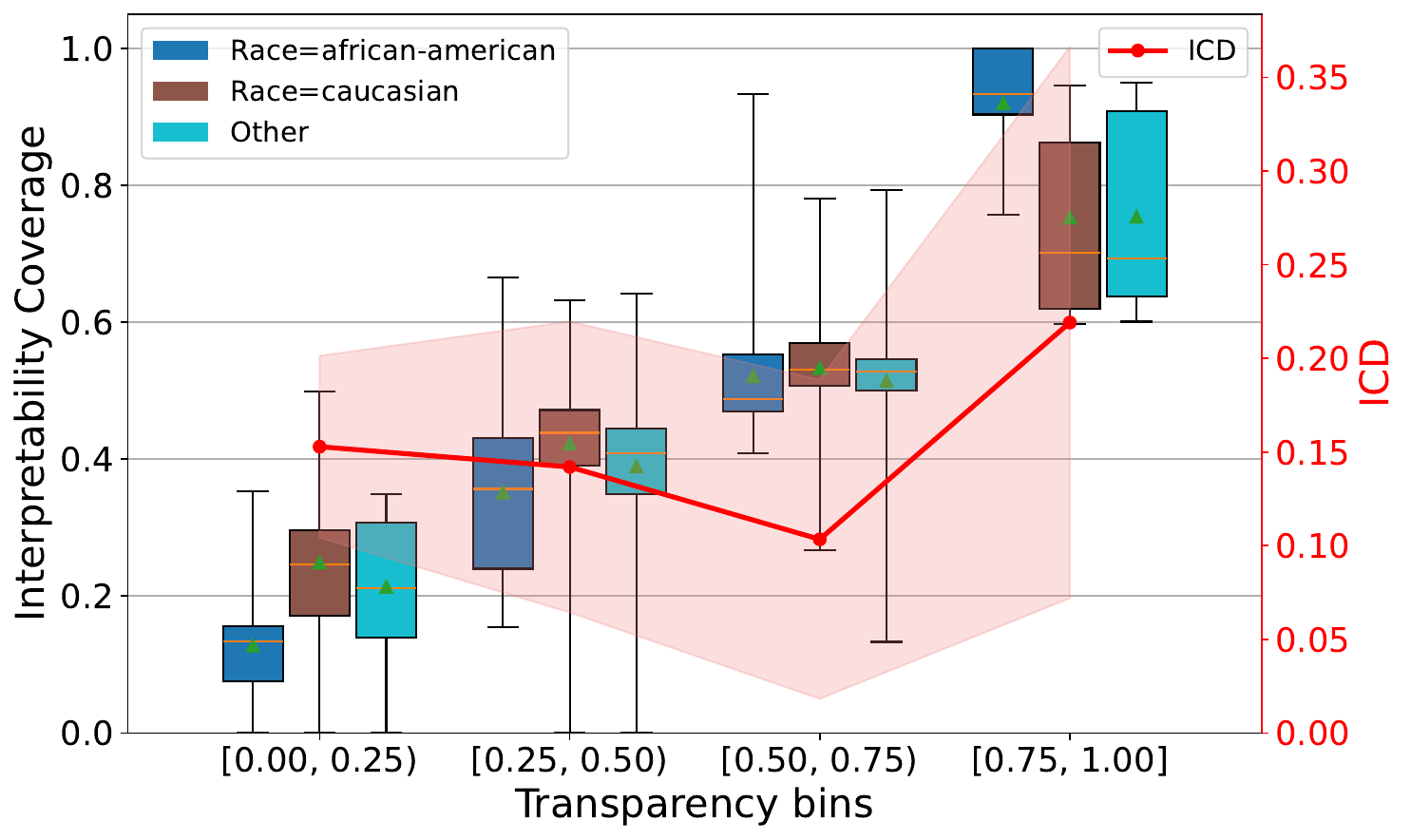}
    \caption{COMPAS dataset}\label{fig:ICF_plot_pre_compas}
\end{subfigure}
\caption{Distribution of test set Interpretability Coverage (IC) across Rashomon sets of HybridCORELSPre for all transparency bins $(\varepsilon = 0.01)$. Results are shown for all datasets and sensitive attributes (Age, Gender, Race), ordered from left to right columns. The red curve represents the average ICD across models within each bin, while the shaded region indicates the corresponding standard deviation.} \label{fig:ICF_plot_pre}
\end{figure*}

\begin{figure*}[t]
\centering

\begin{subfigure}{0.85\textwidth}
    \centering
    \includegraphics[width=0.32\linewidth ]{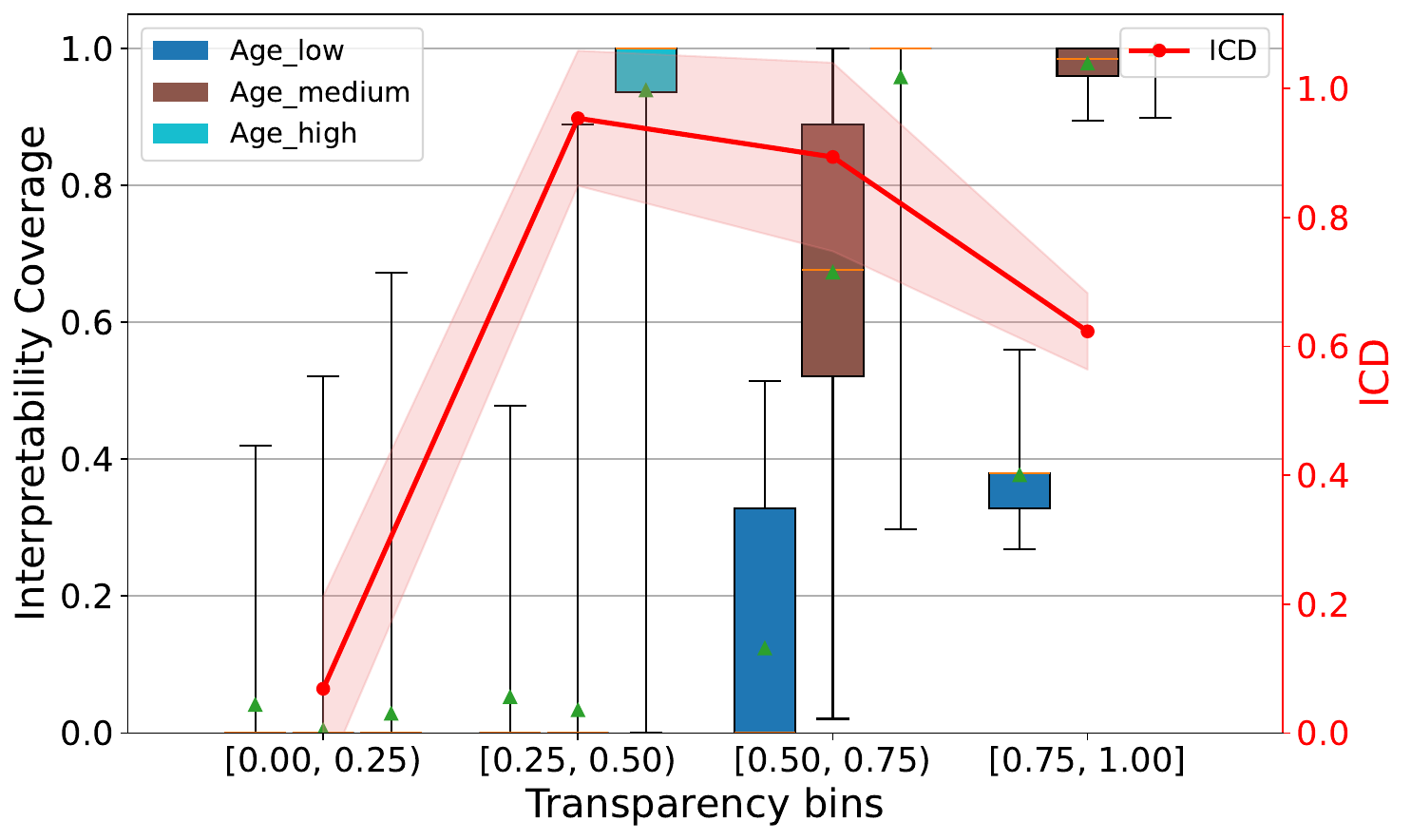}
    \includegraphics[width=0.32\linewidth]{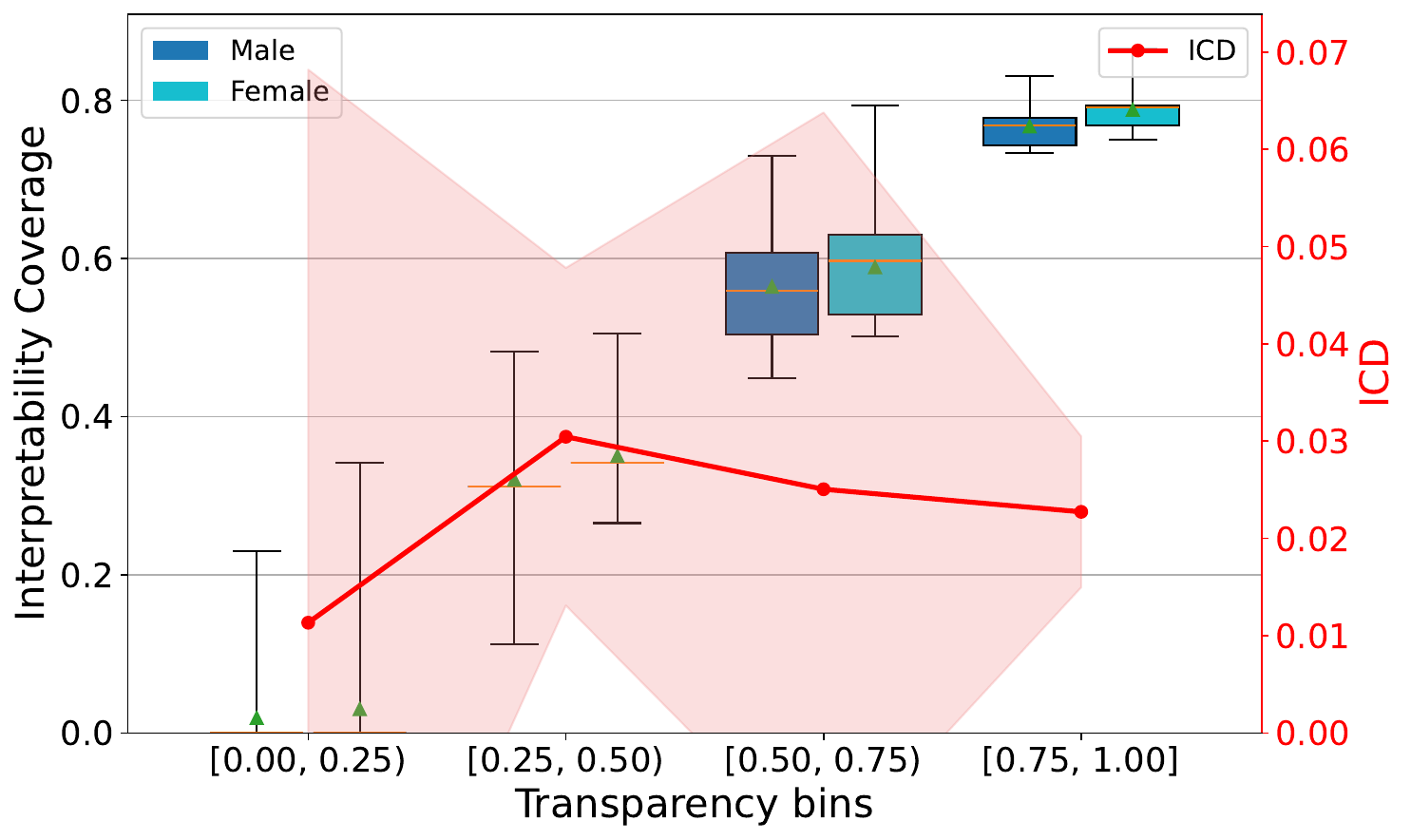}
    \includegraphics[width=0.32\linewidth]{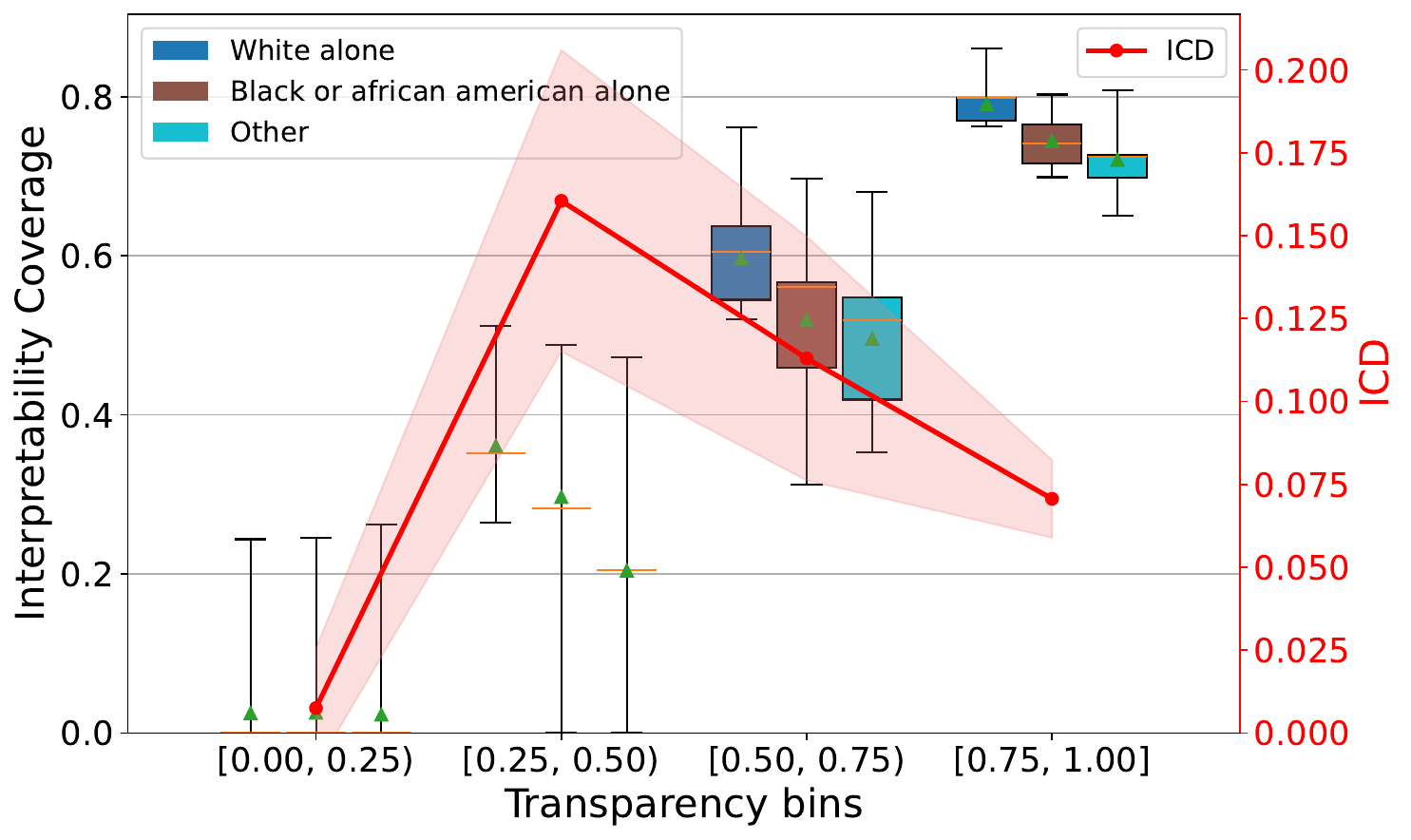}
    \caption{ACS Employment dataset}\label{fig:ICF_plot_HyRS_acs_employ}
\end{subfigure}

\vspace{0.5em}

\begin{subfigure}{0.85\textwidth}
    \centering
    \includegraphics[width=0.32\linewidth]{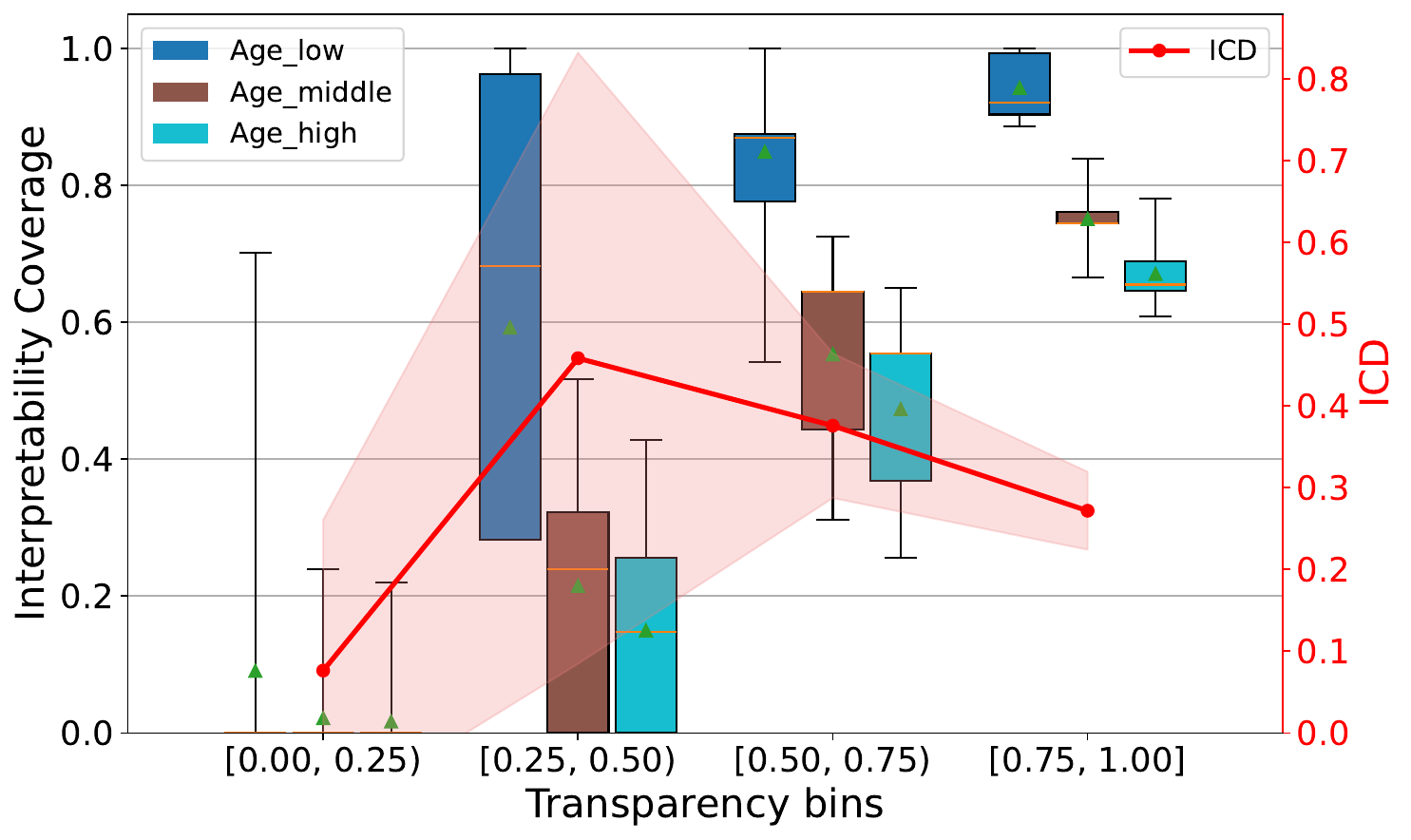}
    \includegraphics[width=0.32\linewidth]{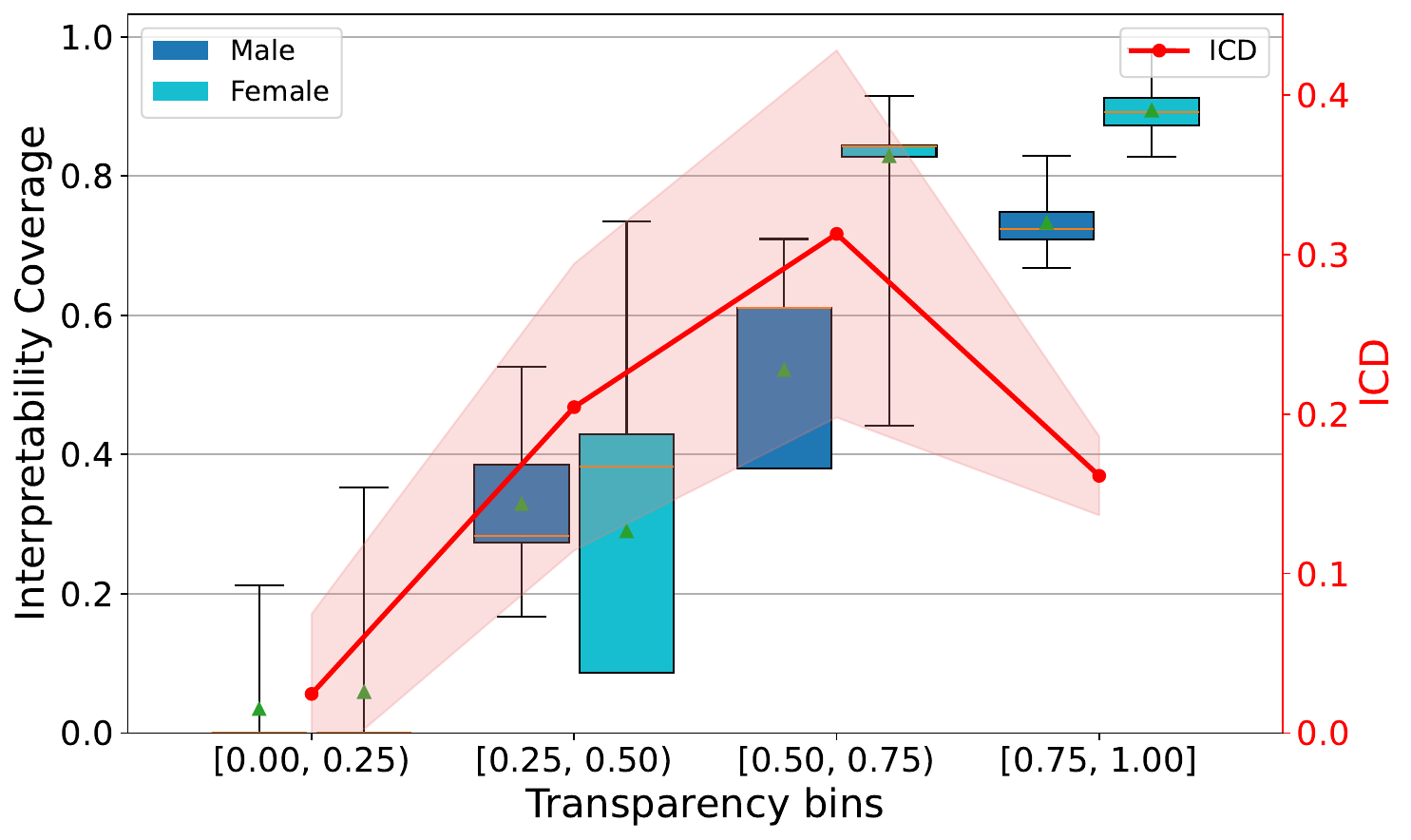}
    \includegraphics[width=0.32\linewidth]{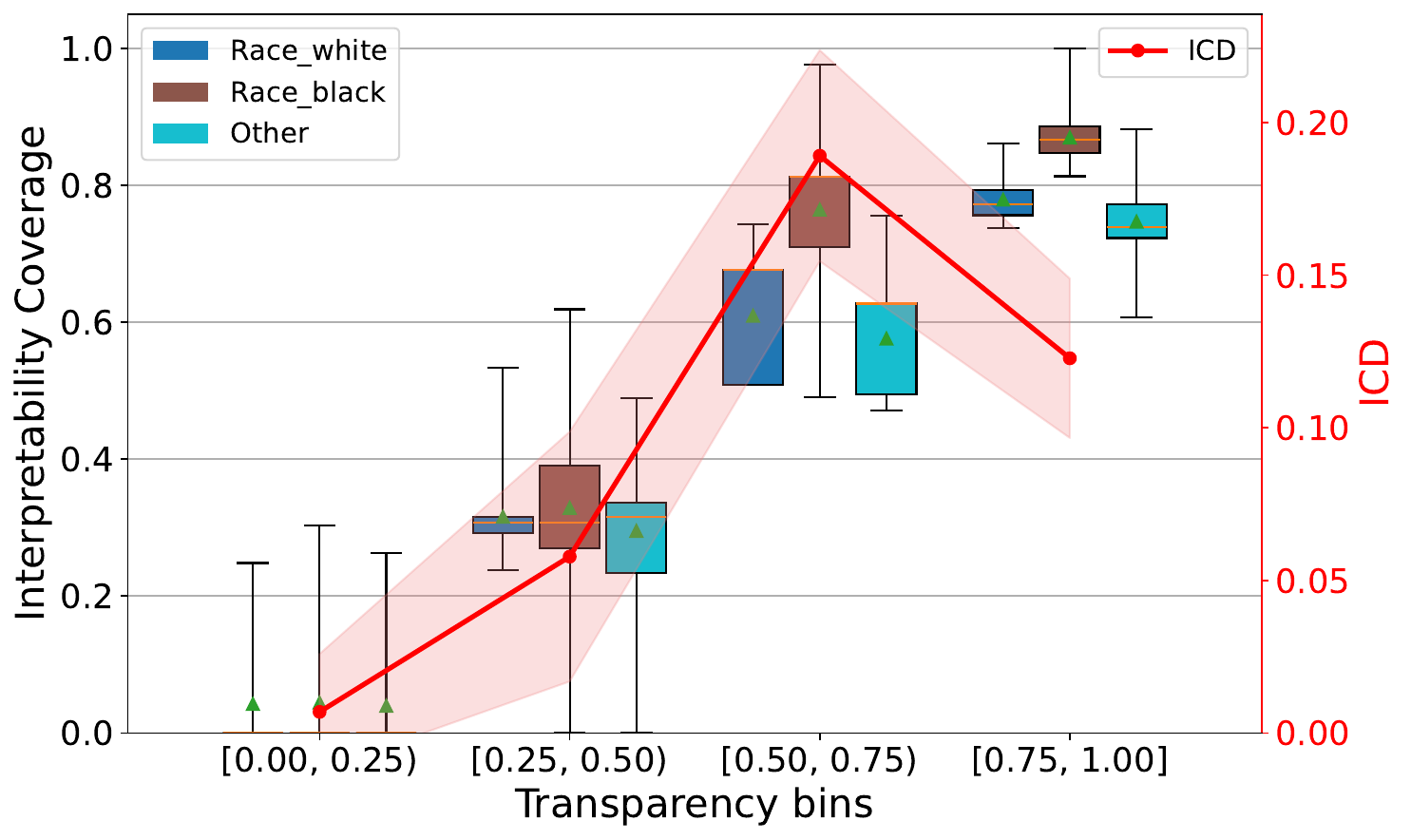}
    \caption{UCI Adult Income dataset}\label{fig:ICF_plot_HyRS_adult}
\end{subfigure}

\vspace{0.5em}

\begin{subfigure}{0.85\textwidth}
    \centering
    \includegraphics[width=0.32\linewidth]{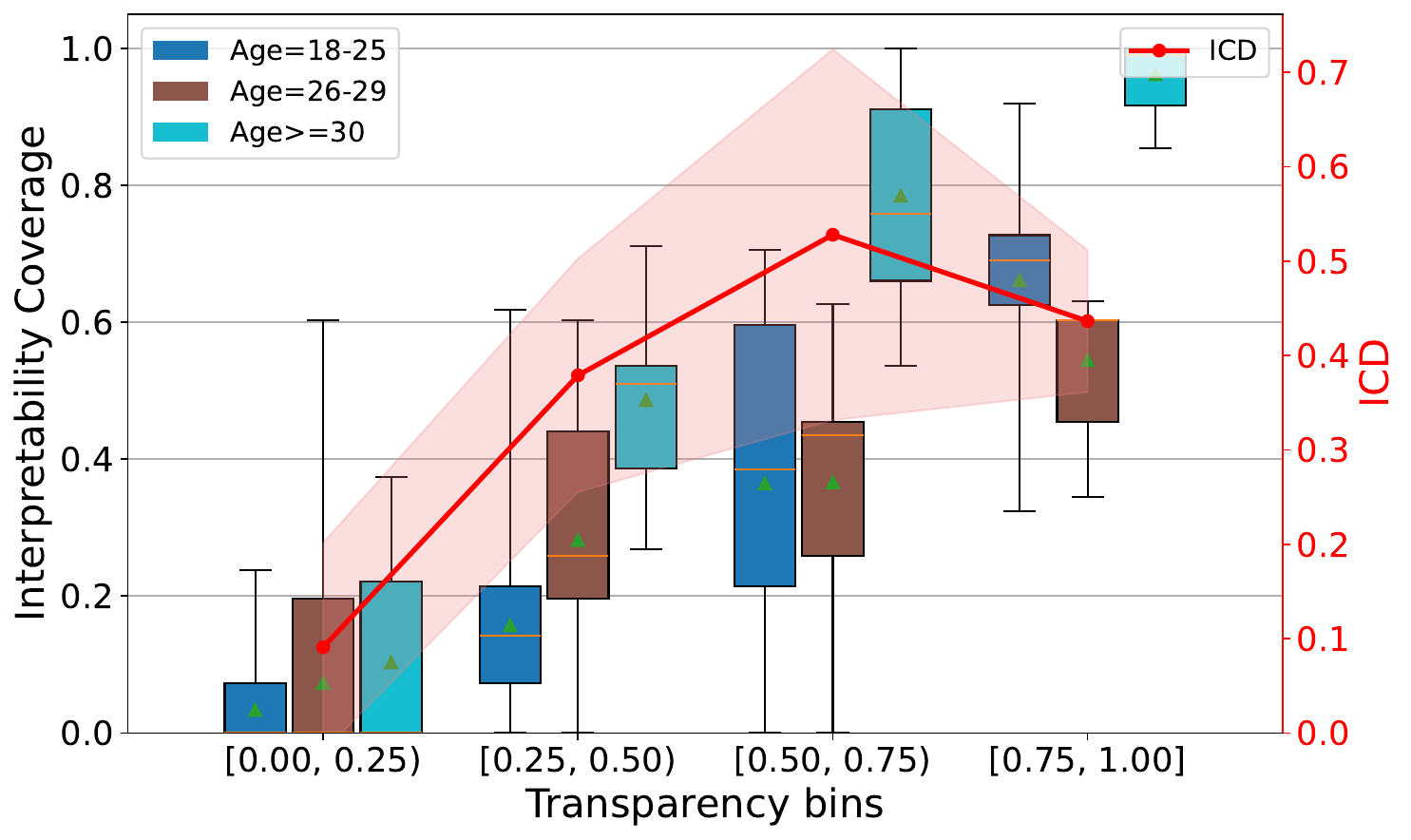}
    \includegraphics[width=0.32\linewidth]{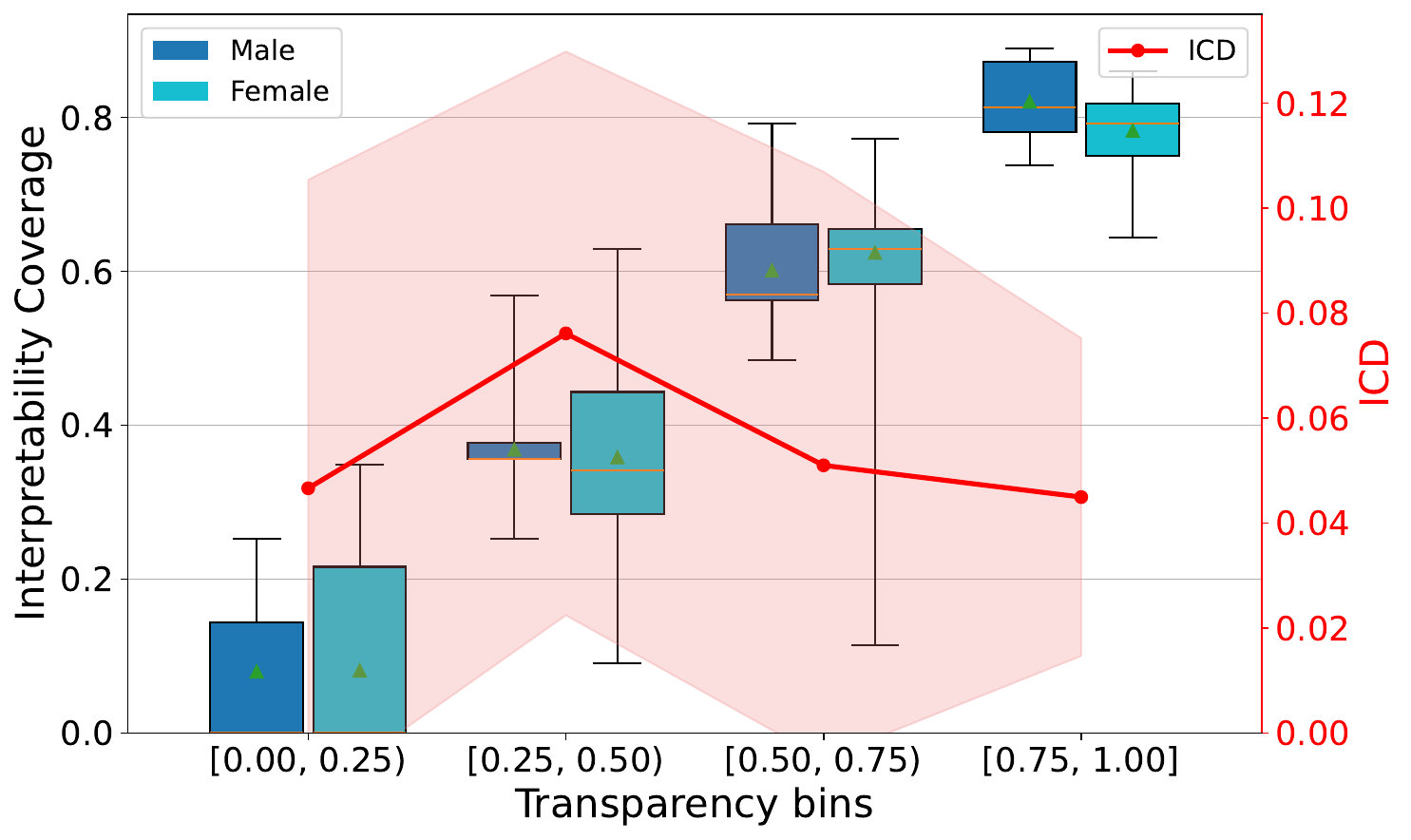}
    \includegraphics[width=0.32\linewidth]{Plots/ICF/Comp_ICF_2axiscompas_HyRS_Race_test_0.01.pdf}
    \caption{COMPAS dataset}\label{fig:ICF_plot_HyRS_compas}
\end{subfigure}
\caption{Distribution of test set Interpretability Coverage (IC) across Rashomon sets of HyRS for all transparency bins $(\varepsilon = 0.01)$. Results are shown for all datasets and sensitive attributes (Age, Gender, Race), ordered from left to right columns. The red curve represents the average ICD across models within each bin, while the shaded region indicates the corresponding standard deviation.} \label{fig:ICF_plot_HyRS}
\end{figure*}

\begin{figure*}[t]
\centering

\begin{subfigure}{0.85\textwidth}
    \centering
    \includegraphics[width=0.32\linewidth]{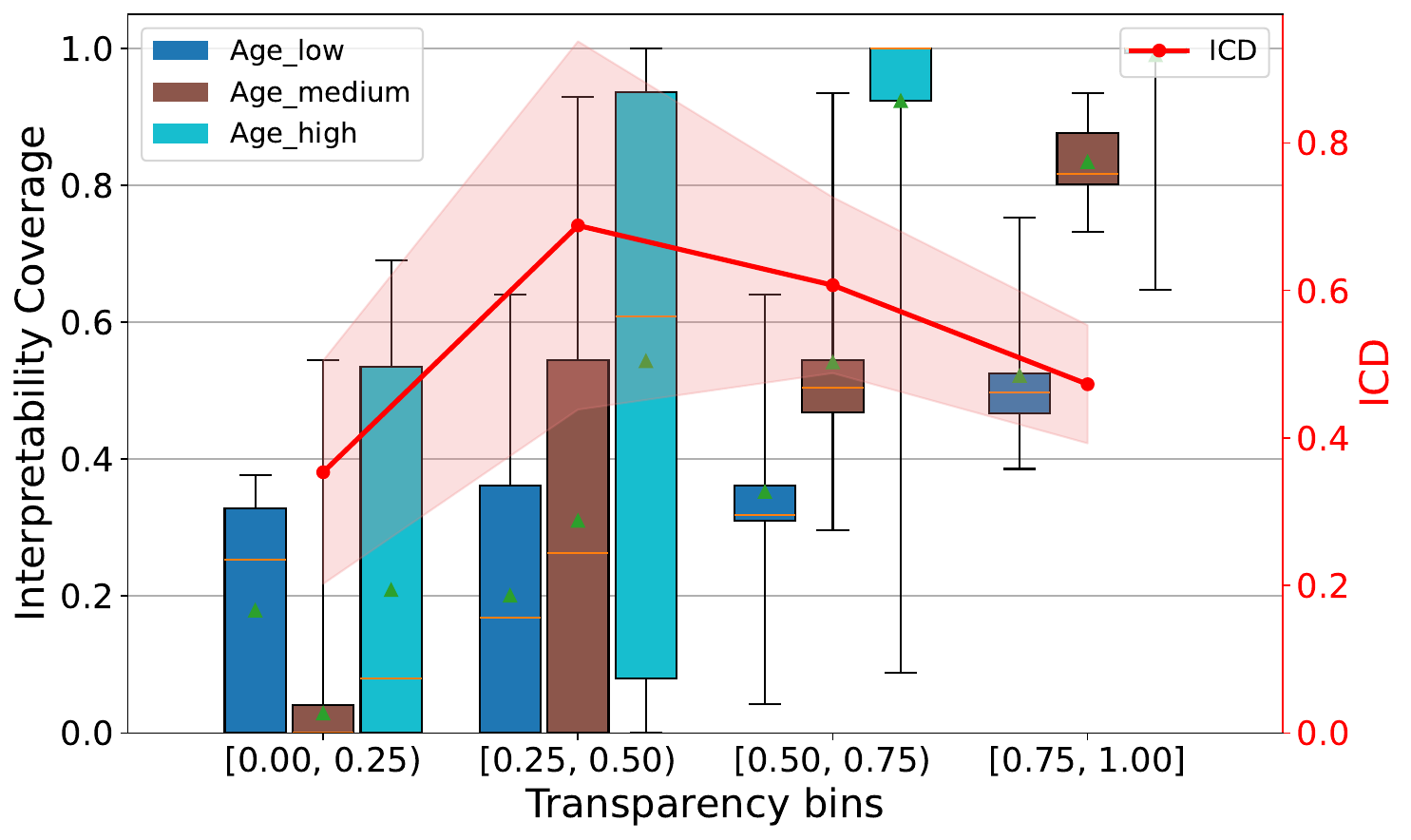}
    \includegraphics[width=0.32\linewidth]{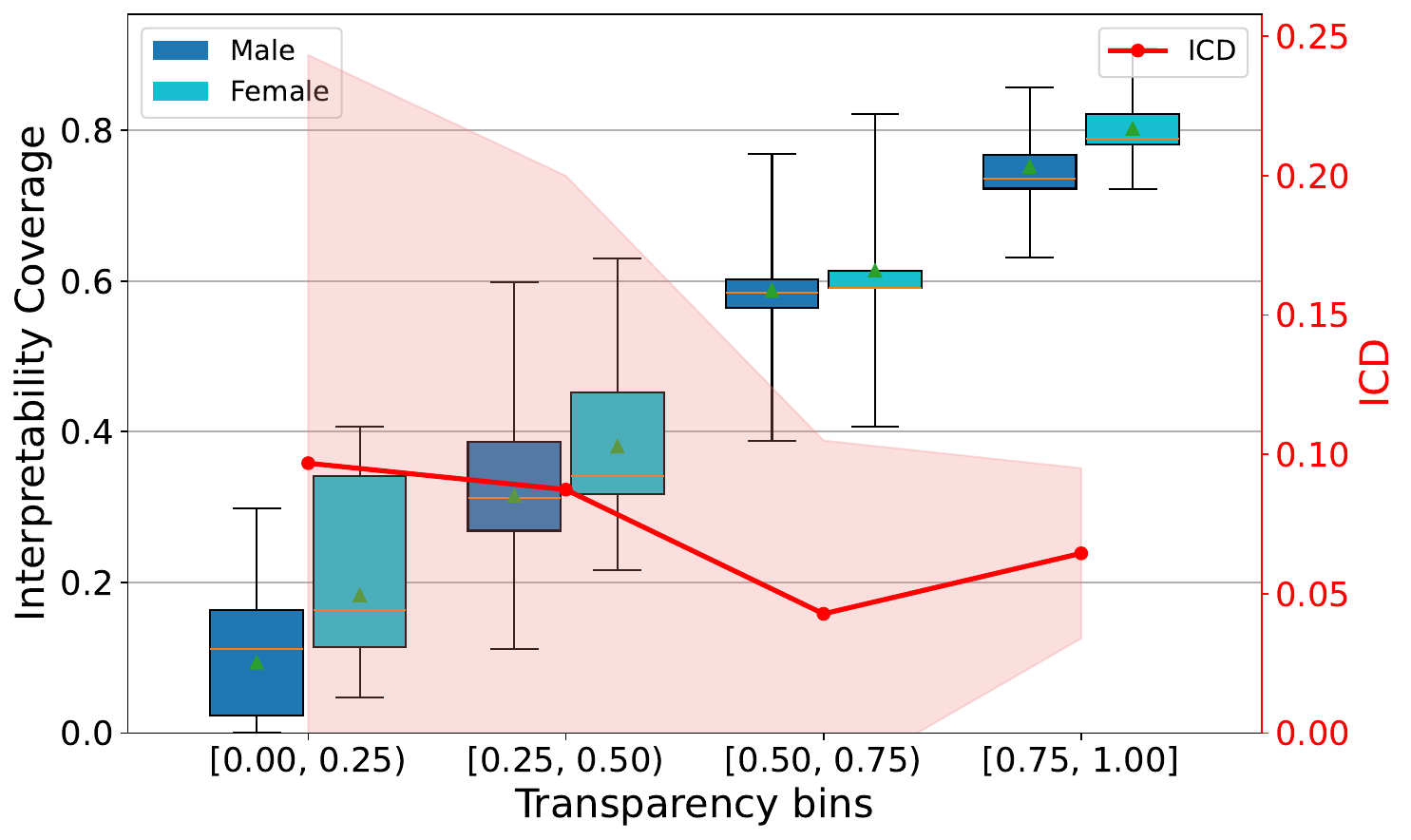}
    \includegraphics[width=0.32\linewidth]{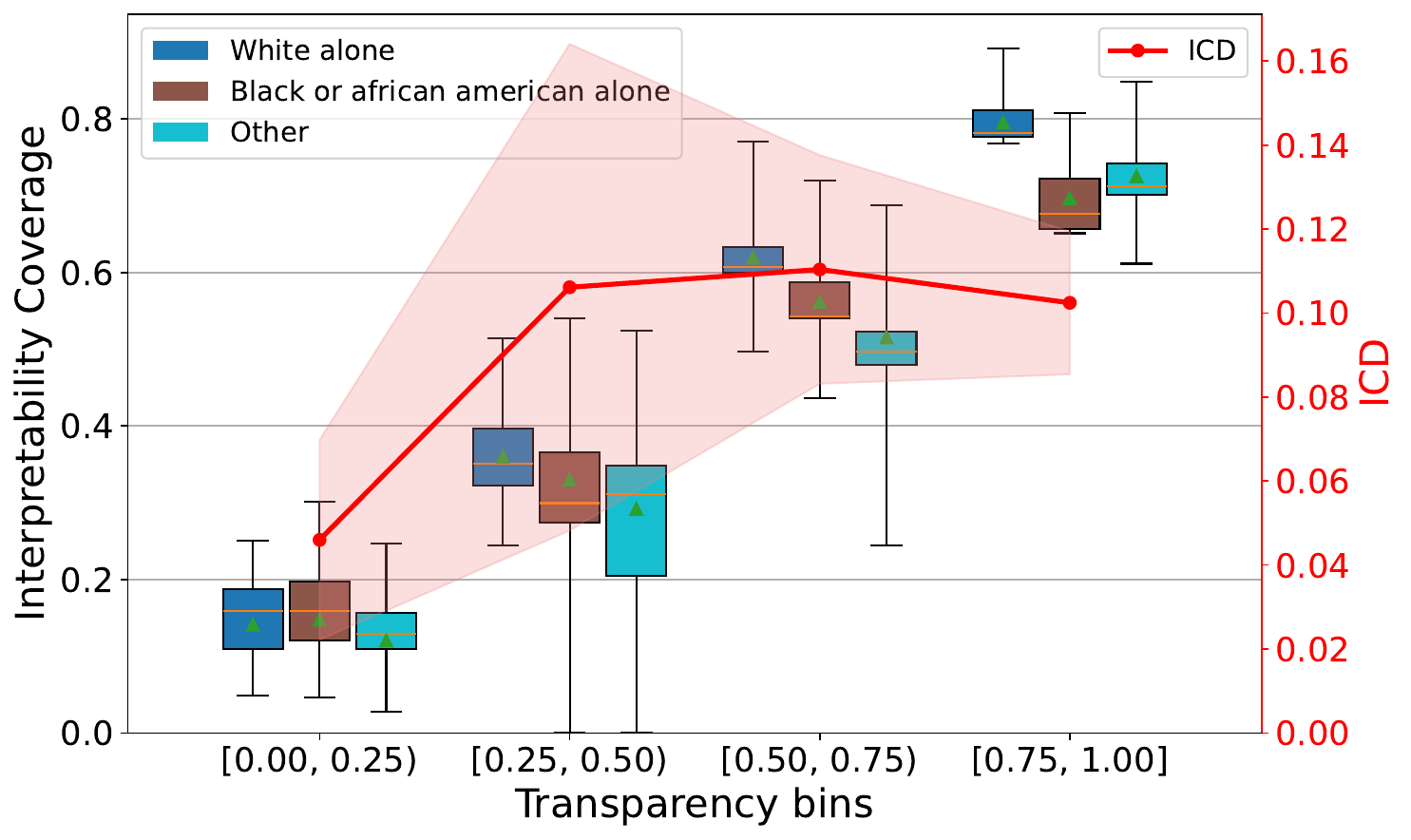}
    \caption{ACS Employment dataset}\label{fig:ICF_plot_CRL_acs_employ}
\end{subfigure}

\vspace{0.5em}

\begin{subfigure}{0.85\textwidth}
    \centering
    \includegraphics[width=0.32\linewidth]{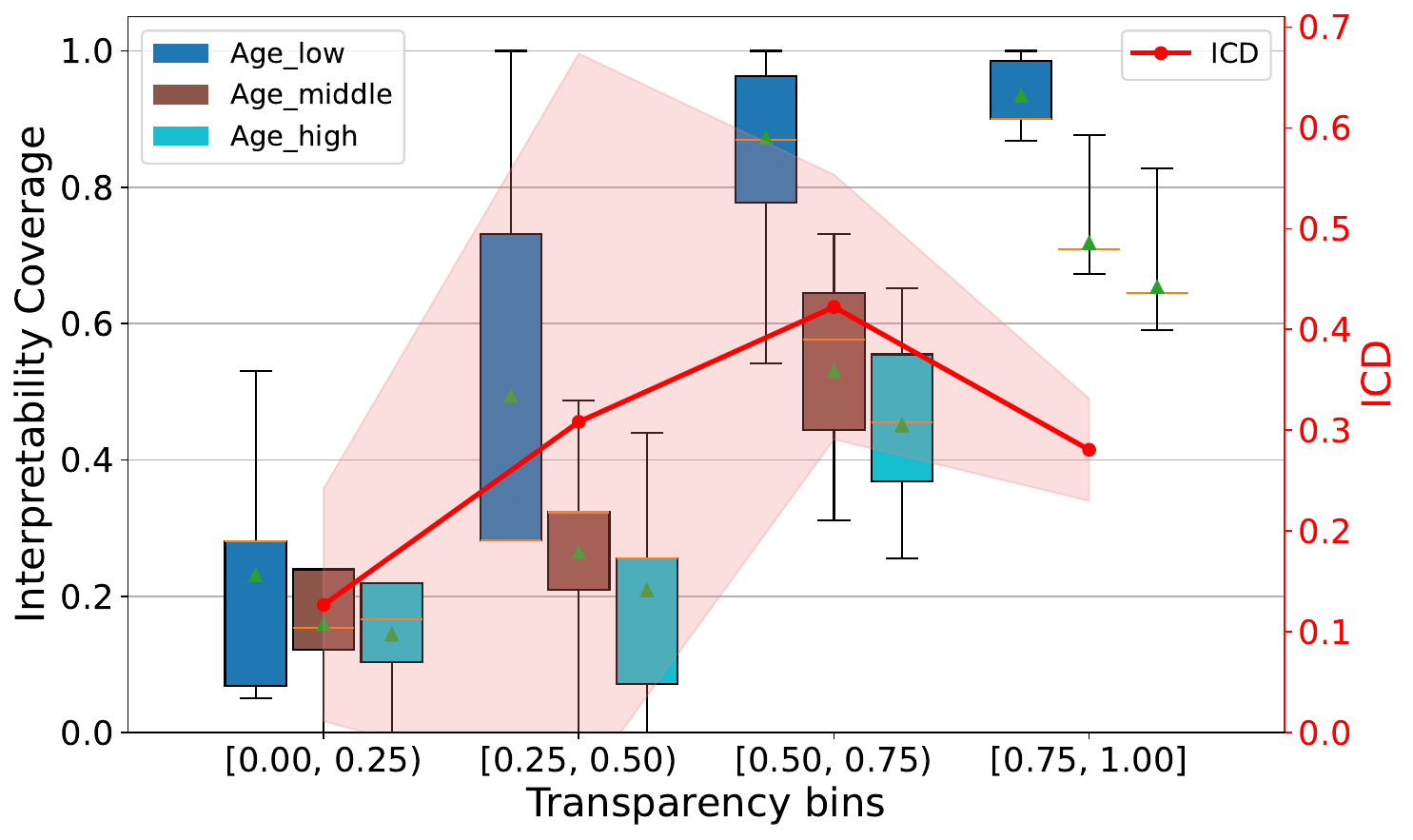}
    \includegraphics[width=0.32\linewidth]{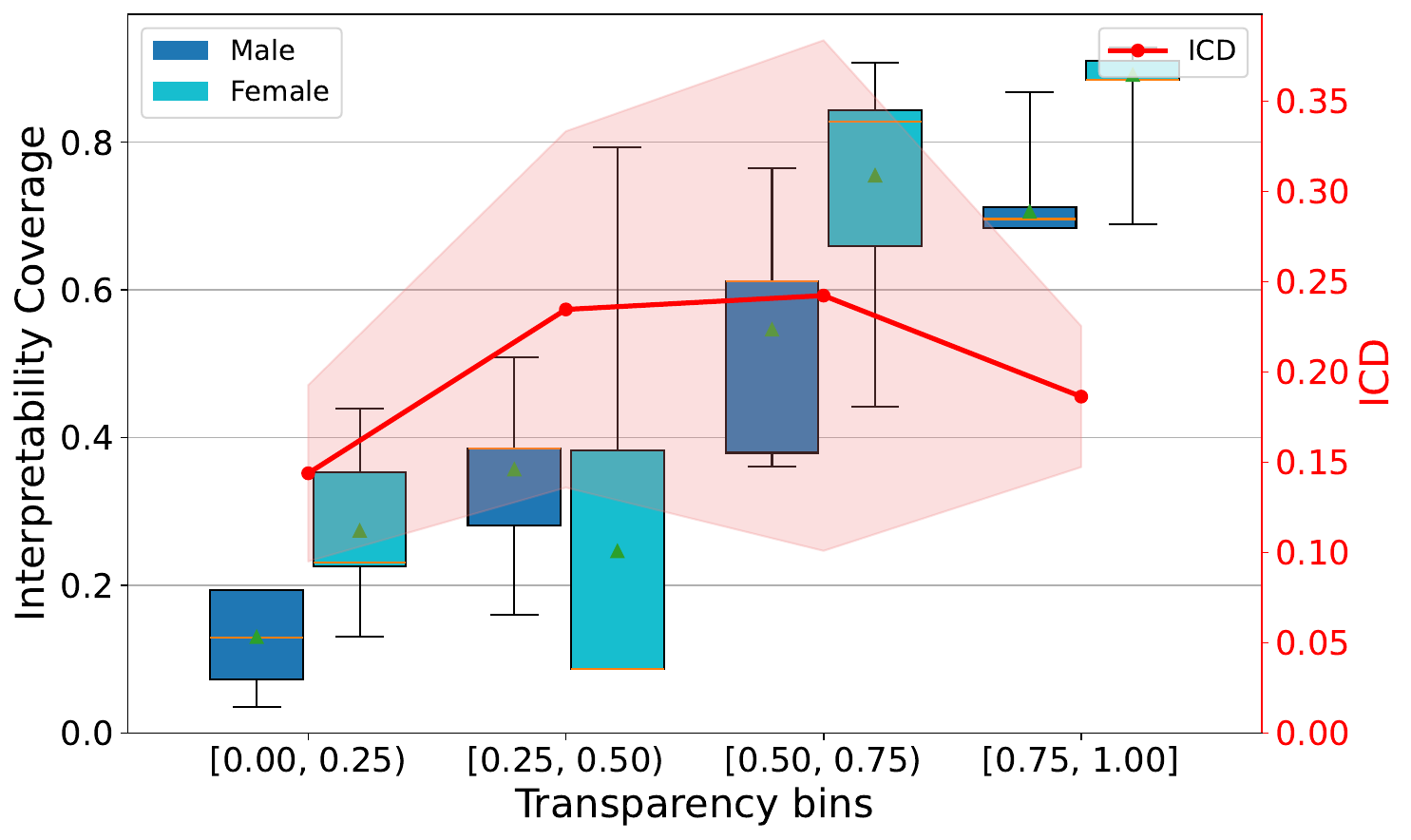}
    \includegraphics[width=0.32\linewidth]{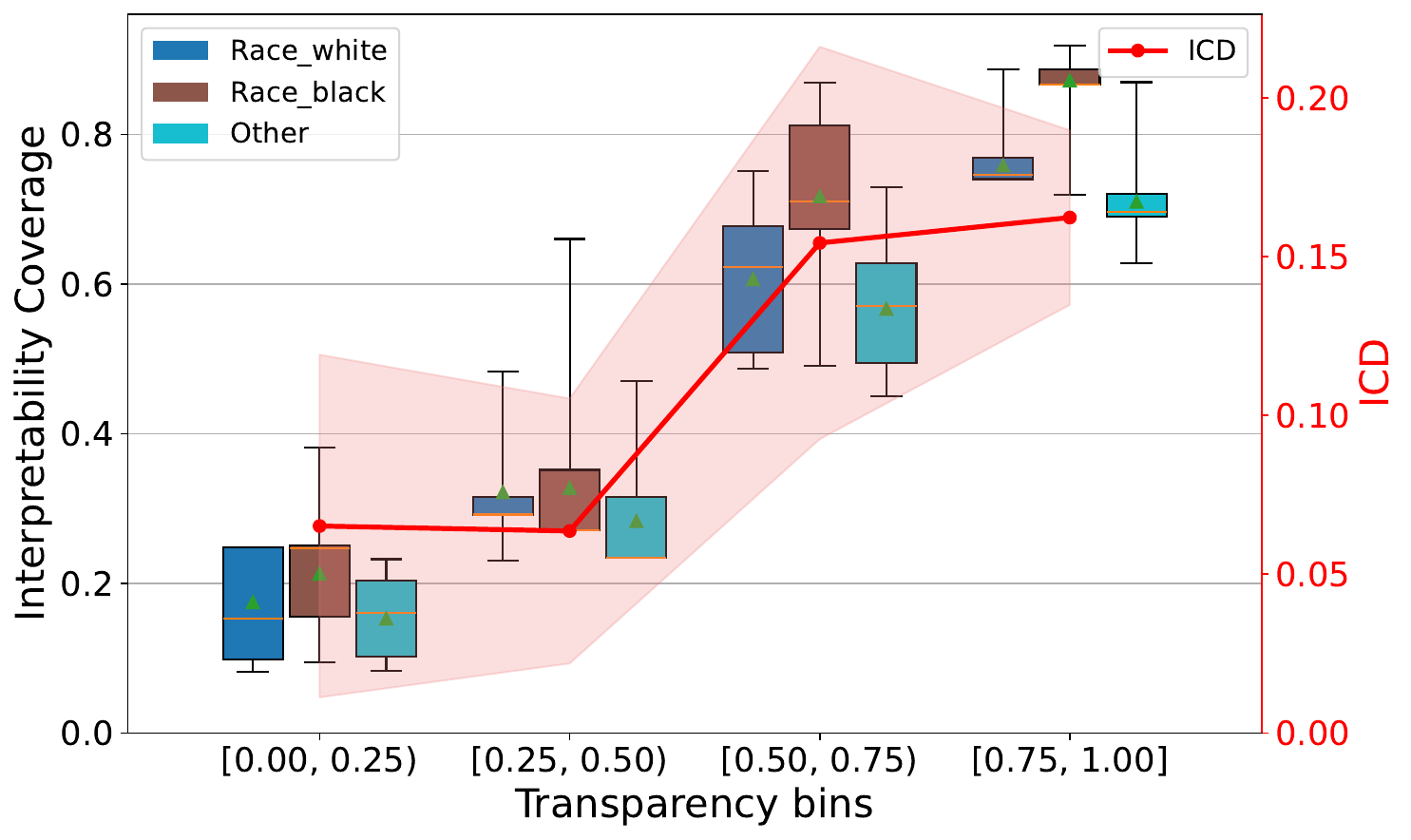}
    \caption{UCI Adult Income dataset}\label{fig:ICF_plot_CRL_adult}
\end{subfigure}

\vspace{0.5em}

\begin{subfigure}{0.85\textwidth}
    \centering
    \includegraphics[width=0.32\linewidth]{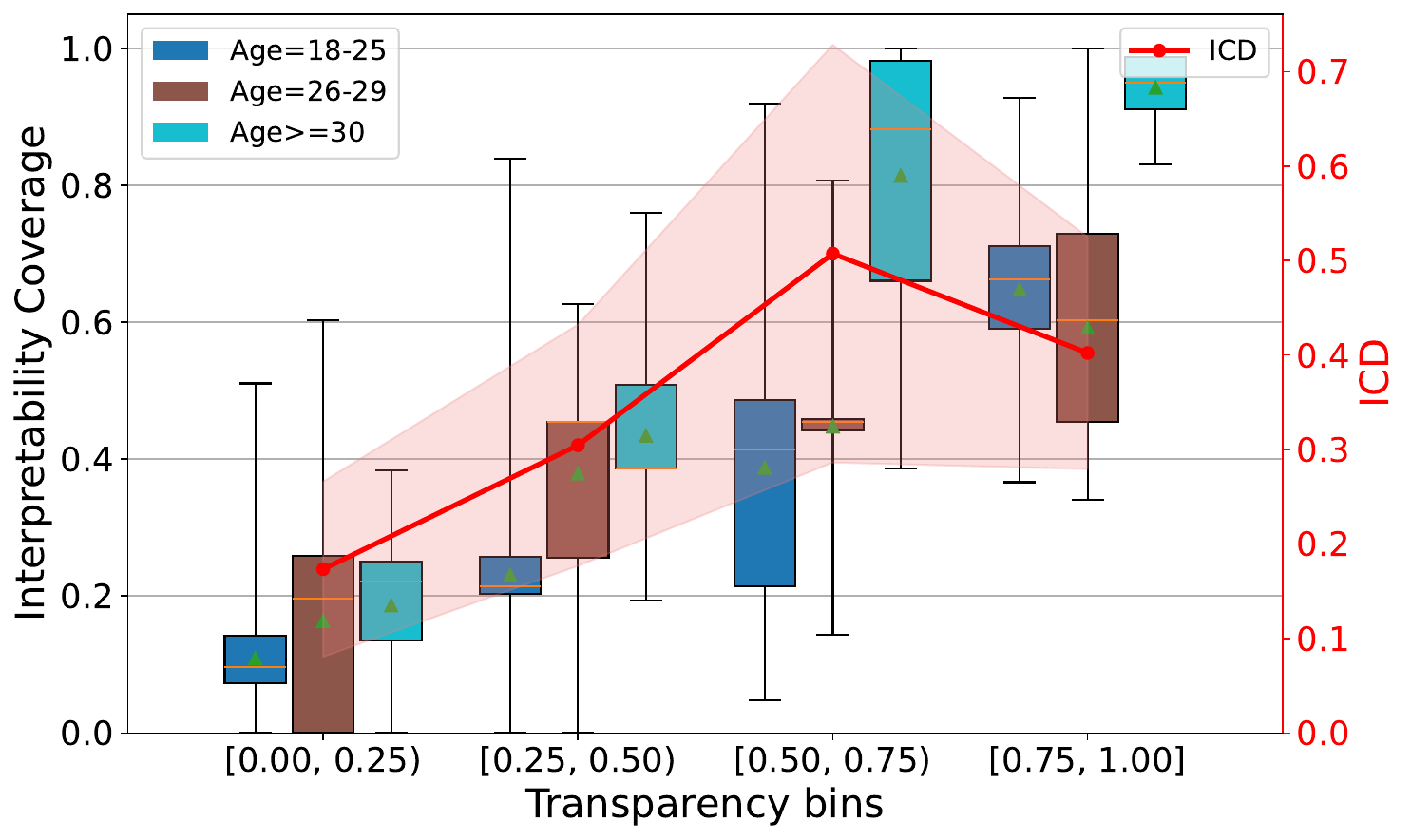}
    \includegraphics[width=0.32\linewidth]{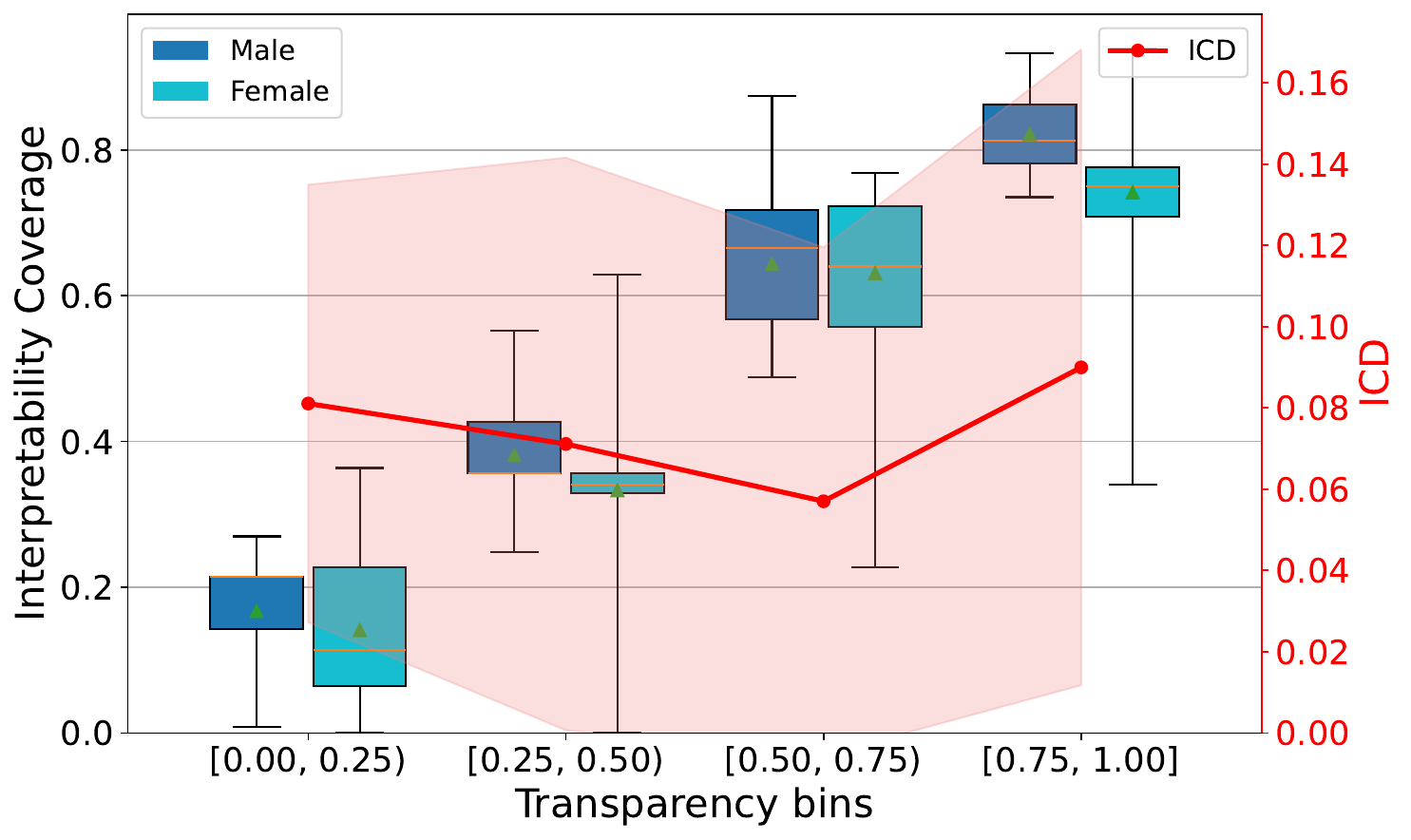}
    \includegraphics[width=0.32\linewidth]{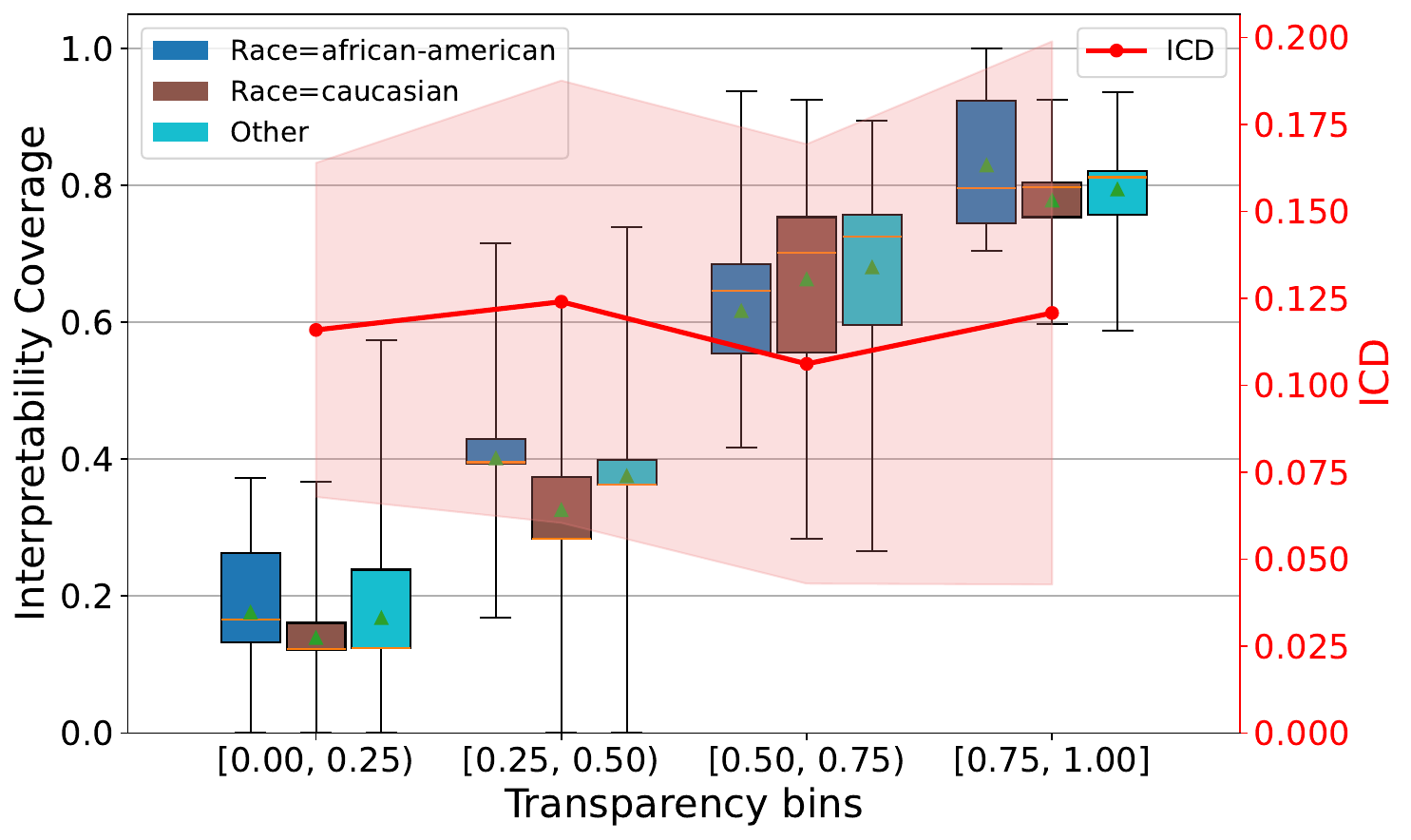}
    \caption{COMPAS dataset}\label{fig:ICF_plot_CRL_compas}
\end{subfigure}
\caption{Distribution of test set Interpretability Coverage (IC) across Rashomon sets of CRL for all transparency bins $(\varepsilon = 0.01)$. Results are shown for all datasets and sensitive attributes (Age, Gender, Race), ordered from left to right columns. The red curve represents the average Interpretability Coverage Disparity (ICD) across models within each bin, while the shaded region indicates the corresponding standard deviation.} \label{fig:ICF_plot_CRL}
\end{figure*}

In support of the findings discussed in Result 1, Figures \ref{fig:ICF_plot_post}--\ref{fig:ICF_plot_CRL} provide a comprehensive overview of Interpretability Coverage (IC) across demographic subgroups for the sensitive attributes Age, Gender, and Race for HybridCORELSPost, HybridCORELSPre, and CRL. The results consistently reveal substantial disparities in interpretability coverage across datasets and transparency bins. In addition, the corresponding ICD curves frequently exhibit bell-like behavior, with disparities peaking in intermediate transparency regimes and decreasing or remaining unchanged at higher transparency levels.

\section{Additional Experimental Results for Section~\ref{sec:mitigation_results}} \label{sec:appendix_mitigation_results}

\begin{figure*}[t]
\centering
\includegraphics[width=0.8\textwidth]{Plots/ICF/HybridCORELS_fairness_shared_legend.pdf}

\begin{subfigure}{0.85\textwidth}
    \centering
    \includegraphics[width=0.32\linewidth]{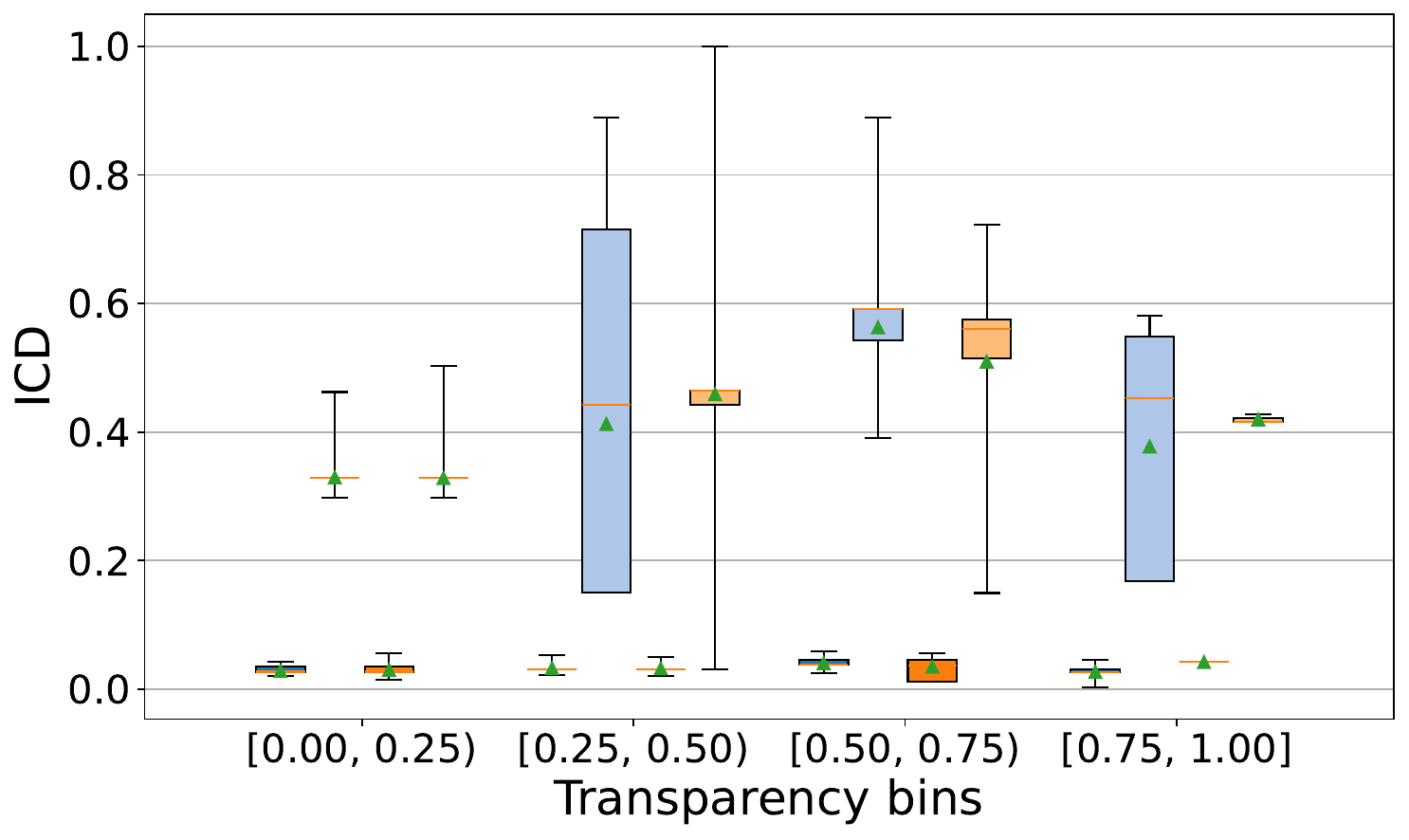}
    \includegraphics[width=0.32\linewidth]{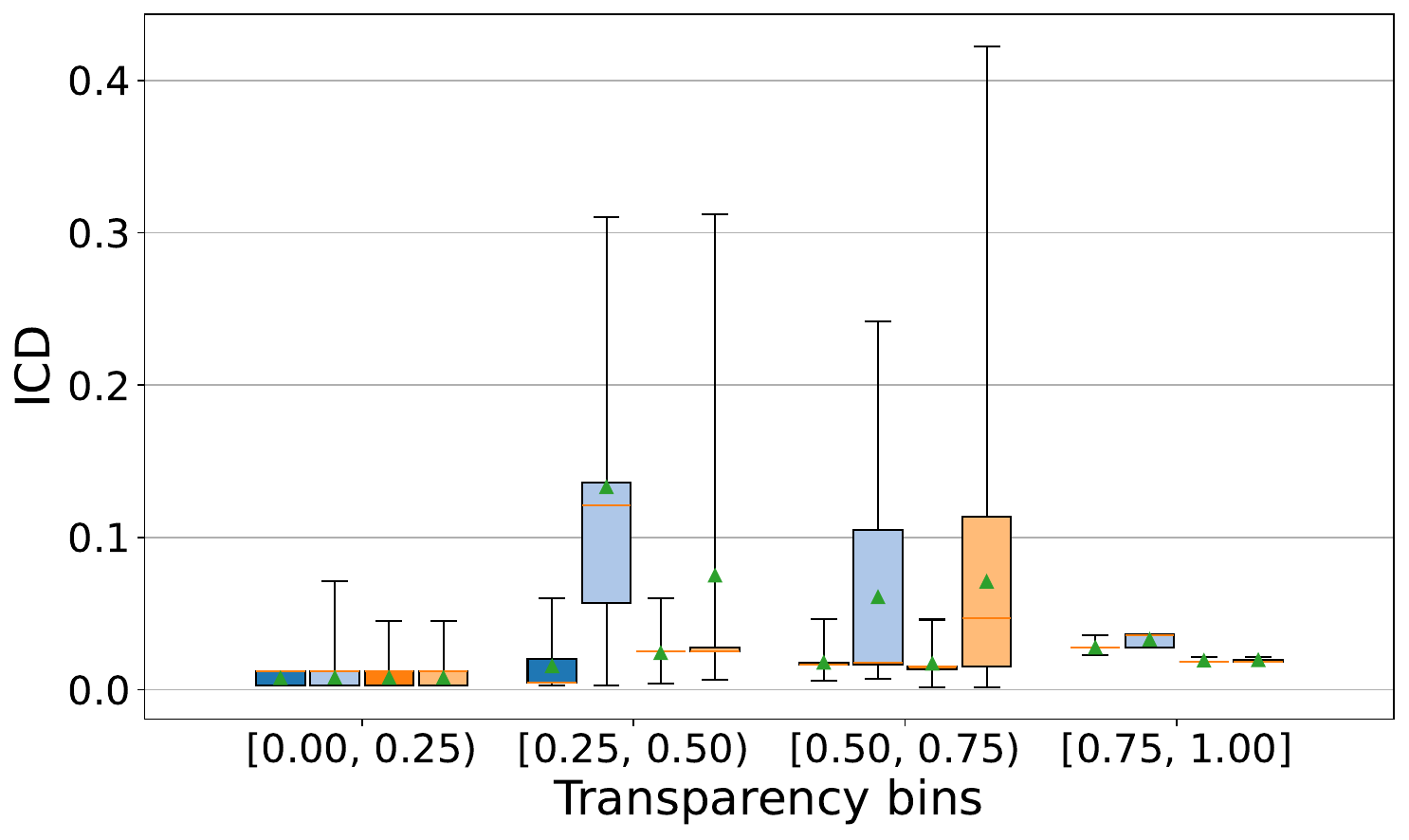}
    \includegraphics[width=0.32\linewidth]{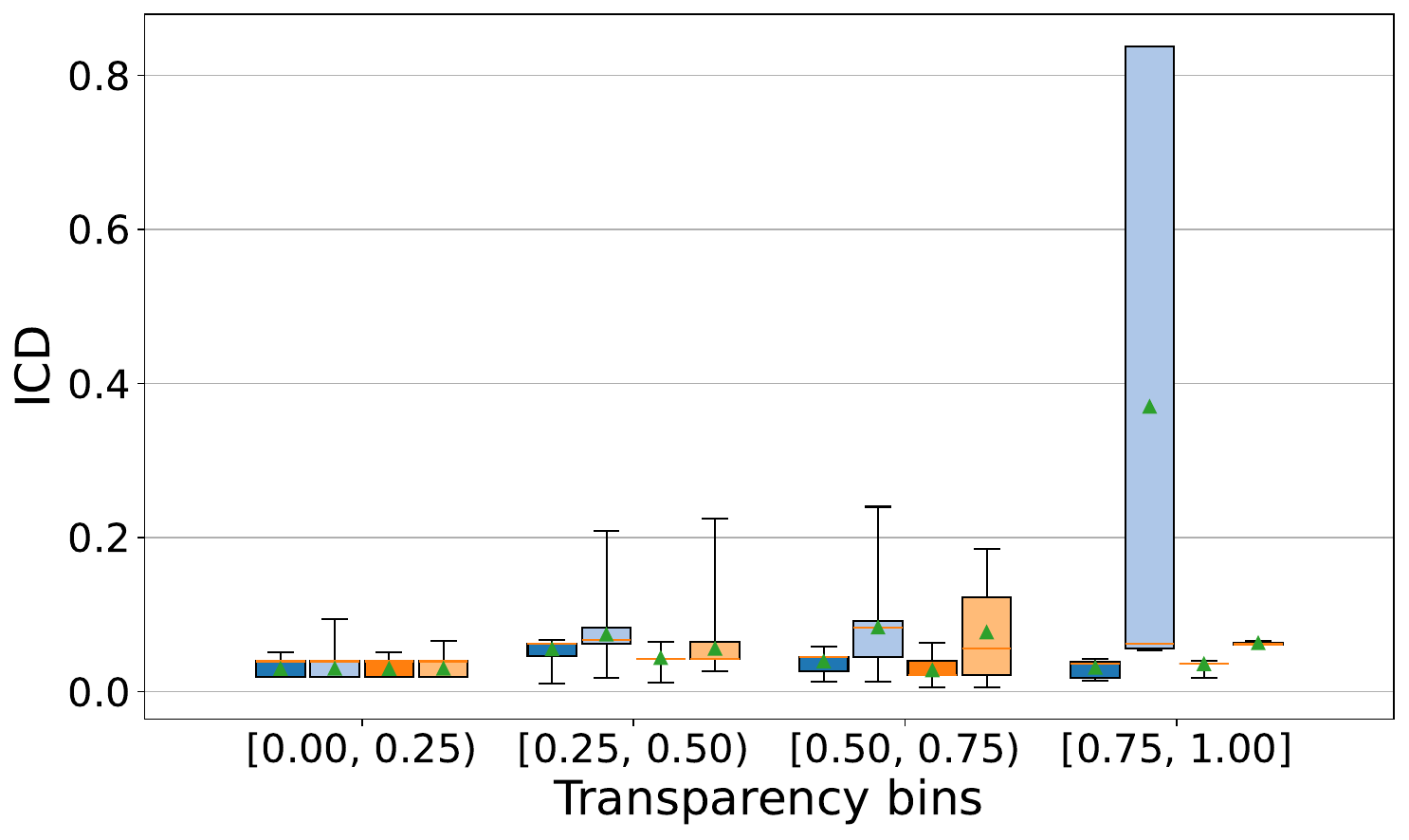}
    \caption{ACS Employment dataset}\label{fig:Fair_max_ICF_acs}
\end{subfigure}

\vspace{0.5em}

\begin{subfigure}{0.85\textwidth}
    \centering
    \includegraphics[width=0.32\linewidth]{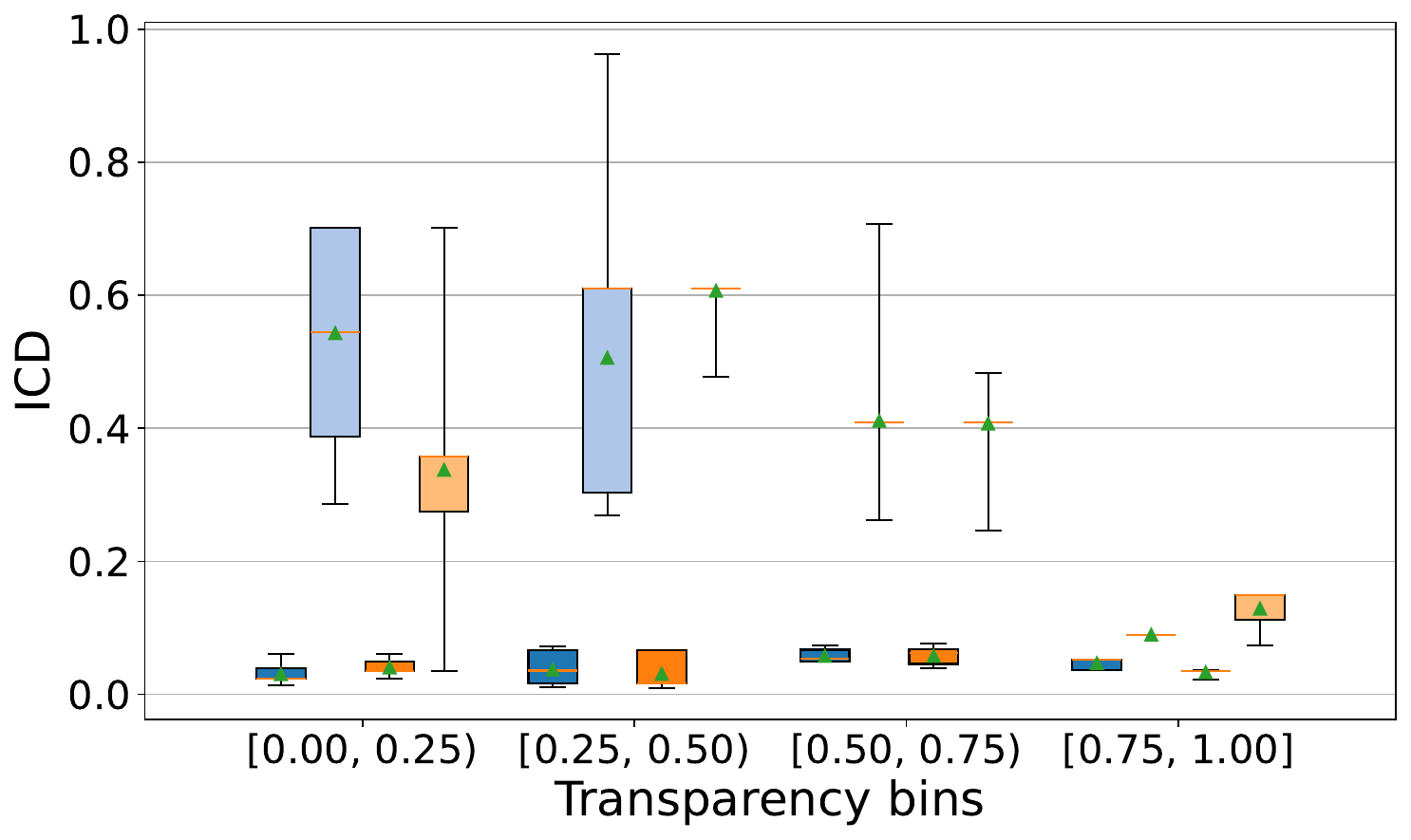}
    \includegraphics[width=0.32\linewidth]{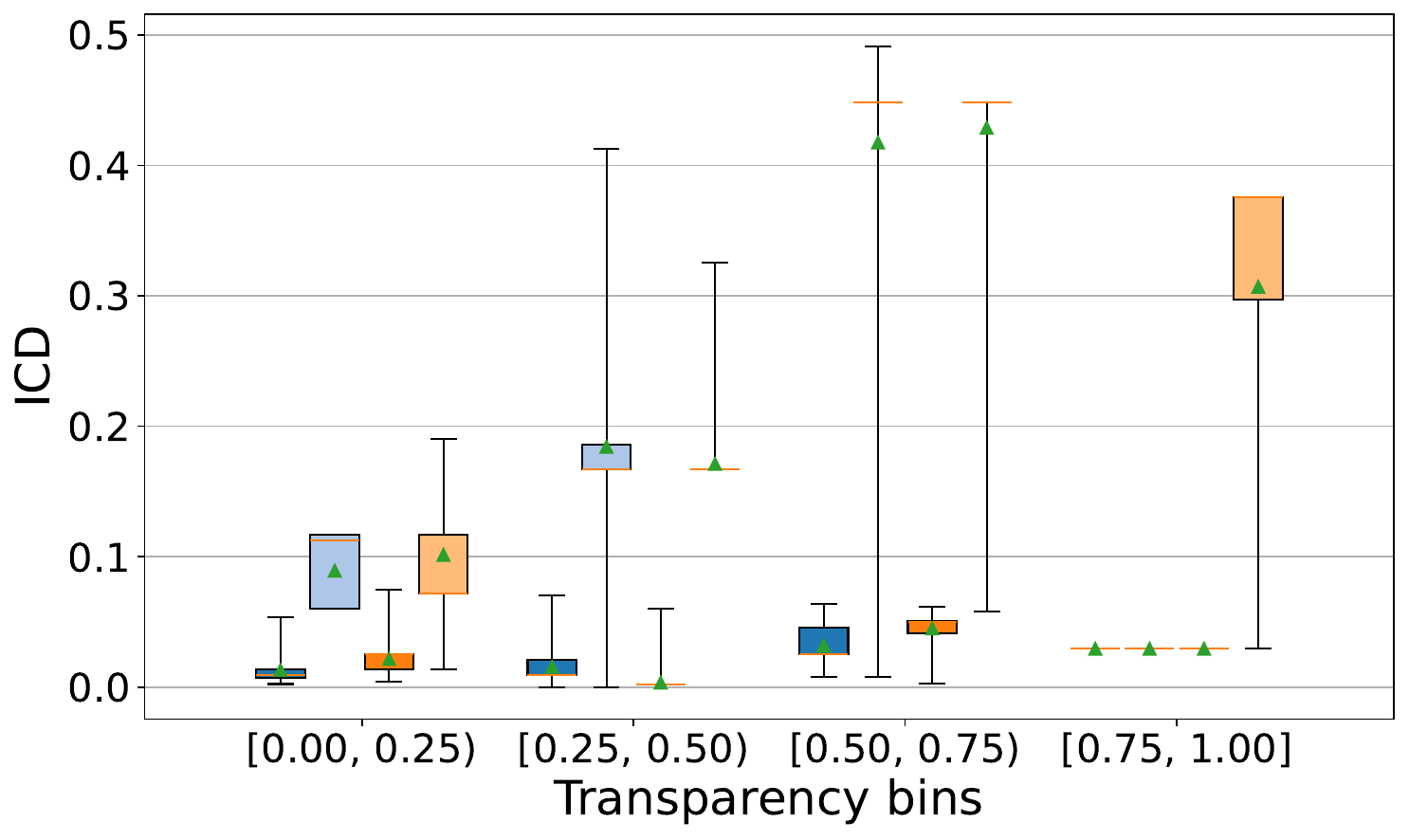}
    \includegraphics[width=0.32\linewidth]{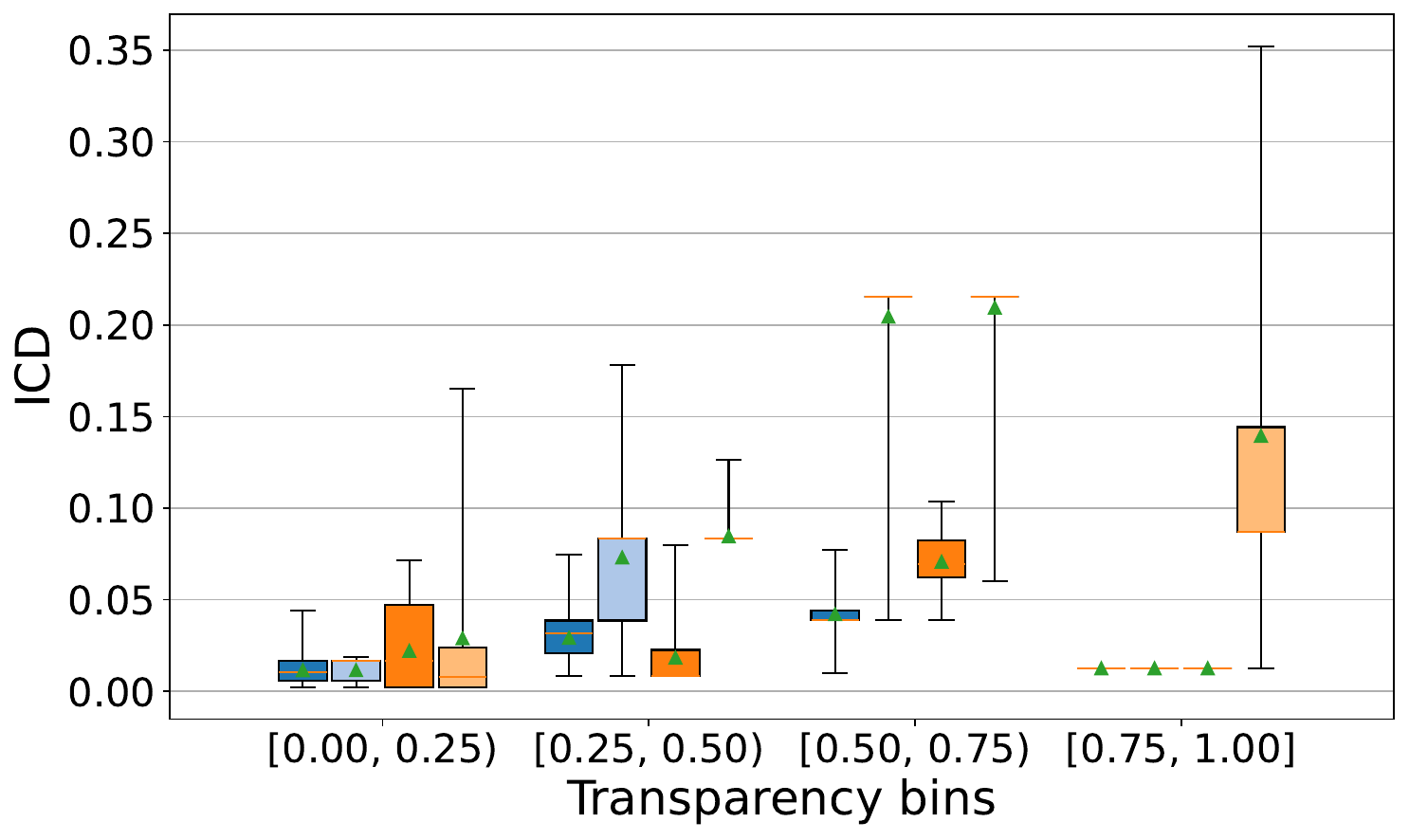}
    \caption{UCI Adult Income dataset}\label{fig:Fair_max_ICF_adult}
\end{subfigure}

\caption{Distribution of test set ICD across Rashomon sets over transparency bins $(\varepsilon = 0.01)$ for HybridCORELSPre and HybridCORELSPost, with or without our proposed ICD mitigation $(\eta = 0.05)$. Each subplot corresponds to mitigation applied with respect to a sensitive attribute (Age, Gender, Race), ordered from left to right columns. Results are reported for the ACS Employment and UCI Adult Income datasets.}

\label{fig:Fair_max_ICF} 
\end{figure*}

Additional experimental analyses evaluating the effectiveness of ICD mitigation are provided in this section, complementing the representative results discussed in Section~\ref{sec:mitigation_results}. In particular, Figure~\ref{fig:Fair_max_ICF} demonstrates the strong generalization ability of the proposed mitigation strategy on unseen test data for the ACS Employment and UCI Adult Income datasets, where both the average magnitude and variability of ICD across Rashomon sets of transparency bins are substantially reduced compared to the unconstrained setting.

\begin{figure*}[t]
\centering

\includegraphics[width=0.8\textwidth]{Plots/ICF/HybridCORELS_fairness_shared_legend.pdf}

\begin{subfigure}{0.85\textwidth}
    \centering
    \includegraphics[width=0.32\linewidth]{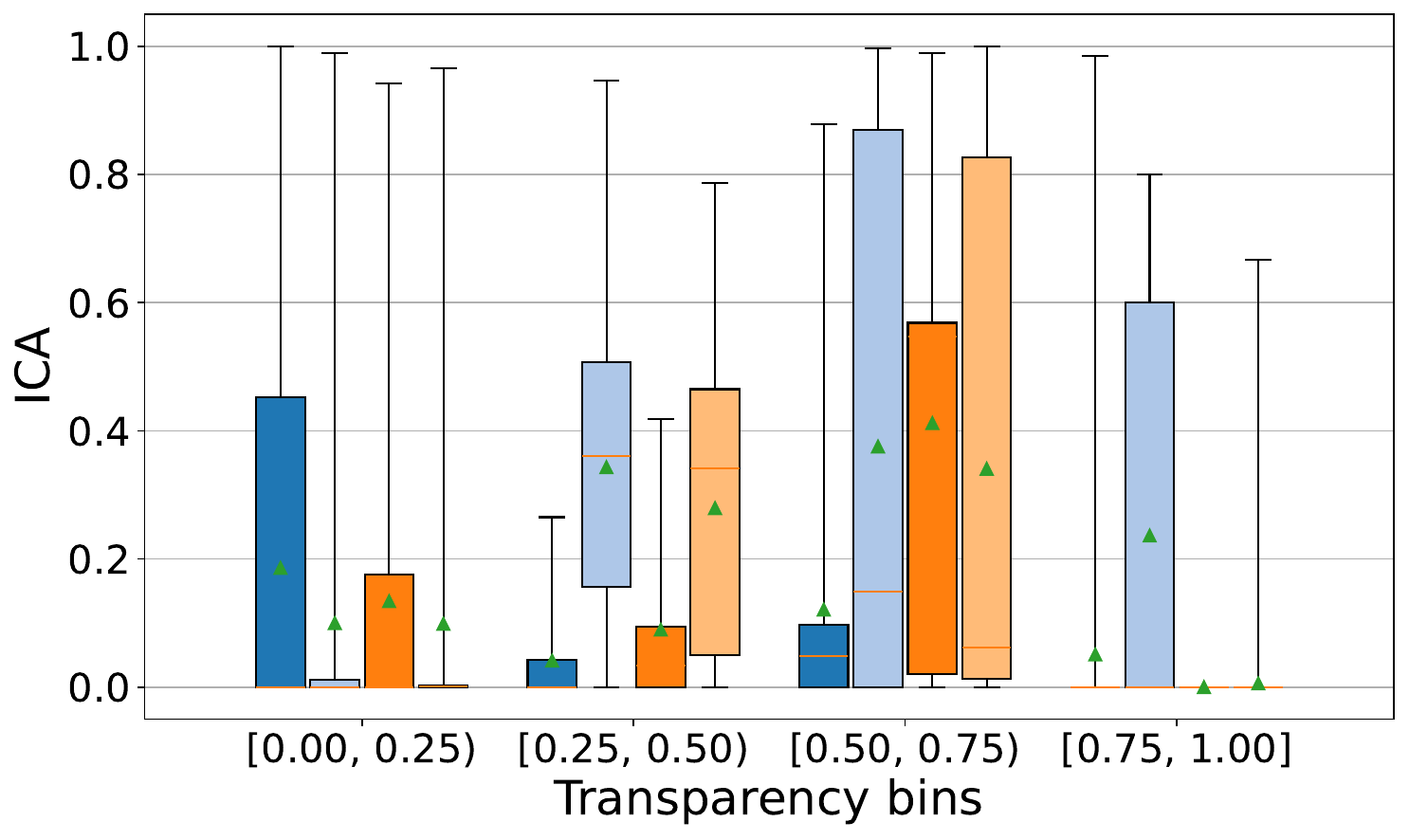}
    \includegraphics[width=0.32\linewidth]{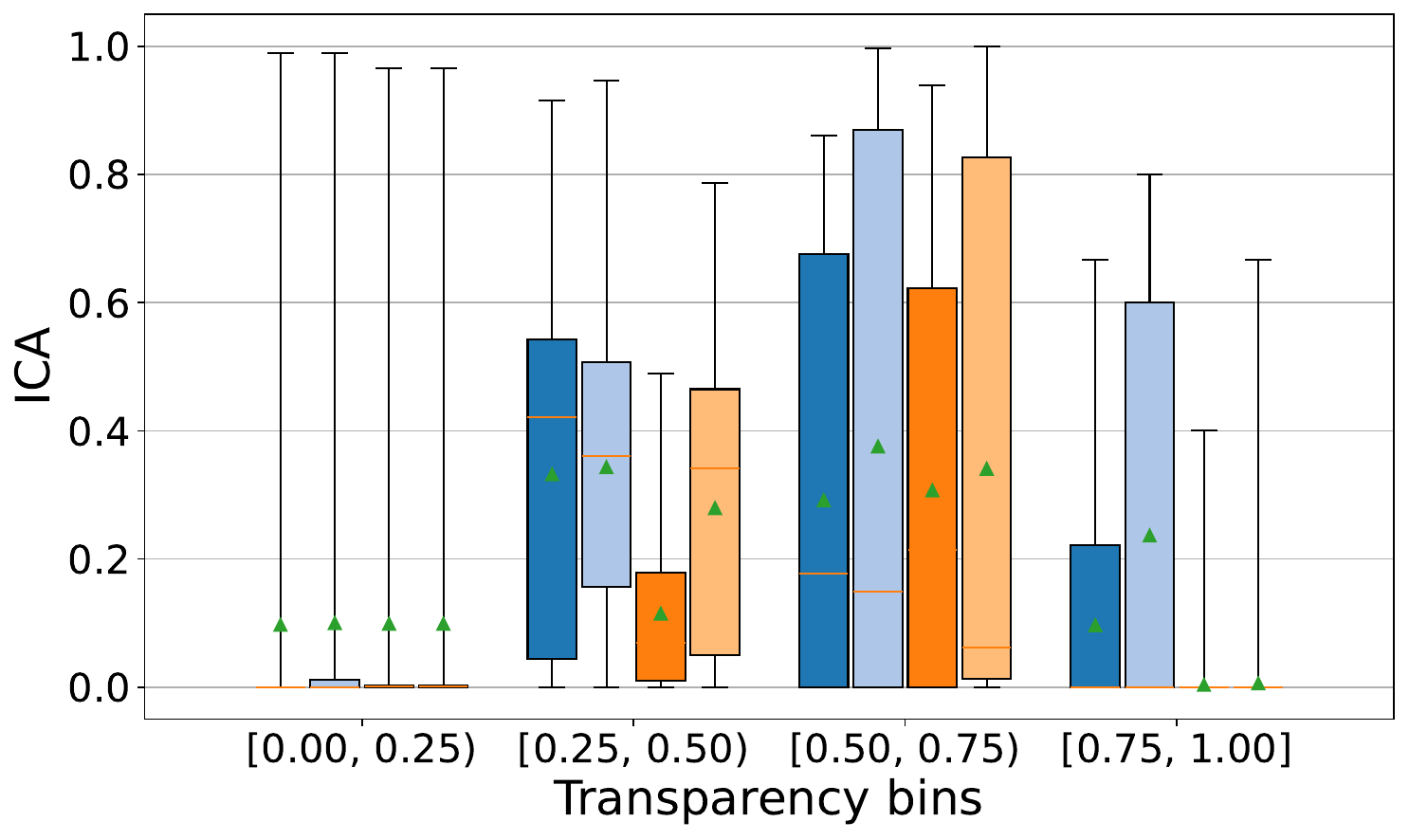}
    \includegraphics[width=0.32\linewidth]{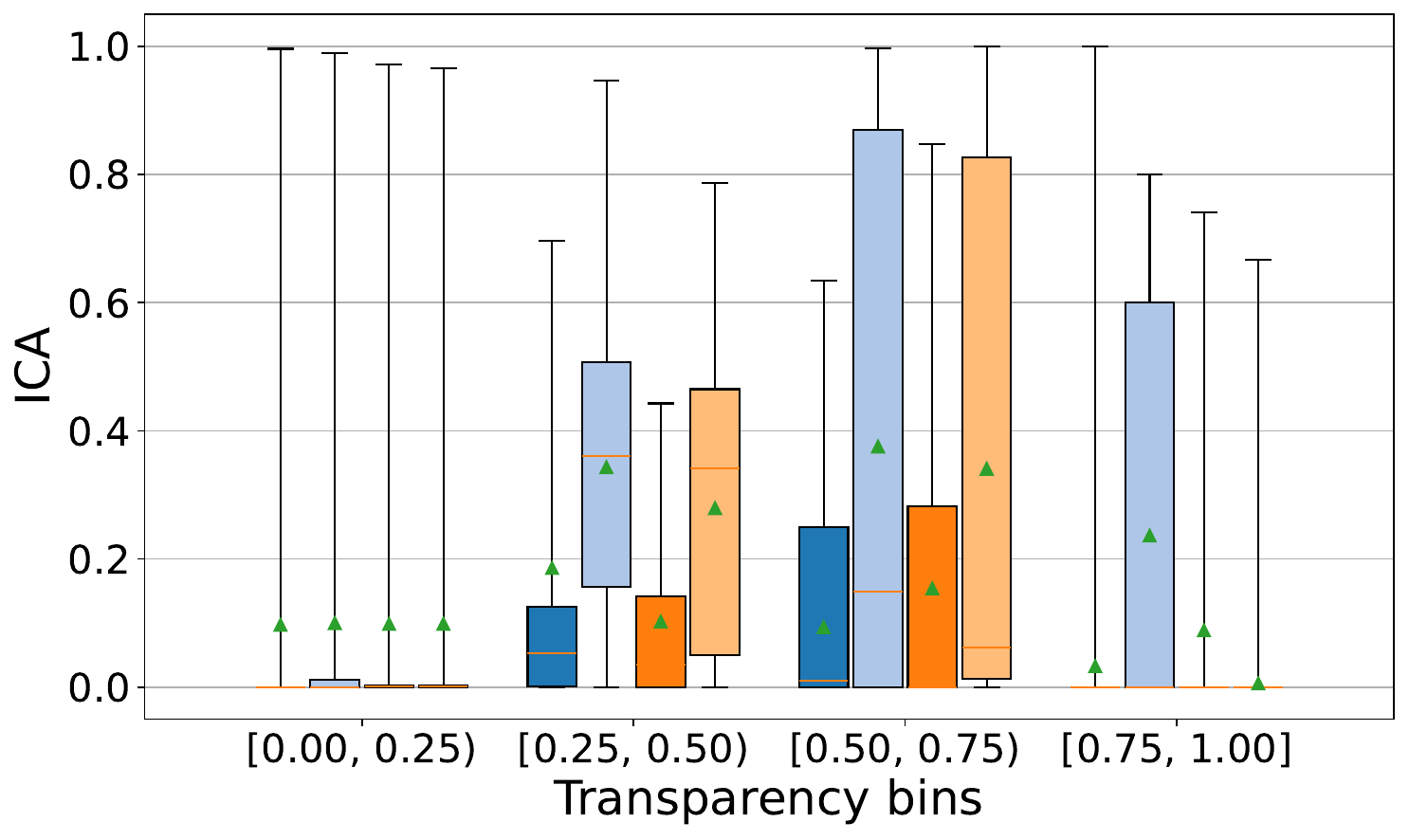}
    \caption{ACS Employment dataset}\label{fig:Fair_ICA_acs}
\end{subfigure}

\vspace{0.5em}

\begin{subfigure}{0.85\textwidth}
    \centering
    \includegraphics[width=0.32\linewidth]{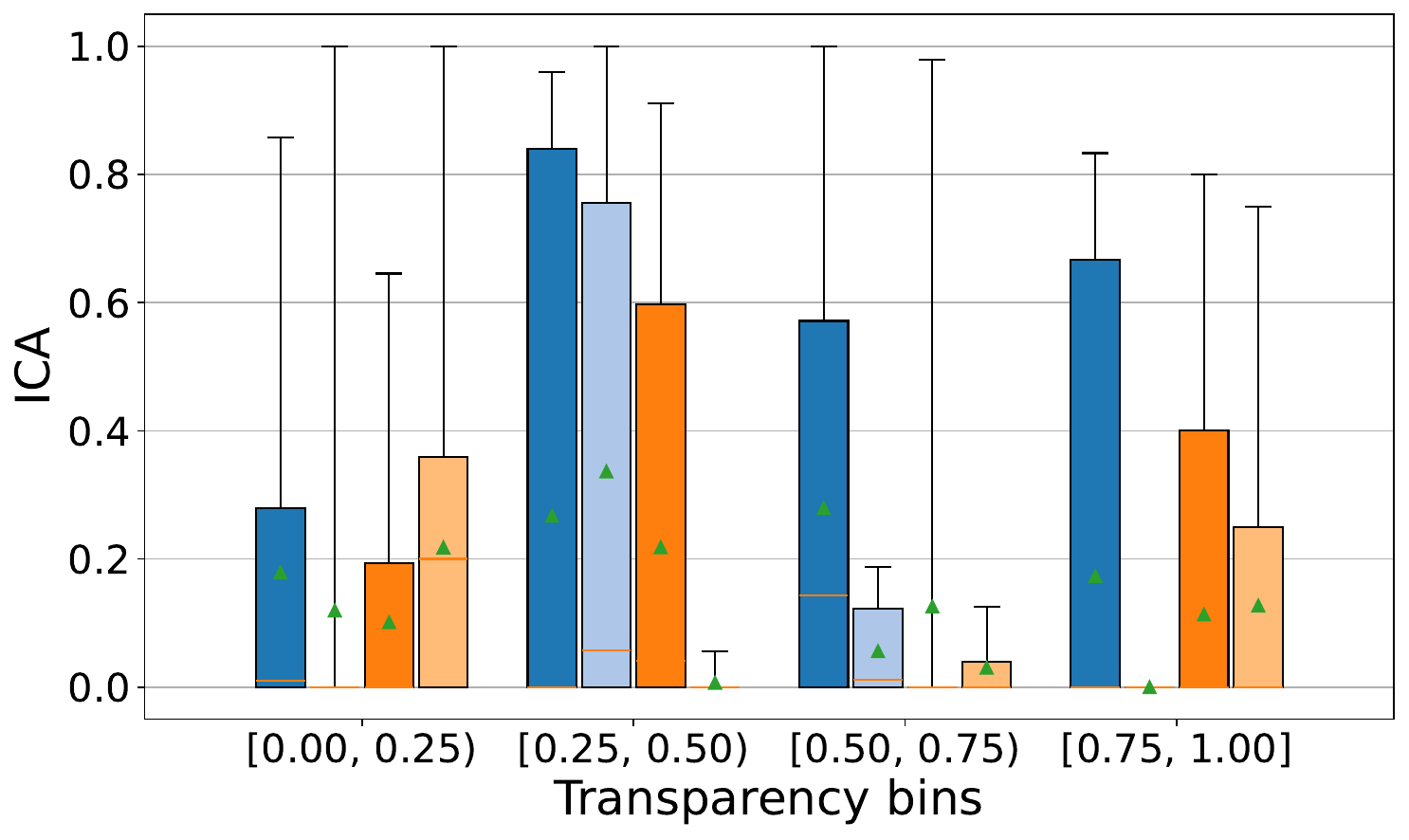}
    \includegraphics[width=0.32\linewidth]{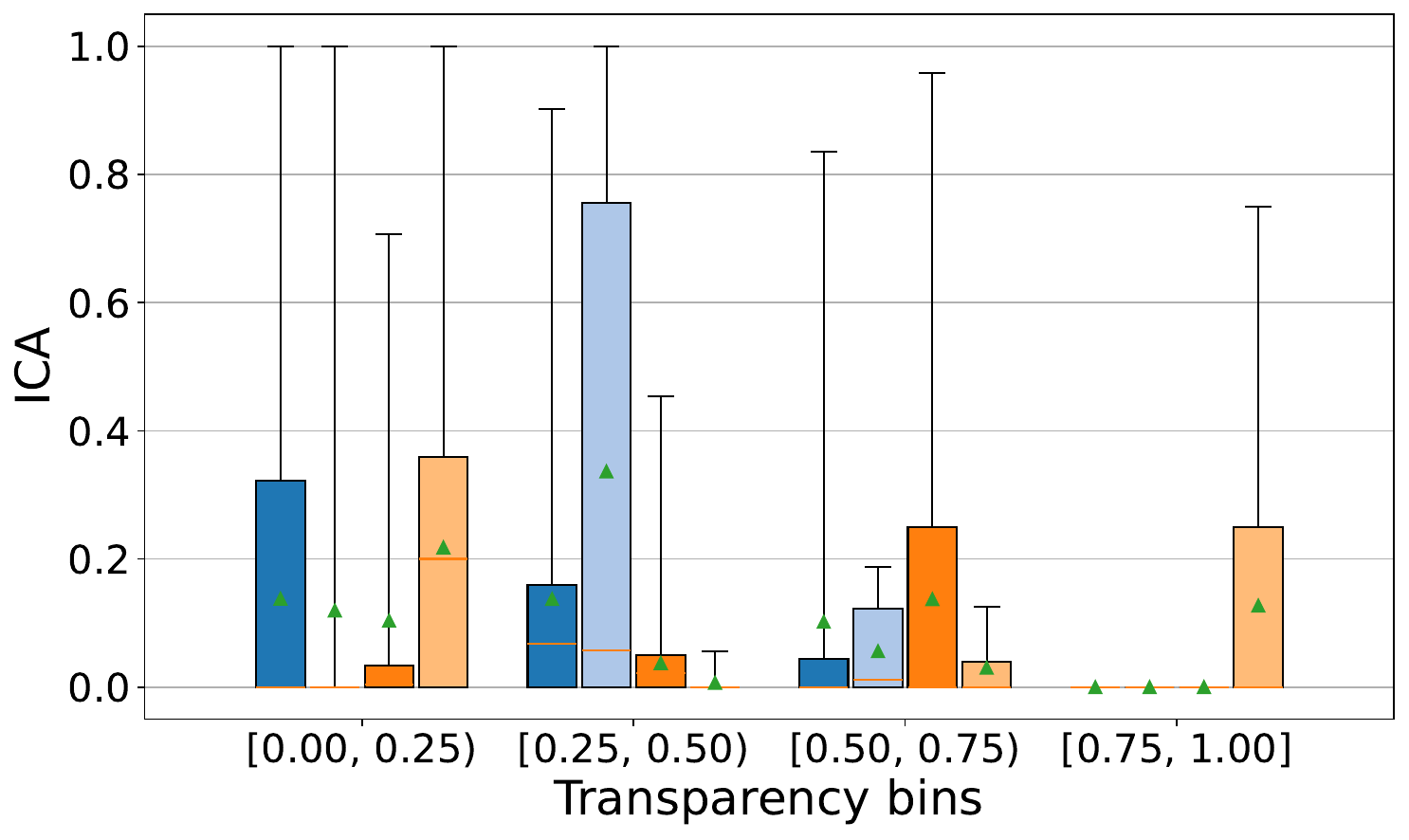}
    \includegraphics[width=0.32\linewidth]{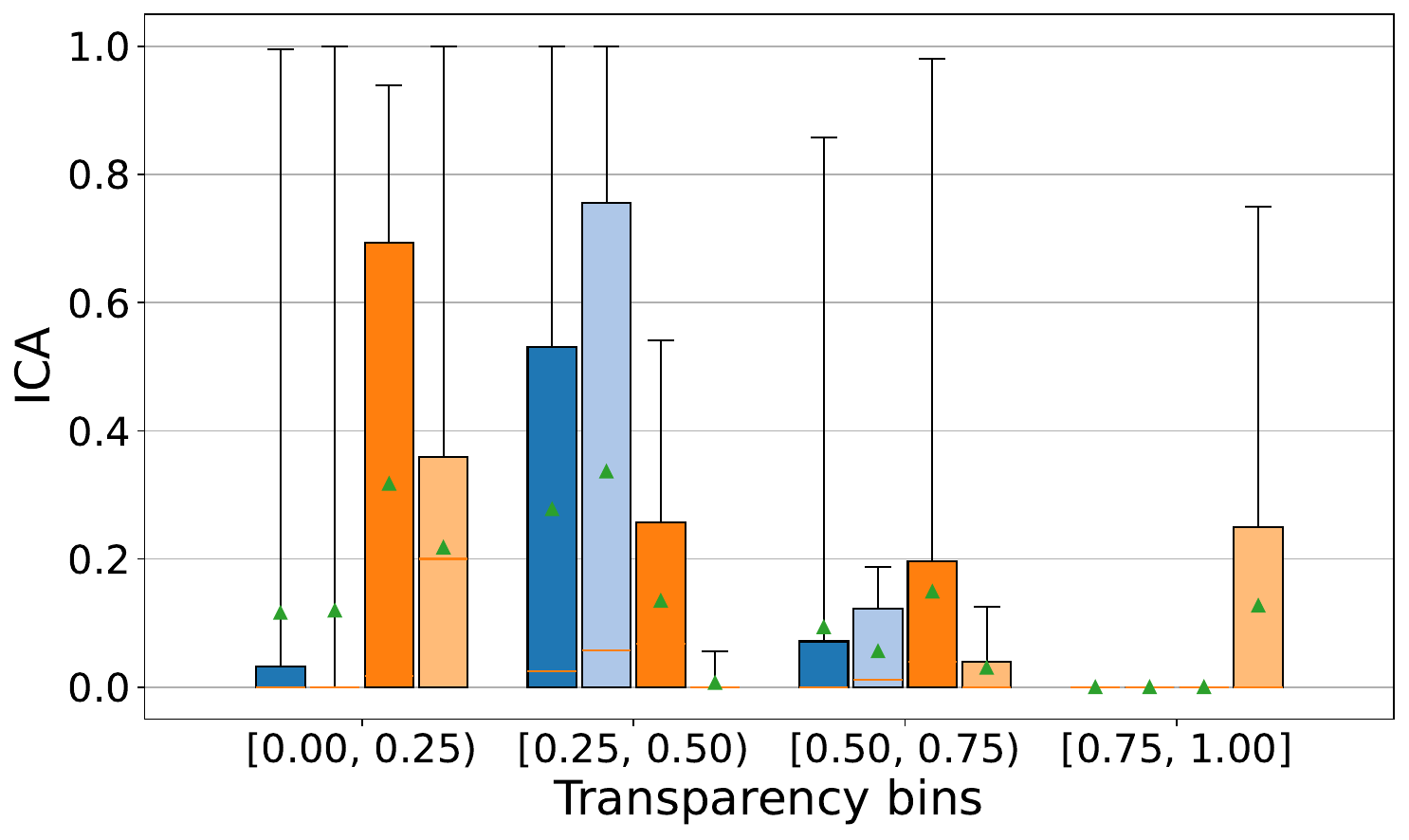}
    \caption{UCI Adult Income dataset}\label{fig:Fair_ICA_adult}
\end{subfigure}

\vspace{0.5em}

\caption{Distribution of test set ICA across Rashomon sets over transparency bins $(\varepsilon = 0.01)$ for HybridCORELSPre and HybridCORELSPost, with or without our proposed ICD mitigation $(\eta = 0.05)$. Each subplot corresponds to mitigation applied with respect to a sensitive attribute (Age, Gender, Race), ordered from left to right columns. Results are reported for the ACS Employment and UCI Adult Income datasets.}
\label{fig:Fair_ICA}
\end{figure*}

Moreover, the distribution of test set ICA within Rashomon sets of different transparency bins, on the ACS Employment and UCI Adult Income datasets are shown in Figure~\ref{fig:Fair_ICA}. The results further confirm that ICD mitigation has only a modest impact on assignment arbitrariness across Rashomon sets.

\begin{figure*}[t]
\centering

\includegraphics[width=0.8\textwidth]{Plots/ICF/HybridCORELS_fairness_shared_legend.pdf}

\begin{subfigure}{0.85\textwidth}
    \centering
    \includegraphics[width=0.32\linewidth]{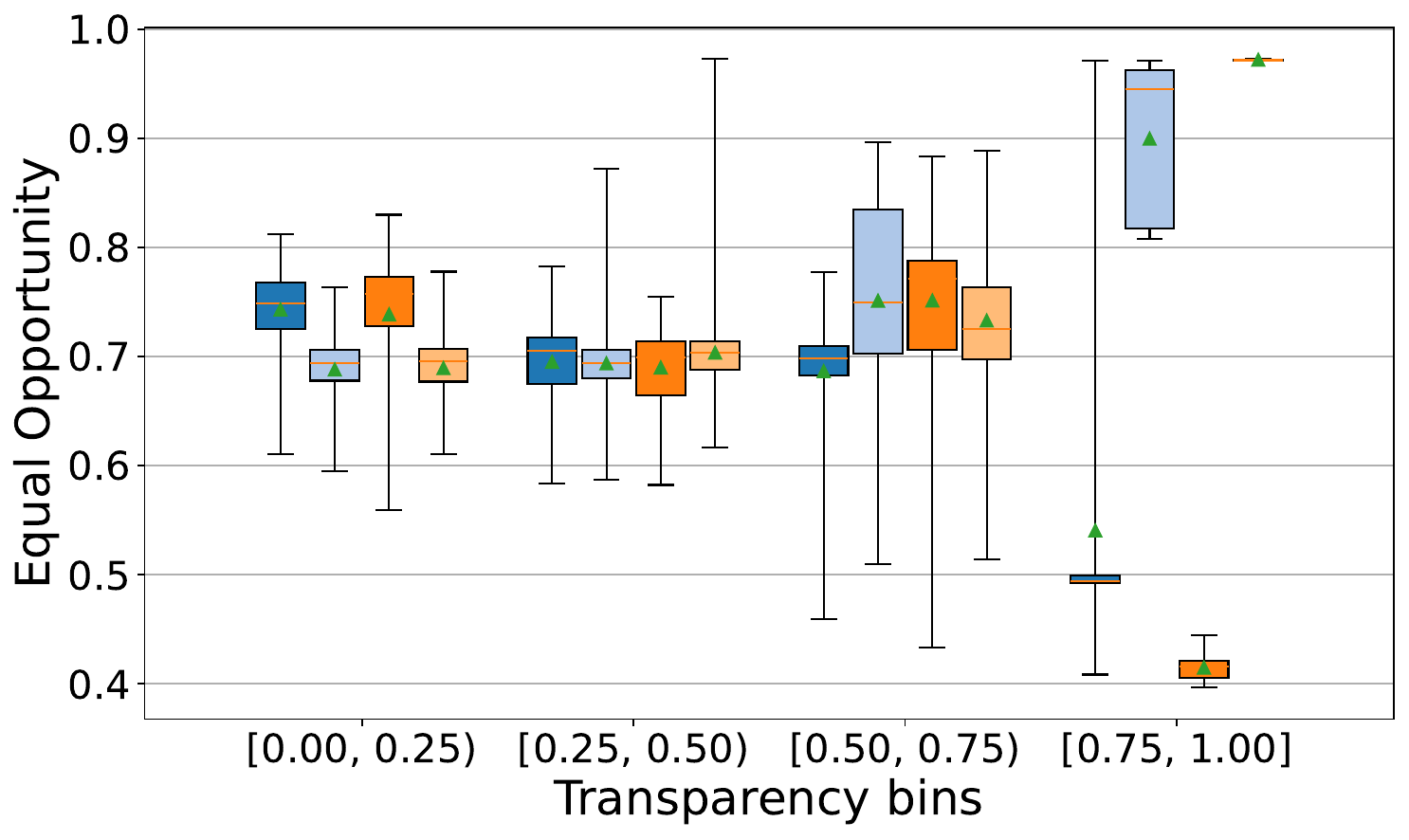}
    \includegraphics[width=0.32\linewidth]{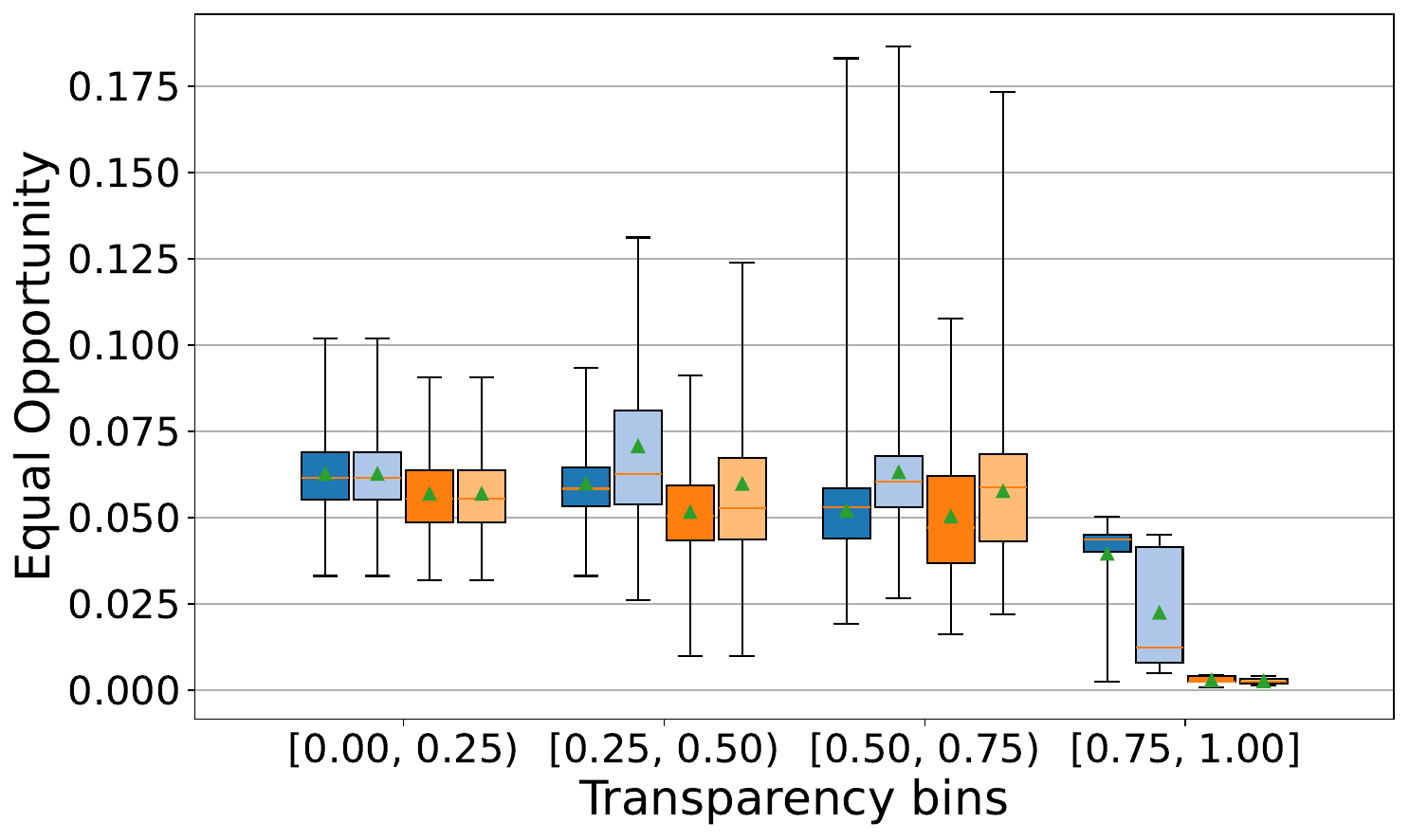}
    \includegraphics[width=0.32\linewidth]{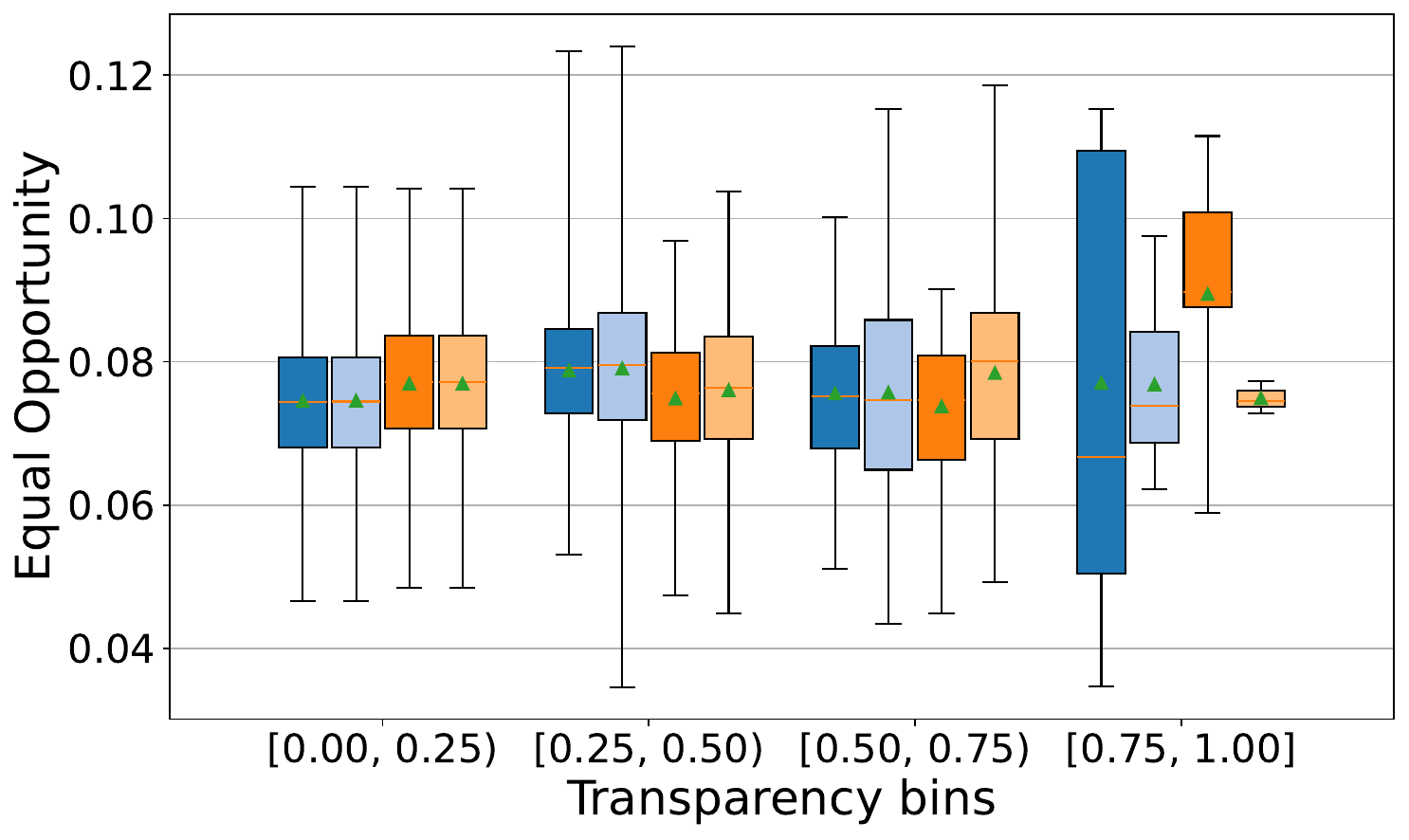}
    \caption{ACS Employment dataset}\label{fig:fig:Fair_max_EO_acs}
\end{subfigure}

\vspace{0.5em}

\begin{subfigure}{0.85\textwidth}
    \centering
    \includegraphics[width=0.32\linewidth]{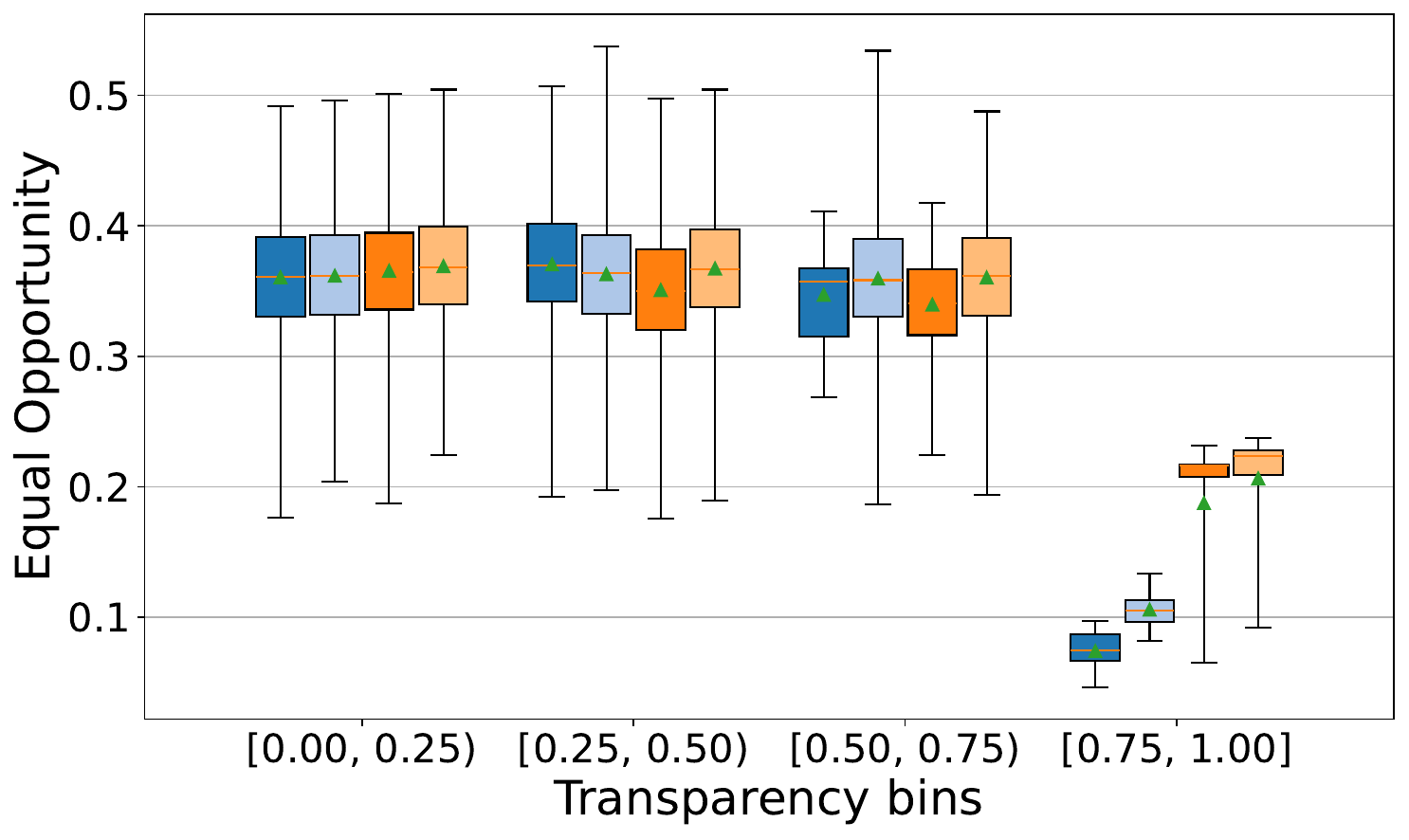}
    \includegraphics[width=0.32\linewidth]{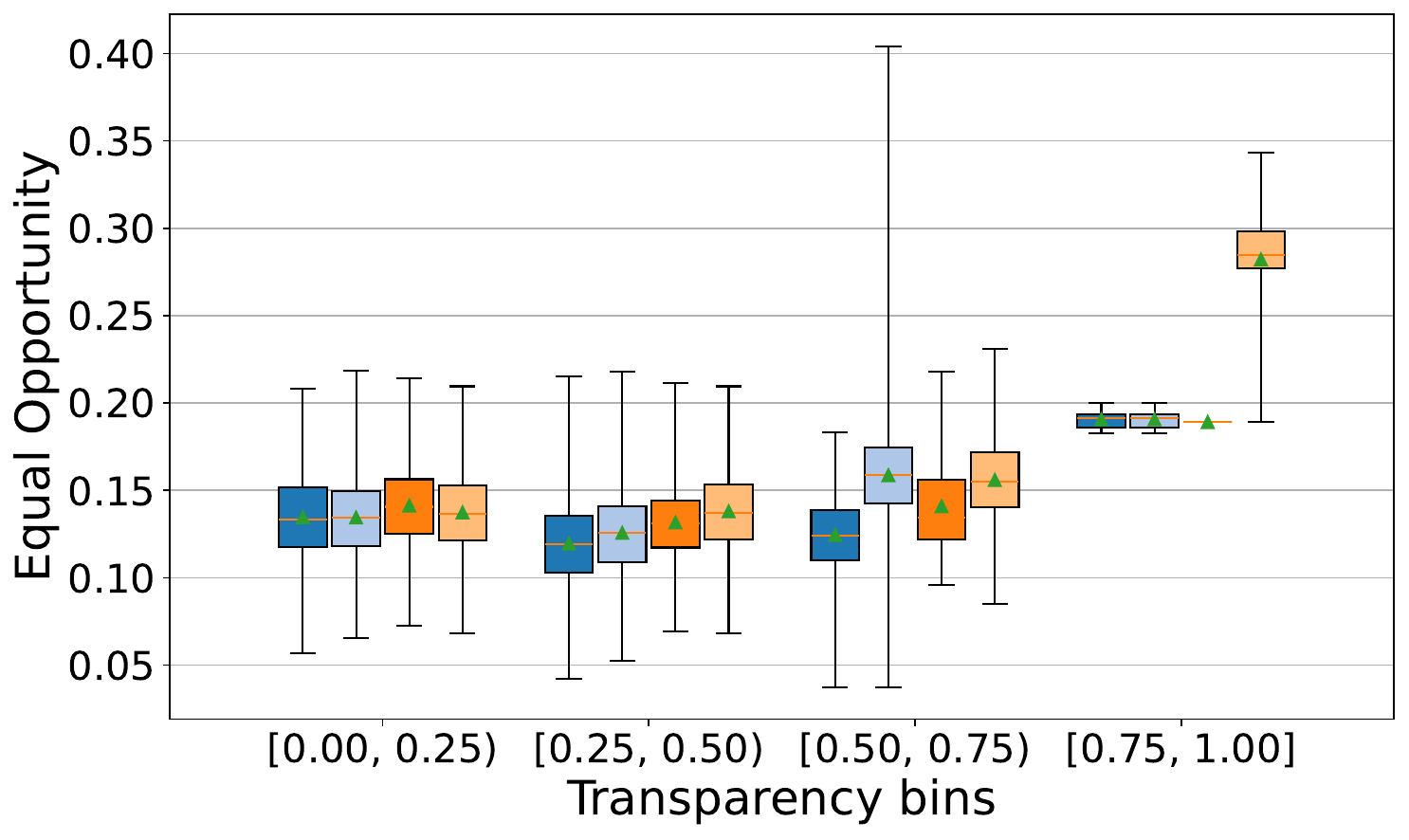}
    \includegraphics[width=0.32\linewidth]{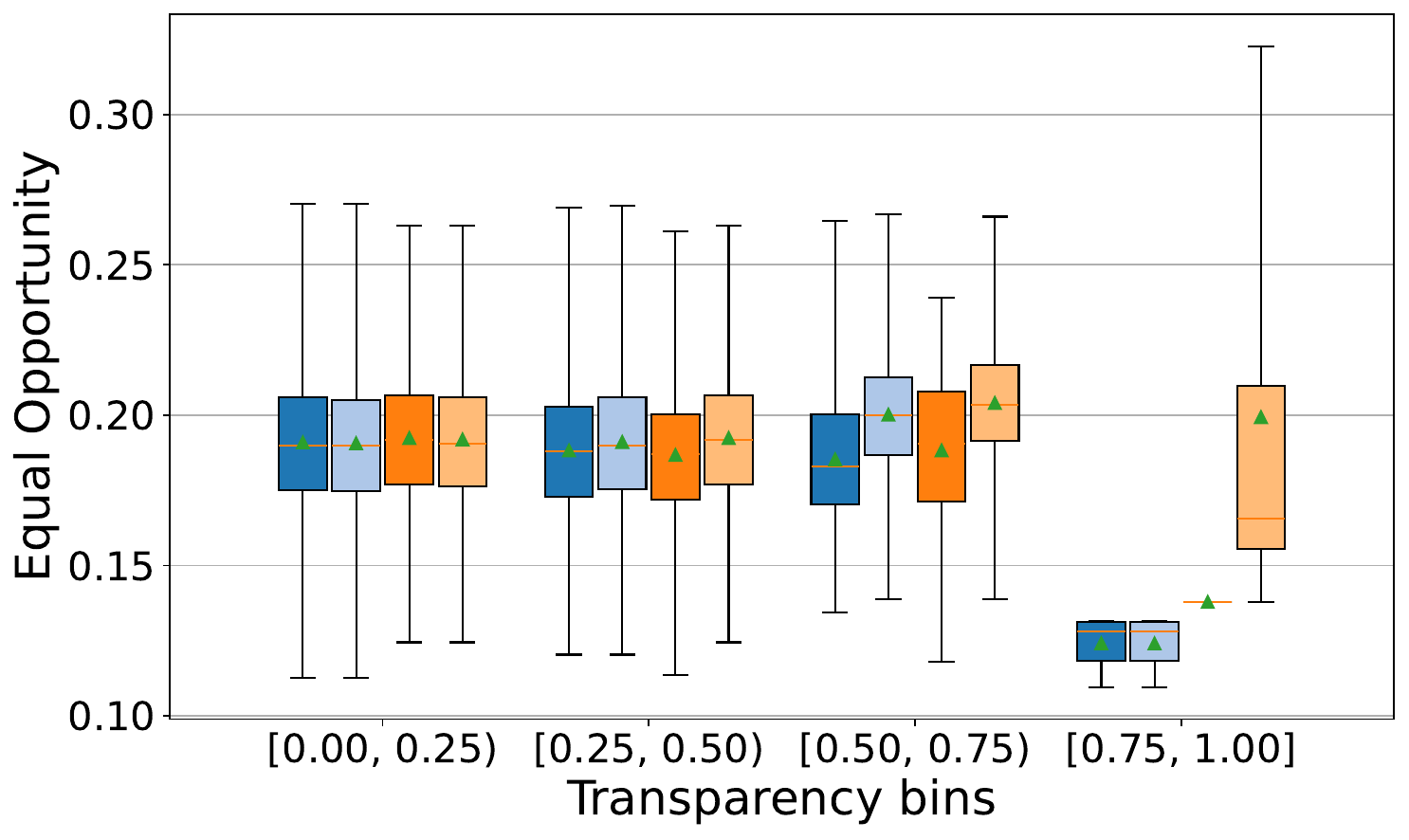}
    \caption{UCI Adult Income dataset}\label{fig:fig:Fair_max_EO_adult}
\end{subfigure}

\vspace{0.5em}

\caption{Distribution of test set Equal Opportunity (EO) across Rashomon sets over transparency bins $(\varepsilon = 0.01)$ for HybridCORELSPre and HybridCORELSPost, before and after enforcing maximum ICD constraints $(\eta = 0.05)$ during training. Each subplot corresponds to mitigation
applied with respect to a sensitive attribute (Age, Gender, Race),ordered from left to right columns. Results are reported for the ACS Employment and UCI Adult Income datasets.}
\label{fig:Fair_max_EO}
\end{figure*}

\begin{figure*}[t]
\centering
\includegraphics[width=0.8\textwidth]{Plots/ICF/HybridCORELS_fairness_shared_legend.pdf}
\begin{subfigure}{0.85\textwidth}
    \centering
    \includegraphics[width=0.32\linewidth]{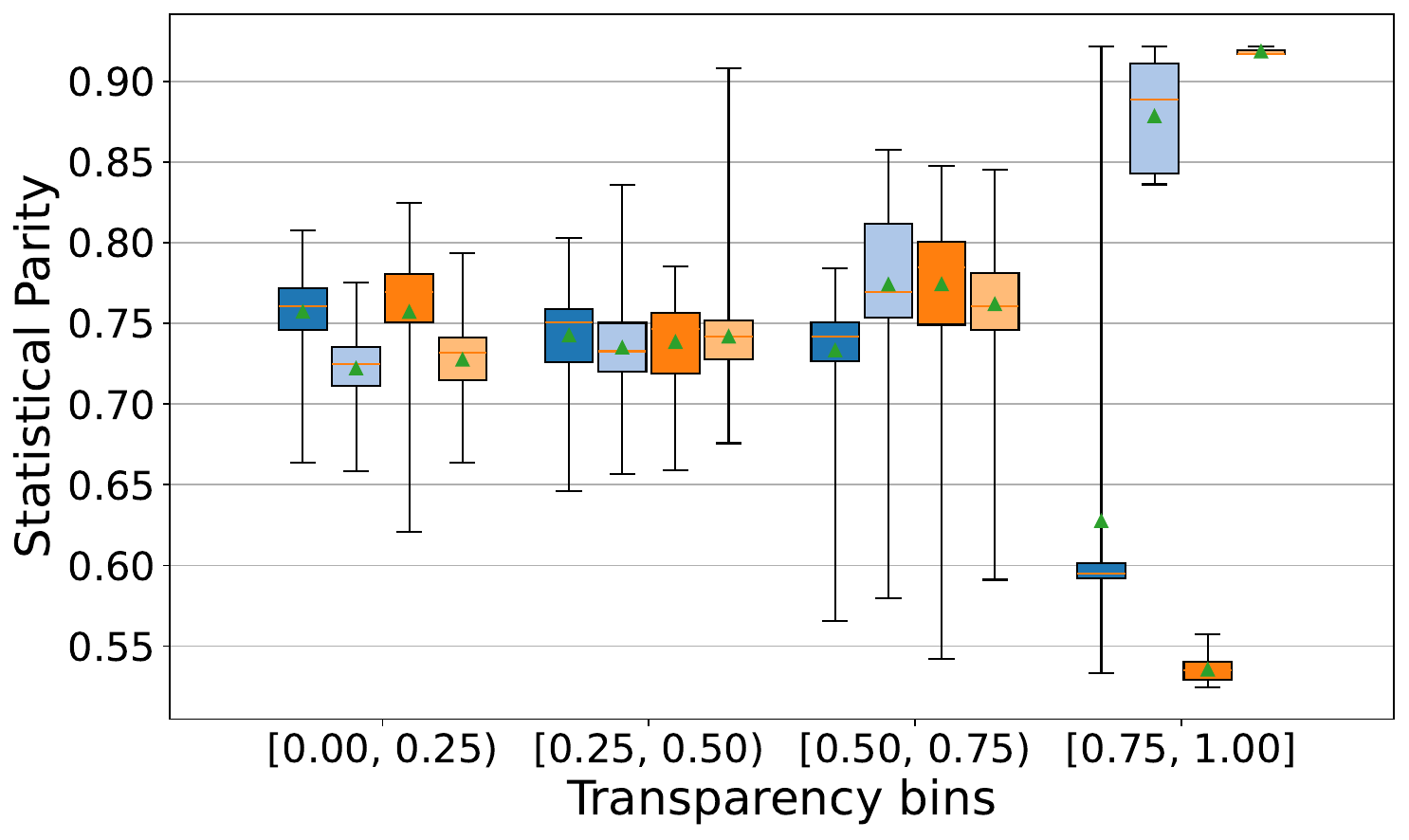}
    \includegraphics[width=0.32\linewidth]{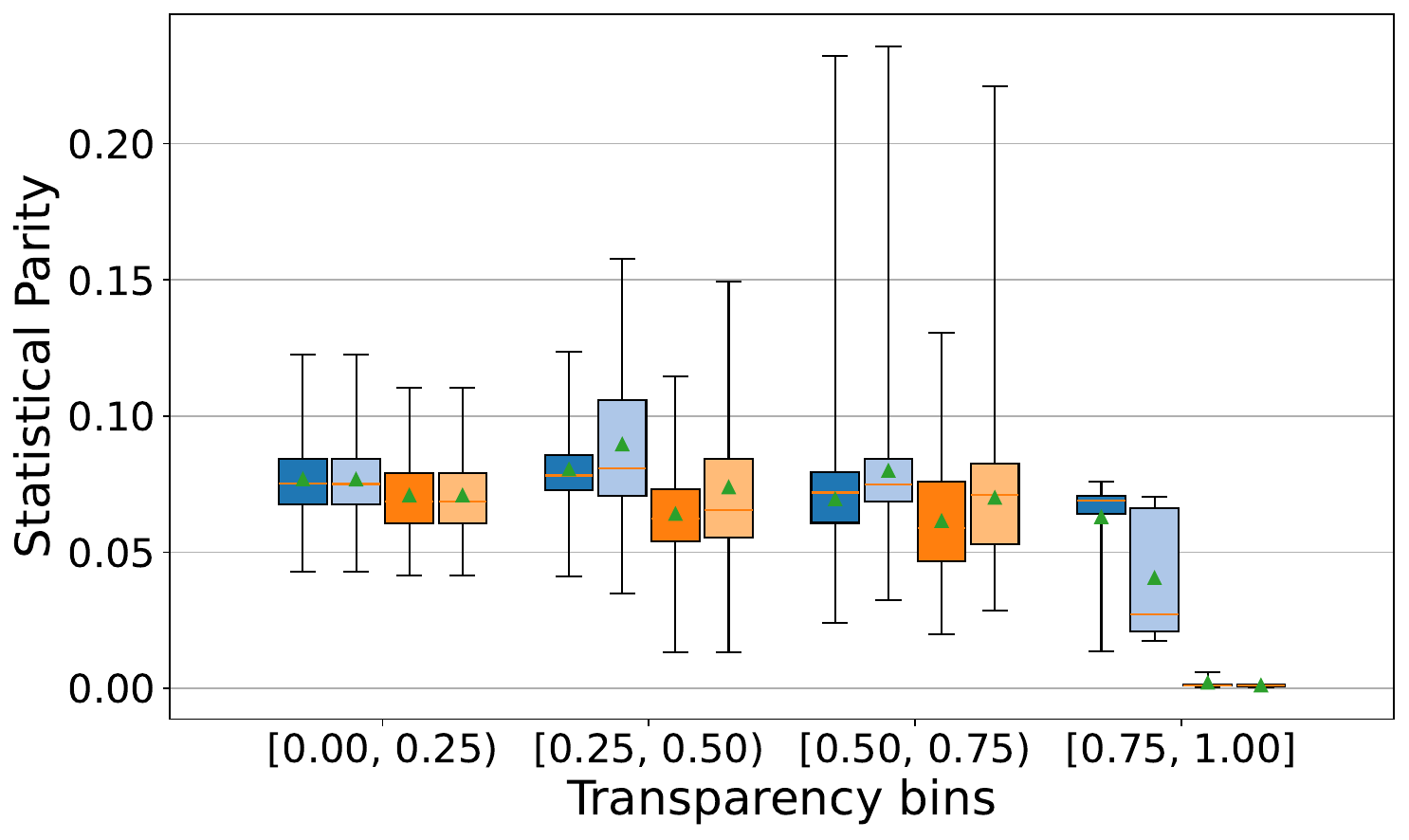}
    \includegraphics[width=0.32\linewidth]{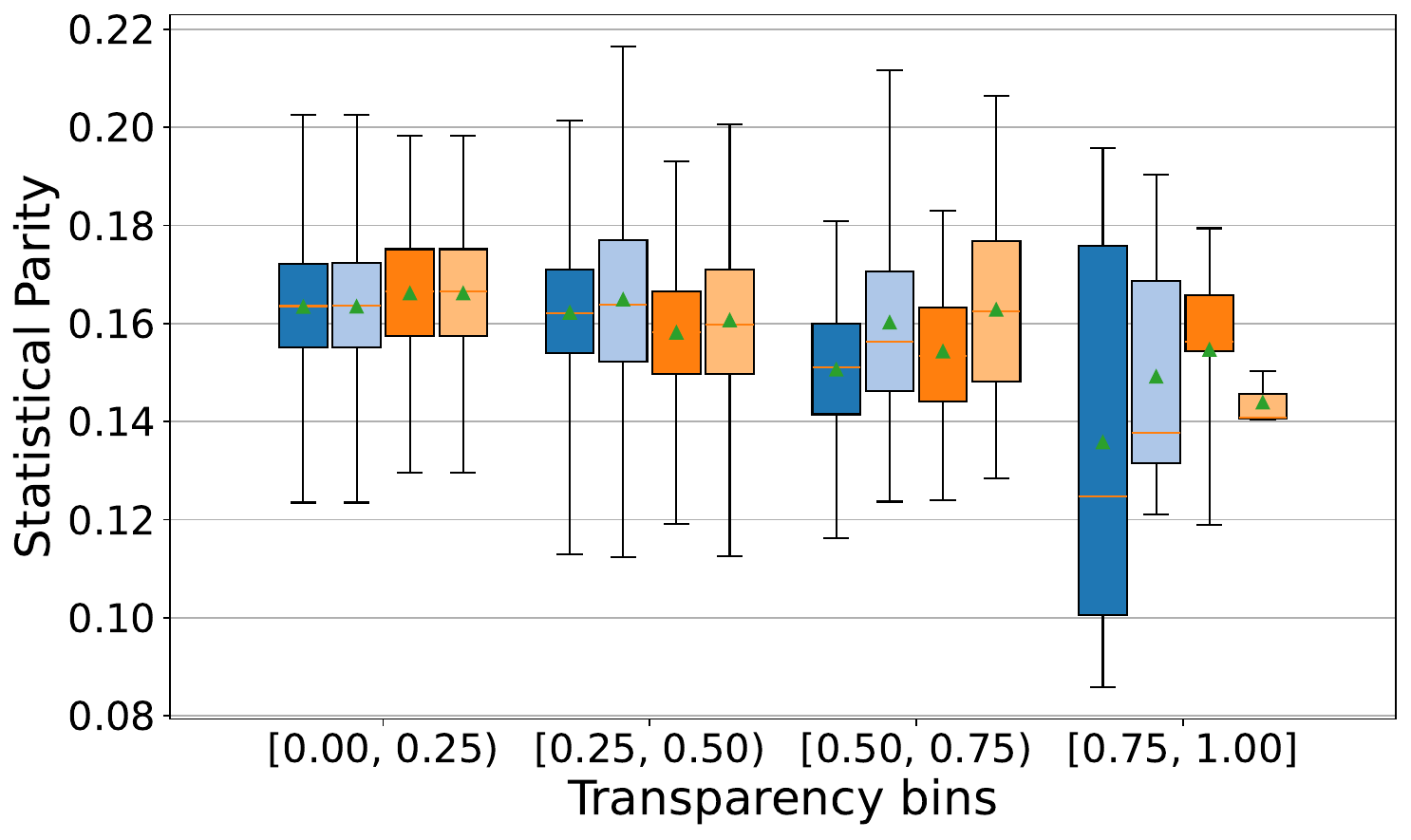}
    \caption{ACS Employment dataset}\label{fig:Fair_max_SP_acs}
\end{subfigure}

\vspace{0.5em}

\begin{subfigure}{0.85\textwidth}
    \centering
    \includegraphics[width=0.32\linewidth]{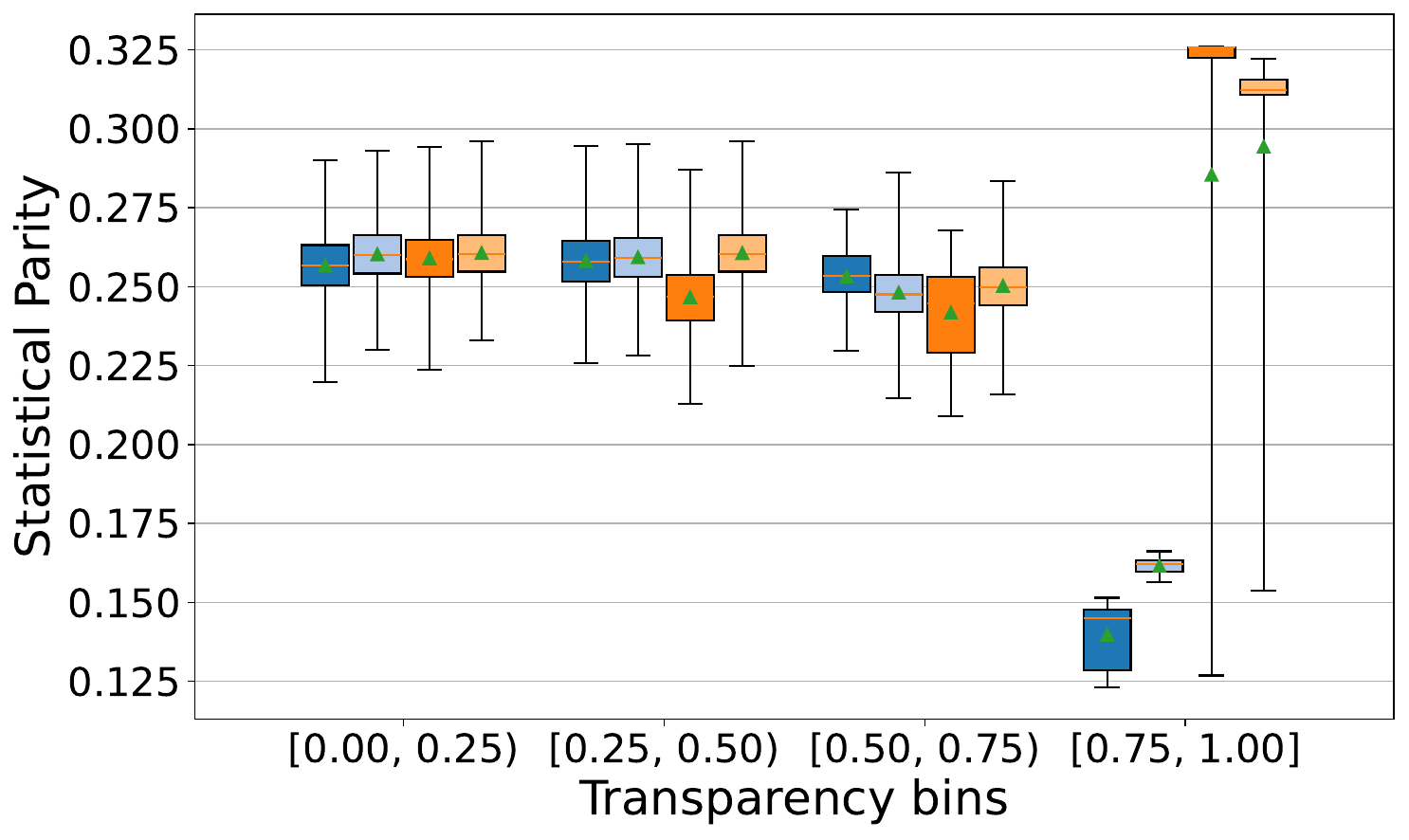}
    \includegraphics[width=0.32\linewidth]{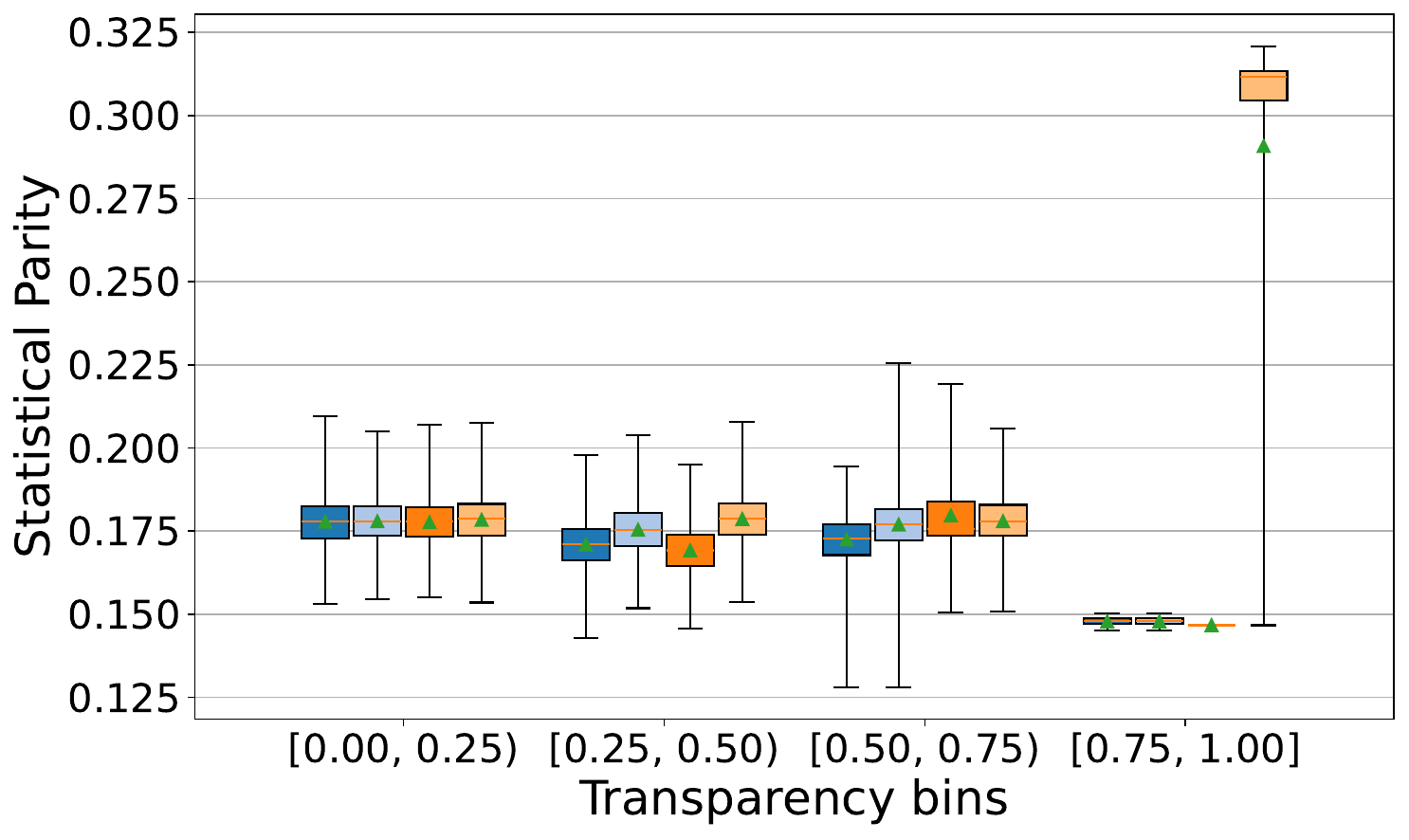}
    \includegraphics[width=0.32\linewidth]{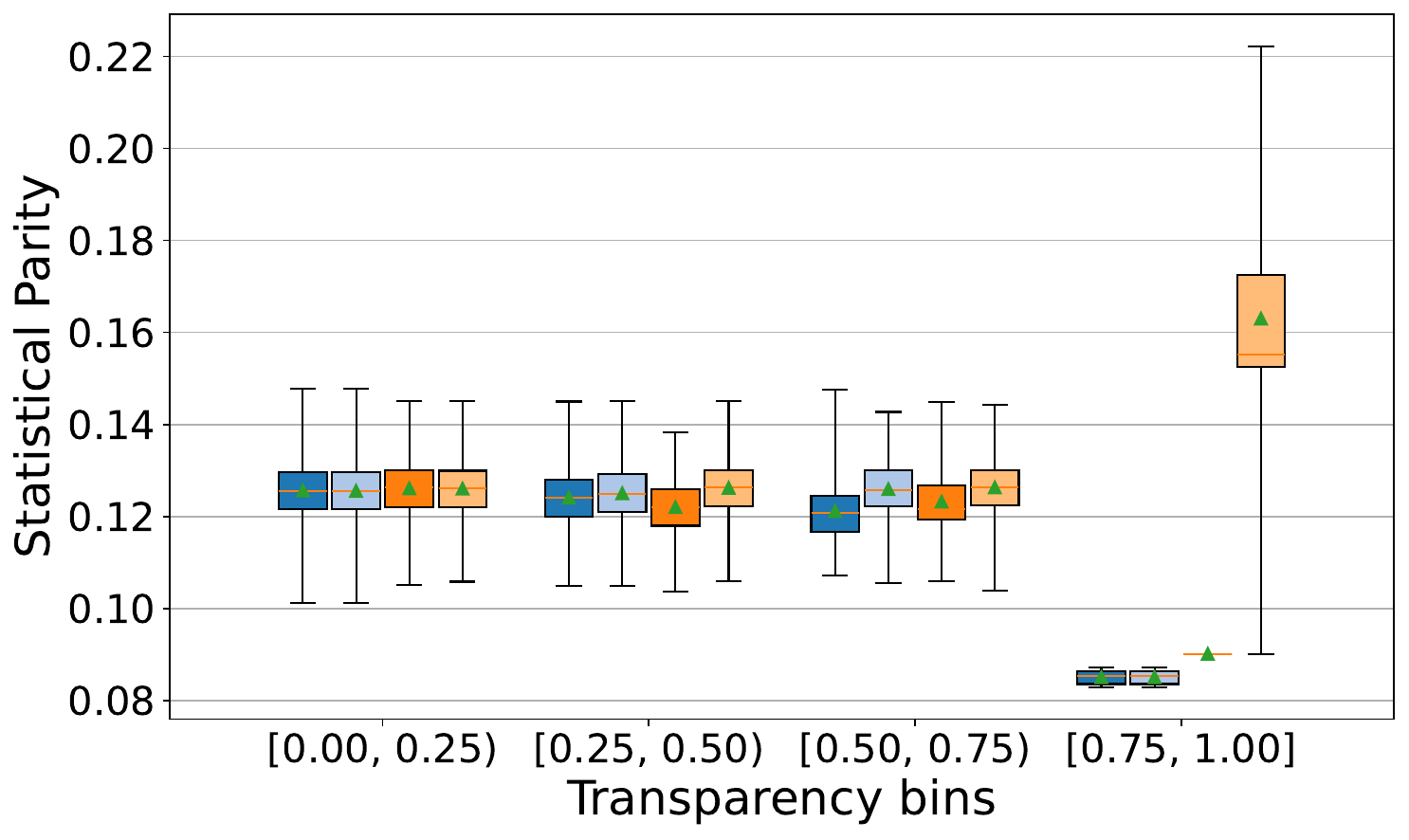}
    \caption{UCI Adult Income dataset}\label{fig:Fair_max_SP_adult}
\end{subfigure}

\vspace{0.5em}

\begin{subfigure}{0.85\textwidth}
    \centering
    \includegraphics[width=0.32\linewidth]{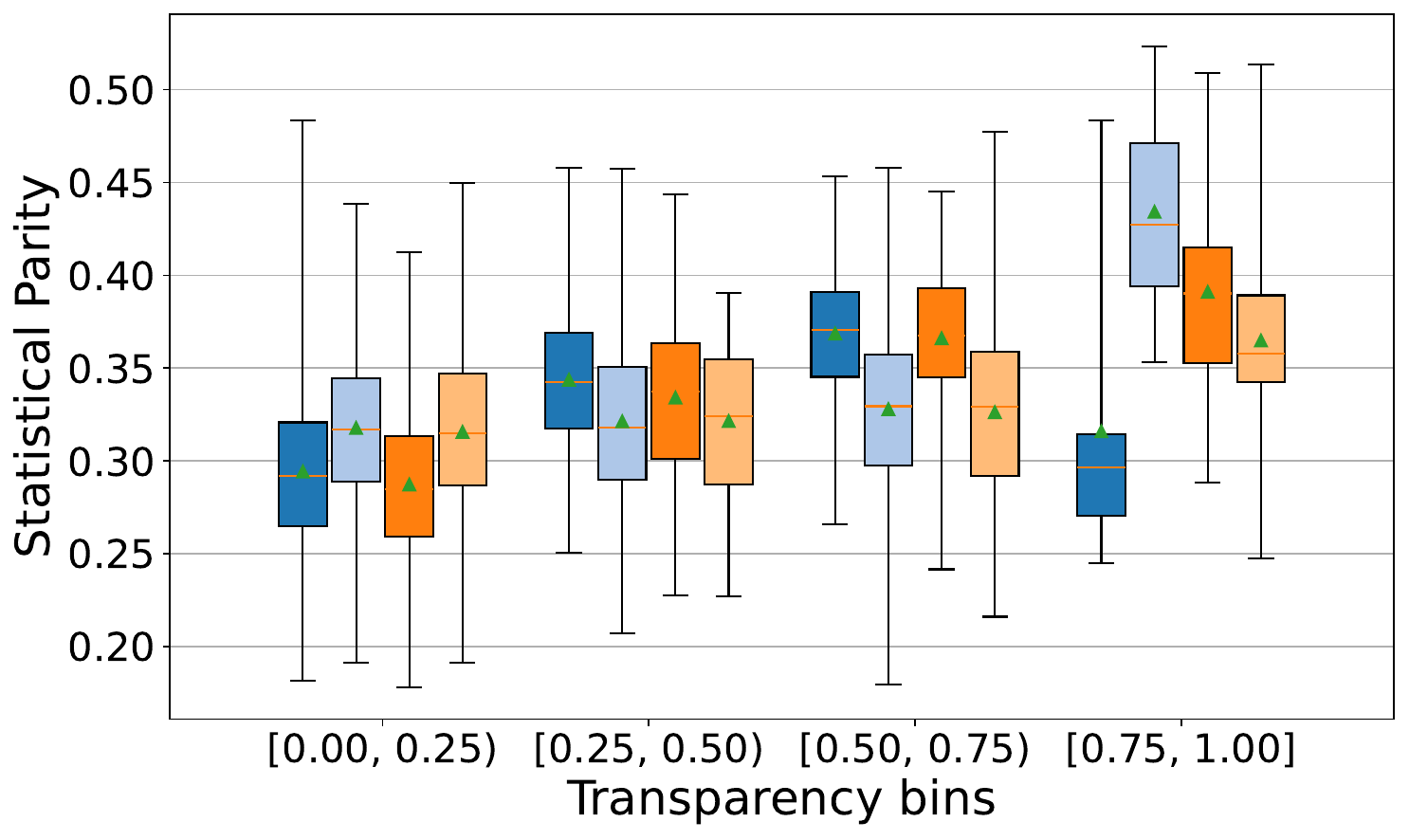}
    \includegraphics[width=0.32\linewidth]{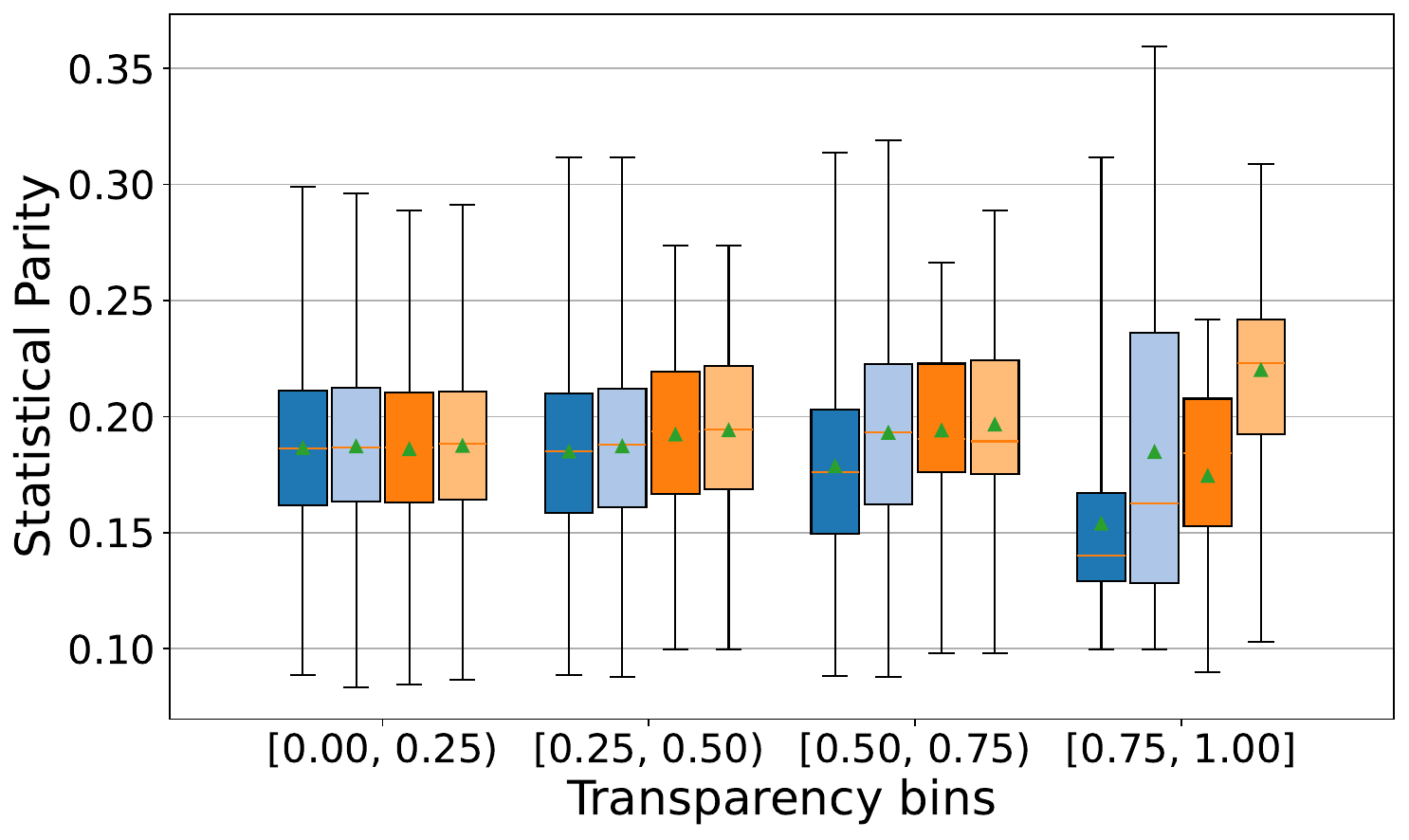}
    \includegraphics[width=0.32\linewidth]{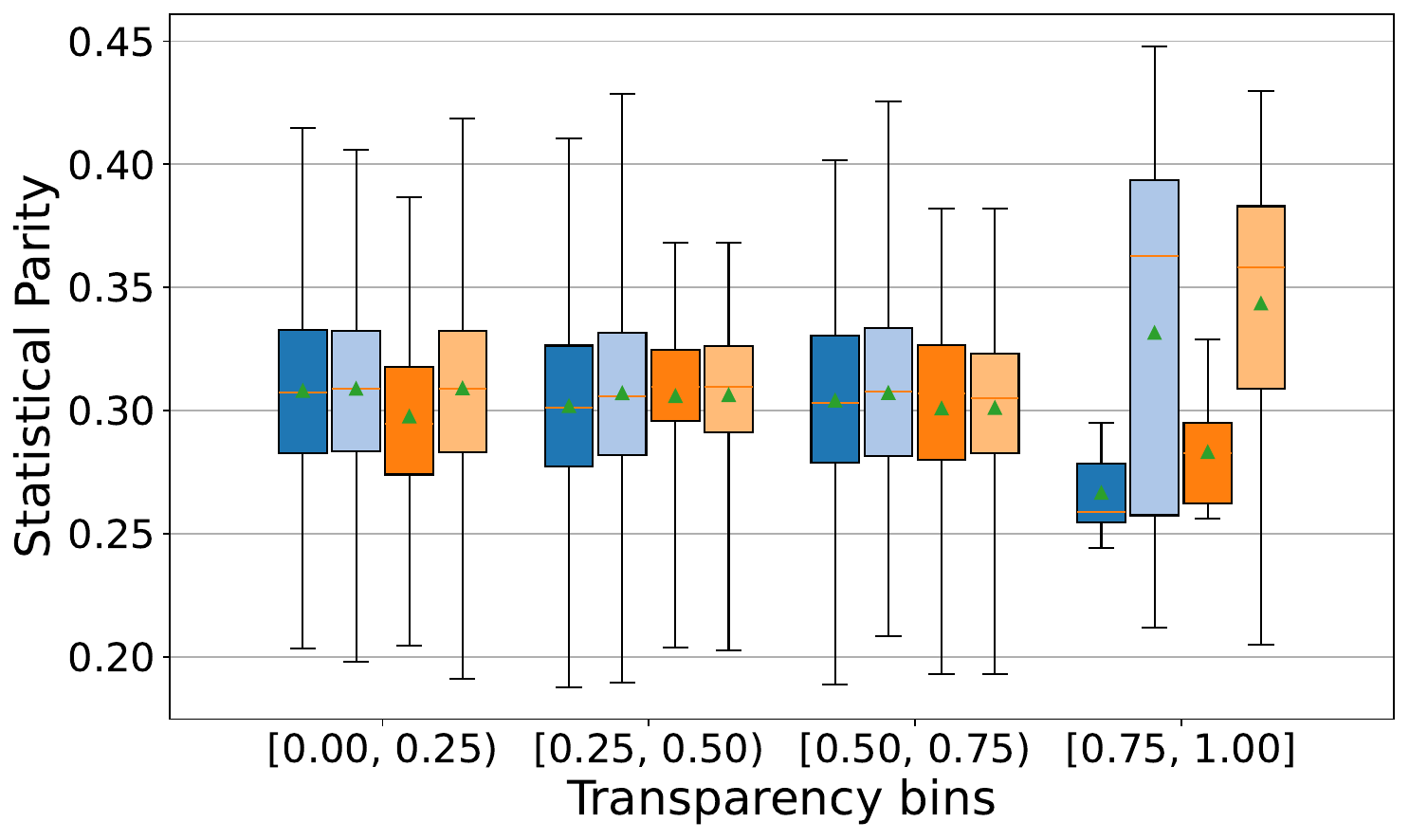}
    \caption{COMPAS dataset}\label{fig:Fair_max_SP_compas}
\end{subfigure}
\caption{Distribution of test set Statistical Parity (SP) across Rashomon sets over transparency bins $(\varepsilon = 0.01)$ for HybridCORELSPre and HybridCORELSPost, with or without our proposed ICD mitigation $(\eta = 0.05)$. Each subplot corresponds to mitigation
applied with respect to a sensitive attribute (Age, Gender, Race),ordered from left to right columns. Results are reported for all datasets.} \label{fig:Fair_max_SP}
\end{figure*}

Figures~\ref{fig:Fair_max_EO} and~\ref{fig:Fair_max_SP} further illustrate the indirect impact of ICD mitigation on algorithmic fairness metrics, namely Equal Opportunity (EO) and Statistical Parity (SP). Depending on the dataset and transparency bin, mitigation leads to varying levels of improvement, including reductions in average disparity, variability, and worst-case observed values.

\begin{figure*}[t]
\centering

\includegraphics[width=0.8\textwidth]{Plots/ICF/HybridCORELS_fairness_shared_legend.pdf}

\begin{subfigure}{0.85\textwidth}
    \centering
    \includegraphics[width=0.32\linewidth]{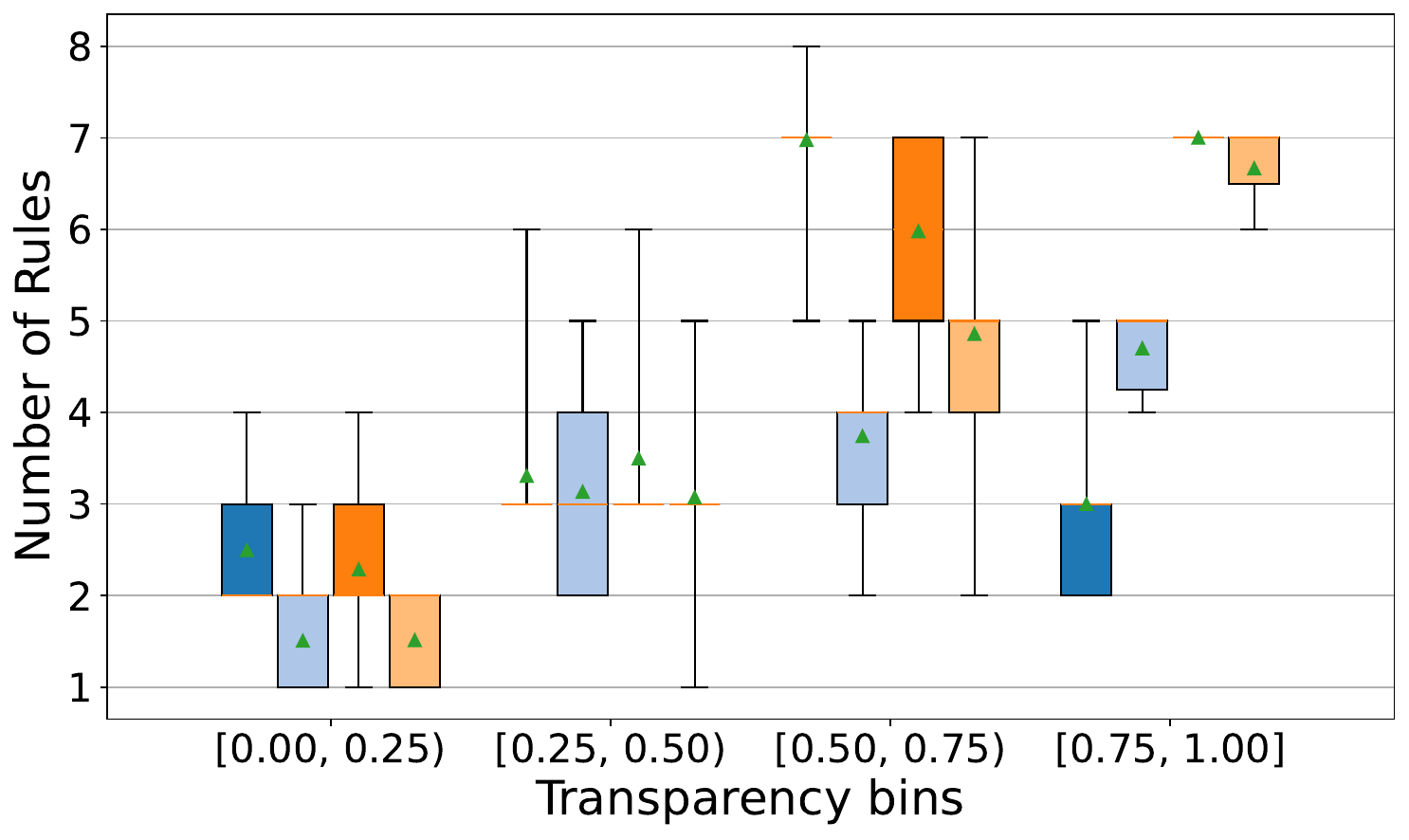}
    \includegraphics[width=0.32\linewidth]{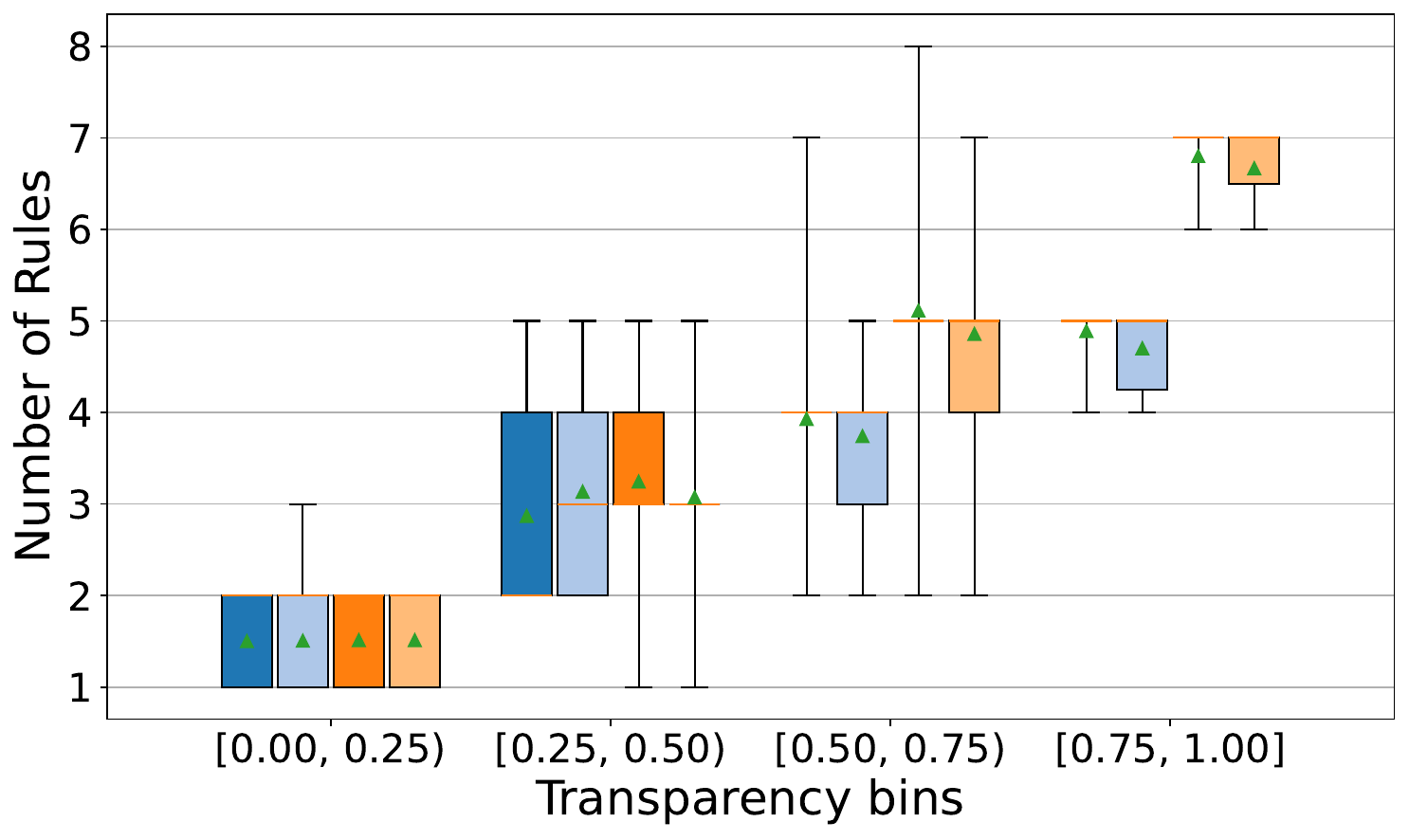}
    \includegraphics[width=0.32\linewidth]{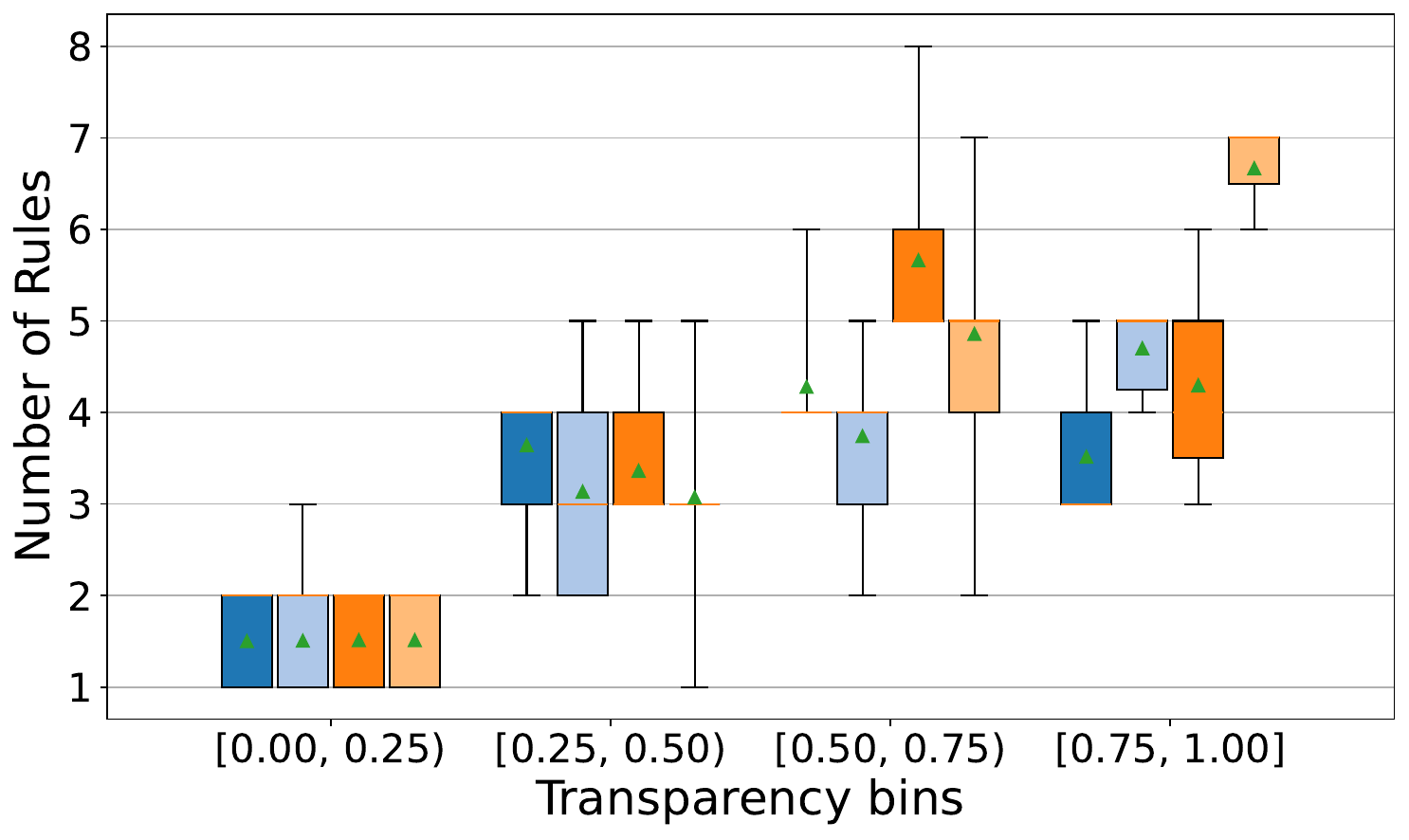}
    \caption{ACS Employment dataset}\label{fig:Fair_Sparsity_acs}
\end{subfigure}

\vspace{0.5em}

\begin{subfigure}{0.85\textwidth}
    \centering
    \includegraphics[width=0.32\linewidth]{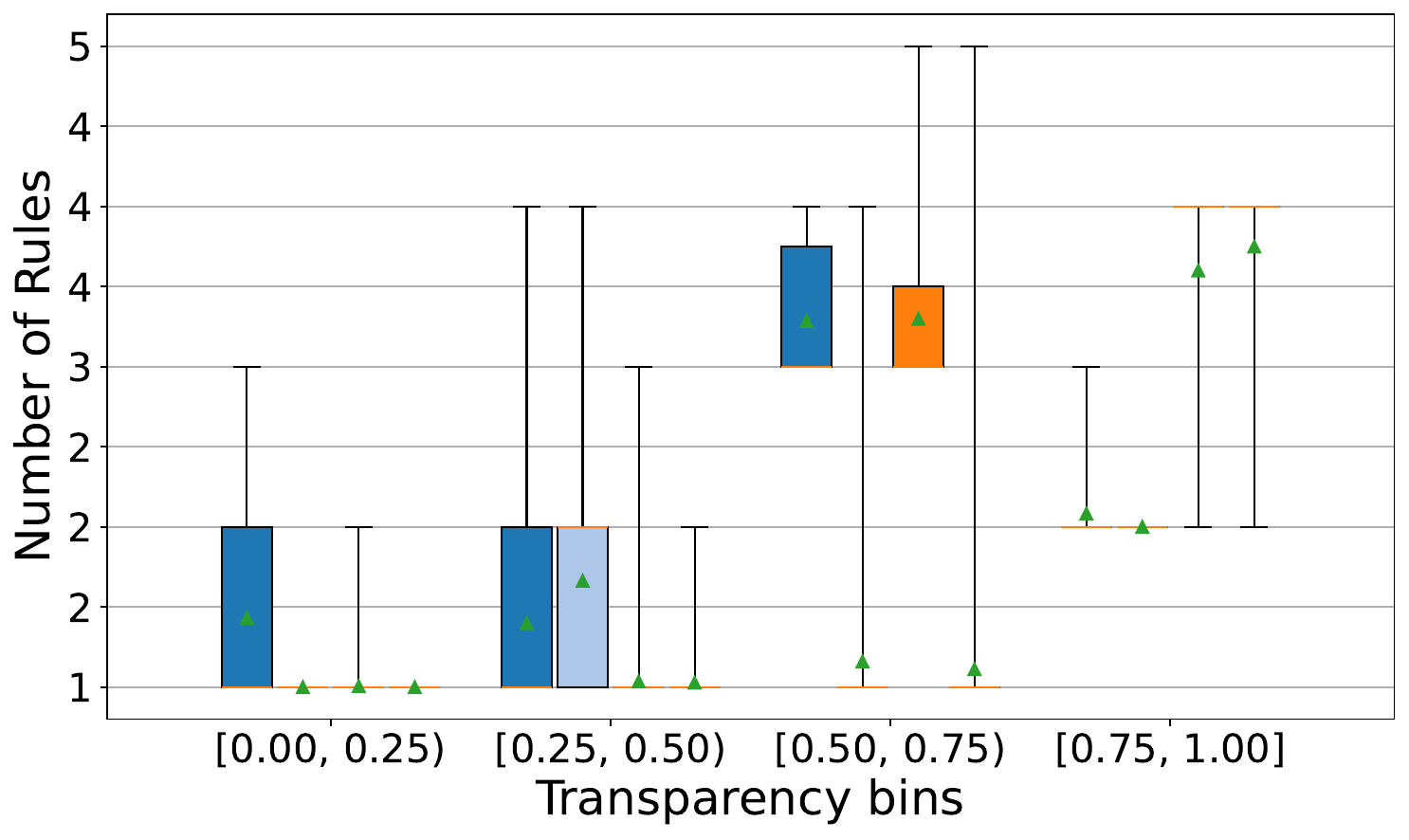}
    \includegraphics[width=0.32\linewidth]{Plots/Sparsity/Fair_Sparsityadult_Age.pdf}
    \includegraphics[width=0.32\linewidth]{Plots/Sparsity/Fair_Sparsityadult_Age.pdf}
    \caption{UCI Adult Income dataset}\label{fig:Fair_Sparsity_adult}
\end{subfigure}

\vspace{0.5em}

\caption{Distribution of model sparsity (number of rules) across Rashomon sets over transparency bins $(\varepsilon = 0.01)$ for HybridCORELSPre and HybridCORELSPost, with or without our proposed ICD mitigation $(\eta = 0.05)$. Each subplot corresponds to mitigation
applied with respect to a sensitive attribute (Age, Gender, Race), ordered from left to right columns. Results are reported for the ACS Employment and UCI Adult Income datasets.}
\label{fig:Fair_Sparsity}
\end{figure*}

\begin{figure*}[t]
\centering

\includegraphics[width=0.8\textwidth]{Plots/ICF/HybridCORELS_fairness_shared_legend.pdf}

\begin{subfigure}{0.85\textwidth}
    \centering
    \includegraphics[width=0.32\linewidth]{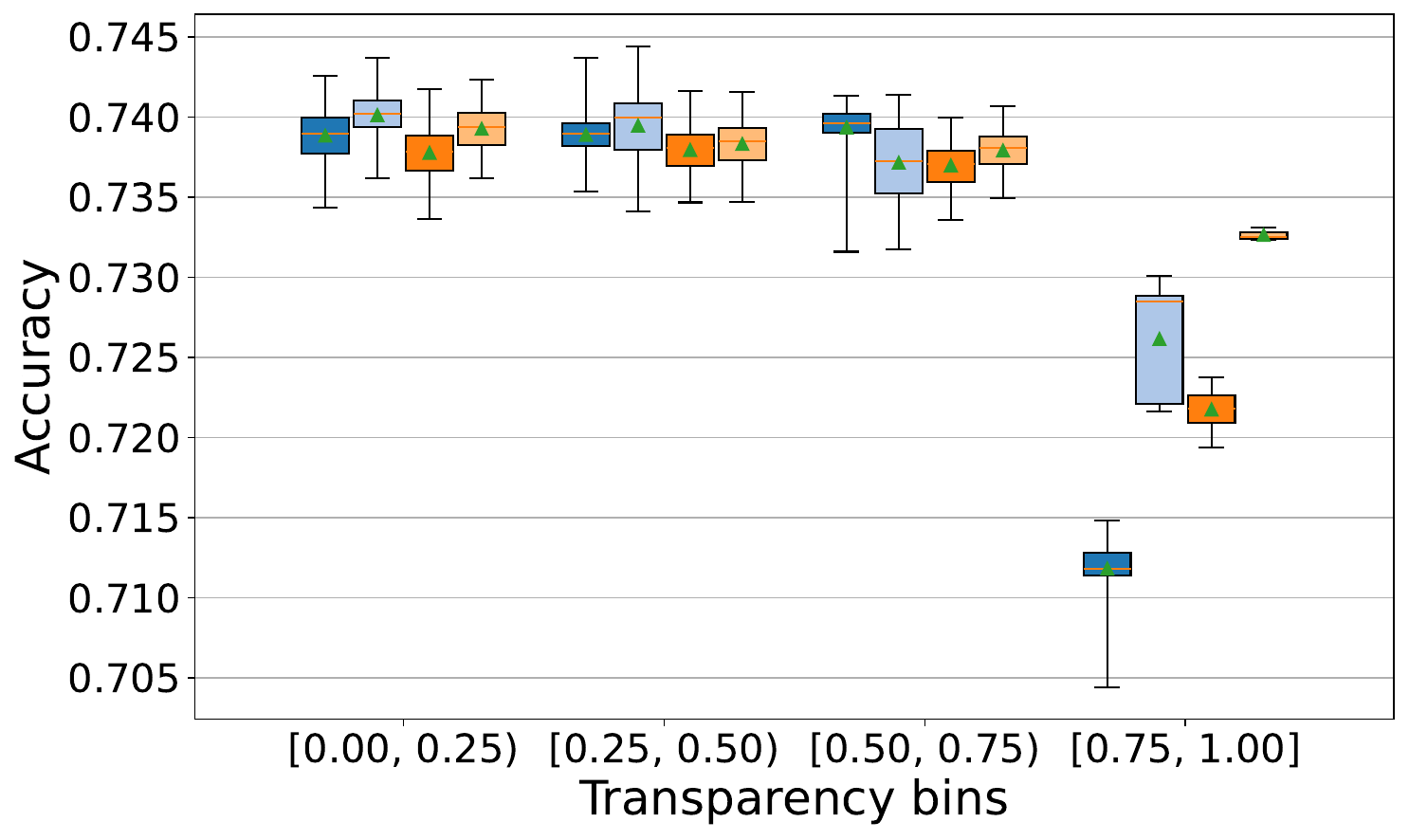}
    \includegraphics[width=0.32\linewidth]{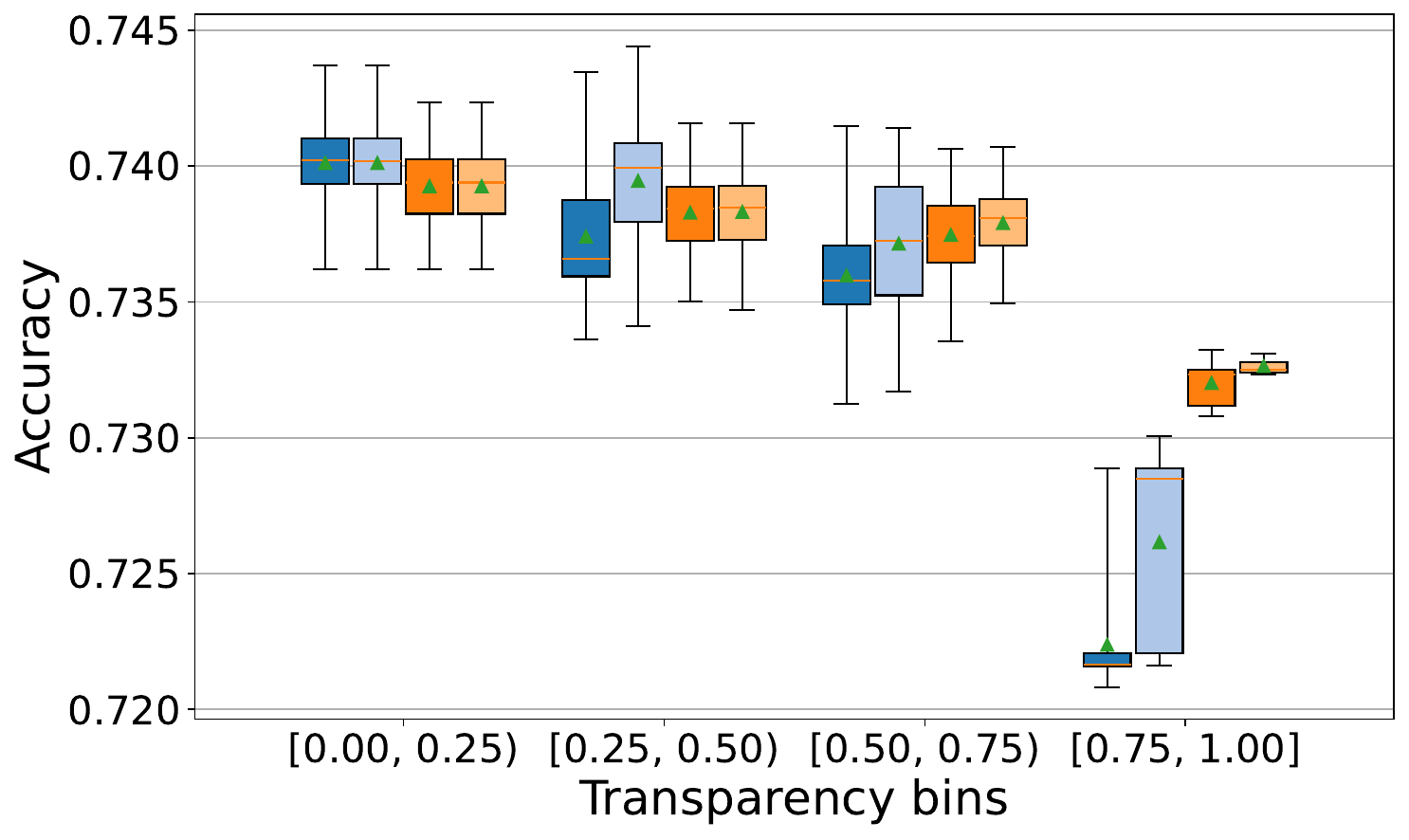}
    \includegraphics[width=0.32\linewidth]{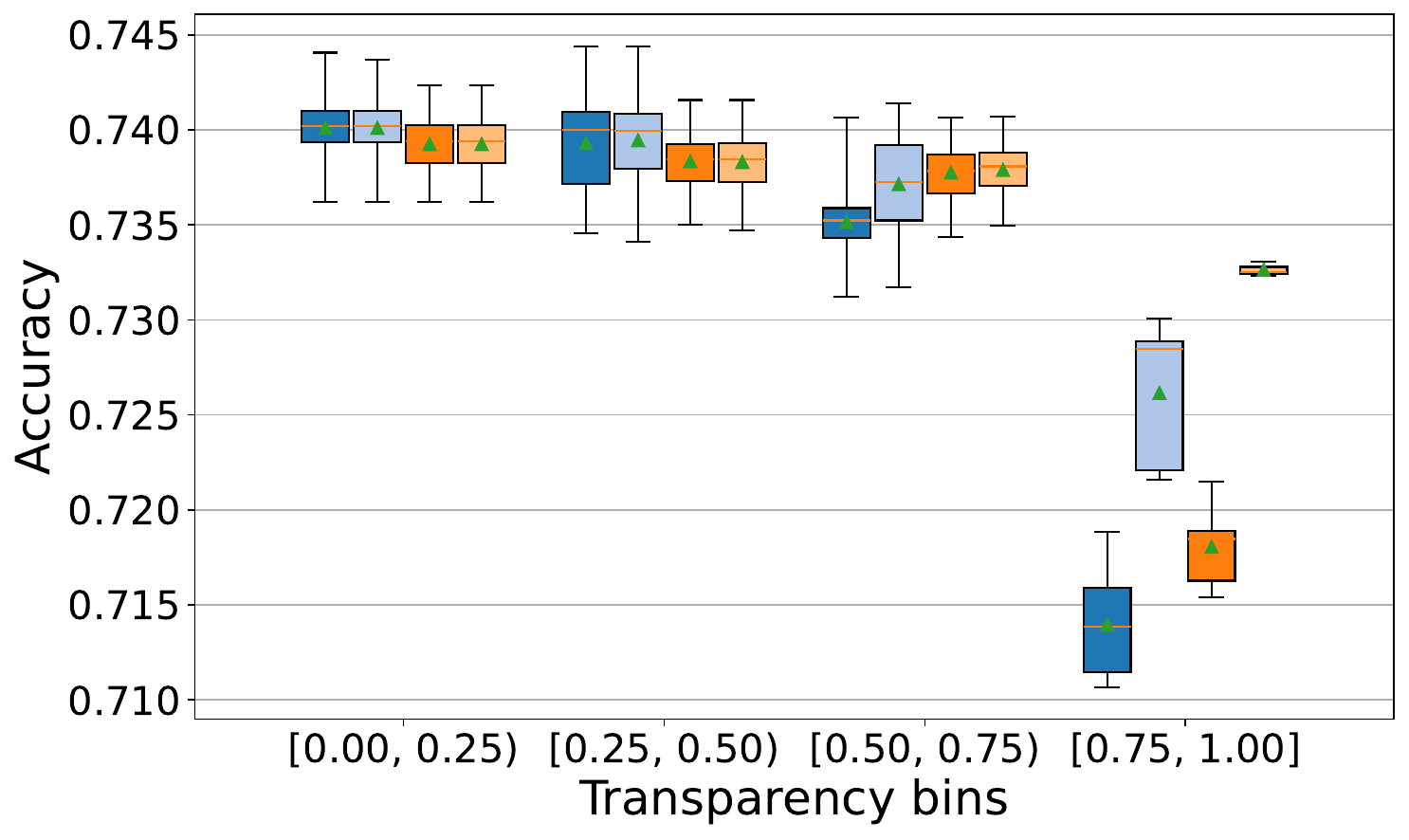}
    \caption{ACS Employment dataset}\label{fig:Fair_Acc_acs}
\end{subfigure}

\vspace{0.5em}

\begin{subfigure}{0.85\textwidth}
    \centering
    \includegraphics[width=0.32\linewidth]{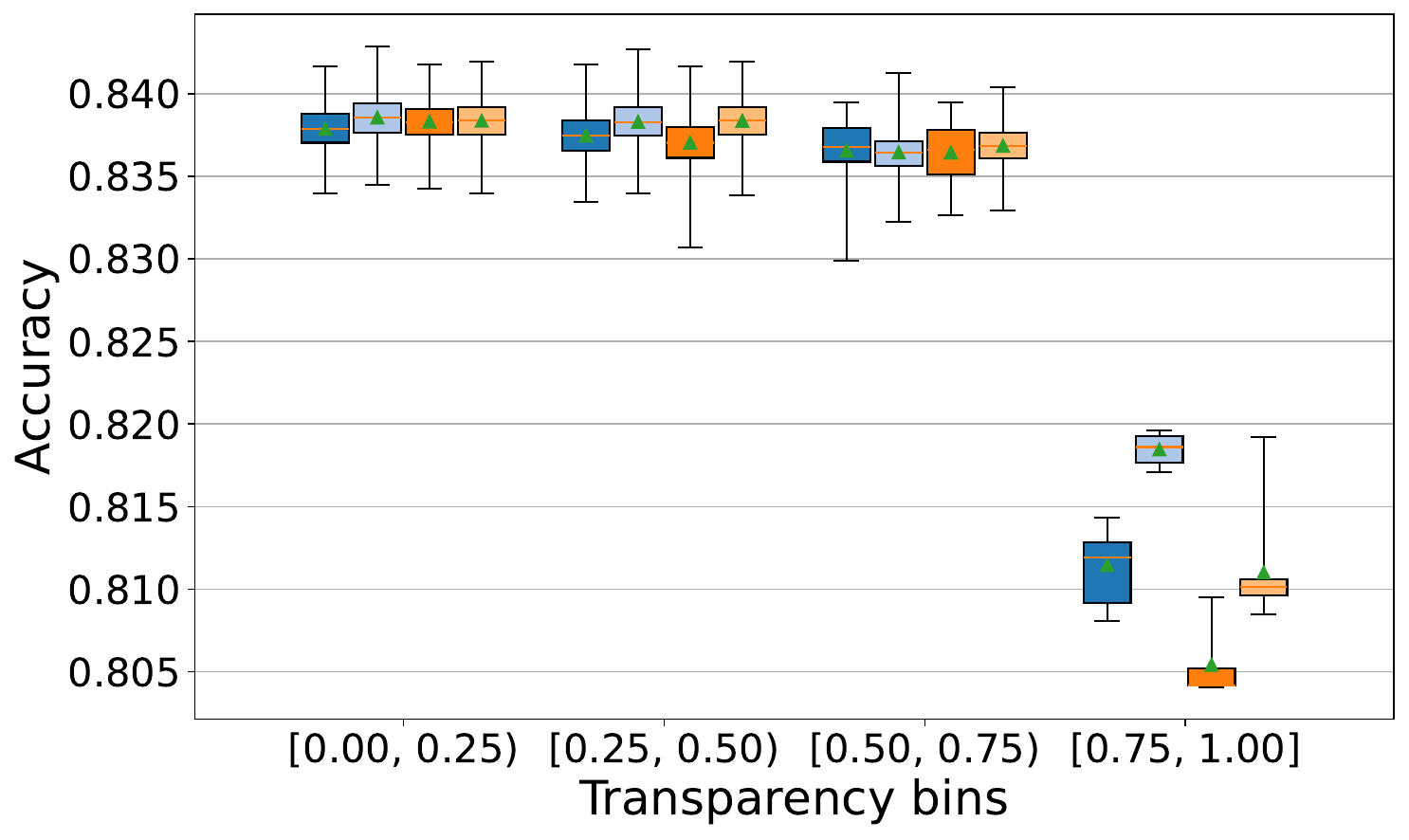}
    \includegraphics[width=0.32\linewidth]{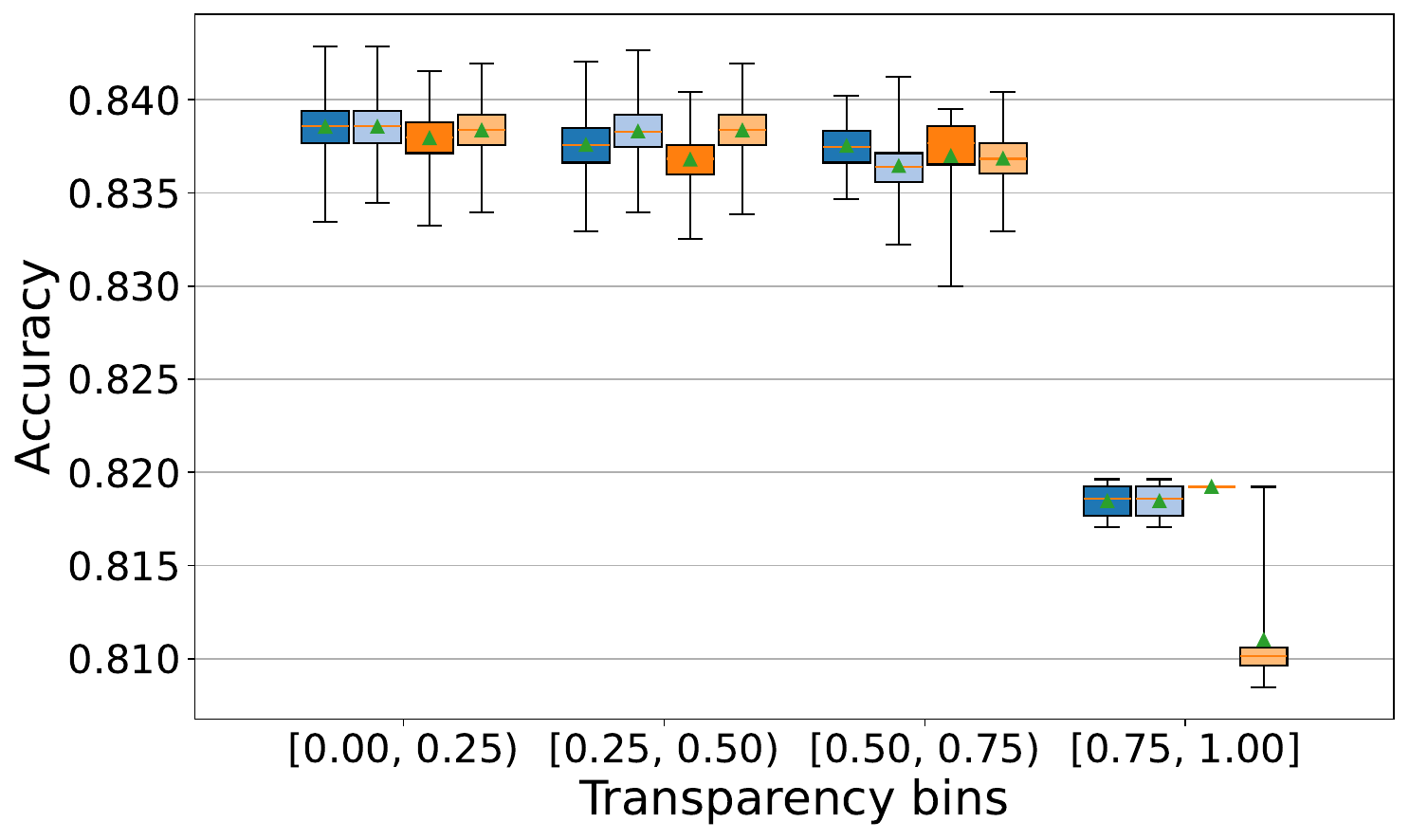}
    \includegraphics[width=0.32\linewidth]{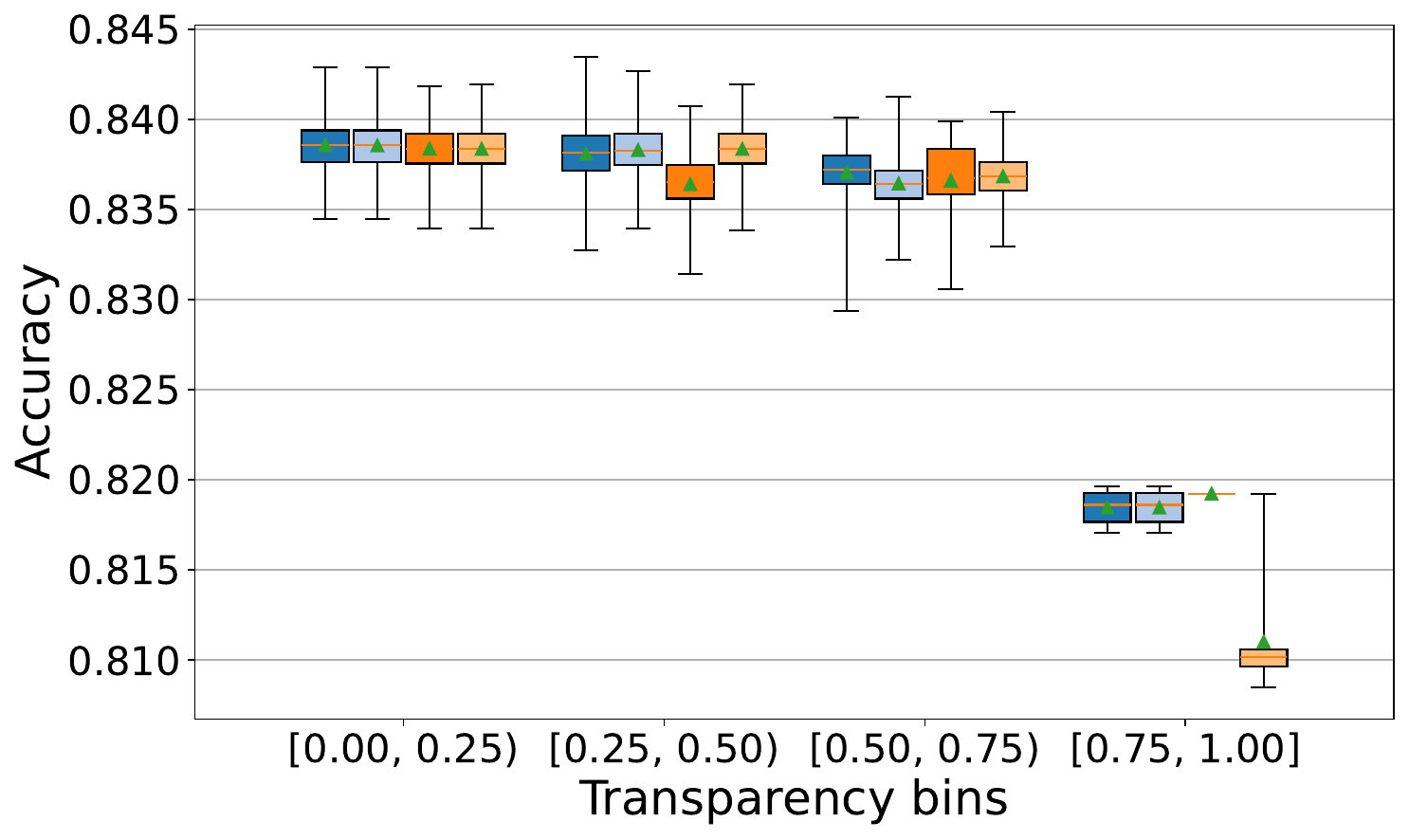}
    \caption{UCI Adult Income dataset}\label{fig:Fair_Acc_adult}
\end{subfigure}

\vspace{0.5em}

\caption{Distribution of test set accuracy across Rashomon sets over transparency bins $(\varepsilon = 0.01)$ for HybridCORELSPre and HybridCORELSPost, with or without our proposed ICD mitigation $(\eta = 0.05)$. Each subplot corresponds to mitigation applied with respect to a sensitive attribute (Age, Gender, Race), ordered from left to right columns. Results are reported for the ACS Employment and UCI Adult Income datasets.}
\label{fig:Fair_Acc}
\end{figure*}

We additionally analyze how sparsity and predictive performance evolve after mitigation across transparency bins. Corresponding results for the number of rules in the interpretable prefix and test accuracy distributions are provided in Figures~\ref{fig:Fair_Sparsity} and~\ref{fig:Fair_Acc}, respectively. Consistent with the observations reported in Result 6, mitigation induces only minor increases in model complexity while largely preserving predictive performance.

\begin{figure*}[t]
\centering
\includegraphics[width=0.6\textwidth]{Plots/Fair/ICD_mitigation_shared_legend.pdf}

\begin{subfigure}{0.35\textwidth}
    \centering
    \includegraphics[width=\linewidth]{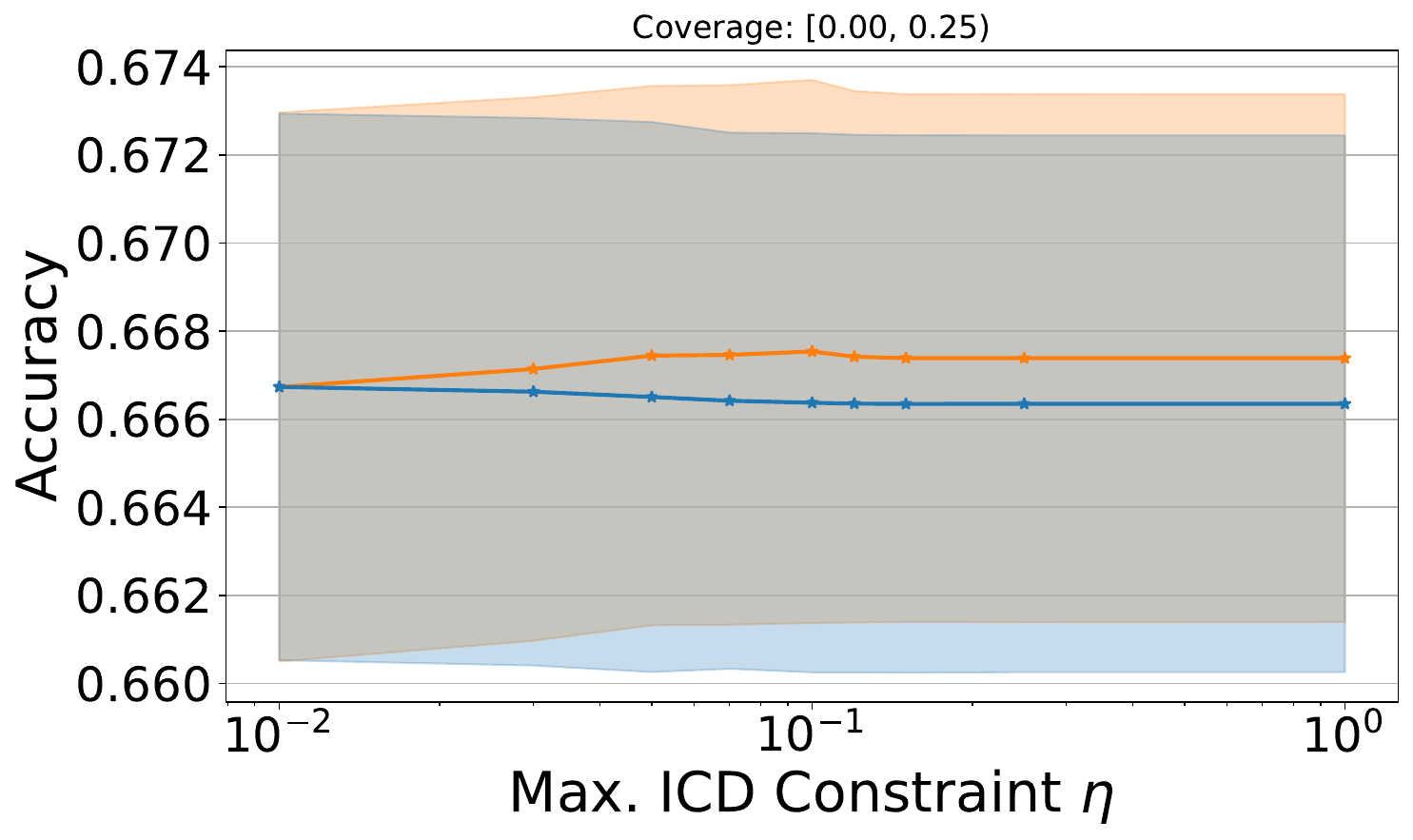}
    \caption{$Q_1$}
\end{subfigure}
\begin{subfigure}{0.35\textwidth}
    \centering
    \includegraphics[width=\linewidth]{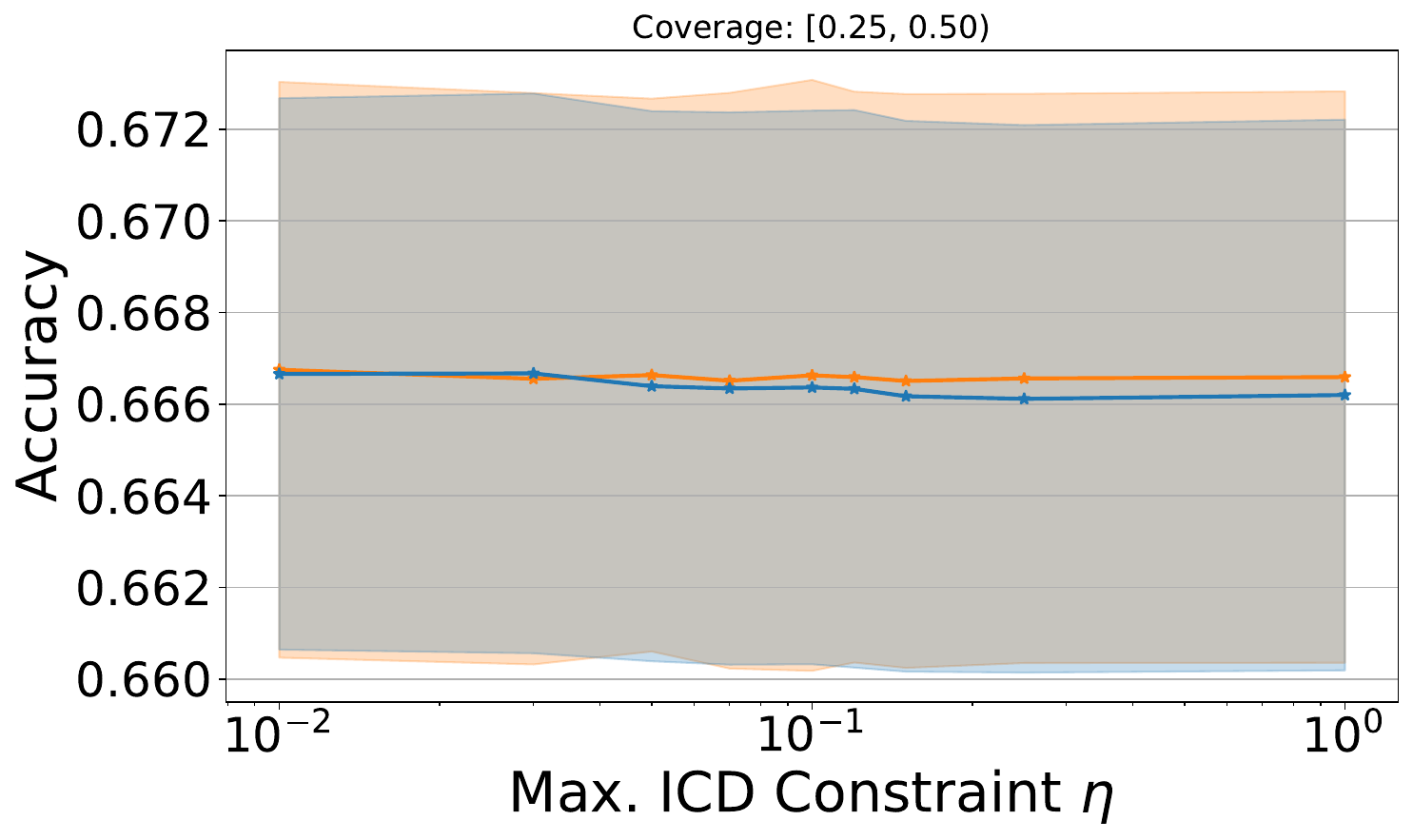}
    \caption{$Q_2$}
\end{subfigure}

\vspace{0.5em} %

\begin{subfigure}{0.35\textwidth}
    \centering
    \includegraphics[width=\linewidth]{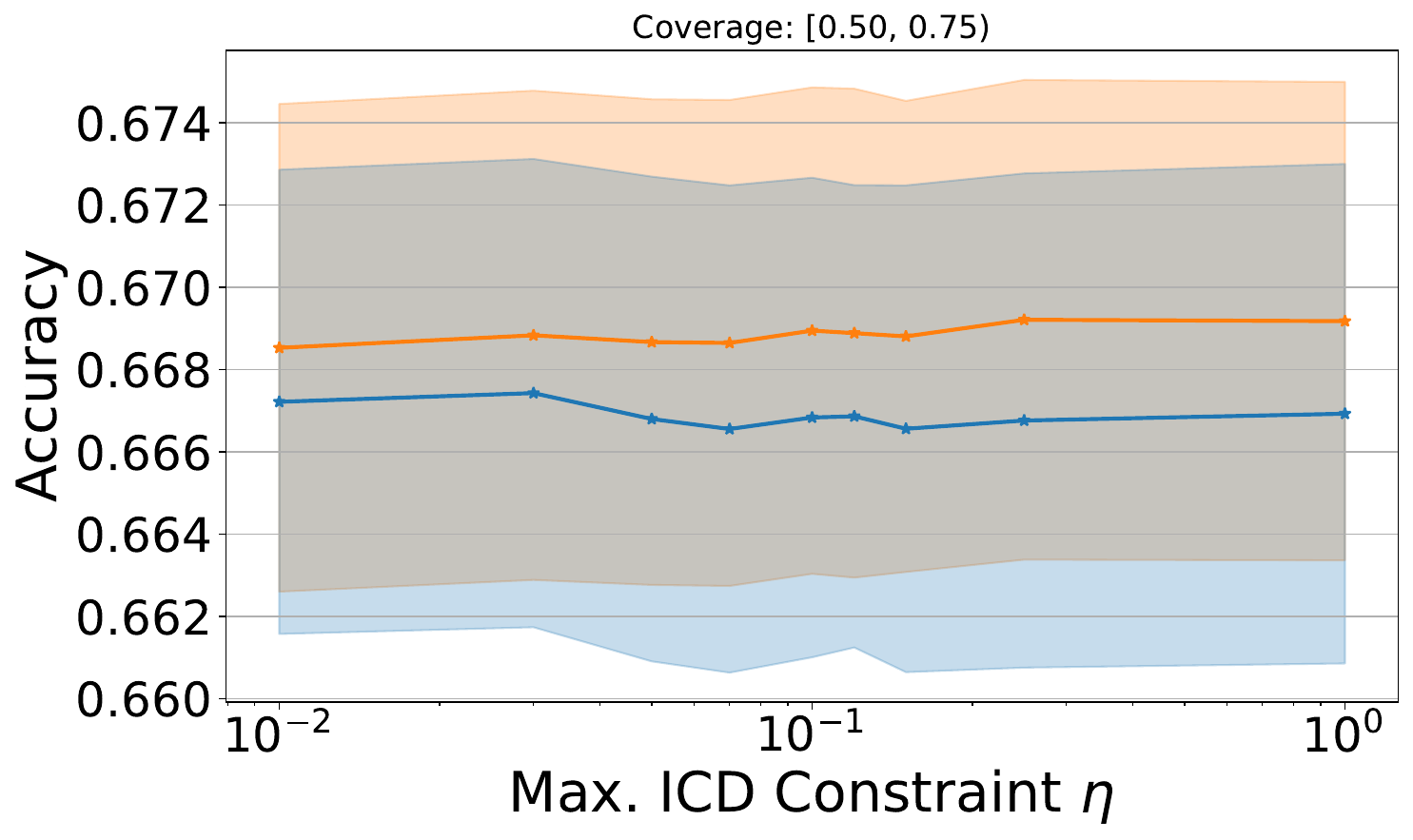}
    \caption{$Q_3$}
\end{subfigure}
\begin{subfigure}{0.35\textwidth}
    \centering
    \includegraphics[width=\linewidth]{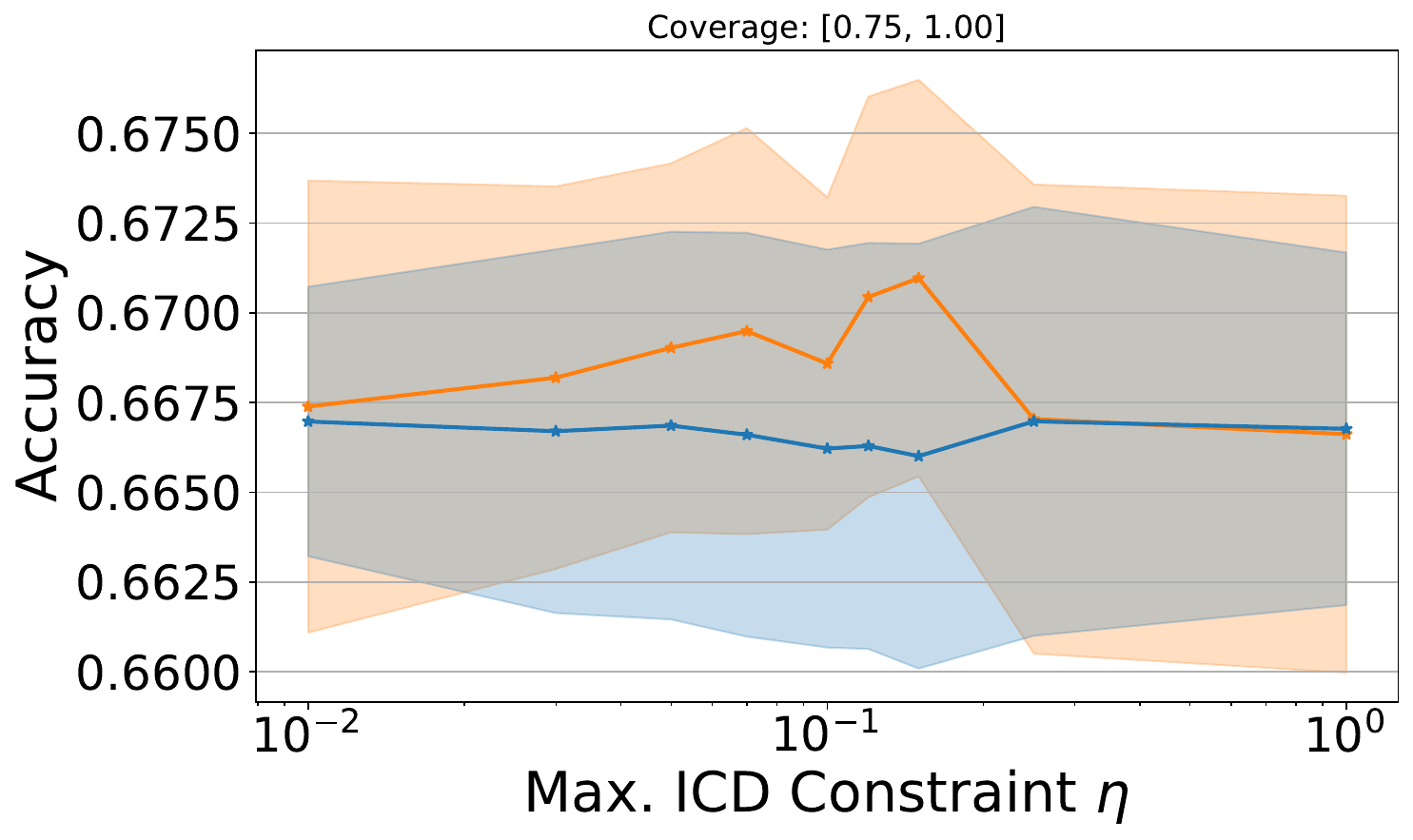}
    \caption{$Q_4$}
\end{subfigure}

\caption{Test set accuracy for HybridCORELSPre and HybridCORELSPost with ICD mitigation across Rashomon sets over transparency bins ($\varepsilon = 0.01$). For different levels of mitigation induced by the maximum ICD constraint ($\eta$), curves report the average and standard deviation across Rashomon models. Results are provided for the COMPAS dataset and Gender as sensitive attribute.}
\label{ACC_MaxCov_4Q}

\end{figure*}

\begin{figure*}[t]
\centering
\includegraphics[width=0.6\textwidth]{Plots/Fair/ICD_mitigation_shared_legend.pdf}

\begin{subfigure}{0.35\textwidth}
    \centering
    \includegraphics[width=\linewidth]{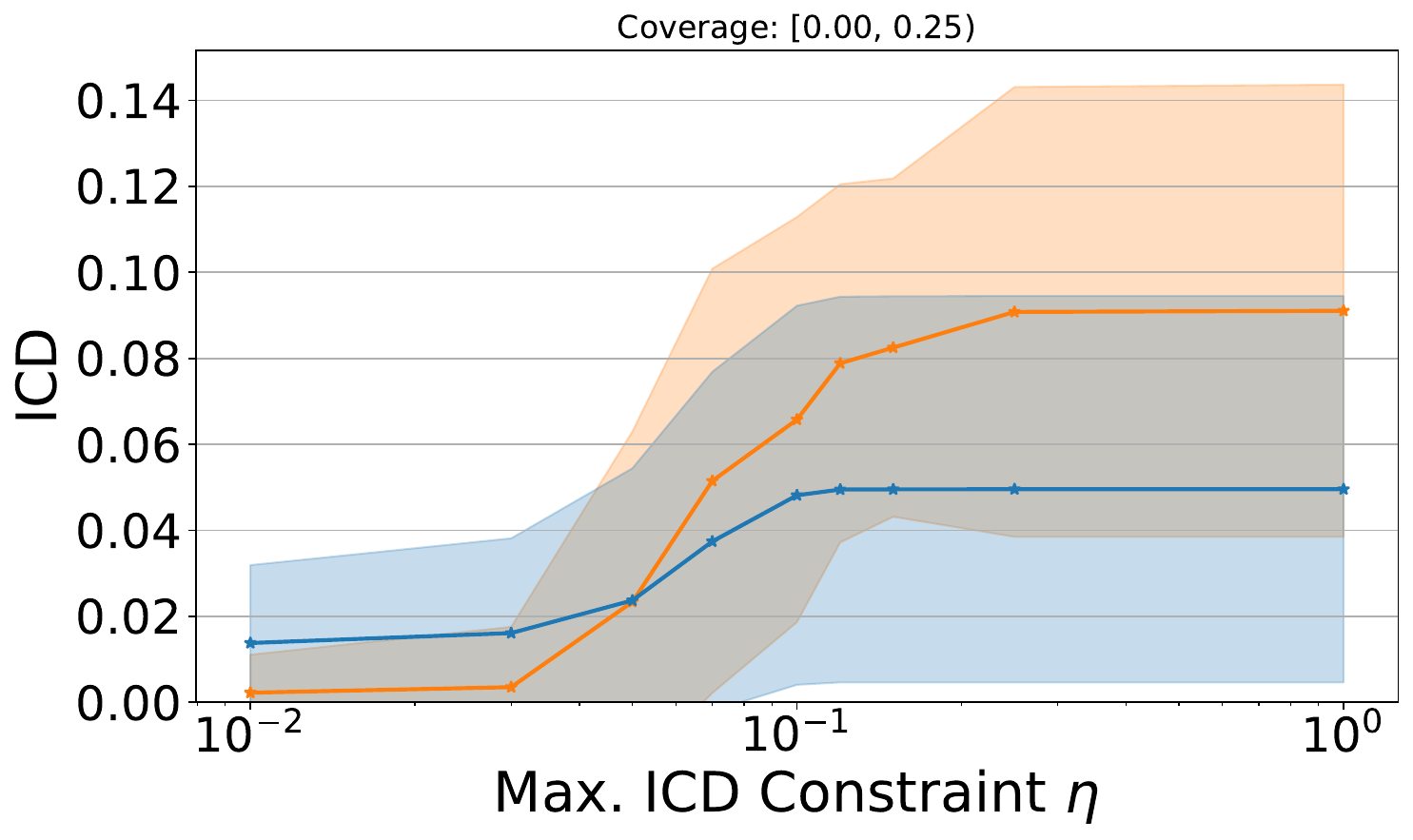}
    \caption{$Q_1$}
\end{subfigure}
\begin{subfigure}{0.35\textwidth}
    \centering
    \includegraphics[width=\linewidth]{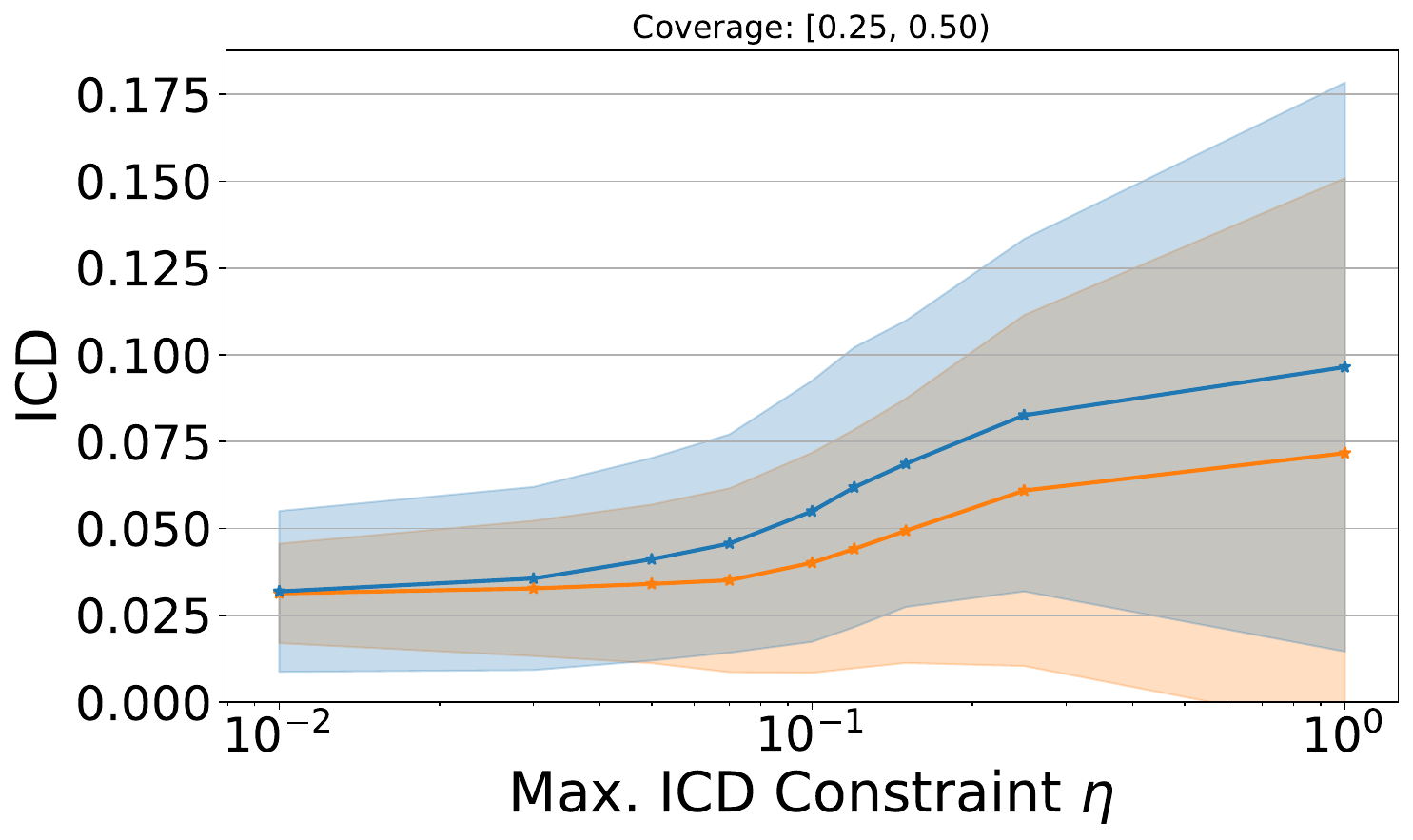}
    \caption{$Q_2$}
\end{subfigure}

\vspace{0.5em} %

\begin{subfigure}{0.35\textwidth}
    \centering
    \includegraphics[width=\linewidth]{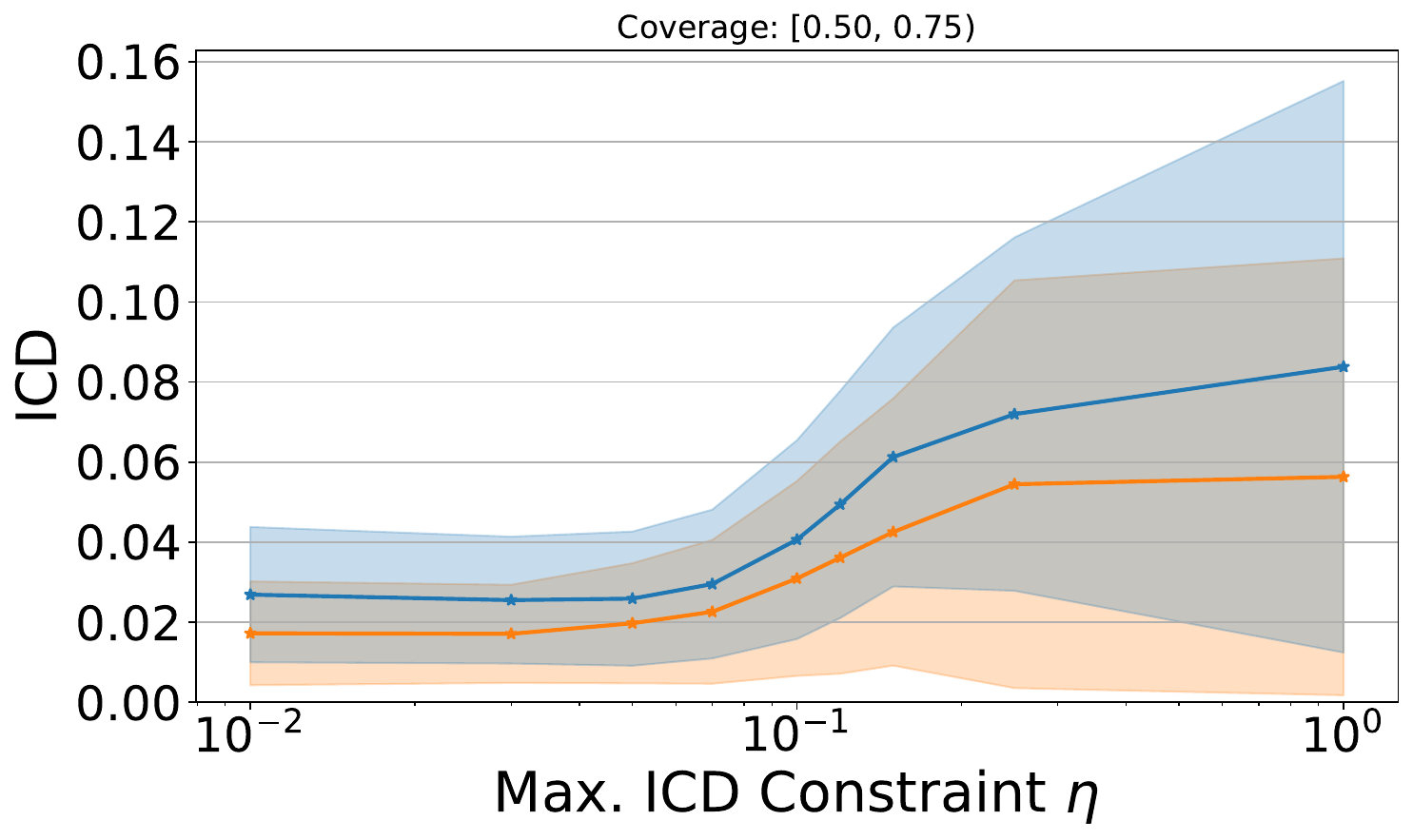}
    \caption{$Q_3$}
\end{subfigure}
\begin{subfigure}{0.35\textwidth}
    \centering
    \includegraphics[width=\linewidth]{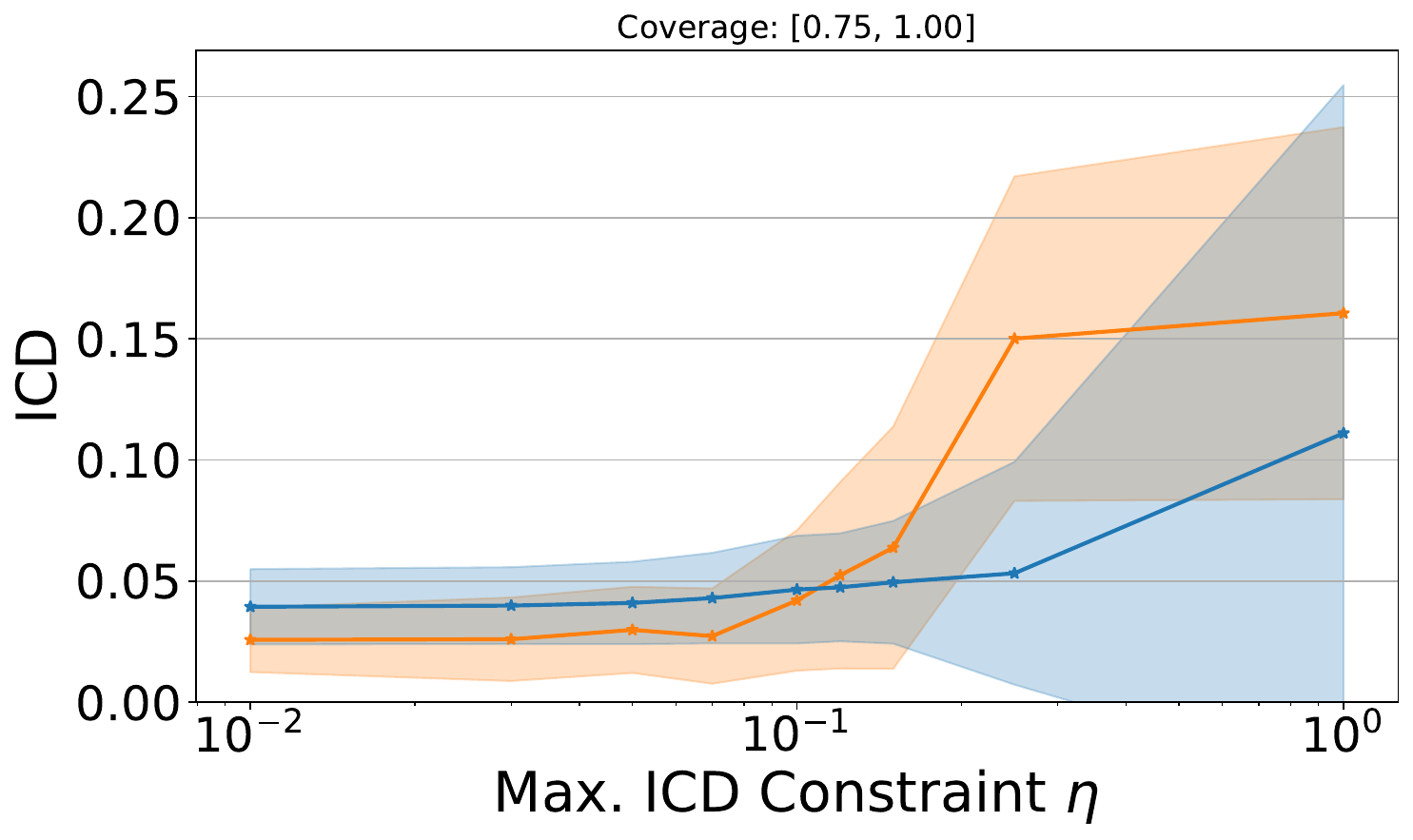}
    \caption{$Q_4$}
\end{subfigure}
\vspace{0.8em}
\caption{Test set ICD for HybridCORELSPre and HybridCORELSPost with ICD mitigation across Rashomon sets over transparency bins ($\varepsilon = 0.01$). For different levels of mitigation induced by the maximum ICD constraint ($\eta$), curves report the average and standard deviation across Rashomon models. Results are provided for the COMPAS dataset and Gender as sensitive attribute.}
\label{ICF_MaxCov_4Q}

\end{figure*}

\begin{figure*}[t]
\centering
\includegraphics[width=0.6\textwidth]{Plots/Fair/ICD_mitigation_shared_legend.pdf}

\begin{subfigure}{0.35\textwidth}
    \centering
    \includegraphics[width=\linewidth]{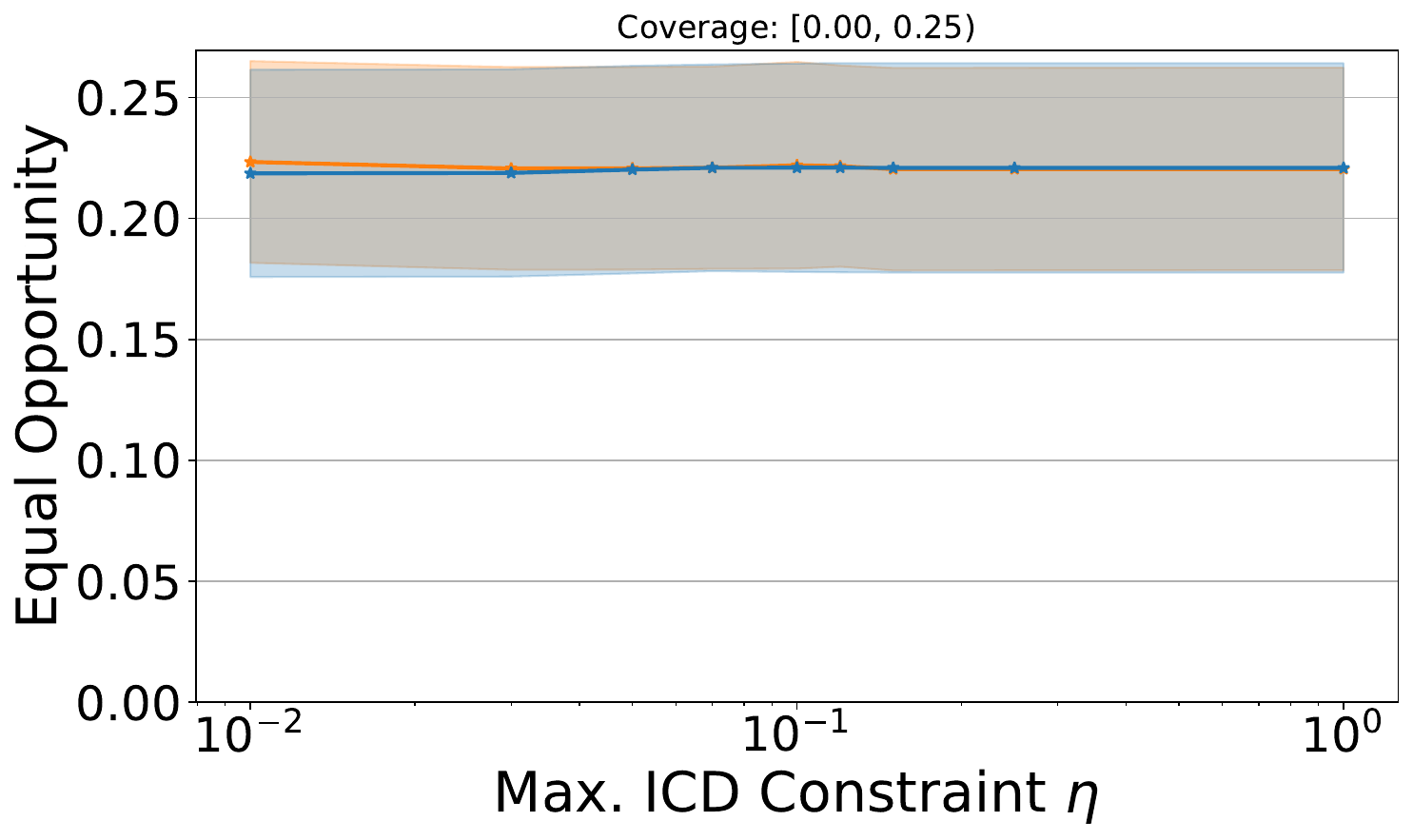}
    \caption{$Q_1$}
\end{subfigure}
\begin{subfigure}{0.35\textwidth}
    \centering
    \includegraphics[width=\linewidth]{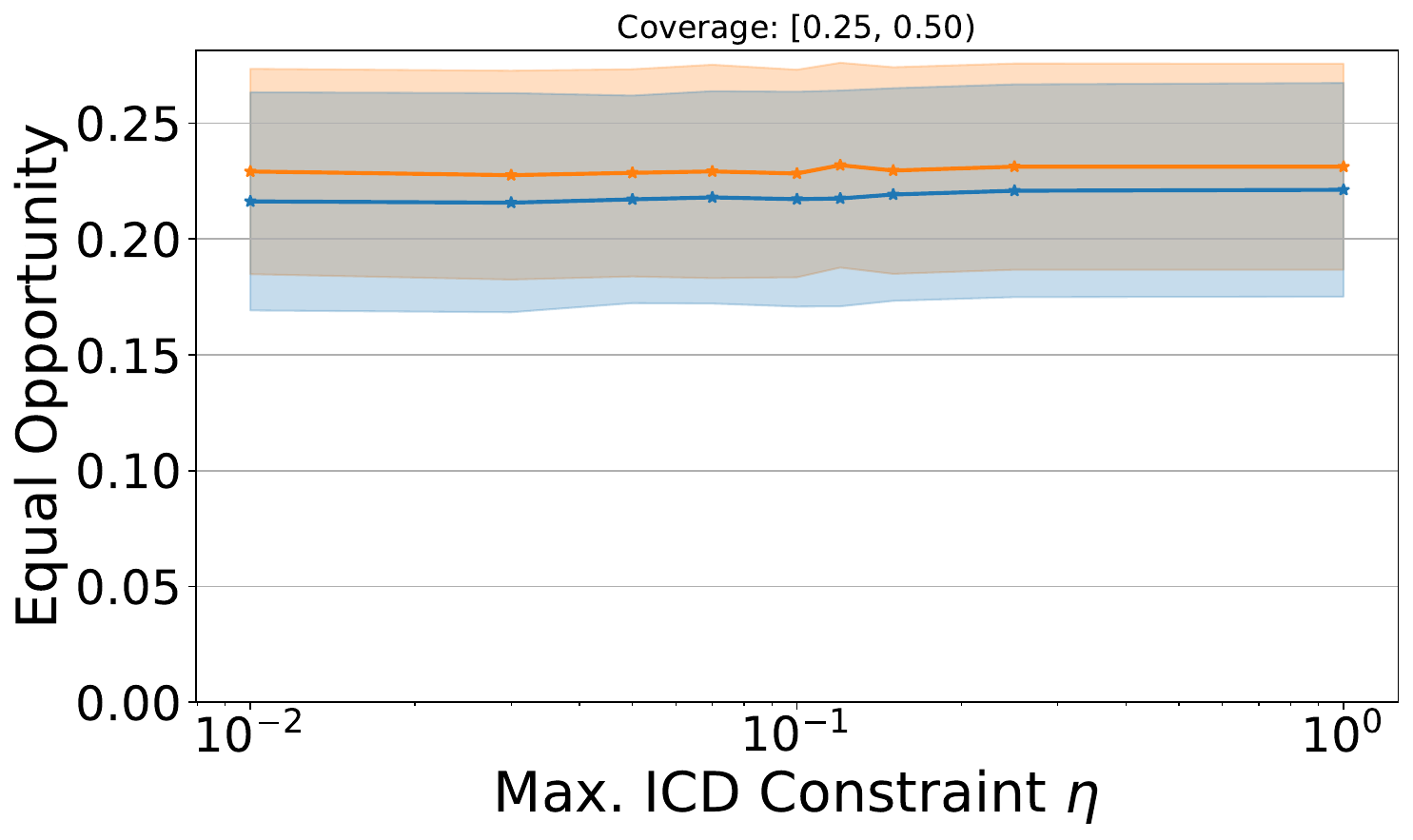}
    \caption{$Q_2$}
\end{subfigure}

\vspace{0.5em} %

\begin{subfigure}{0.35\textwidth}
    \centering
    \includegraphics[width=\linewidth]{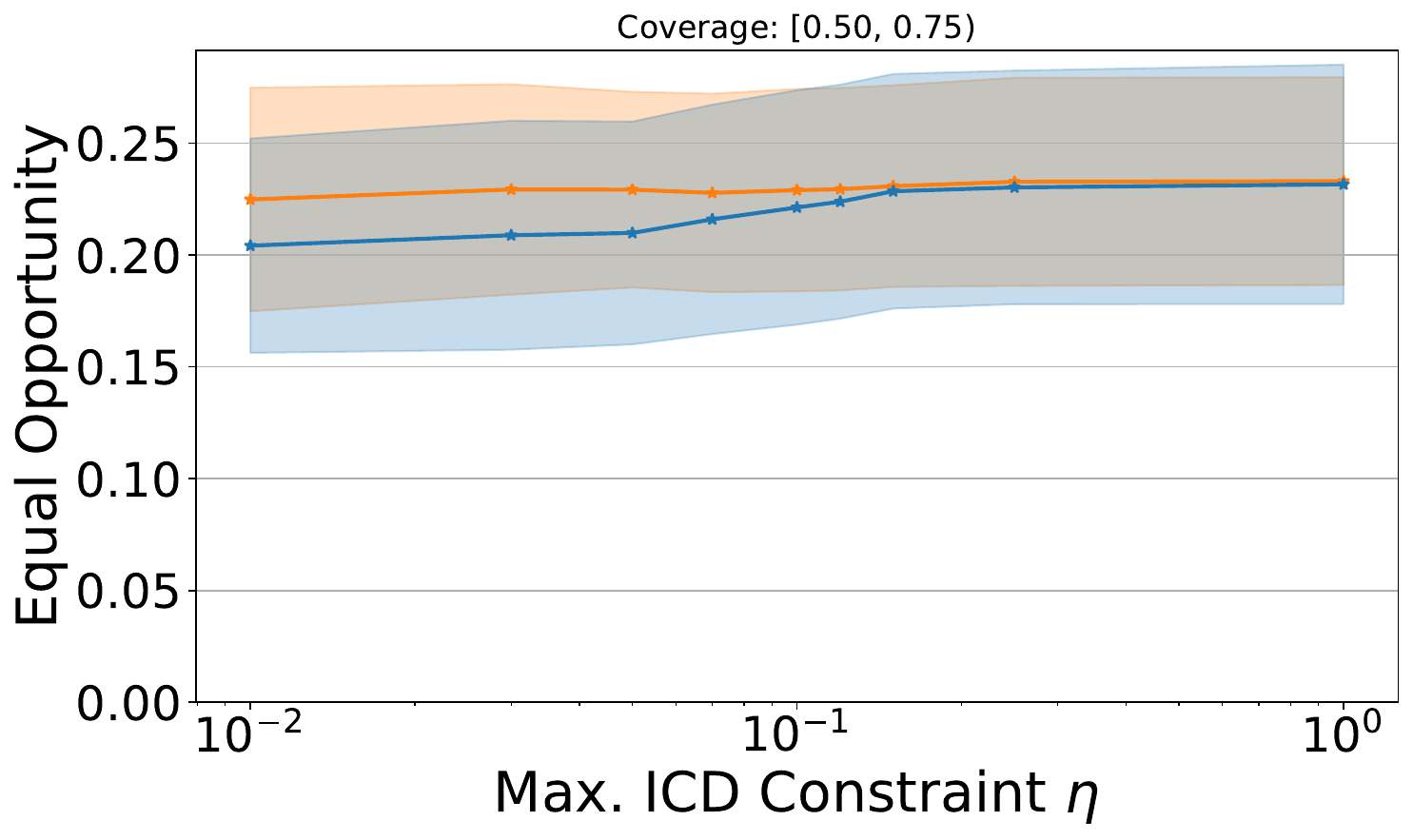}
    \caption{$Q_3$}
\end{subfigure}
\begin{subfigure}{0.35\textwidth}
    \centering
    \includegraphics[width=\linewidth]{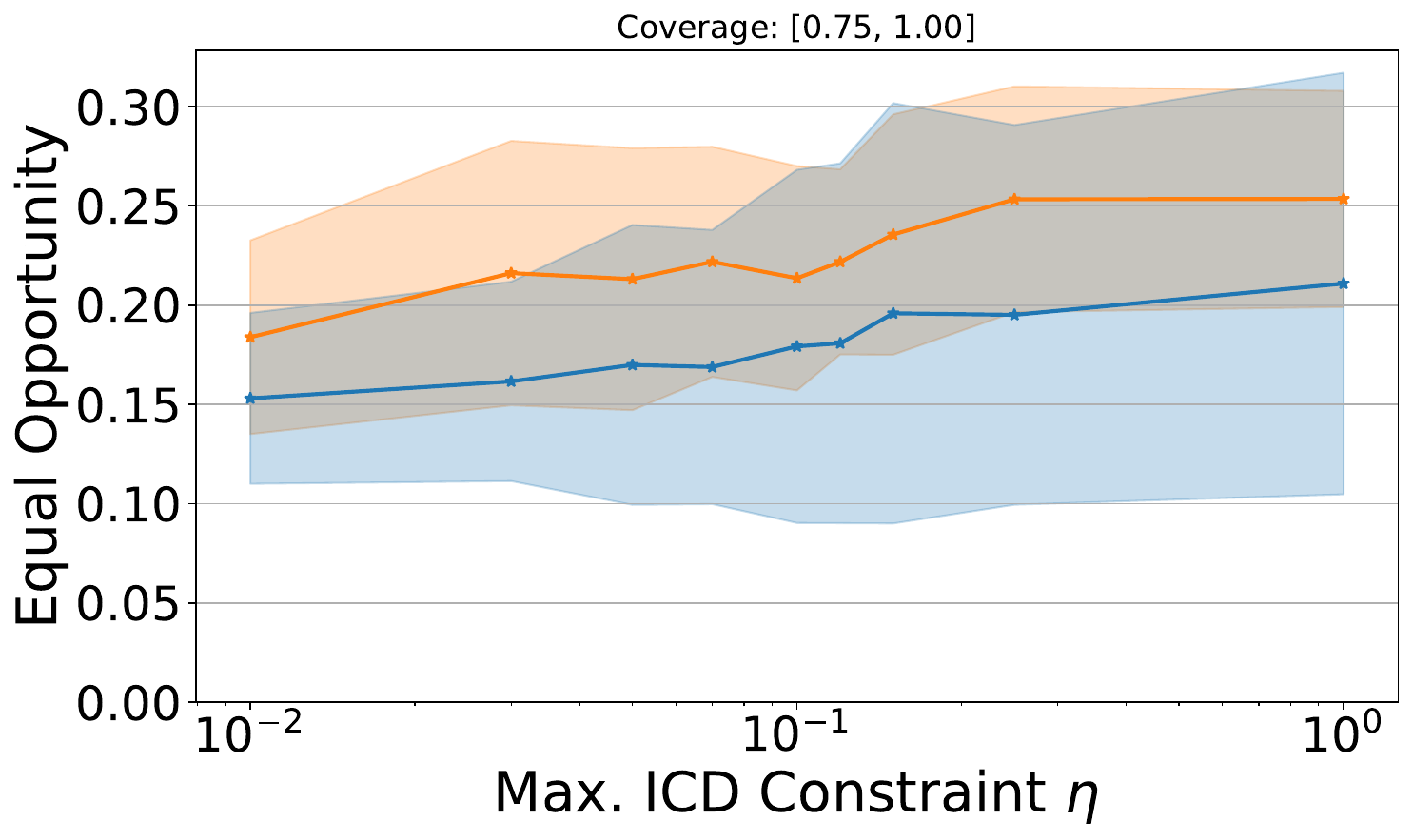}
    \caption{$Q_4$}
\end{subfigure}
\caption{Test set Equal Opportunity (EO) for HybridCORELSPre and HybridCORELSPost with ICD mitigation across Rashomon sets over transparency bins ($\varepsilon = 0.01$). For different levels of mitigation induced by the maximum ICD constraint ($\eta$), curves report the average and standard deviation across Rashomon models. Results are provided for the COMPAS dataset and Gender as sensitive attribute.}
\label{EO_MaxCov_4Q}

\end{figure*}

\begin{figure*}[t]
\centering
\includegraphics[width=0.6\textwidth]{Plots/Fair/ICD_mitigation_shared_legend.pdf}

\begin{subfigure}{0.35\textwidth}
    \centering
    \includegraphics[width=\linewidth]{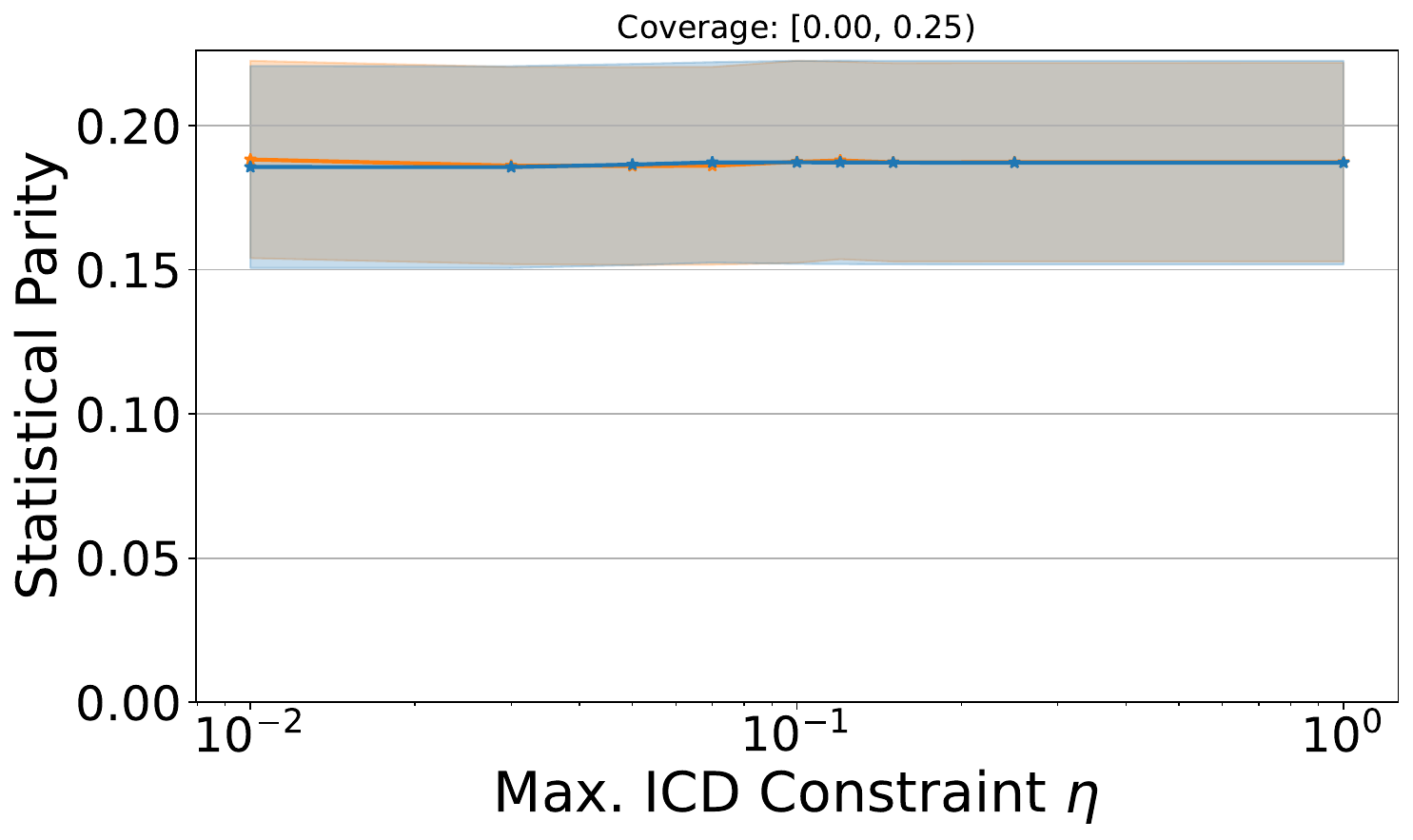}
    \caption{$Q_1$}
\end{subfigure}
\begin{subfigure}{0.35\textwidth}
    \centering
    \includegraphics[width=\linewidth]{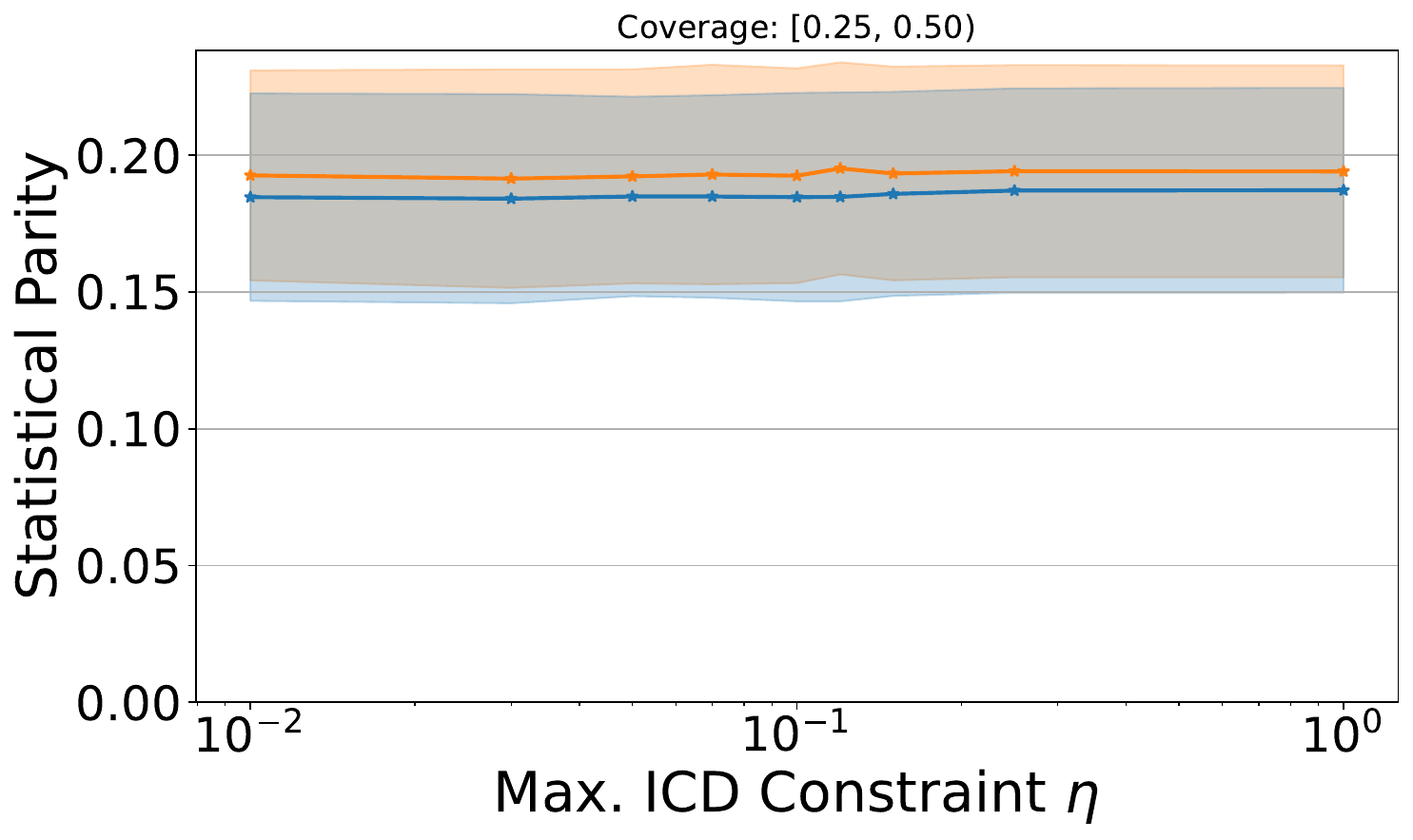}
    \caption{$Q_2$}
\end{subfigure}

\vspace{0.5em} %

\begin{subfigure}{0.35\textwidth}
    \centering
    \includegraphics[width=\linewidth]{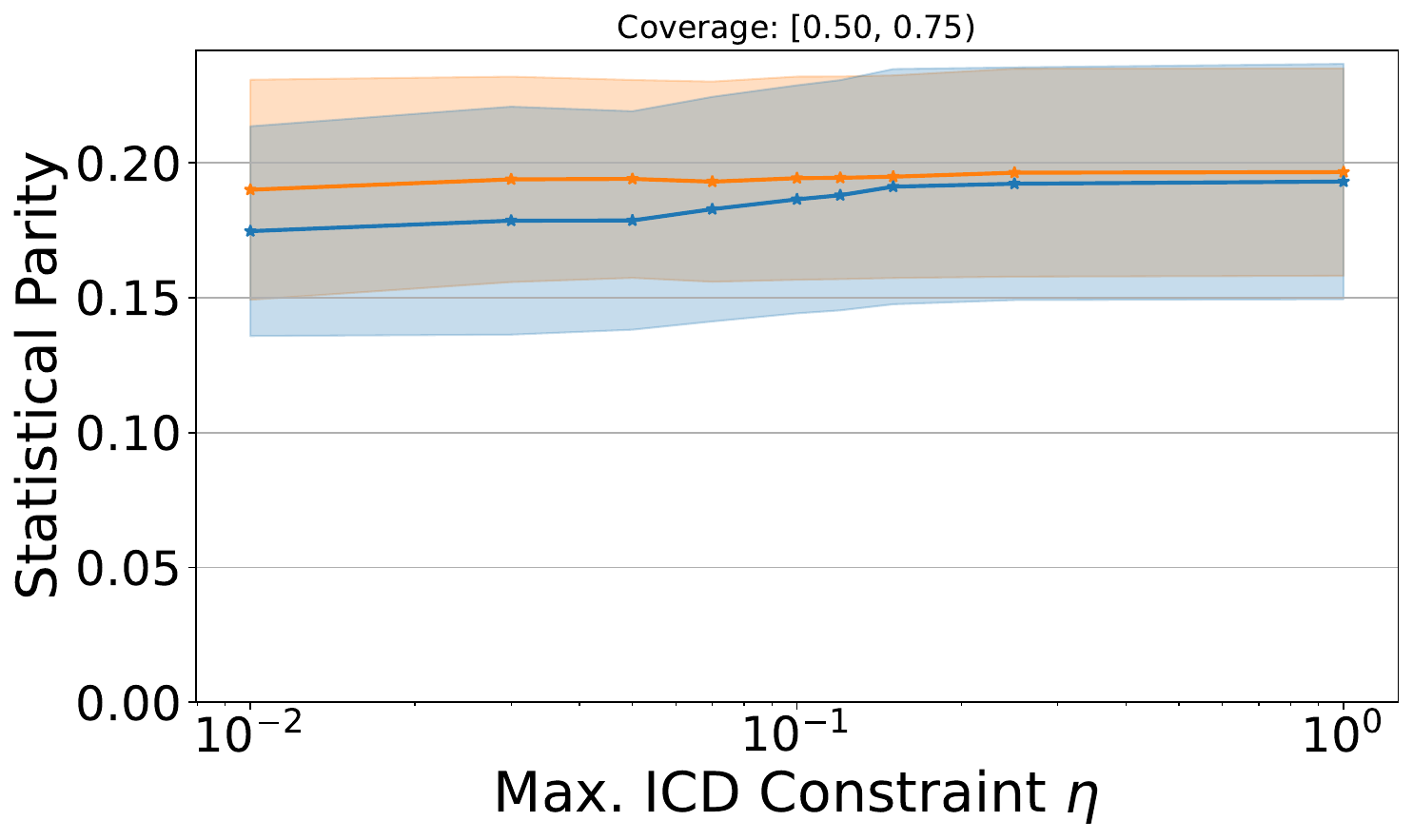}
    \caption{$Q_3$}
\end{subfigure}
\begin{subfigure}{0.35\textwidth}
    \centering
    \includegraphics[width=\linewidth]{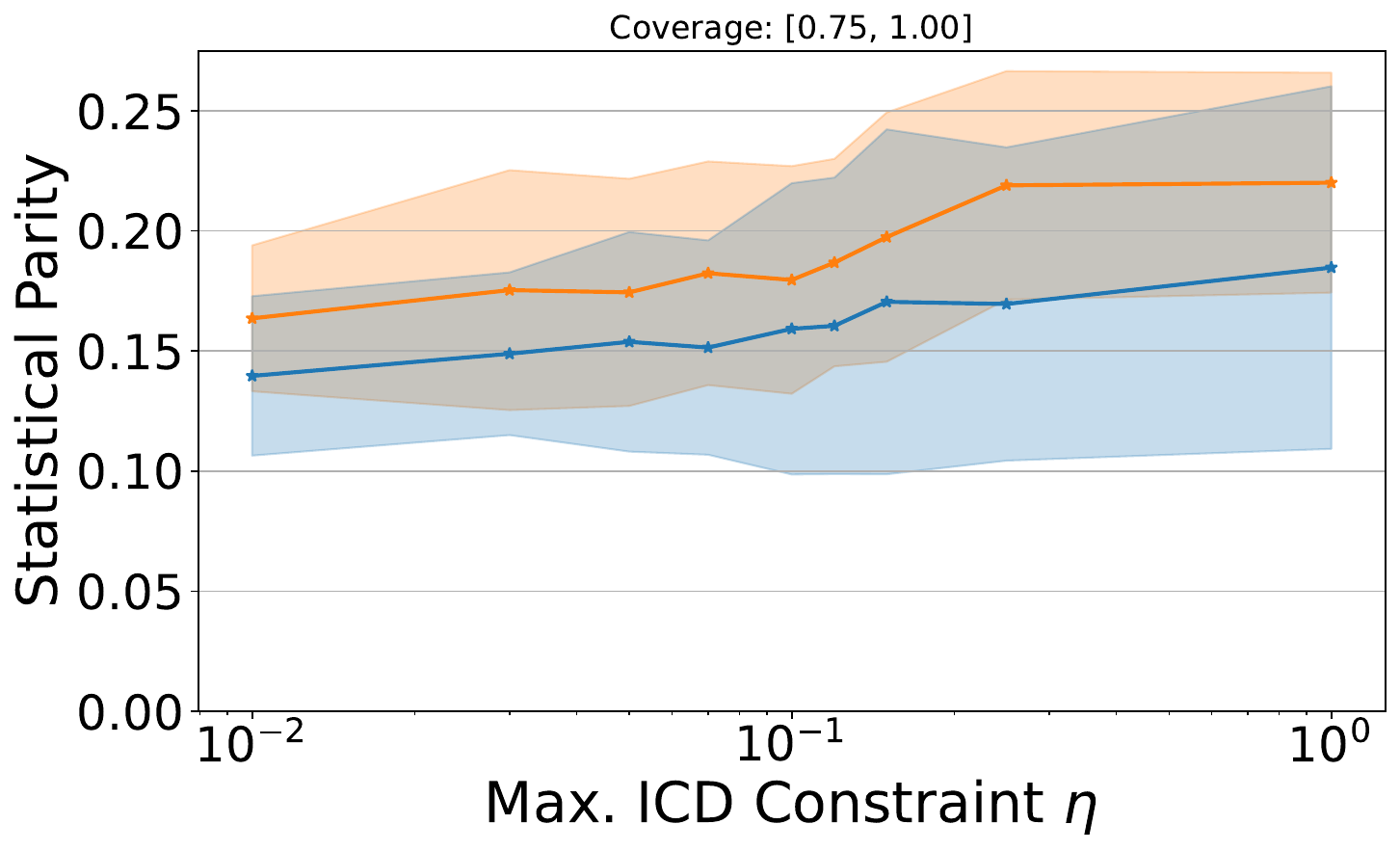}
    \caption{$Q_4$}
\end{subfigure}
\caption{Test set Statistical Parity (SP) for HybridCORELSPre and HybridCORELSPost with ICD mitigation across Rashomon sets over transparency bins ($\varepsilon = 0.01$). For different levels of mitigation induced by the maximum ICD constraint ($\eta$), curves report the average and standard deviation across Rashomon models. Results are provided for the COMPAS dataset and Gender as sensitive attribute.}
\label{SP_MaxCov_4Q}

\end{figure*}

\begin{figure*}[t]
\centering
\includegraphics[width=0.6\textwidth]{Plots/Fair/ICD_mitigation_shared_legend.pdf}

\begin{subfigure}{0.35\textwidth}
    \centering
    \includegraphics[width=\linewidth]{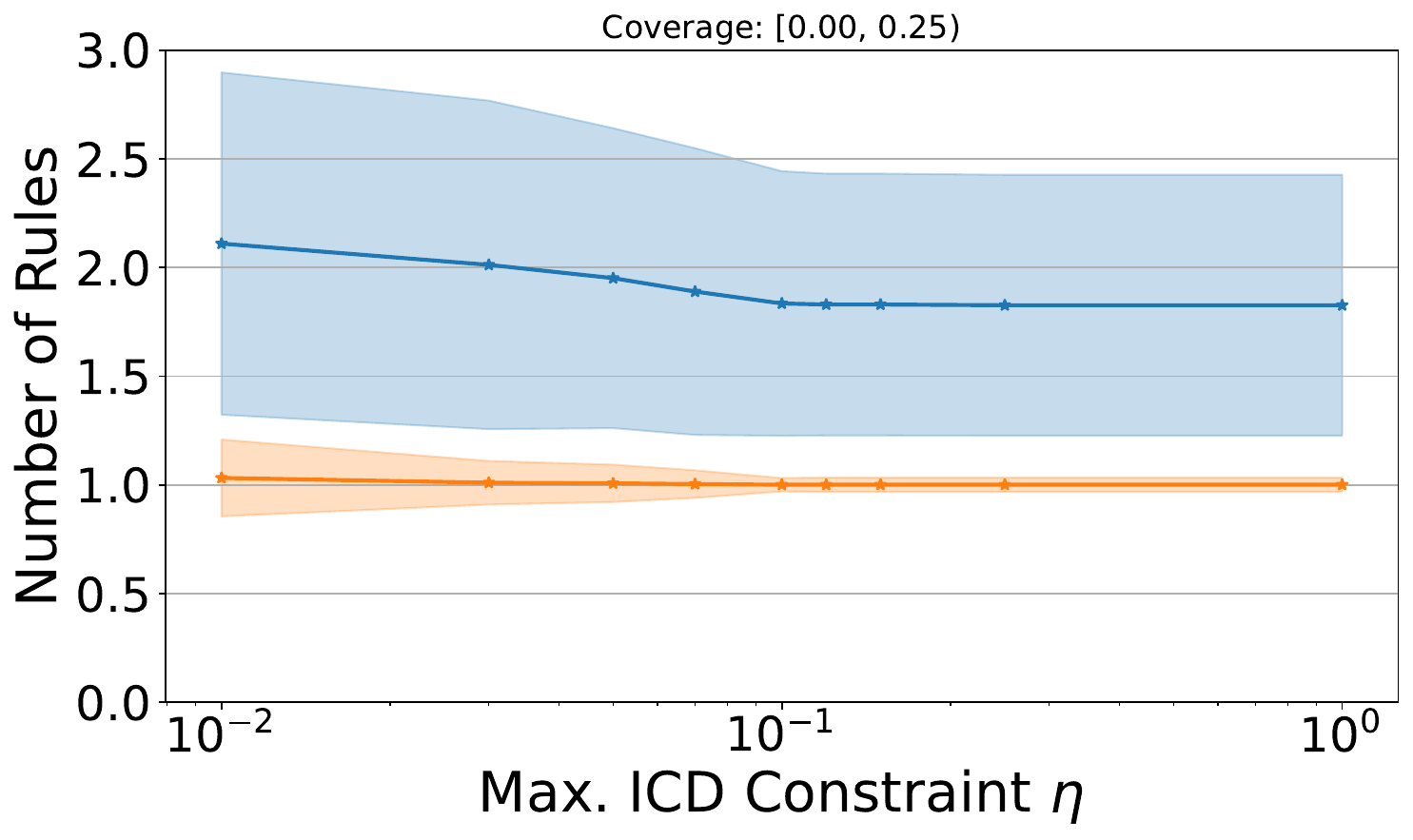}
    \caption{$Q_1$}
\end{subfigure}
\begin{subfigure}{0.35\textwidth}
    \centering
    \includegraphics[width=\linewidth]{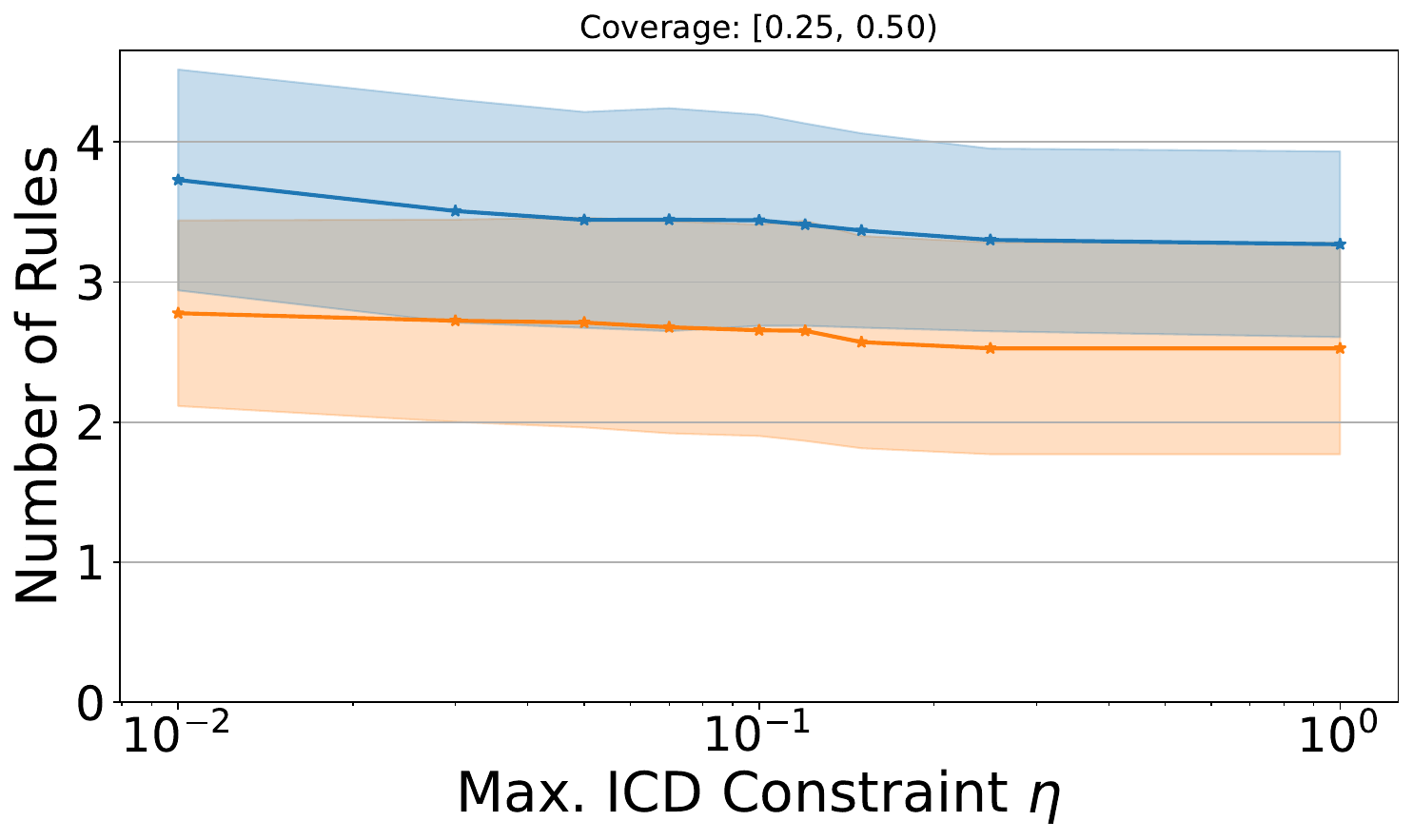}
    \caption{$Q_2$}
\end{subfigure}

\vspace{0.5em} %

\begin{subfigure}{0.35\textwidth}
    \centering
    \includegraphics[width=\linewidth]{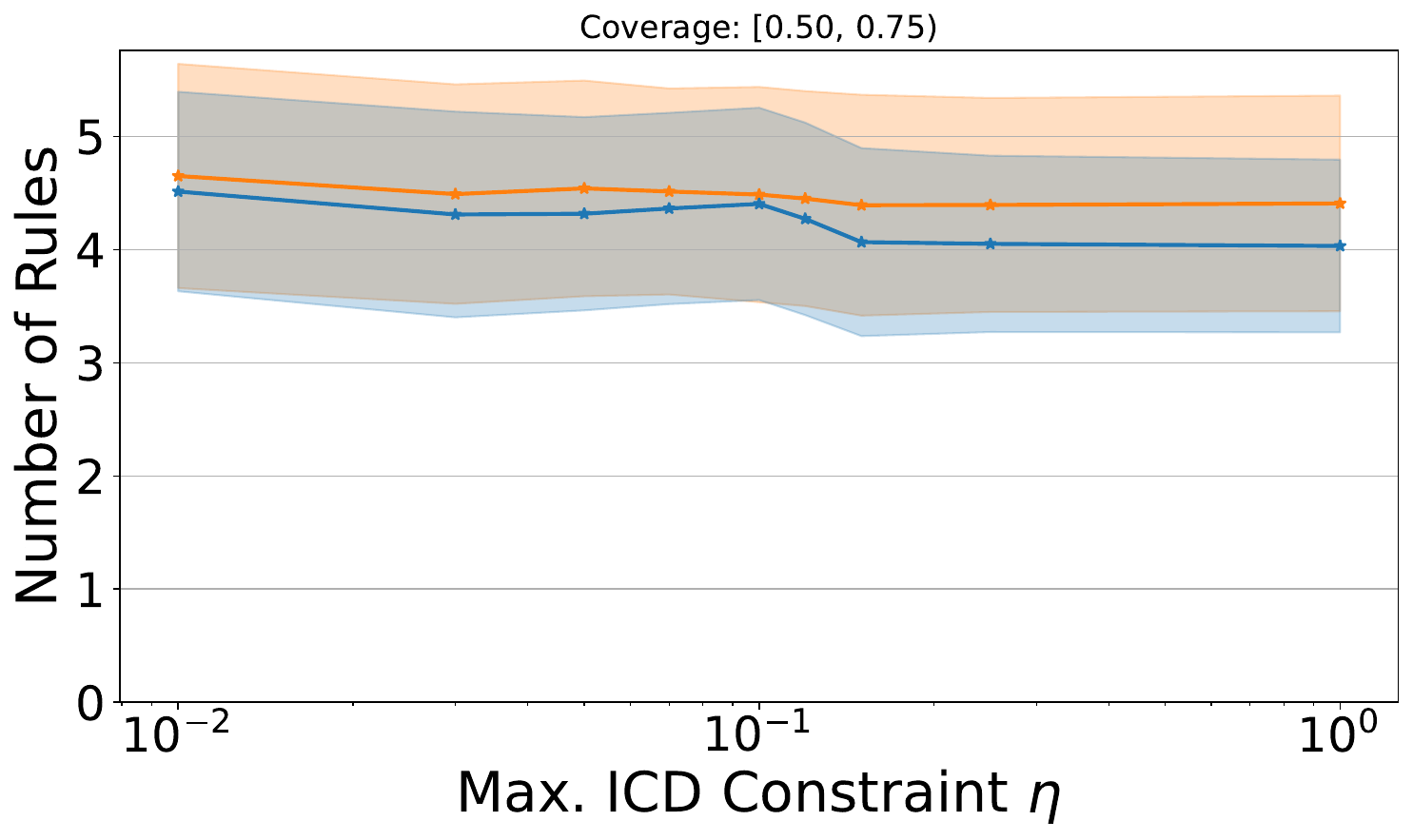}
    \caption{$Q_3$}
\end{subfigure}
\begin{subfigure}{0.35\textwidth}
    \centering
    \includegraphics[width=\linewidth]{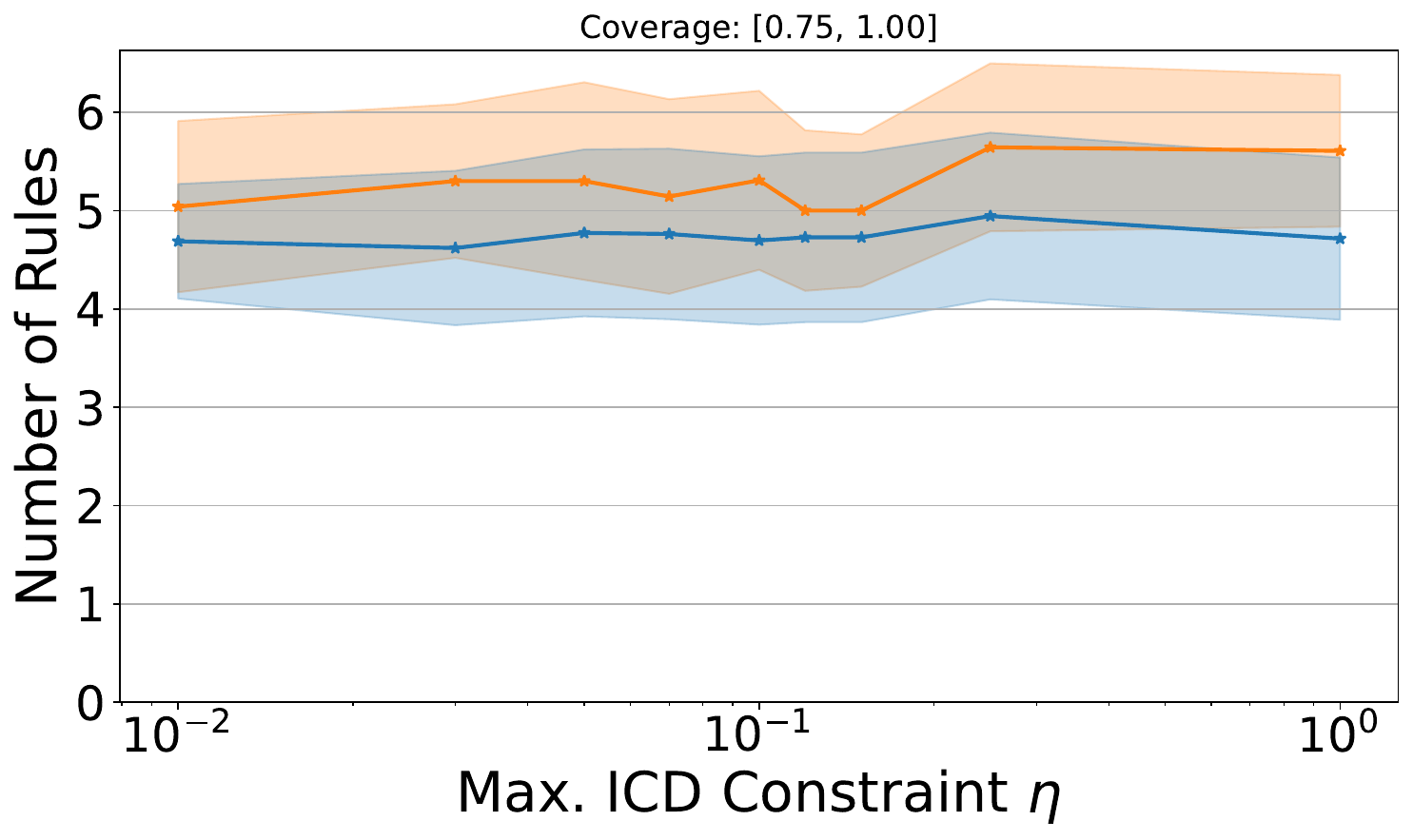}
    \caption{$Q_4$}
\end{subfigure}
\caption{Model sparsity for HybridCORELSPre and HybridCORELSPost with ICD mitigation across Rashomon sets over transparency bins ($\varepsilon = 0.01$). For different levels of mitigation induced by the maximum ICD constraint ($\eta$), curves report the average and standard deviation across Rashomon models. Results are provided for the COMPAS dataset and Gender as sensitive attribute.}
\label{Sparsity_MaxCov_4Q}

\end{figure*}

Finally, we provide in Figures~\ref{ACC_MaxCov_4Q} to~\ref{Sparsity_MaxCov_4Q} detailed analyses of how test accuracy, ICD, EO, SP, and sparsity evolve as the strength of mitigation is gradually relaxed from highly restrictive maximum ICD constraints $(\eta = 0.01)$ to the unconstrained setting $(\eta=1)$. In contrast to the aggregated analysis presented in Figure~\ref{allmetrics_MaxCov_1Q}, these additional results are evaluated across Rashomon sets over transparency bins, providing a more detailed view of the impact of mitigation at different transparency regimes. Overall, the observed trends are consistent with the discussion presented in Result 7, showing that substantial reductions in ICD can be achieved with limited impact on predictive performance and sparsity, while frequently yielding indirect improvements in algorithmic fairness metrics.

\end{document}